\documentclass[10pt,twocolumn,letterpaper]{article}

\usepackage{cvpr}
\usepackage{times}
\usepackage{epsfig}
\usepackage{graphicx}
\usepackage{amsmath}
\usepackage{amssymb}

\usepackage{booktabs}
\usepackage[table]{xcolor}

\usepackage[pagebackref=true,breaklinks=true,letterpaper=true,colorlinks,bookmarks=false]{hyperref}

\cvprfinalcopy 

\def\cvprPaperID{2195} 

\ifcvprfinal\fi

\definecolor{yelloworange}{RGB}{255, 153, 0}
\definecolor{ultramarineblue}{RGB}{65, 102, 245}

\def\be {\begin{equation}}
\def\ee {\end{equation}}
\def\beas {\begin{eqnarray*}}
\def\eeas {\end{eqnarray*}}
\def\bea {\begin{eqnarray}}
\def\eea {\end{eqnarray}}
\def\bes {\begin{equation*}}
\def\ees {\end{equation*}}
\def\ba {\begin{align}}
\def\ea {\end{align}}
\def\barr {\begin{array}}
\def\earr {\end{array}}

\newtheorem{lemma-ap}{Lemma}

\newtheorem{claim-ap}{Claim}

\makeatletter

\usepackage{xspace}
\def\@onedot{\ifx\@let@token.\else.\null\fi\xspace}
\DeclareRobustCommand\onedot{\futurelet\@let@token\@onedot}

\newcommand{\figref}[1]{Fig\onedot~\ref{#1}}
\newcommand{\equref}[1]{Eq\onedot~\eqref{#1}}

\newcommand{\tabref}[1]{Tab\onedot~\ref{#1}}

\newcommand{\by}[2]{\ensuremath{#1 \! \times \! #2}}

\def\eg{\emph{e.g}\onedot} 
\def\ie{\emph{i.e}\onedot} 
 
\def\etc{\emph{etc}\onedot} 
\def\wrt{w.r.t\onedot} 
\def\etal{\emph{et al}\onedot}

\begin{document}

\title{Attention to Scale: Scale-aware Semantic Image Segmentation}

\author{
\begin{tabular}[t]{c c c}
Liang-Chieh Chen\footnotemark & Yi Yang, Jiang Wang, Wei Xu & Alan L. Yuille \\
lcchen@cs.ucla.edu & \{yangyi05, wangjiang03, wei.xu\}@baidu.com  & yuille@stat.ucla.edu \\
& & alan.yuille@jhu.edu \\
\end{tabular}
}

\maketitle

\footnotetext{$^*$Work done in part during an internship at Baidu USA.}

\begin{abstract}
Incorporating multi-scale features in fully convolutional neural networks (FCNs) has been a key element to achieving state-of-the-art performance on semantic image segmentation.
One common way to extract multi-scale features is to feed multiple resized input images to a shared deep network and then merge the resulting features for pixel-wise classification.
In this work, we propose an attention mechanism that learns to softly weight the multi-scale features at each pixel location.
We adapt a state-of-the-art semantic image segmentation model, which we jointly train with multi-scale input images and the attention model. 
The proposed attention model not only outperforms average- and max-pooling, but allows us to diagnostically visualize the importance of features at different positions and scales. 
Moreover, we show that adding extra supervision to the output at each scale is essential to achieving excellent performance when merging multi-scale features. 
We demonstrate the effectiveness of our model with extensive experiments on three challenging datasets, including PASCAL-Person-Part, PASCAL VOC 2012 and a subset of MS-COCO 2014.

\end{abstract}

\section{Introduction}

\begin{figure} 
  \centering          
   \includegraphics[width=0.99\linewidth]{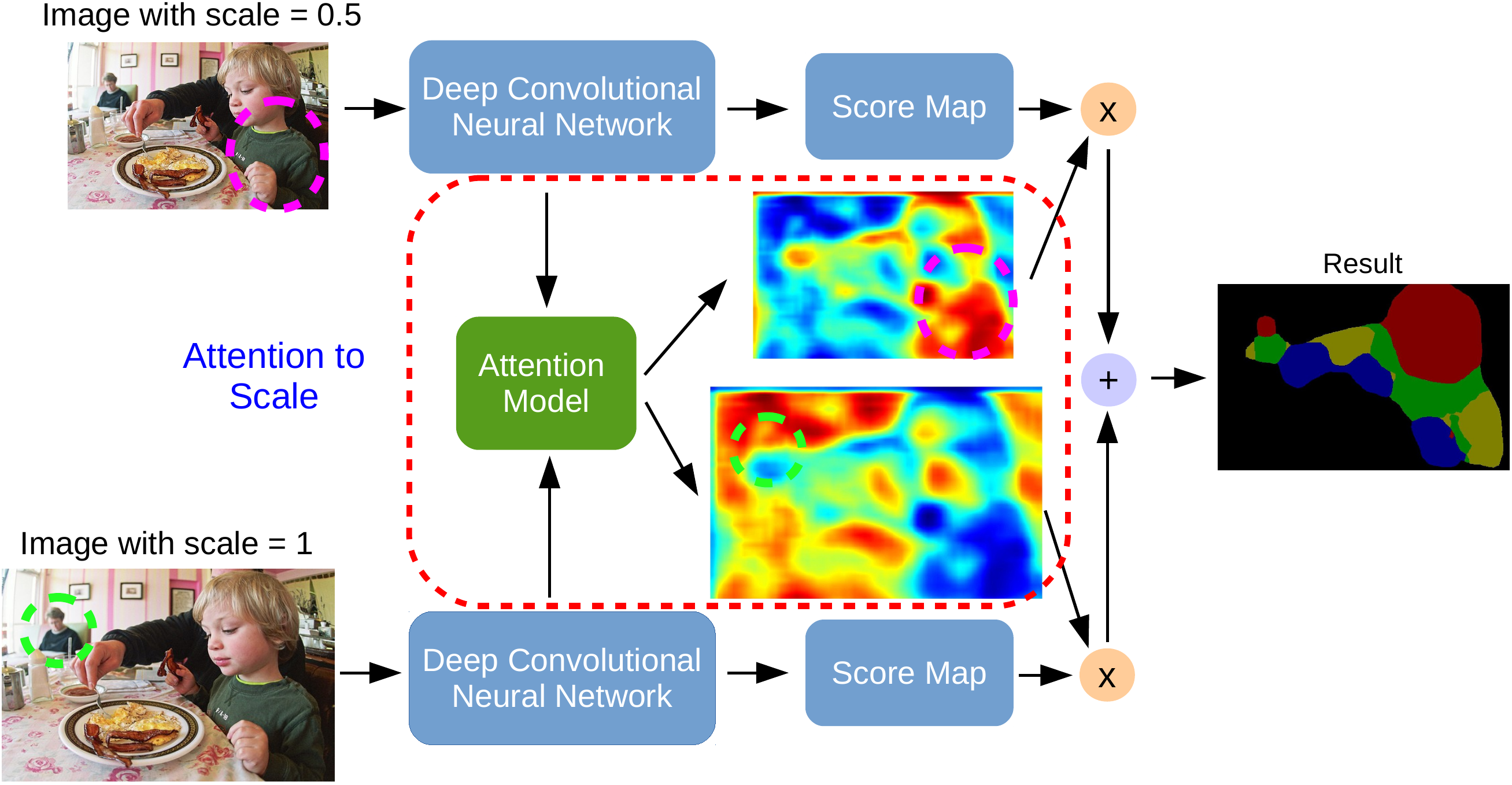} \\
  \caption{Model illustration. The attention model learns to put different weights on objects of different scales. For example, our model learns to put large weights on the small-scale person (green dashed circle) for features from scale = 1, and large weights on the large-scale child (magenta dashed circle) for features from scale = 0.5. We jointly train the network component and the attention model.}
  \label{fig:model_illustration}
\end{figure}  

Semantic image segmentation, also known as image labeling or scene parsing, relates to the problem of assigning semantic labels (\eg, ``person'' or ``dog'') to every pixel in the image. It is a very challenging task in computer vision and one of the most crucial steps towards scene understanding \cite{everingham2014pascal}. Successful image segmentation techniques could facilitate a large group of applications such as image editing \cite{evening2005adobe}, augmented reality \cite{azuma1997survey} and self-driving vehicles \cite{fritsch2013new}.

Recently, various methods  \cite{chen2014semantic, dai2015boxsup, liu2015semantic, noh2015learning, zheng2015conditional, lin2015efficient} based on {\it Fully Convolutional Networks} (FCNs) \cite{long2014fully} demonstrate astonishing results on several semantic segmentation benchmarks. Among these models, one of the key elements to successful semantic segmentation is the use of multi-scale features \cite{farabet2013learning, pinheiro2013recurrent, hariharan2014hypercolumns, long2014fully, mostajabi2014feedforward, lin2015efficient}. In the FCNs setting, there are mainly two types of network structures that exploit multi-scale features \cite{xie2015holistically}. 

The first type, which we refer to as {\it skip-net}, combines features from the intermediate layers of FCNs \cite{hariharan2014hypercolumns, long2014fully, mostajabi2014feedforward, chen2014semantic}.  Features within a skip-net are multi-scale in nature due to the increasingly large receptive field sizes. During training, a skip-net usually employs a two-step process \cite{hariharan2014hypercolumns, long2014fully, mostajabi2014feedforward, chen2014semantic}, where it first trains the deep network backbone and then fixes or slightly fine-tunes during multi-scale feature extraction. The problem with this strategy is that the training process is not ideal (\ie, classifier training and feature-extraction are separate) and the training time is usually long (\eg, three to five days \cite{long2014fully}).

The second type, which we refer to as {\it share-net}, resizes the input image to several scales and passes each through a shared deep network. It then computes the final prediction based on the fusion of the resulting multi-scale features \cite{farabet2013learning, lin2015efficient}. A share-net does not need the two-step training process mentioned above. It usually employs average- or max-pooling over scales \cite{felzenszwalb2010object, ciresan2012multi, papandreou2014untangling, dai2015boxsup}. Features at each scale are either equally important or sparsely selected.

Recently, attention models have shown great success in several computer vision and natural language processing tasks \cite{bahdanau2014neural, mnih2014recurrent, xu2015show, chen2015abc}. Rather than compressing an entire image or sequence into a static representation, attention allows the model to focus on the most relevant features as needed. 
In this work, we incorporate an attention model for semantic image segmentation.
Unlike previous work that employs attention models in the 2D spatial and/or temporal dimension \cite{sharma2015action, yao2015describing}, we explore its effect in the scale dimension.

In particular, we adapt a state-of-the-art semantic segmentation model \cite{chen2014semantic} to a share-net and employ a soft attention model \cite{bahdanau2014neural} to generalize average- and max-pooling over scales, as shown in \figref{fig:model_illustration}. 
The proposed attention model learns to weight the multi-scale features according to the object scales presented in the image (\eg, the model learns to put large weights on features at coarse scale for large objects). 
For each scale, the attention model outputs a {\it weight map} which weights features pixel by pixel, and the weighted sum of FCN-produced score maps across all scales is then used for classification. 

Motivated by \cite{bengio2007greedy, lee2014deeply, szegedy2014going, xie2015holistically}, we further introduce extra supervision to the output of FCNs at each scale, which we find essential for a better performance. We jointly train the attention model and the multi-scale networks. We demonstrate the effectiveness of our model on several challenging datasets, including PASCAL-Person-Part \cite{chen_cvpr14}, PASCAL VOC 2012 \cite{everingham2014pascal}, and a
subset of MS-COCO 2014 \cite{lin2014microsoft}. Experimental results show that our proposed method consistently improves over strong baselines. The attention component also gives a non-trivial improvement over average- and max-pooling methods. More importantly, the proposed attention model provides diagnostic visualization, unveiling the black box network operation by visualizing the importance of features at each scale for every image position.

\section{Related Work}
Our model draws success from several areas, including deep networks, multi-scale features for semantic segmentation, and attention models.

\textbf{Deep networks:} Deep Convolutional Neural Networks (DCNNs) \cite{lecun1998gradient} have demonstrated state-of-the-art performance on several computer vision tasks, including image classification \cite{krizhevsky2012imagenet, sermanet2013overfeat, szegedy2014going, simonyan2014very, papandreou2014untangling} and object detection \cite{girshick2014rich, he2014spatial}. For the semantic image segmentation task, state-of-the-art methods are variants of the fully convolutional
neural networks (FCNs) \cite{long2014fully}, including \cite{chen2014semantic, dai2015boxsup, lin2015efficient, noh2015learning, zheng2015conditional}. In particular, our method builds upon the current state-of-the-art DeepLab model \cite{chen2014semantic}. 

\begin{figure}    
  \centering
  \addtolength{\tabcolsep}{-5pt}
  \begin{tabular}{cc}
   \includegraphics[height=0.59\linewidth]{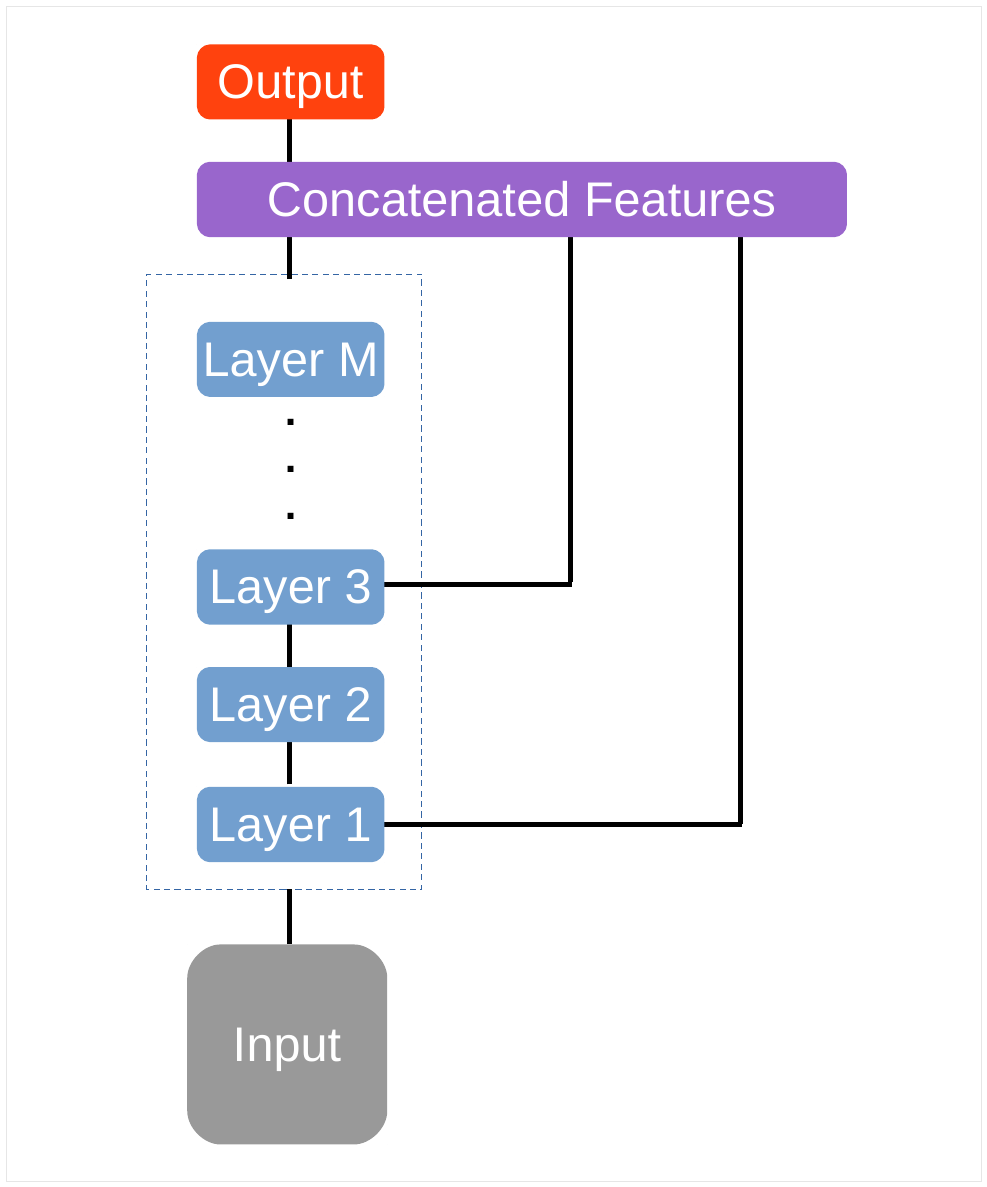} &
   \includegraphics[height=0.59\linewidth]{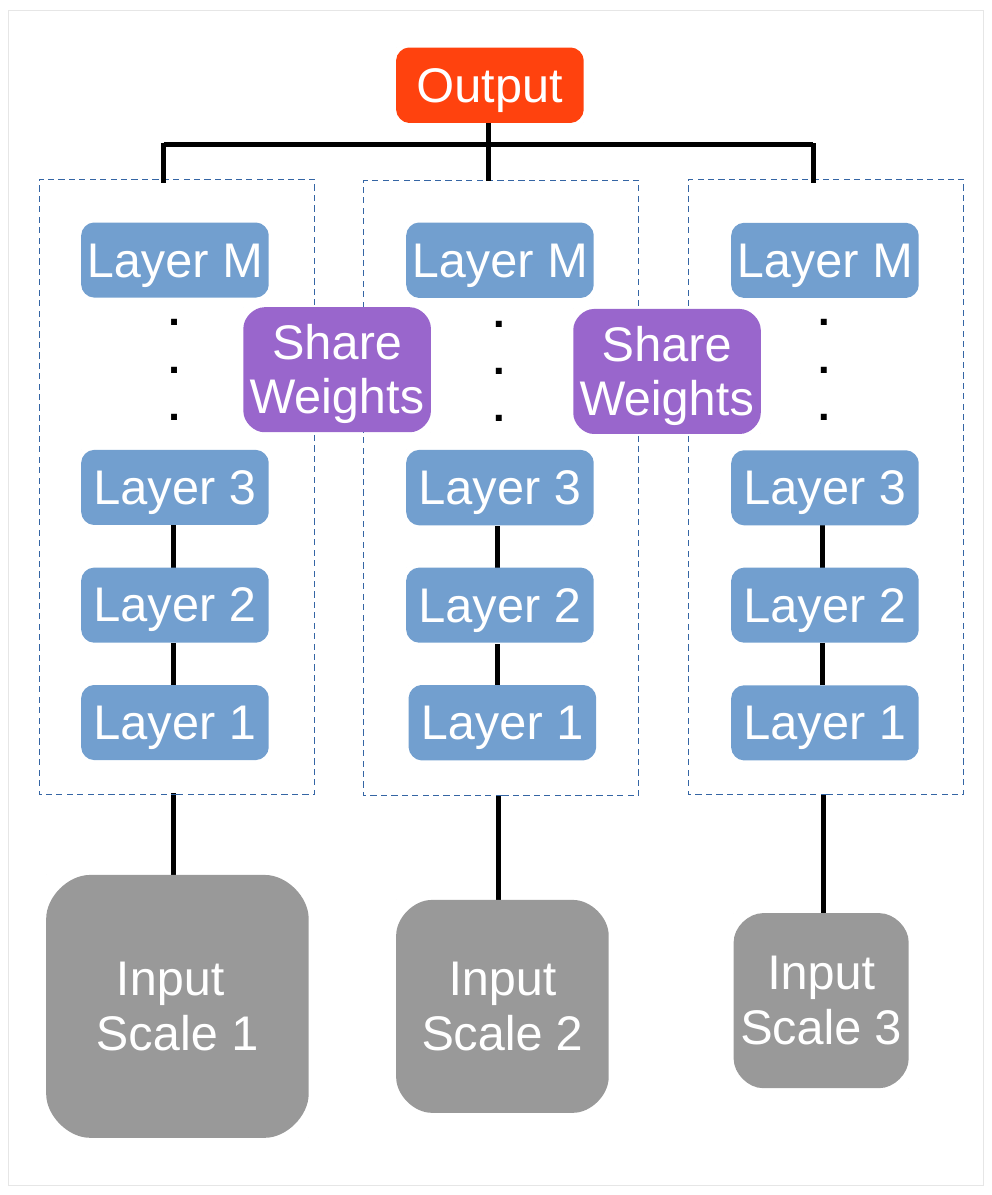} \\
   (a) skip-net &
   (b) share-net \\
  \end{tabular}
  \caption{Different network structures for extracting multi-scale features: (a) Skip-net: features from intermediate layers are fused to produce the final output. (b) Share-net: multi-scale inputs are applied to a shared network for prediction. In this work, we demonstrate the effectiveness of the share-net when combined with attention mechanisms over scales.}
  \label{fig:nets}
\end{figure}  

\textbf{Multi-scale features:} It is known that multi-scale features are useful for computer vision tasks, \eg, \cite{florack1996gaussian, arbelaez2011contour}. 
In the context of deep networks for semantic segmentation, we mainly discuss two types of networks that exploit multi-scale features. The first type, {\it skip-net}, exploits features from different levels of the network. For example, FCN-8s \cite{long2014fully} gradually learns finer-scale prediction from lower layers (initialized with coarser-scale prediction). Hariharan \etal \cite{hariharan2014hypercolumns} classified a pixel with hypercolumn representation (\ie, concatenation of
features from intermediate layers). Mostajabi \etal \cite{mostajabi2014feedforward} classified a superpixel with features extracted at zoom-out spatial levels from a small proximal neighborhood to the whole image region. DeepLab-MSc (DeepLab with Multi-Scale features) \cite{chen2014semantic} applied Multi-Layer Perceptrons (MLPs) to the input image and to the outputs of pooling layers, in order to extract multi-scale features. ParseNet \cite{liu2015parsenet} aggregated features over the whole image to provide global contextual information. 

The second type, {\it share-net}, applies multi-scale input images to a shared network. For example, Farabet \etal \cite{farabet2013learning} employed a Laplacian pyramid, passed each scale through a shared network, and fused the features from all the scales. Lin \etal \cite{lin2015efficient} resized the input image for three scales and concatenated the resulting three-scale features to generate the unary and pairwise potentials of a Conditional Random Field (CRF). Pinheiro \etal
\cite{pinheiro2013recurrent}, instead of applying multi-scale input images at once, fed multi-scale images at different stages in a recurrent convolutional neural network. This share-net strategy has also been employed during the test stage for a better performance by Dai \etal \cite{dai2015boxsup}. In this work, we extend DeepLab \cite{chen2014semantic} to be a type of {\it share-net} and demonstrate its effectiveness on three challenging datasets. Note that Eigen and Fergus
\cite{eigen2014predicting} fed input images to DCNNs at three scales from coarse to fine sequentially. The DCNNs at different scales have different structures, and a two-step training process is required for their model.

\textbf{Attention models for deep networks:} In computer vision, attention models have been used widely used for image classification \cite{cao2015look, gregor2015draw, xiao2015application} and object detection \cite{ba2014multiple, caicedo2015active, yoo2015attentionnet}. Mnih \etal \cite{mnih2014recurrent} learn an attention model that adaptively selects image regions for processing. However, their attention model is not differentiable, which is necessary for standard backpropagation during training. On the other hand, Gregor \etal \cite{gregor2015draw} employ a differentiable attention model to specify where to read/write image regions for image generation. 

Bahdanau \etal \cite{bahdanau2014neural} propose an attention model that softly weights the importance of input words in a source sentence when predicting a target word for machine translation. Following this, Xu \etal \cite{xu2015show} and Yao \etal \cite{yao2015describing} use attention models for image captioning and video captioning respectively. These methods apply attention in the 2D spatial and/or temporal dimension while we use attention to identify the most relevant scales. 

\textbf{Attention to scale:} To merge the predictions from multi-scale features, there are two common approachs: average-pooling \cite{ciresan2012multi, dai2015boxsup} or max-pooling \cite{felzenszwalb2010object, papandreou2014untangling} over scales. Motivated by \cite{bahdanau2014neural}, we propose to jointly learn an attention model that softly weights the features from different input scales when predicting the semantic label of a pixel. The final output of our model is produced by the
weighted sum of score maps across all the scales. We show that the proposed attention model not only improves performance over average- and max-pooling, but also allows us to diagnostically {\it visualize} the importance of features at different positions and scales, separating us from existing work that exploits multi-scale features for semantic segmentation.

\section{Model}
\label{sec:model}

\subsection{Review of DeepLab}
FCNs have proven successful in semantic image segmentation \cite{dai2015boxsup, liu2015semantic, zheng2015conditional}. In this subsection, we briefly review the DeepLab model \cite{chen2014semantic}, which is a variant of FCNs \cite{long2014fully}. 

DeepLab adopts the 16-layer architecture of state-of-the-art classification network of \cite{simonyan2014very} (\ie, VGG-16 net). The network is modified to be fully convolutional \cite{long2014fully}, producing dense feature maps. In particular, the last fully-connected layers of the original VGG-16 net are turned into convolutional layers (\eg, the last layer has a spatial convolutional kernel with size $\by{1}{1}$). The spatial decimation factor of the original VGG-16 net is 32 because of
the employment of five max-pooling layers each with stride $2$. DeepLab reduces it to 8 by using the {\`a} trous (with holes) algorithm \cite{Mall99}, and employs linear interpolation to upsample by a factor of 8 the score maps of the final layer to original image resolution. There are several variants of DeepLab \cite{chen2014semantic}. In this work, we mainly focus on DeepLab-LargeFOV. The suffix, LargeFOV, comes from the fact that the model adjusts the filter weights at the
convolutional variant of $fc_6$ ($fc_6$ is the original first fully connected layer in VGG-16 net) with {\`a} trous algorithm so that its Field-Of-View is larger. 


\subsection{Attention model for scales}

Herein, we discuss how to merge the multi-scale features for our proposed model. We propose an attention model that learns to weight the multi-scale features. Average pooling \cite{ciresan2012multi, dai2015boxsup} or max pooling \cite{felzenszwalb2010object, papandreou2014untangling} over scales to merge features can be considered as special cases of our method.

\begin{figure}
  \centering   
  \addtolength{\tabcolsep}{-5pt}       
  \begin{tabular}{cc}
   \includegraphics[height=0.53\linewidth]{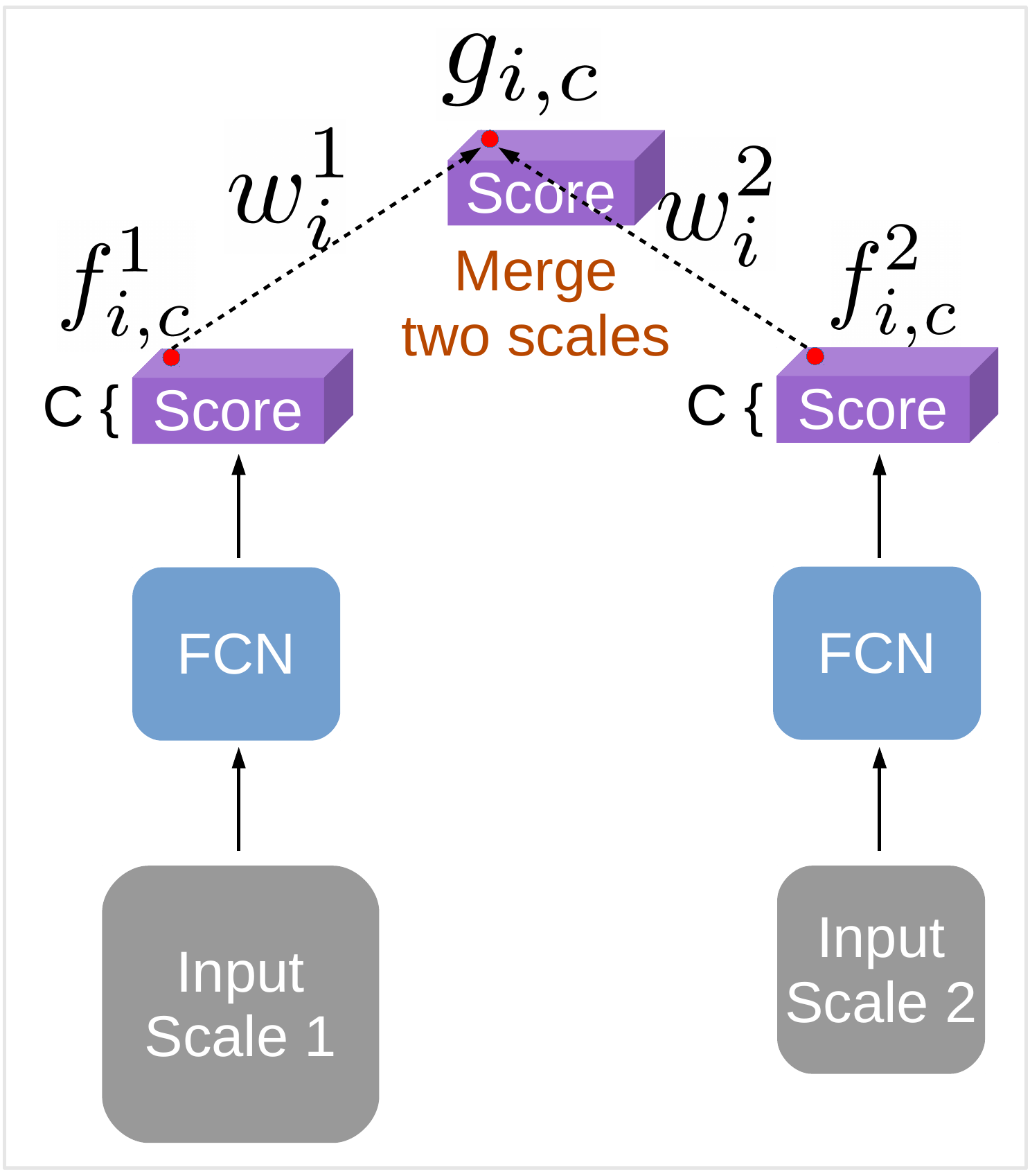} &
   \includegraphics[height=0.53\linewidth]{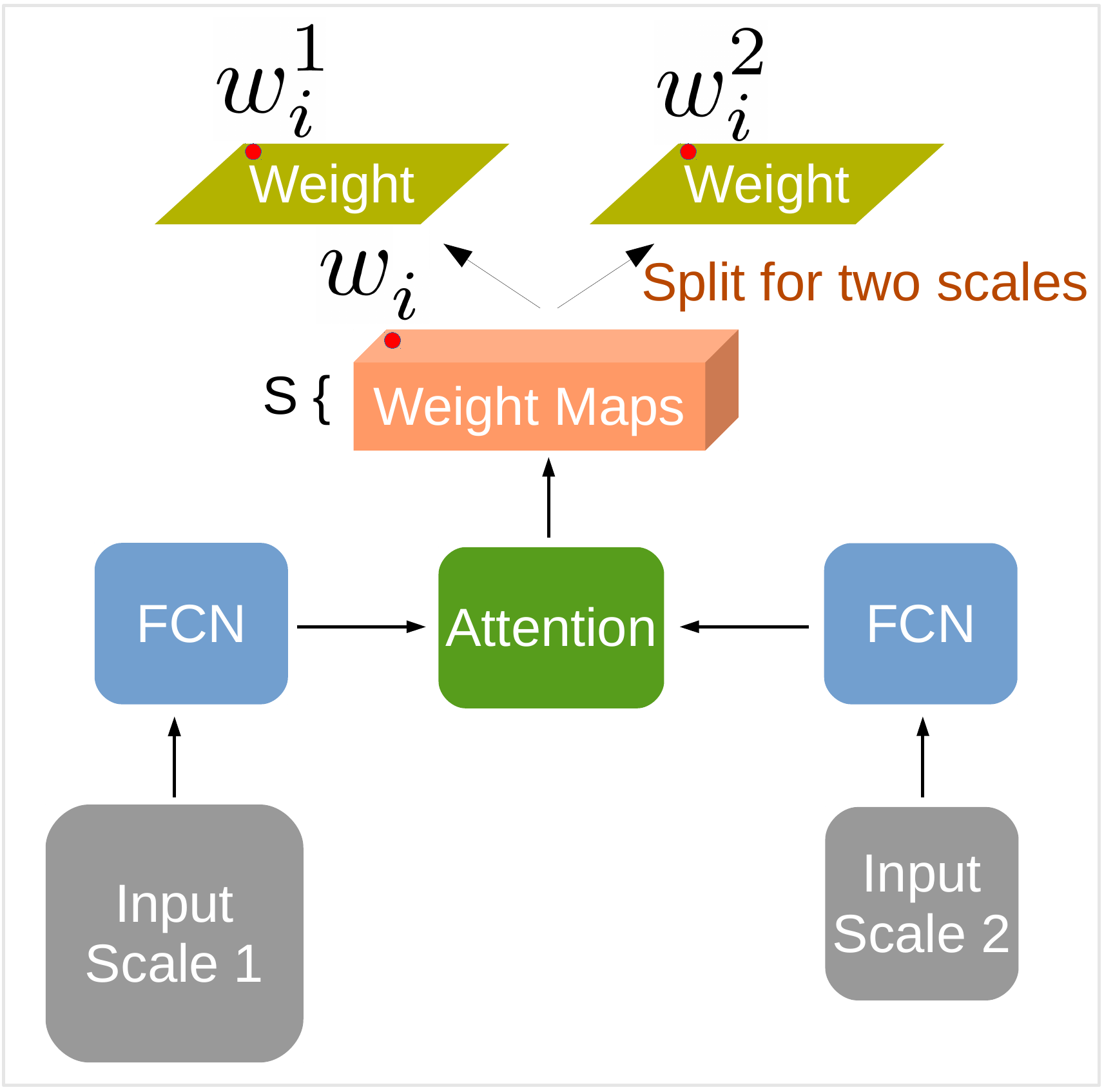} \\
   (a)  &
   (b)  \\
  \end{tabular}
  \caption{(a) Merging score maps (\ie, last layer output before SoftMax) for two scales. (b) Our proposed attention model makes use of features from FCNs and produces weight maps, reflecting how to do a weighted merge of the FCN-produced score maps at different scales and at different positions.}
  \label{fig:models}
\end{figure}  

Based on share-net, suppose an input image is resized to several scales $s \in \{1, ..., S\}$. Each scale is passed through the DeepLab (the FCN weights are shared across all scales) and produces a score map for scale $s$, denoted as $f^s_{i,c}$ where $i$ ranges over all the spatial positions (since it is fully convolutional) and $c \in \{1, ..., C\}$ where $C$ is the number of classes of interest. The score maps $f^s_{i,c}$ are resized to have the same resolution (with respect to the finest scale) by bilinear interpolation. We denote $g_{i,c}$ to be the weighted sum of score maps at $(i,c)$ for all scales, \ie,

\begin{equation}
\label{eq:weighted_sum}
g_{i,c} = \sum_{s=1}^S w^s_{i} \cdot f^s_{i,c}
\end{equation}

The weight $w^s_{i}$ is computed by
\begin{equation}
w^s_{i} = \frac{\exp(h^s_{i})}{\sum_{t=1}^S \exp(h^t_{i})}
\end{equation}
where $h^s_{i}$ is the score map (\ie, last layer output before SoftMax) produced by the attention model at position $i$ for scale $s$. Note $w^s_{i}$ is shared across all the channels. The attention model is parameterized by another FCN so that dense maps are produced. The proposed attention model takes as input the convolutionalized $fc_7$ features from VGG-16 \cite{simonyan2014very}, and it consists of two layers (the first layer has 512 filters with kernel size $\by{3}{3}$ and second layer has $S$ filters with kernel size $\by{1}{1}$ where $S$ is the number of scales employed). We will discuss this design choice in the experimental results.

The weight $w^s_{i}$ reflects the importance of feature at position $i$ and scale $s$. As a result, the attention model decides how much attention to pay to features at different positions and scales. It further enables us to visualize the attention for each scale by visualizing $w^s_{i}$. Note in our formulation, average-pooling or max-pooling over scales are two special cases. In particular, the weights $w^s_{i}$ in \equref{eq:weighted_sum} will be replaced by $1/S$ for average-pooling, while the summation in \equref{eq:weighted_sum} becomes the $\max$ operation and $w^s_{i}=1 \ \forall s $ and $i$ in the case of max-pooling.


We emphasize that the attention model computes a soft weight for each scale and position, and it allows the gradient of the loss function to be backpropagated through, similar to \cite{bahdanau2014neural}. Therefore, we are able to jointly train the attention model as well as the FCN (\ie, DeepLab) part end-to-end. One advantage of the proposed joint training is that tedious annotations of the ``ground truth scale'' for each pixel is avoided, letting the model adaptively find the best weights on scales.

\subsection{Extra supervision} 
We learn the network parameters using training images annotated at the pixel-level. The final output is produced by performing a softmax operation on the merged score maps across all the scales. We minimize the cross-entropy loss averaged over all image positions with Stochastic Gradient Descent (SGD). The network parameters are initialized from the ImageNet-pretrained VGG-16 model of \cite{simonyan2014very}.

In addition to the supervision introduced to the final output, we add extra supervision to the FCN for each scale \cite{lee2014deeply}. The motivation behind this is that we would like to merge {\it discriminative} features (after pooling or attention model) for the final classifier output. As pointed out by \cite{lee2014deeply}, discriminative classifiers trained with discriminative features demonstrate better performance for classification tasks. Instead of adding extra supervision to the intermediate layers \cite{bengio2007greedy, lee2014deeply, szegedy2014going, xie2015holistically}, we inject extra supervision to the final output of DeepLab for each scale so that the features to be merged are trained to be more discriminative. Specifically, the total loss function contains $1+S$ cross entropy loss functions (one for final output and one for each scale) with weight one for each. The ground truths are downsampled properly \wrt. the output resolutions during training. 


\section{Experimental Evaluations}
\label{sec:exp}

In this section, after presenting the common setting for all the experiments, we evaluate our method on three datasets, including PASCAL-Person-Part \cite{chen_cvpr14}, PASCAL VOC 2012 \cite{everingham2014pascal}, and a subset of MS-COCO 2014 \cite{lin2014microsoft}.

\textbf{Network architectures:} Our network is based on the publicly available model, DeepLab-LargeFOV \cite{chen2014semantic}, which modifies VGG-16 net \cite{simonyan2014very} to be FCN \cite{long2014fully}. We employ the same settings for DeepLab-LargeFOV as \cite{chen2014semantic}.


\textbf{Training:} SGD with mini-batch is used for training. We set the mini-batch size of 30 images and initial learning rate of 0.001 (0.01 for the final classifier layer). The learning rate is multiplied by 0.1 after 2000 iterations. We use the momentum of 0.9 and weight decay of 0.0005. Fine-tuning our network on all the reported experiments takes about 21 hours on an NVIDIA Tesla K40 GPU. During training, our model takes all scaled inputs and performs training jointly. Thus, the total training time is twice that of a vanilla DeepLab-LargeFOV. The average inference time for one PASCAL image is 350 ms.

\textbf{Evaluation metric:} The performance is measured in terms of pixel intersection-over-union (IOU) averaged across classes \cite{everingham2014pascal}.

\textbf{Reproducibility:} The proposed methods are implemented by extending Caffe framework \cite{jia2014caffe}. The code and models are available at \url{http://liangchiehchen.com/projects/DeepLab.html}.

\textbf{Experiments:} To demonstrate the effectiveness of our model, we mainly experiment along three axes: (1) multi-scale inputs (from one scale to three scales with $s \in \{1, 0.75, 0.5\}$), (2) different methods (average-pooling, max-pooling, or attention model) to merge multi-scale features, and (3) training with or without extra supervision.


\subsection{PASCAL-Person-Part}

\begin{table}
  \centering
  \rowcolors{2}{}{yelloworange!25}
  \addtolength{\tabcolsep}{2.5pt}
    \begin{tabular}{l c c}
      \toprule[0.2 em]
      \multicolumn{2}{l}{Baseline: DeepLab-LargeFOV} & 51.91  \\
      \toprule[0.2 em]
      {\bf Merging Method} &  & w/ E-Supv \\
      \midrule
      \midrule
      {\it Scales = \{1, 0.5\}} & & \\
      Max-Pooling & 52.90 & 55.26 \\
      Average-Pooling & 52.71 & 55.17 \\
      Attention & 53.49 & 55.85 \\
      \midrule
      {\it Scales = \{1, 0.75, 0.5\}} & & \\
      Max-Pooling & 53.02 & 55.78\\
      Average-Pooling & 52.56 & 55.72 \\
      Attention & 53.12 & {\bf 56.39} \\
      \bottomrule[0.1 em]
    \end{tabular}
    \caption{Results on PASCAL-Person-Part {\it validation} set. E-Supv: extra supervision.}
    \label{tab:deeplab_part}
\end{table}

\begin{table}
  \centering
  \small
  \rowcolors{2}{}{yelloworange!25}
  \addtolength{\tabcolsep}{-2.5pt}
    \begin{tabular}{lccccccc}
    \toprule[0.2 em]
    Head & Torso & U-arms & L-arms & U-legs & L-legs & Bkg & Avg \\
    \midrule
    81.47 & 59.06 & 44.15 & 42.50 & 38.28 & 35.62 & 93.65 & 56.39 \\
    \bottomrule[0.1 em]
    \end{tabular}
    \caption{Per-part results on PASCAL-Person-Part {\it validation} set with our attention model.}
    \label{tab:deeplab_everypart}
\end{table}

\textbf{Dataset:} We perform experiments on semantic part segmentation, annotated by \cite{chen_cvpr14} from the PASCAL VOC 2010 dataset. Few works \cite{wang2014semantic, wang2015joint} have worked on the animal part segmentation for the dataset. On the other hand, we focus on the {\it person} part for the dataset, which contains more training data and large scale variation. Specifically, the dataset contains detailed part annotations for every person, including eyes, nose, \etc. We merge the annotations to be Head, Torso, Upper/Lower Arms and Upper/Lower Legs, resulting in six person part classes and one background class. We only use those images containing persons for training (1716 images) and validation (1817 images).

\begin{figure*} 
  \centering          
  \begin{tabular}{c c c | c c c}
    & \includegraphics[height=0.1\linewidth, width=0.1\linewidth]{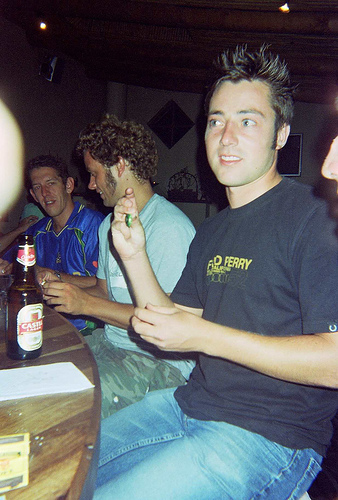} & & & \includegraphics[height=0.1\linewidth, width=0.1\linewidth]{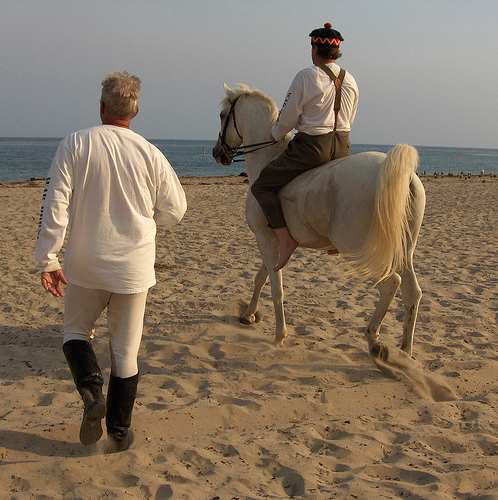} & \\
   \includegraphics[height=0.1\linewidth, width=0.14\linewidth]{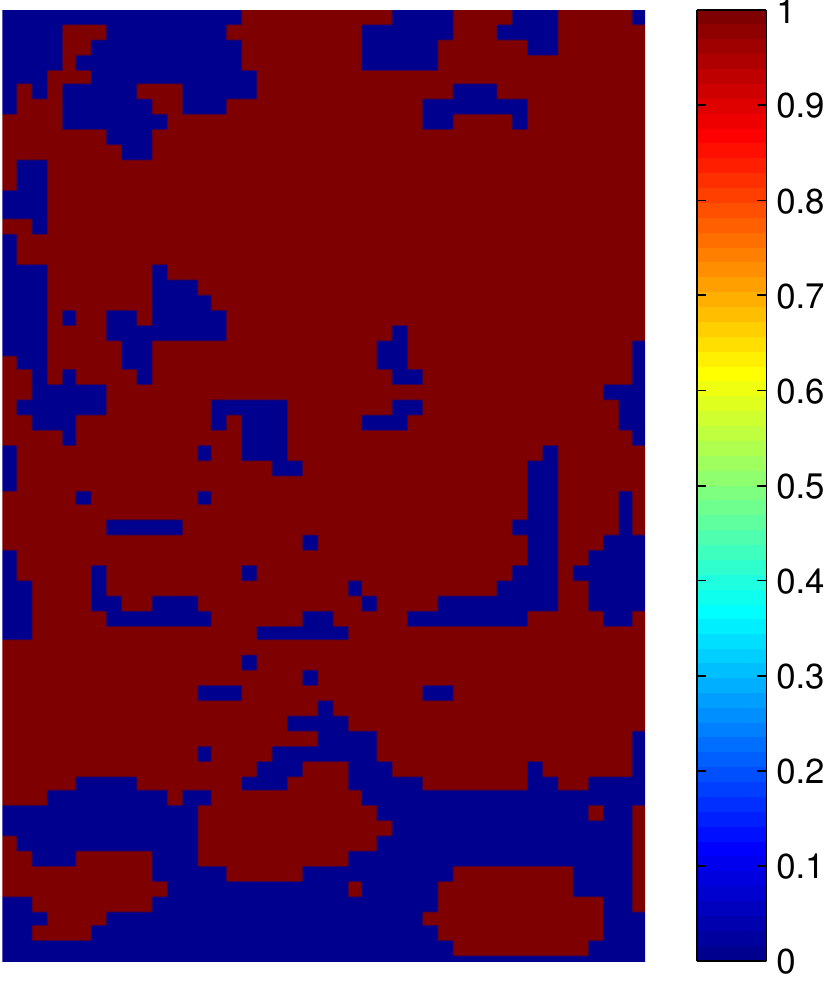} &
   \includegraphics[height=0.1\linewidth, width=0.14\linewidth]{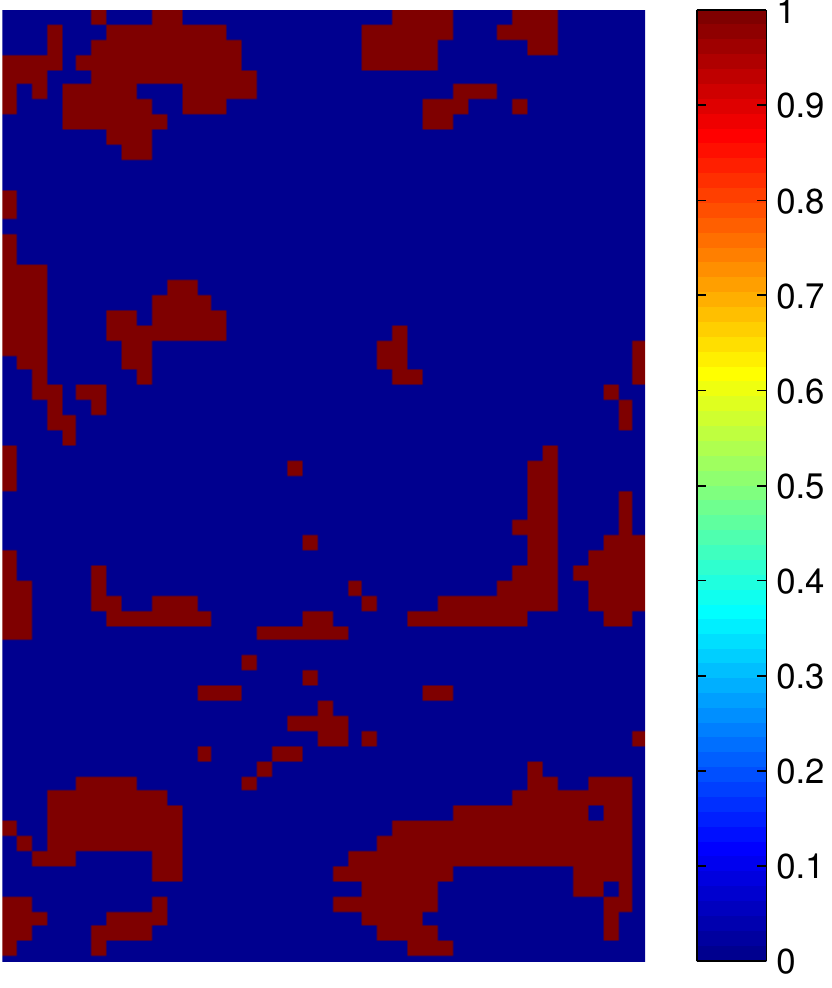} &
   \includegraphics[height=0.1\linewidth, width=0.14\linewidth]{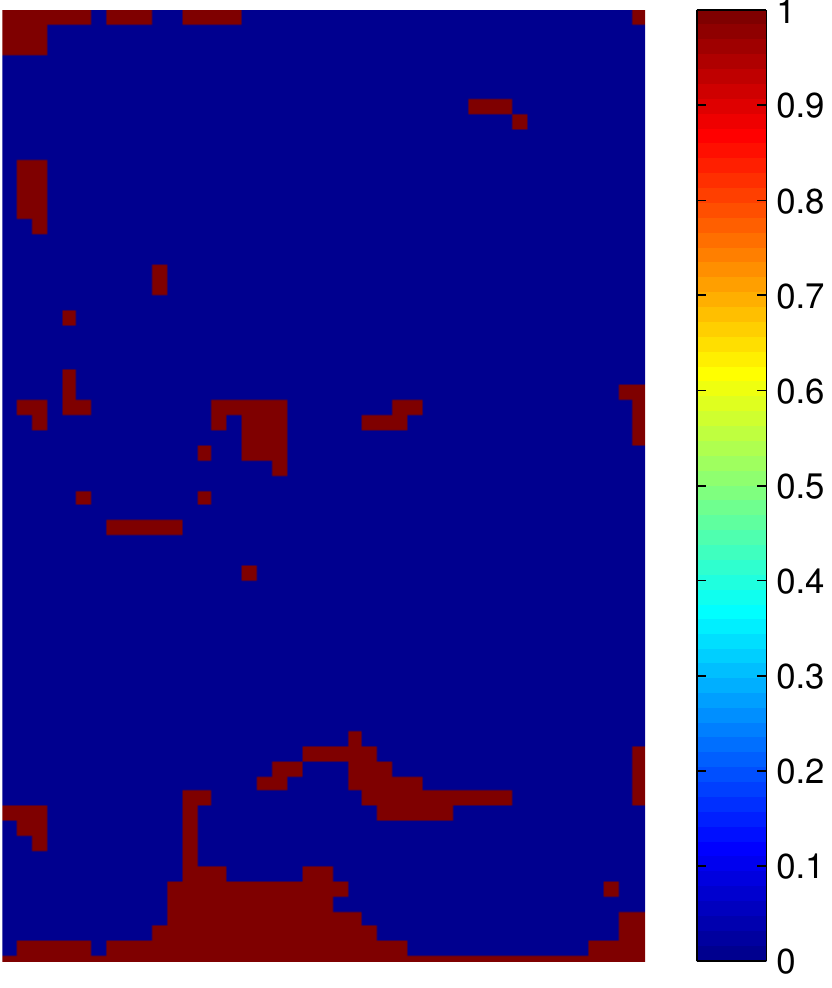} &
   \includegraphics[height=0.1\linewidth, width=0.14\linewidth]{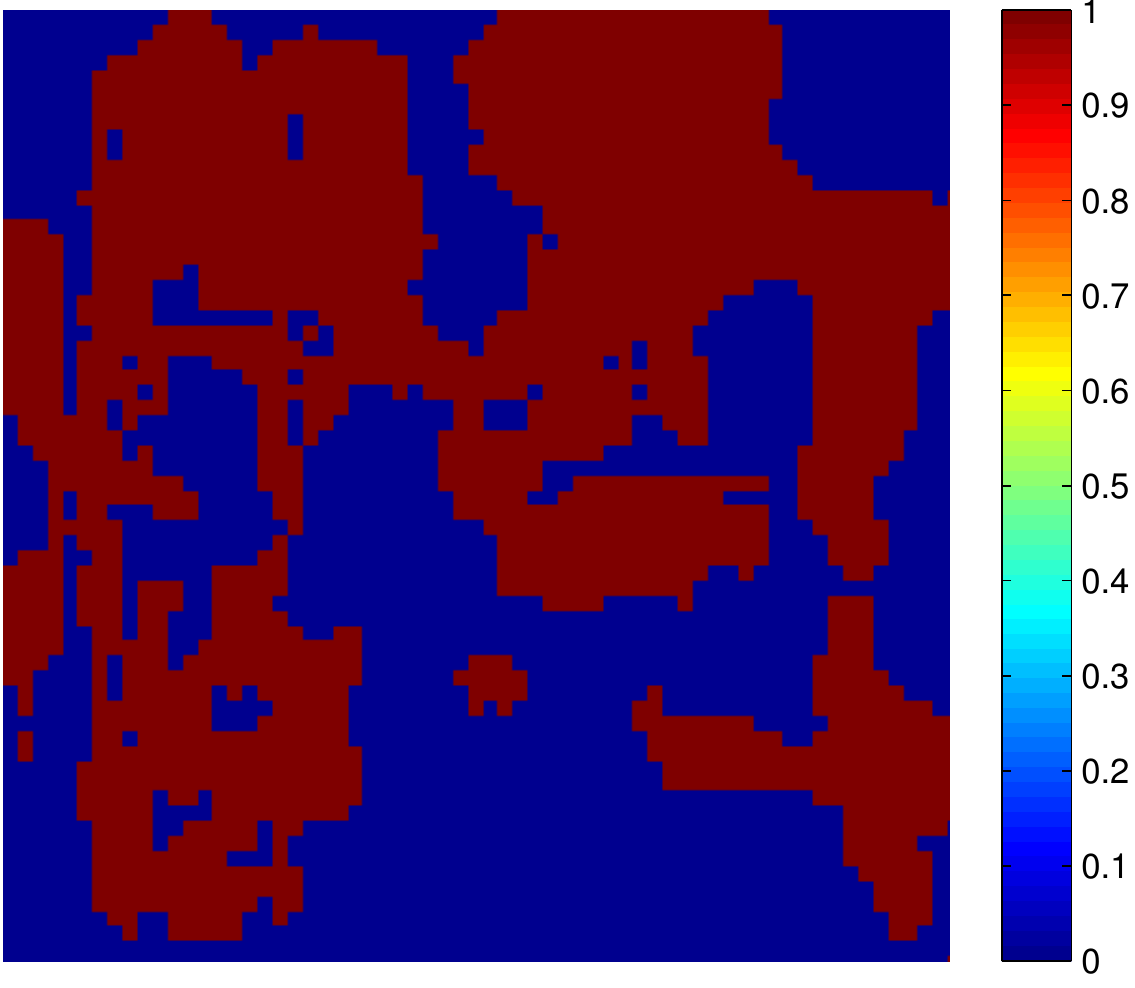} &
   \includegraphics[height=0.1\linewidth, width=0.14\linewidth]{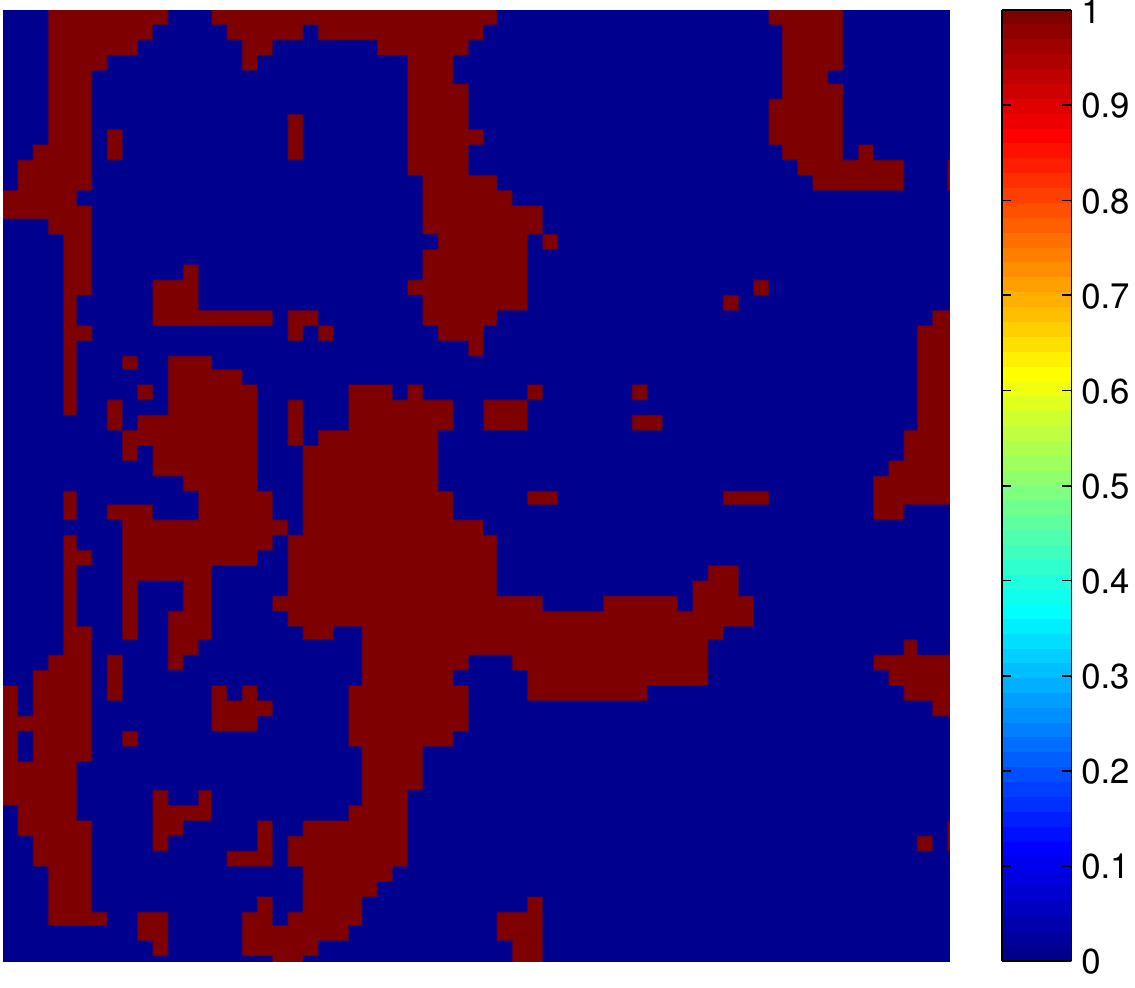} &
   \includegraphics[height=0.1\linewidth, width=0.14\linewidth]{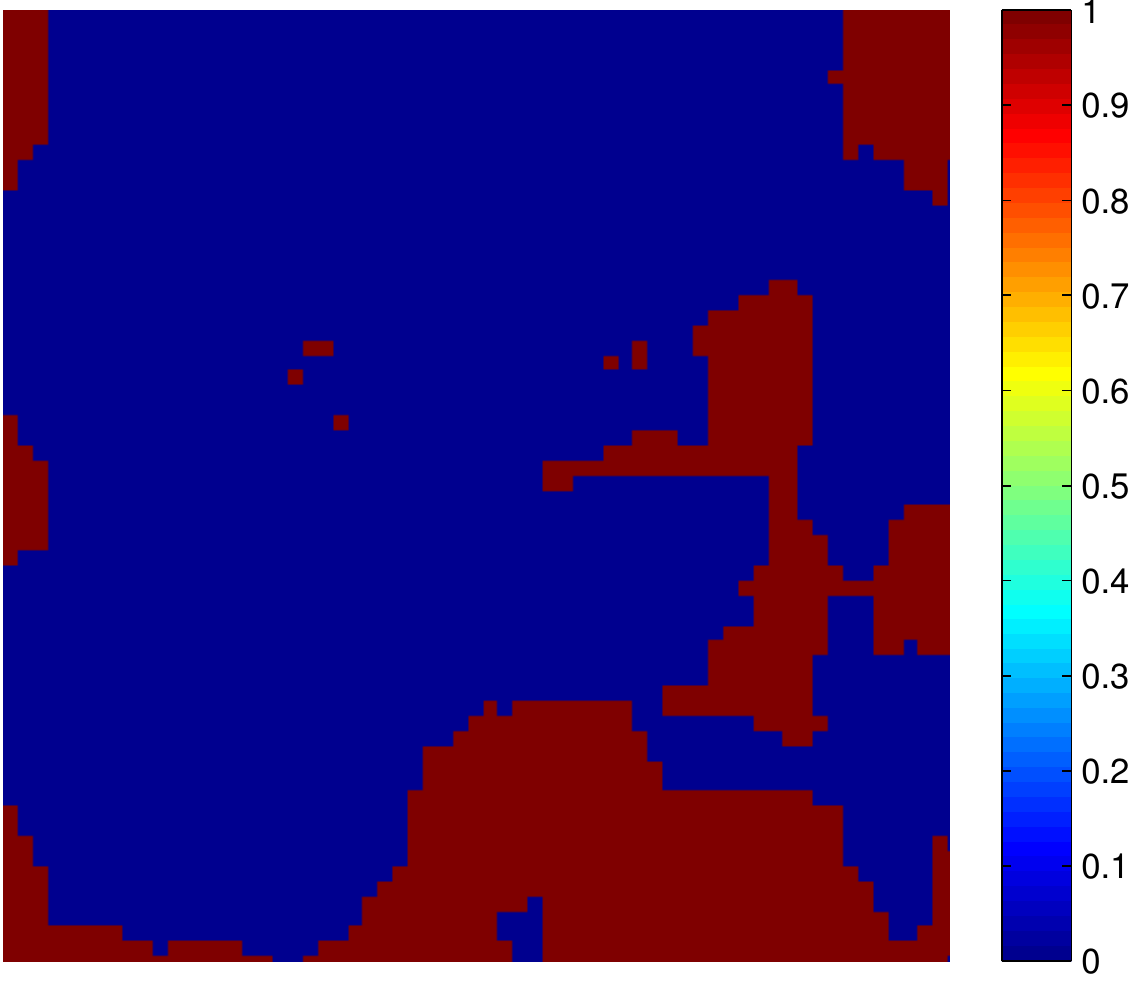} \\
   \includegraphics[height=0.1\linewidth, width=0.14\linewidth]{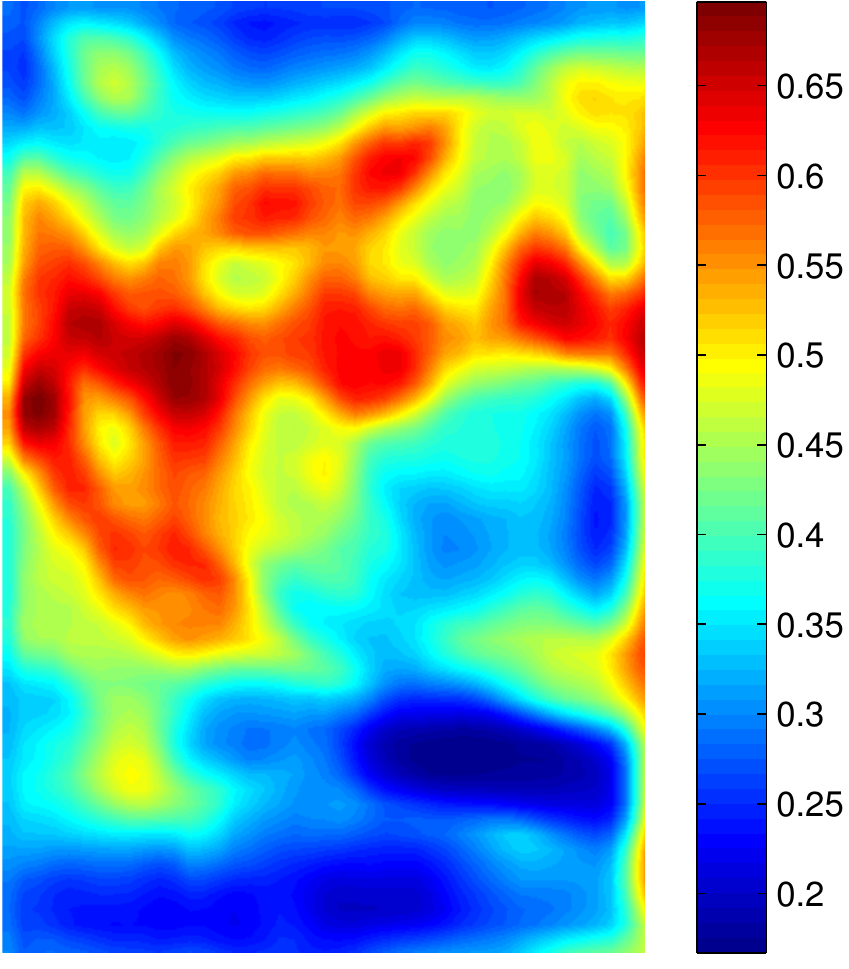} &
   \includegraphics[height=0.1\linewidth, width=0.14\linewidth]{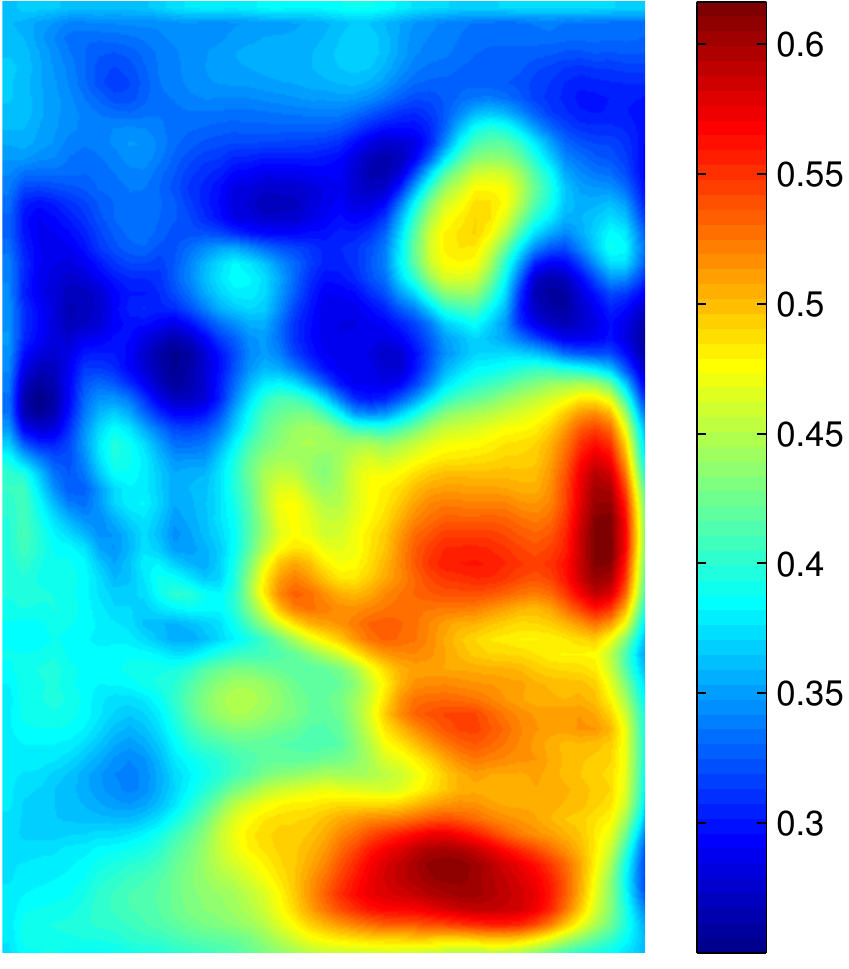} &
   \includegraphics[height=0.1\linewidth, width=0.14\linewidth]{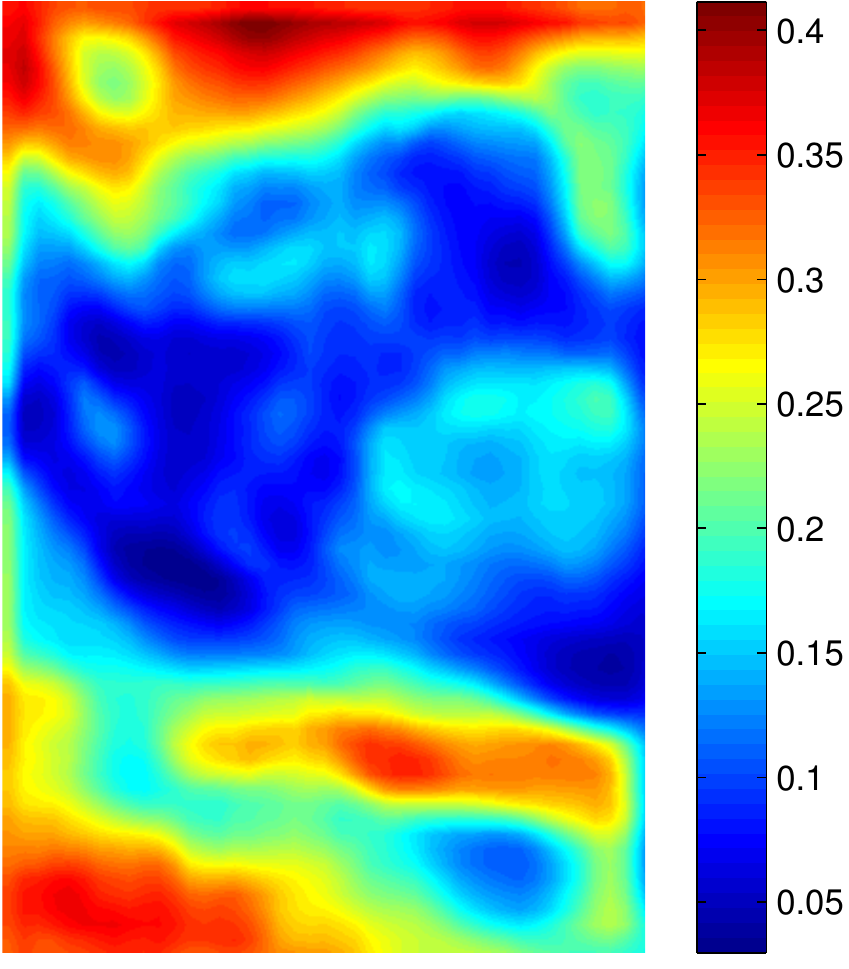} &
   \includegraphics[height=0.1\linewidth, width=0.14\linewidth]{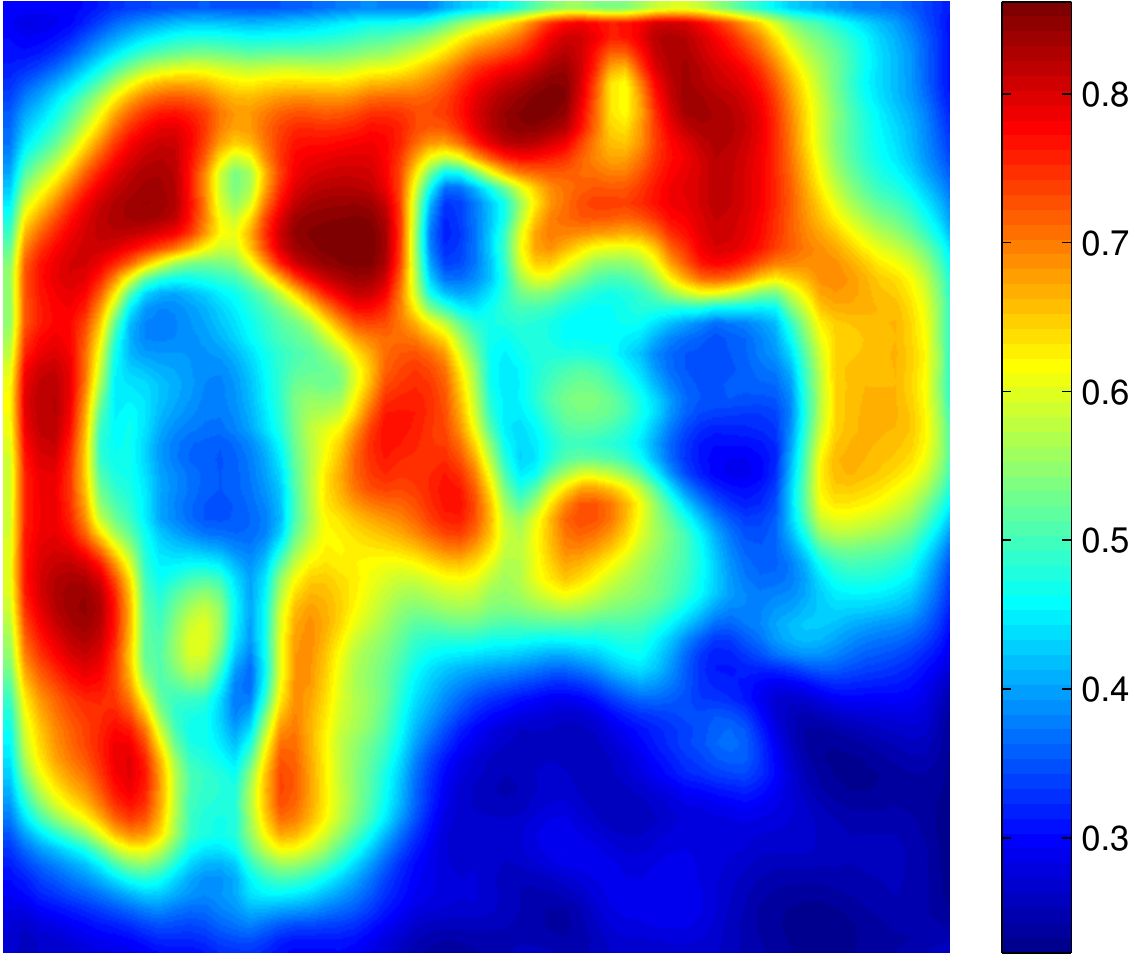} &
   \includegraphics[height=0.1\linewidth, width=0.14\linewidth]{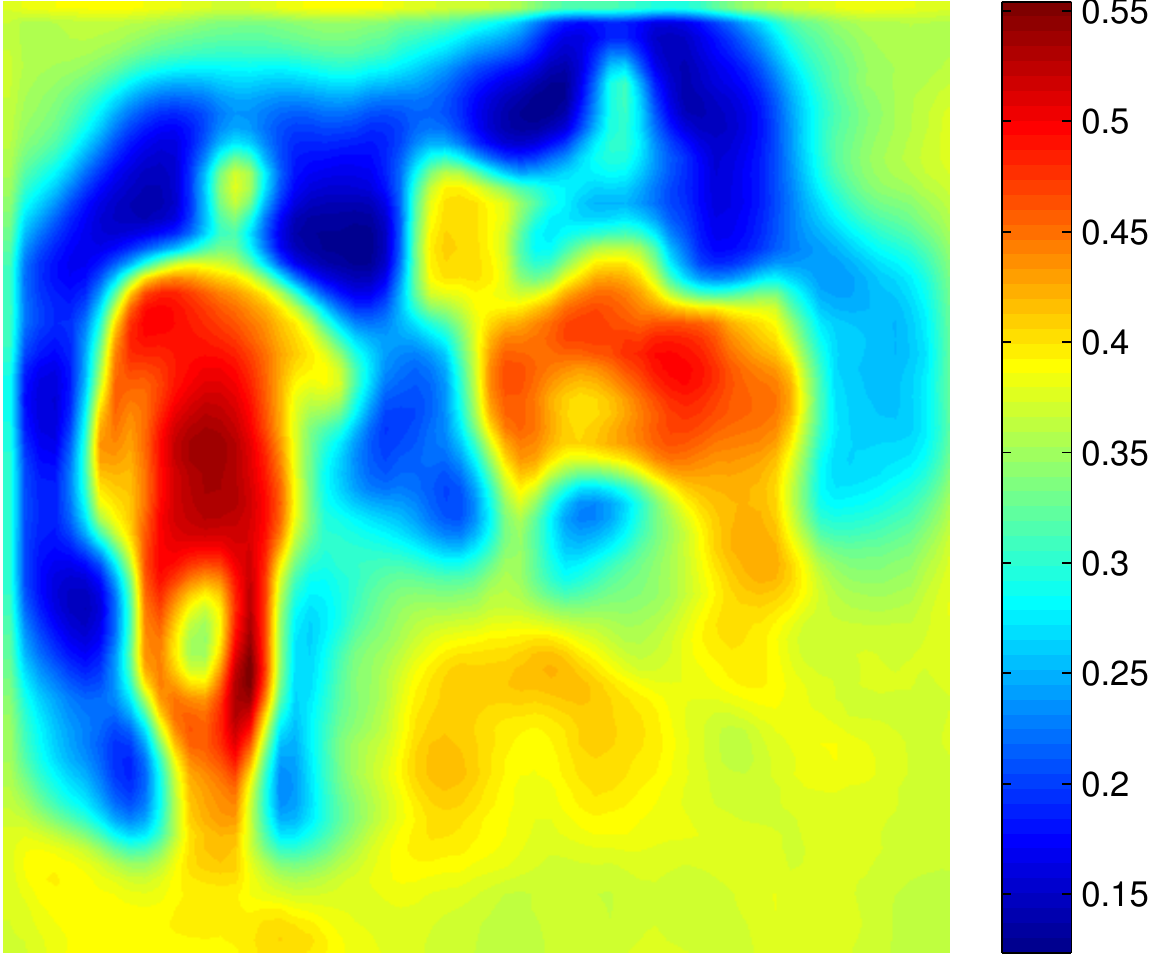} &
   \includegraphics[height=0.1\linewidth, width=0.14\linewidth]{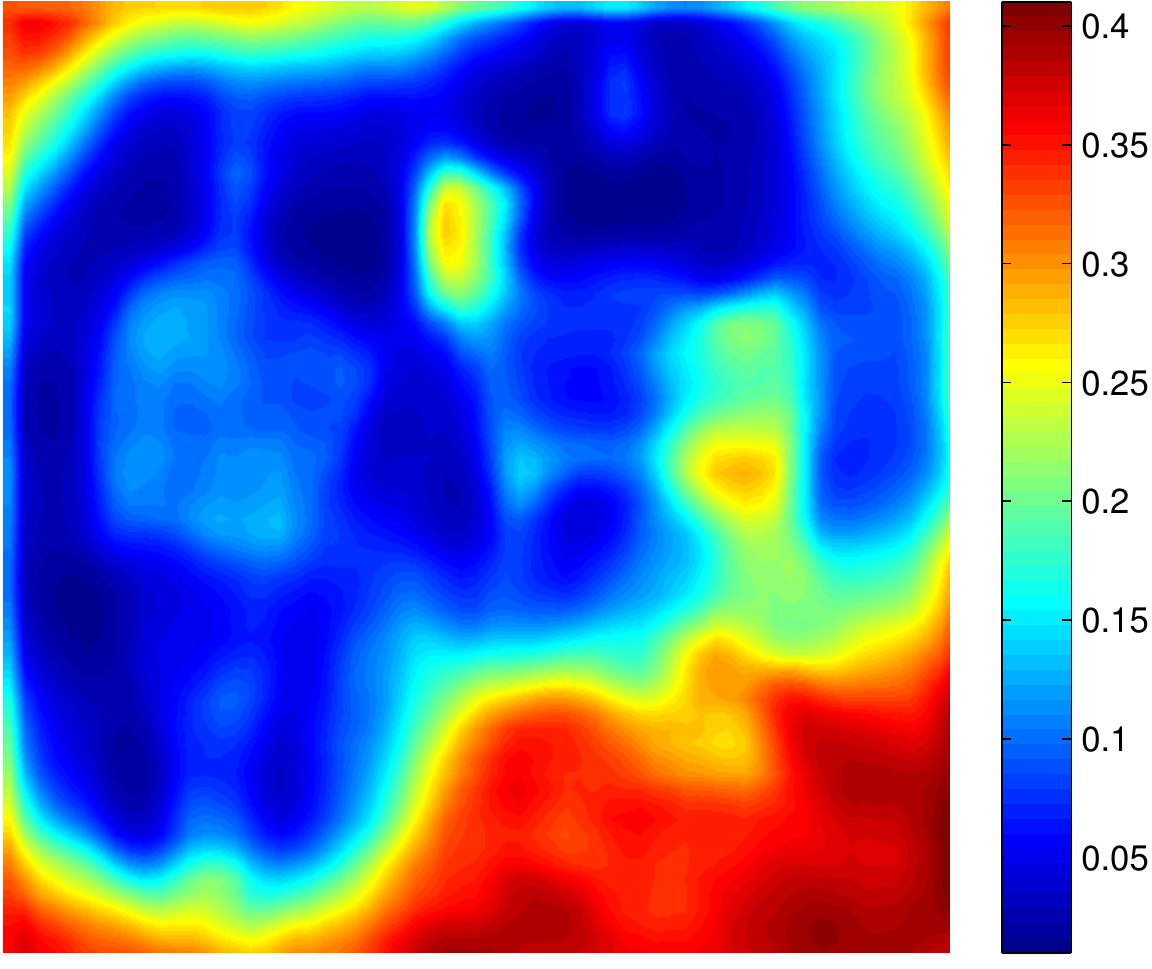} \\
   {\scriptsize (a) Scale-1 Attention} &
   {\scriptsize (b) Scale-0.75 Attention} &
   {\scriptsize (c) Scale-0.5 Attention} &
   {\scriptsize (a) Scale-1 Attention} &
   {\scriptsize (b) Scale-0.75 Attention} &
   {\scriptsize (c) Scale-0.5 Attention} \\
  \end{tabular}
  \caption{Weight maps produced by max-pooling (row 2) and by attention model (row 3). Notice that our attention model learns better interpretable weight maps for different scales. (a) Scale-1 attention (\ie, weight map for scale $s=1$) captures small-scale objects, (b) Scale-0.75 attention usually focuses on middle-scale objects, and (c) Scale-0.5 attention emphasizes on background contextual information.}
  \label{fig:max_vs_ours}
\end{figure*}

\textbf{Improvement over DeepLab:} We report the results in \tabref{tab:deeplab_part} when employing DeepLab-LargeFOV as the baseline. We find that using two input scales improves over using only one input scale, and it is also slightly better than using three input scales combined with average-pooling or attention model. We hypothesize that when merging three scale inputs, the features to be merged must be sufficiently discriminative or direct fusion degrades performance. On the other hand,
max-pooling seems robust to this effect. No matter how many scales are used, our attention model yields better results than average-pooling and max-pooling. We further visualize the weight maps produced by max-pooling and our attention model in \figref{fig:max_vs_ours}, which clearly shows that our attention model learns better interpretable weight maps for different scales. Moreover, we find that by introducing extra supervision to the FCNs for
each scale significantly improves the performance (see the column {\it w/ E-Supv}), regardless of what merging scheme is employed. The results show that adding extra supervision is essential for merging multi-scale features. Finally, we compare our proposed method with DeepLab-MSc-LargeFOV, which exploits the features from the intermediate layers for classification (MSc denotes Multi-Scale features). Note DeepLab-MSc-LargeFOV is a type of {\it skip-net}. Our best model ($56.39\%$) attains $2.67\%$ better performance than DeepLab-MSc-LargeFOV ($53.72\%$).

\begin{figure*}
  \centering
  \begin{tabular}{c c c c c c}
   \multicolumn{6}{c}{\includegraphics[width=0.5\linewidth]{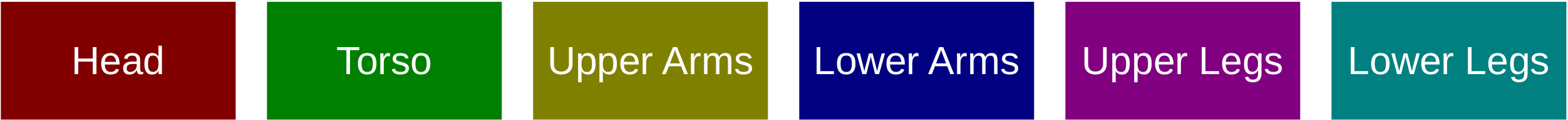}} \\
   \includegraphics[height=0.1\linewidth]{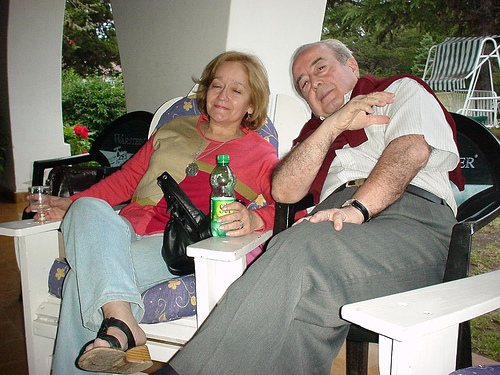} &
   \includegraphics[height=0.1\linewidth]{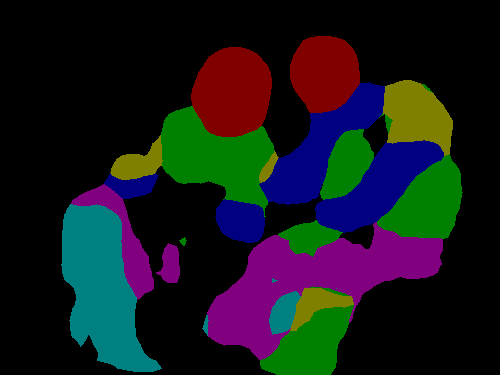} &
   \includegraphics[height=0.1\linewidth]{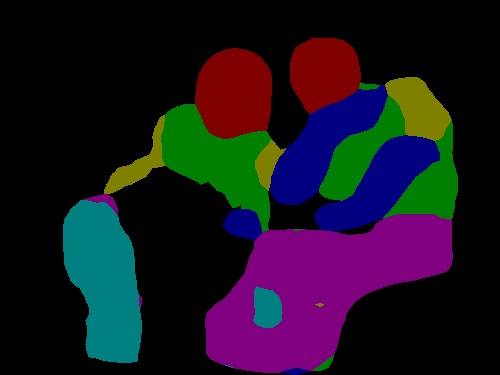} &
   \includegraphics[height=0.1\linewidth]{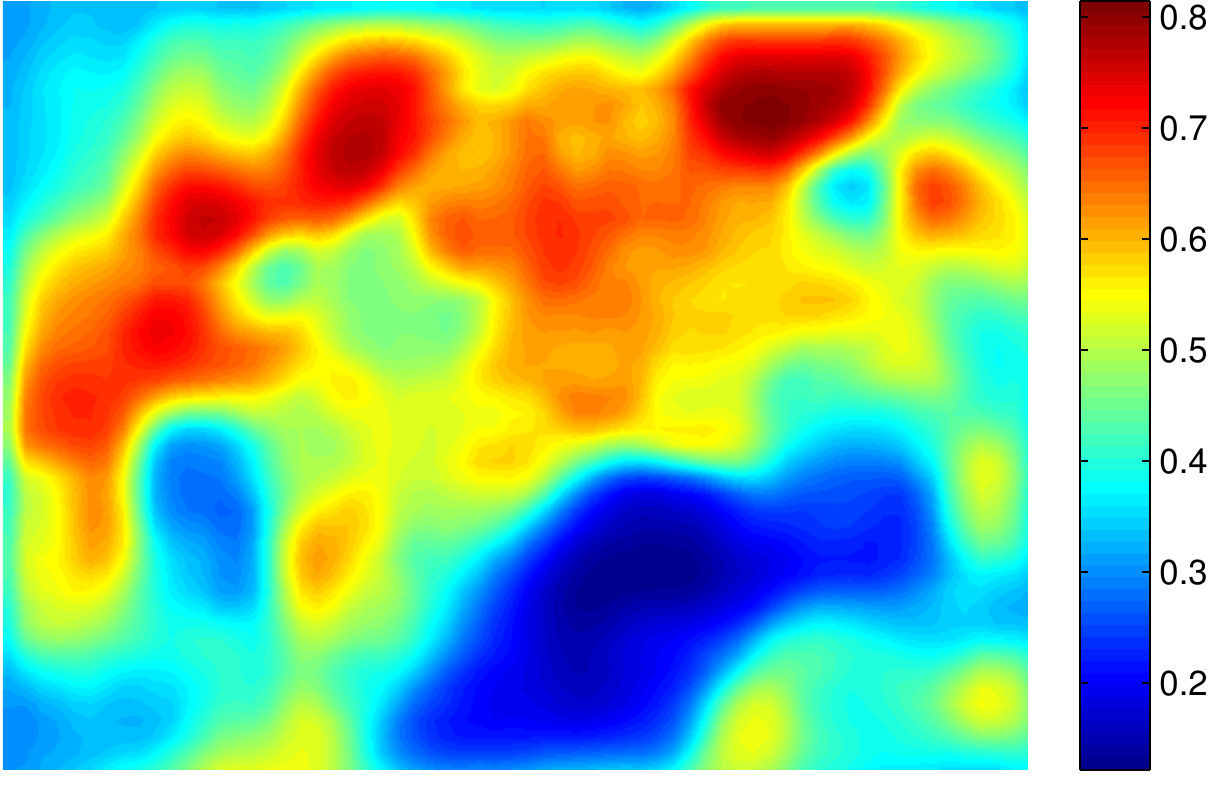} &
   \includegraphics[height=0.1\linewidth]{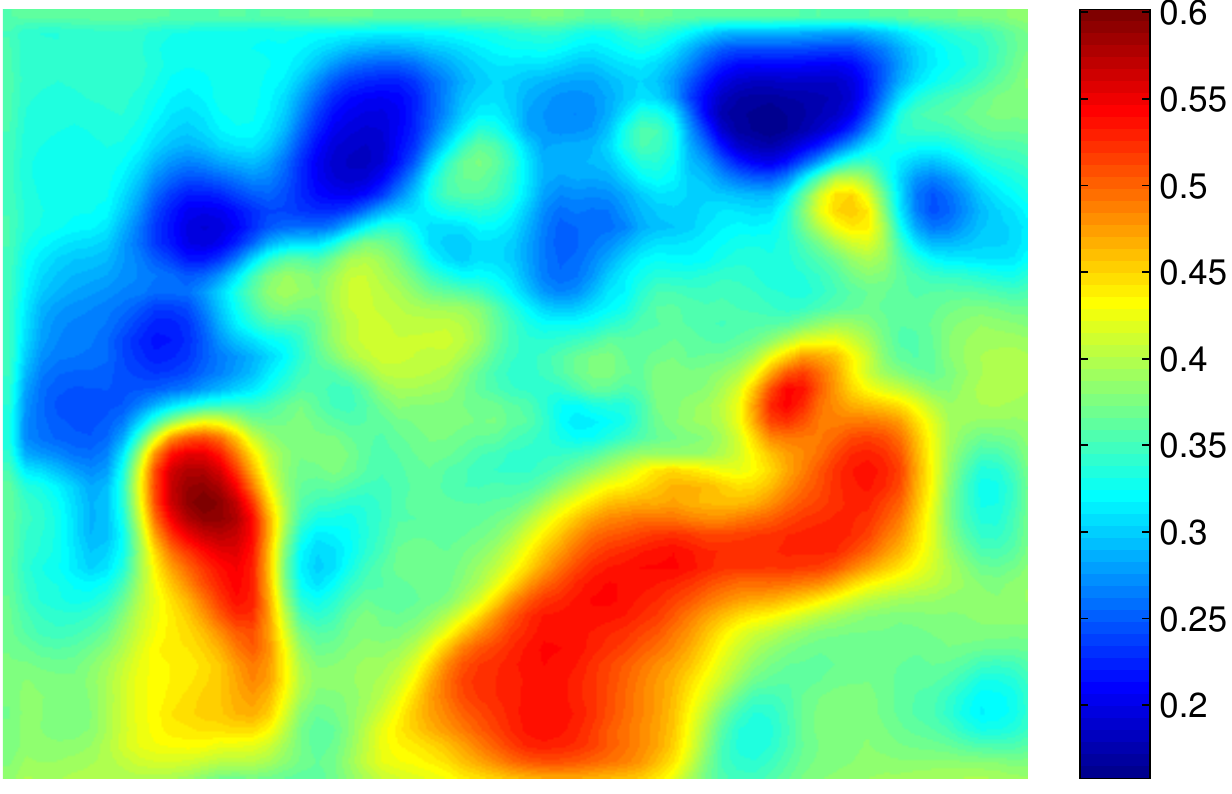} &
   \includegraphics[height=0.1\linewidth]{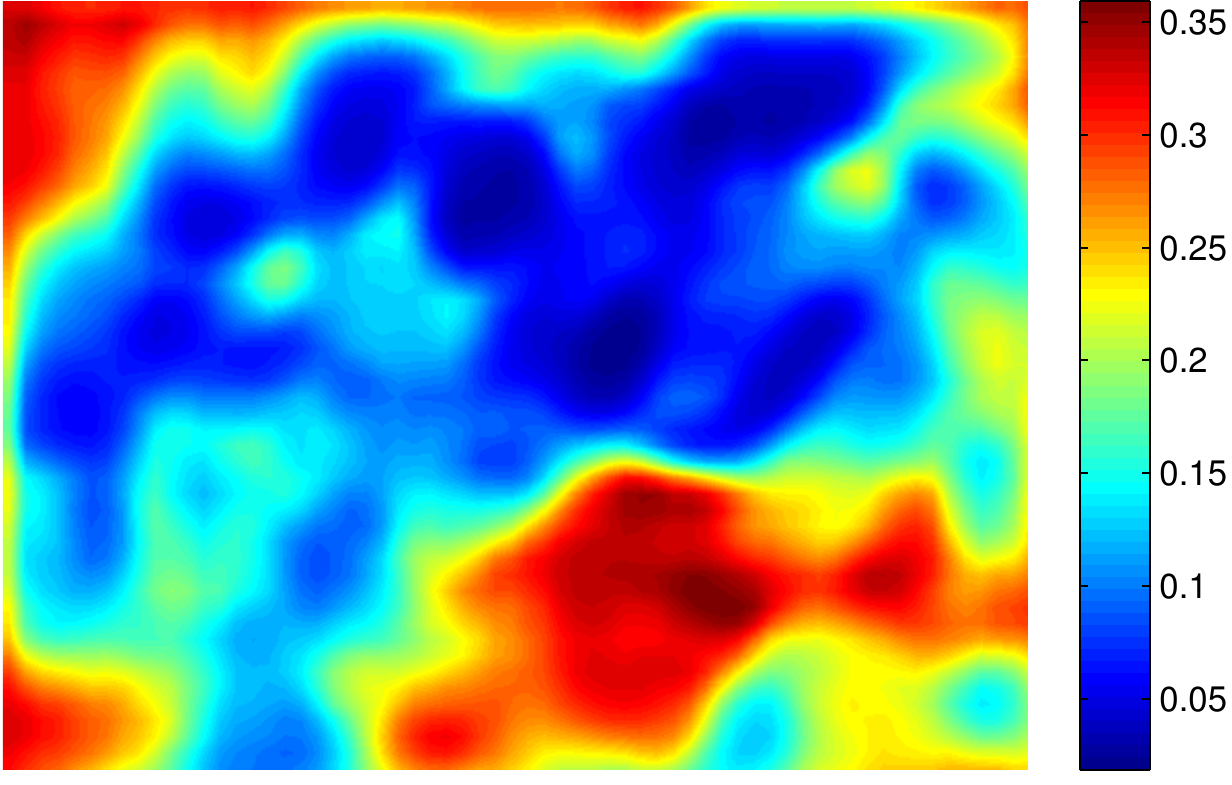} \\
   \includegraphics[height=0.1\linewidth]{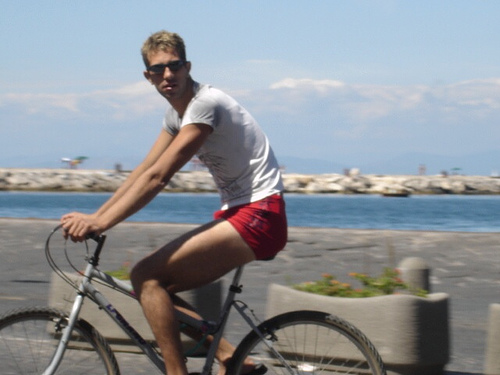} &
   \includegraphics[height=0.1\linewidth]{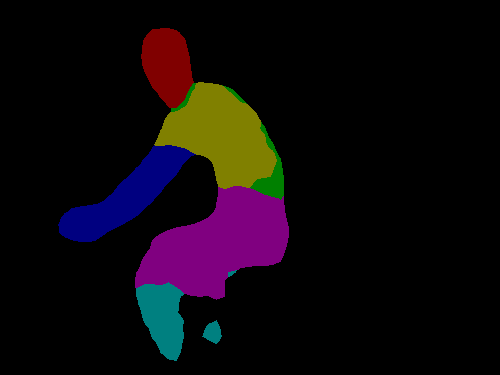} &
   \includegraphics[height=0.1\linewidth]{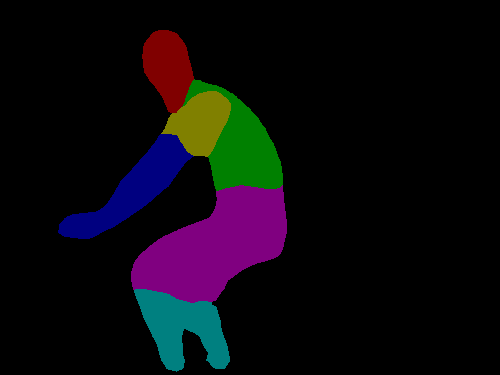} &
   \includegraphics[height=0.1\linewidth]{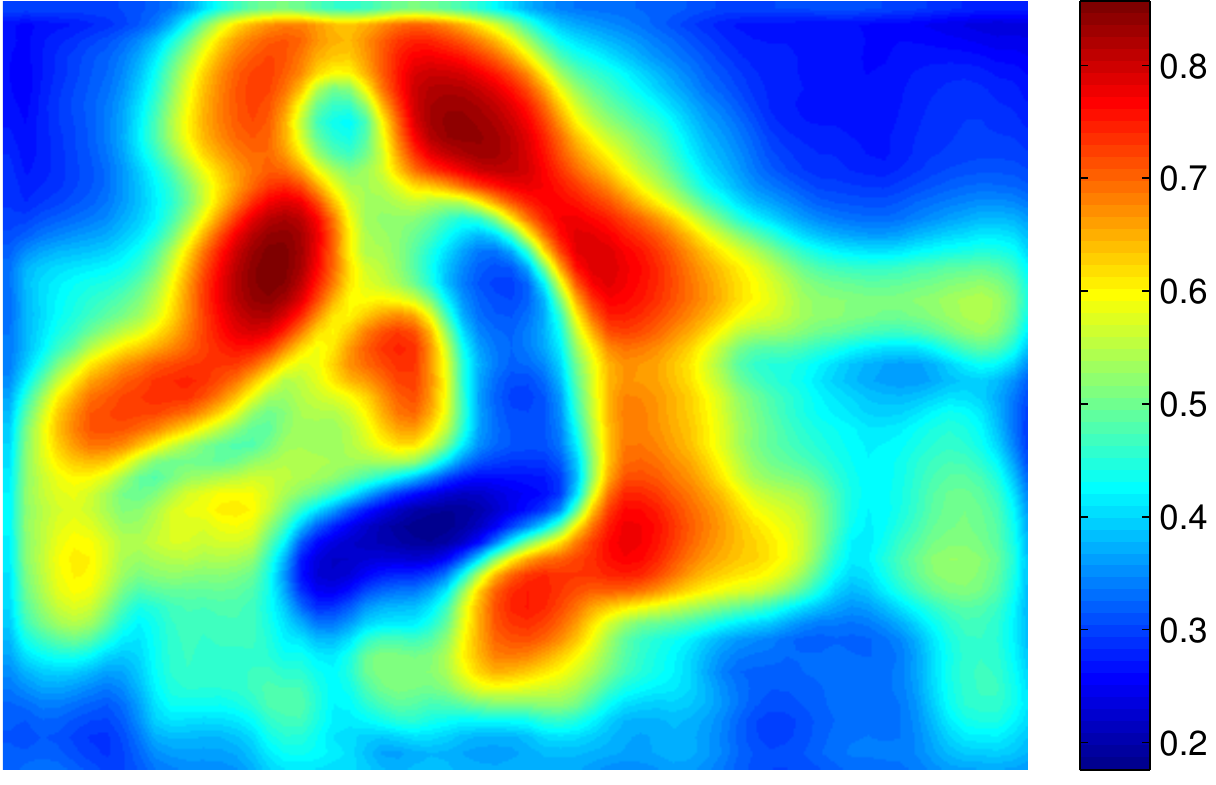} &
   \includegraphics[height=0.1\linewidth]{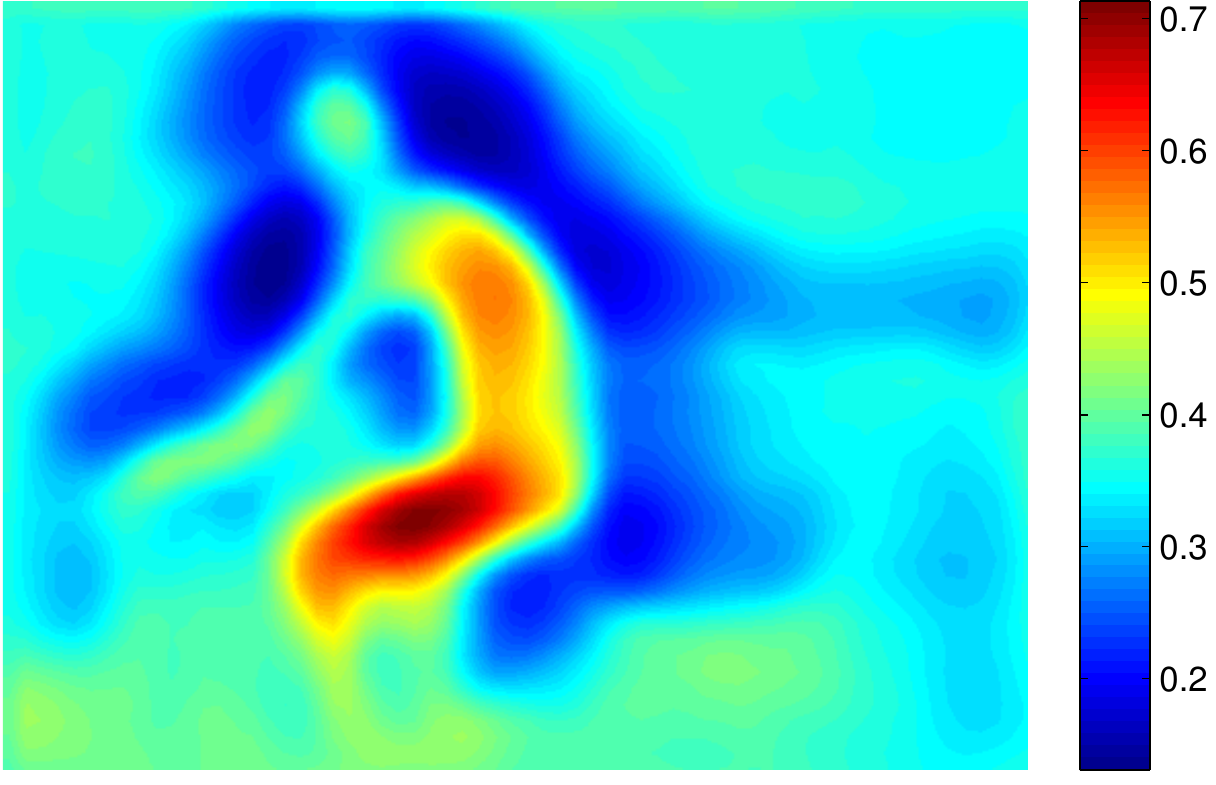} &
   \includegraphics[height=0.1\linewidth]{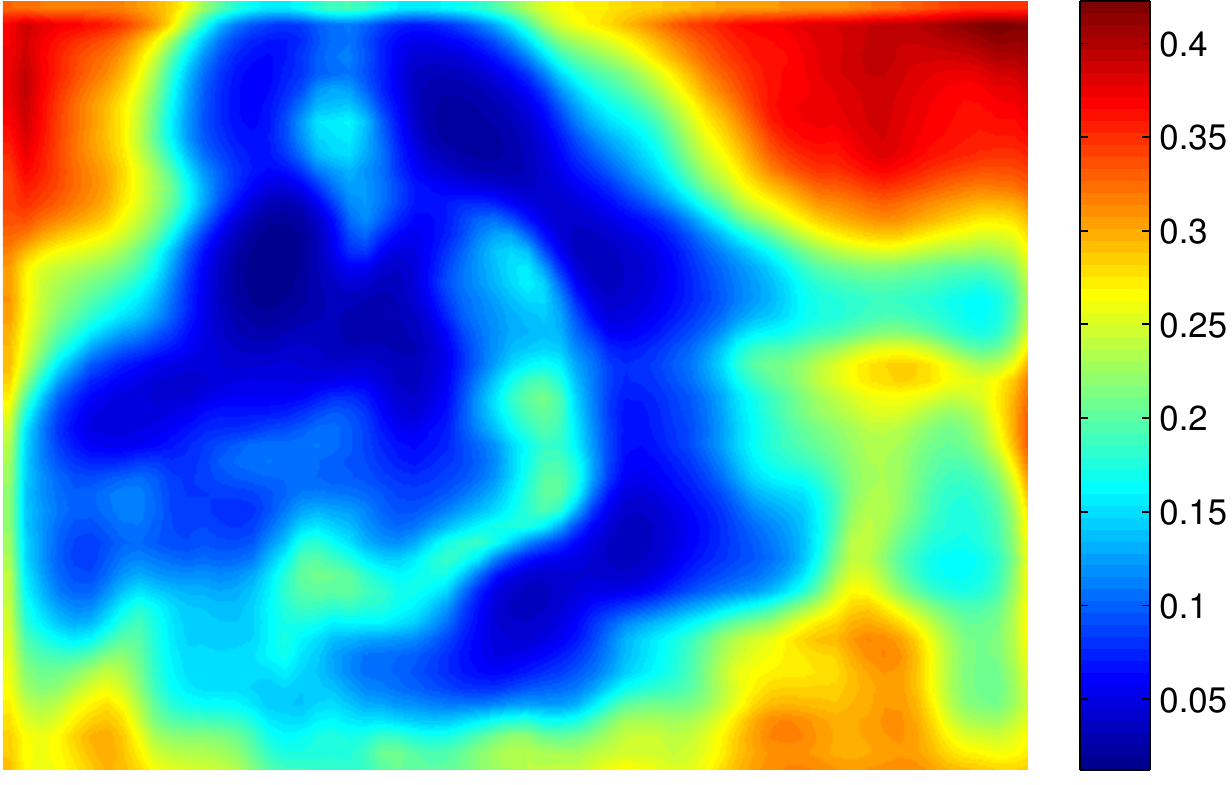} \\
   \includegraphics[height=0.1\linewidth]{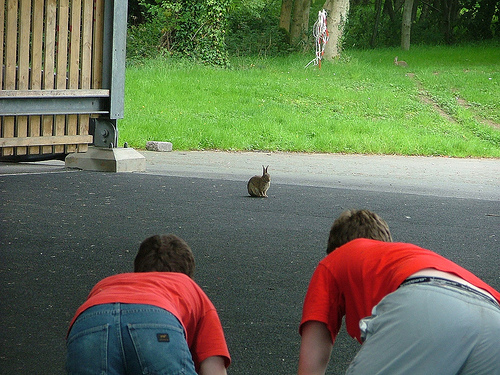} &
   \includegraphics[height=0.1\linewidth]{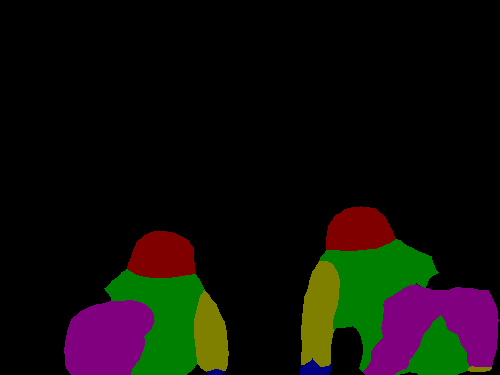} &
   \includegraphics[height=0.1\linewidth]{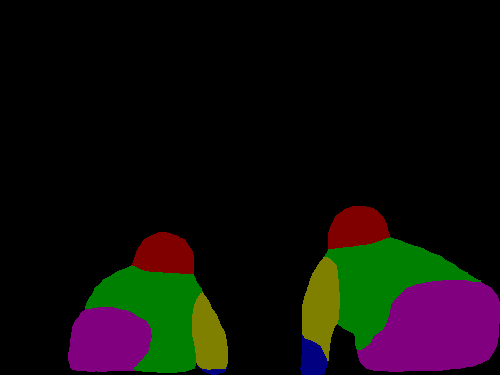} &
   \includegraphics[height=0.1\linewidth]{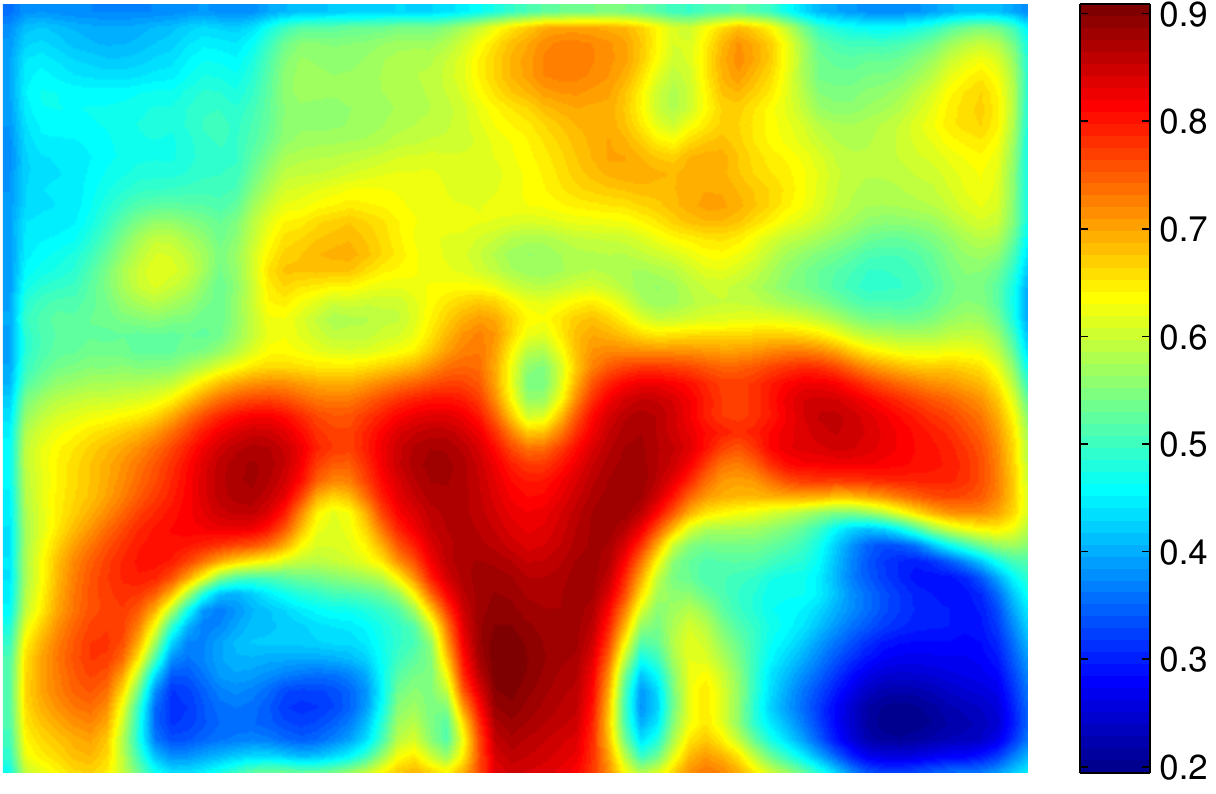} &
   \includegraphics[height=0.1\linewidth]{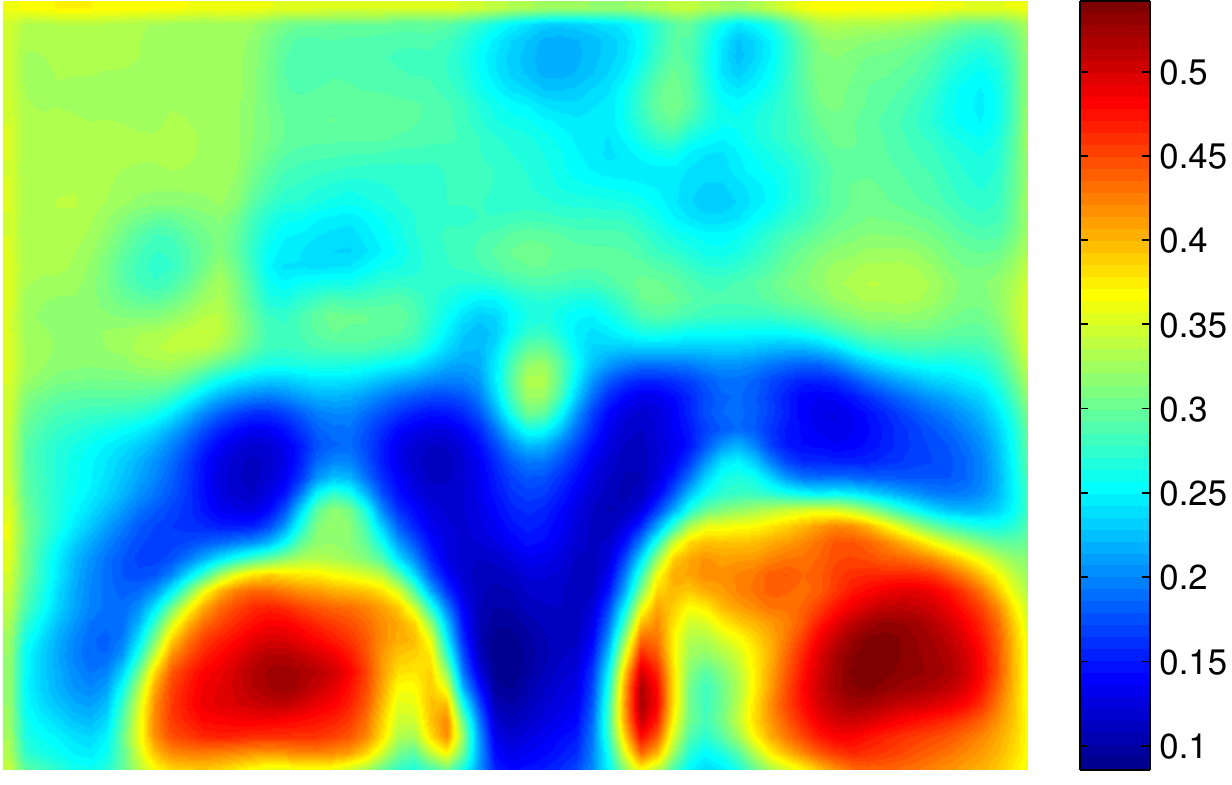} &
   \includegraphics[height=0.1\linewidth]{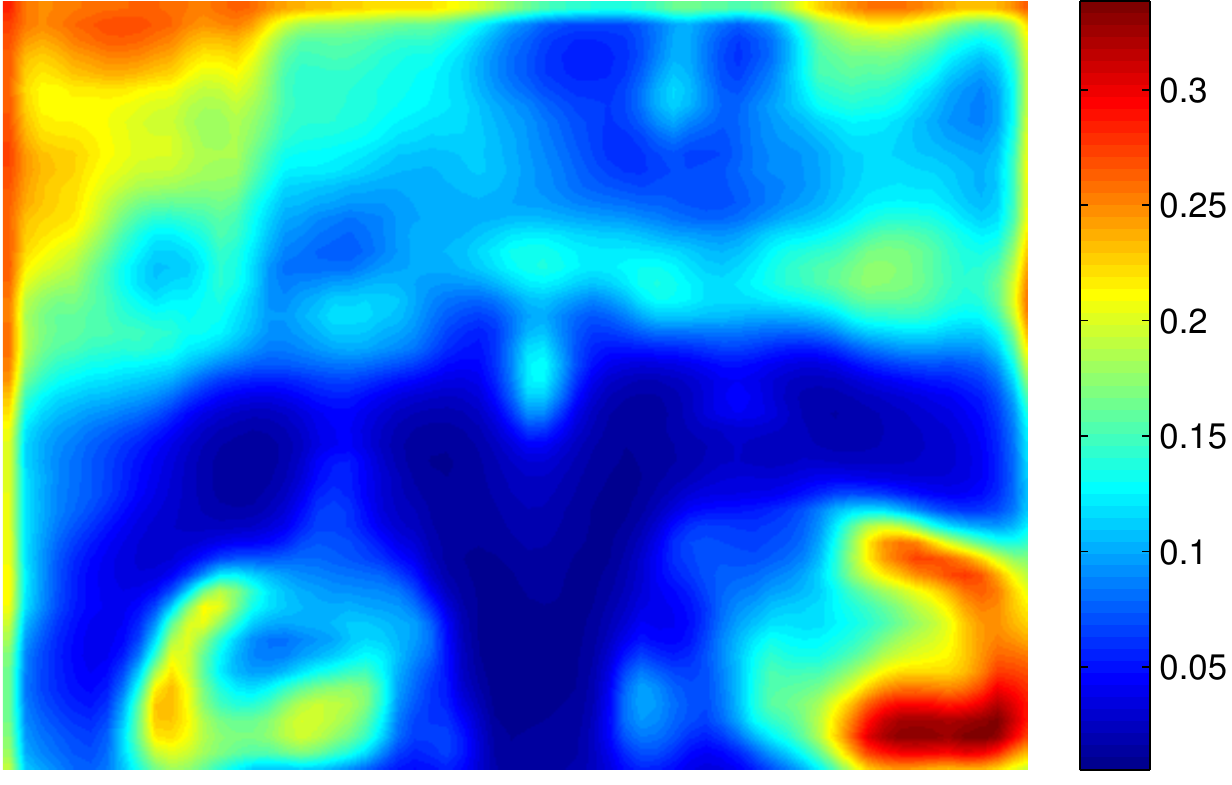} \\

   \hline
   \includegraphics[height=0.10\linewidth]{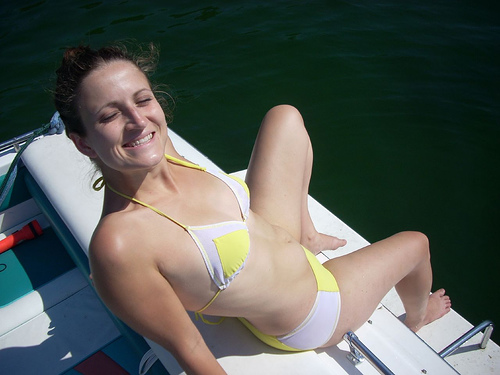} &
   \includegraphics[height=0.10\linewidth]{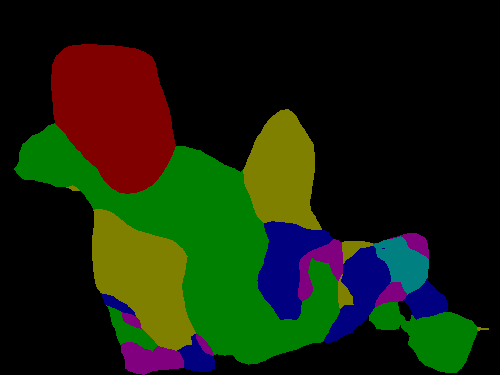} &
   \includegraphics[height=0.10\linewidth]{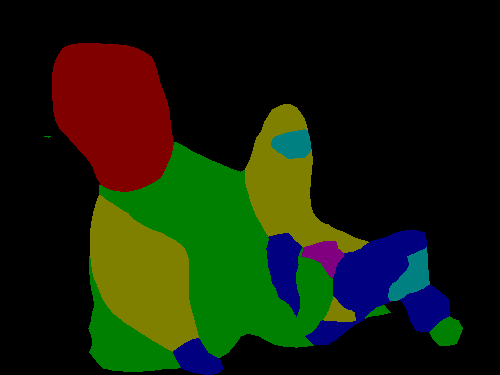} &
   \includegraphics[height=0.10\linewidth]{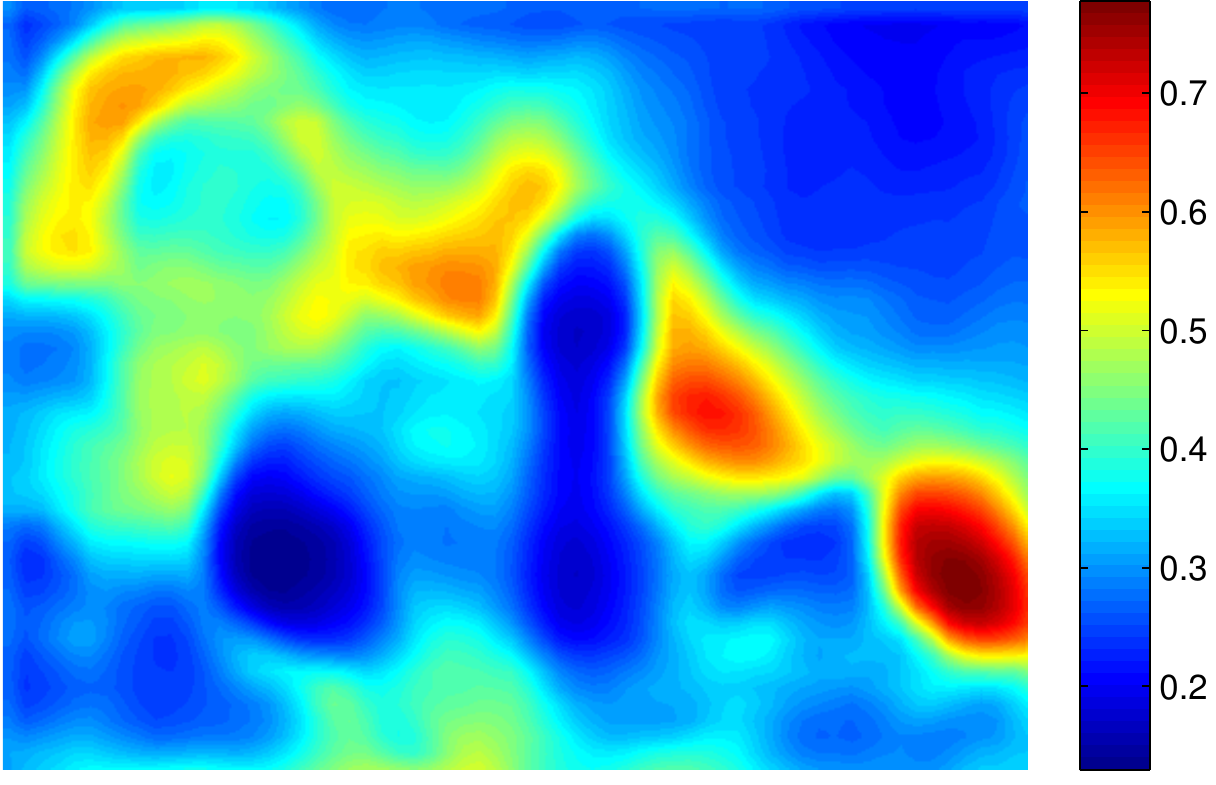} &
   \includegraphics[height=0.10\linewidth]{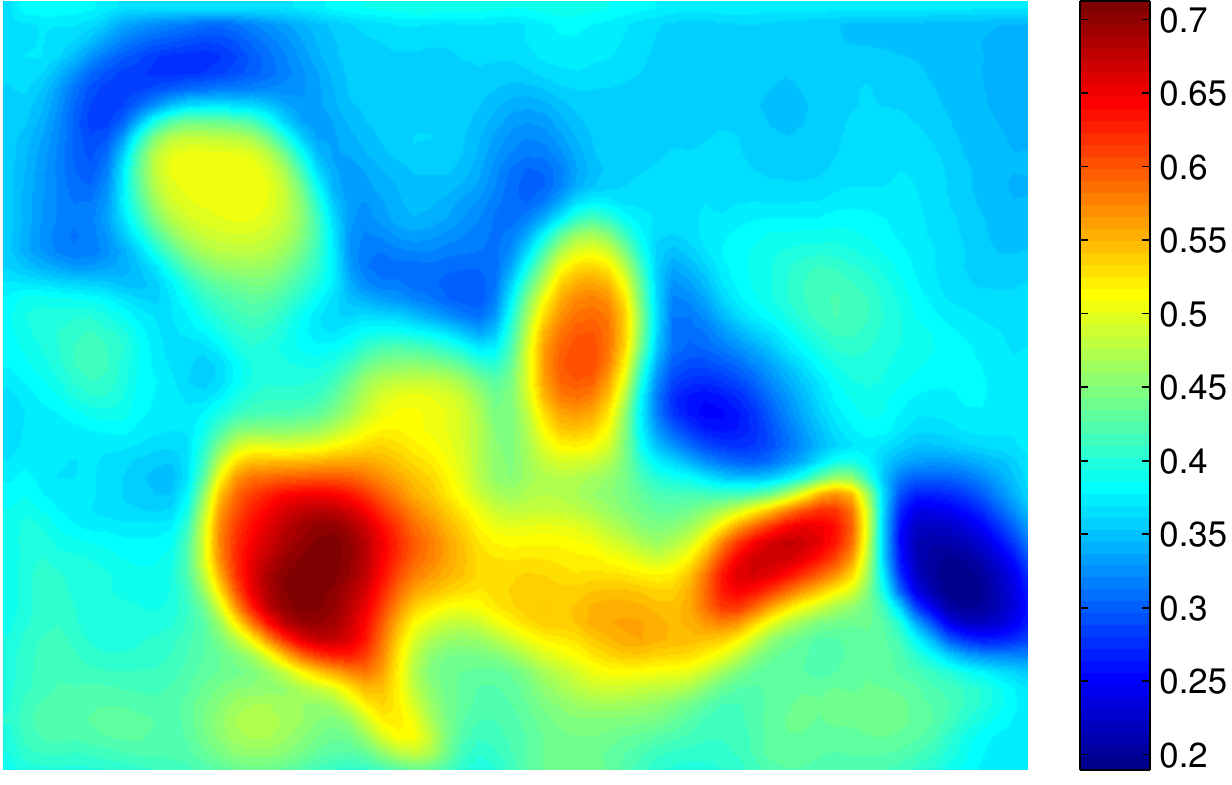} &
   \includegraphics[height=0.10\linewidth]{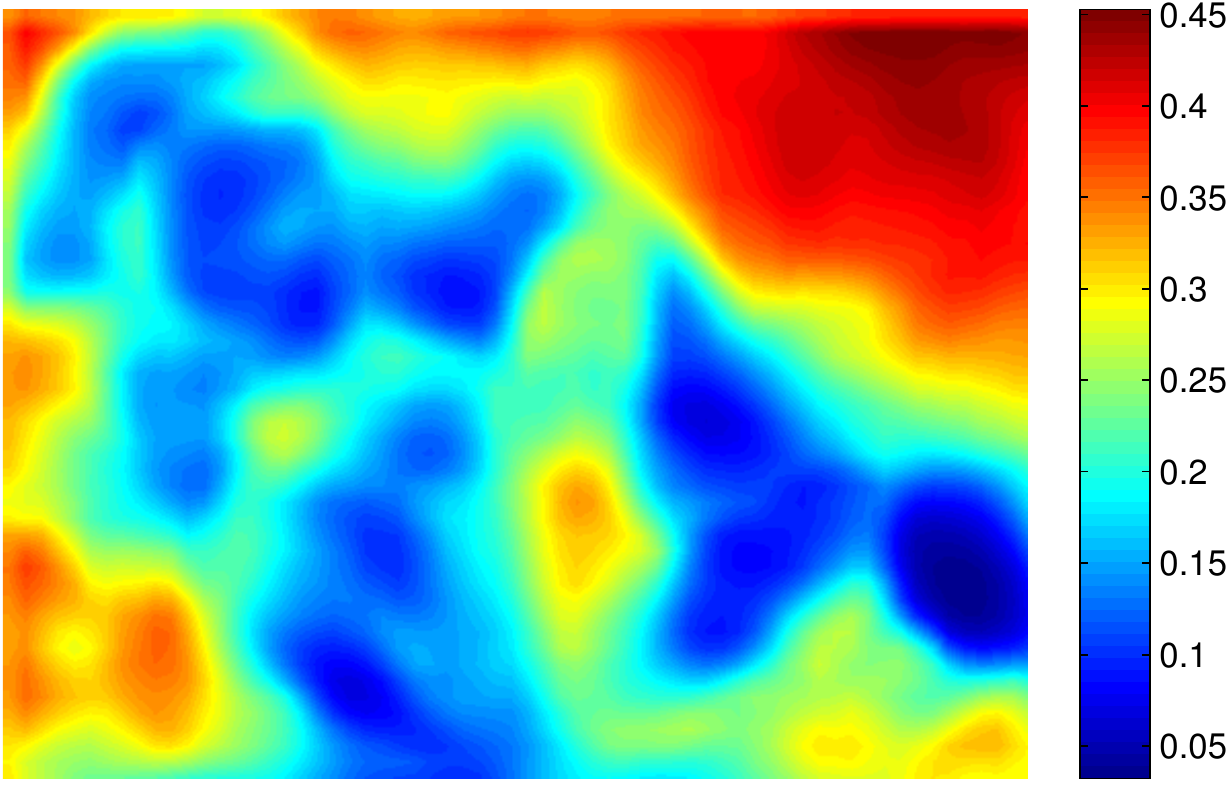} \\
   \includegraphics[height=0.10\linewidth]{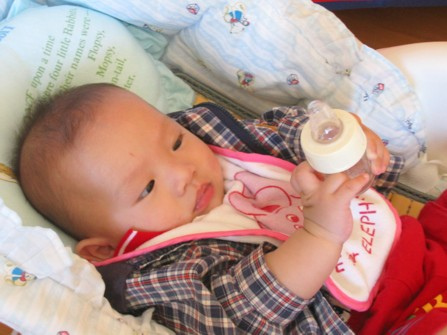} &
   \includegraphics[height=0.10\linewidth]{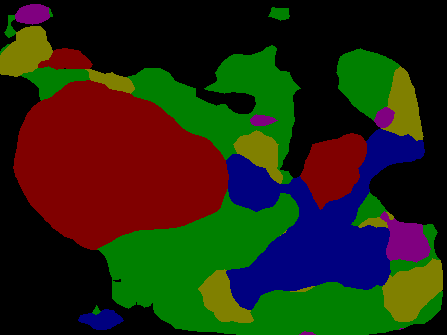} &
   \includegraphics[height=0.10\linewidth]{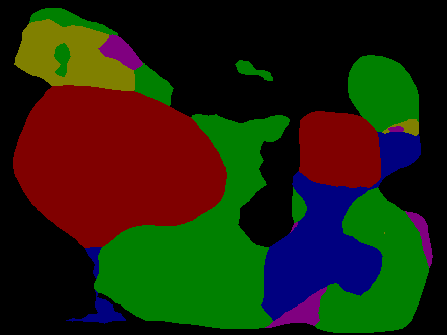} &
   \includegraphics[height=0.10\linewidth]{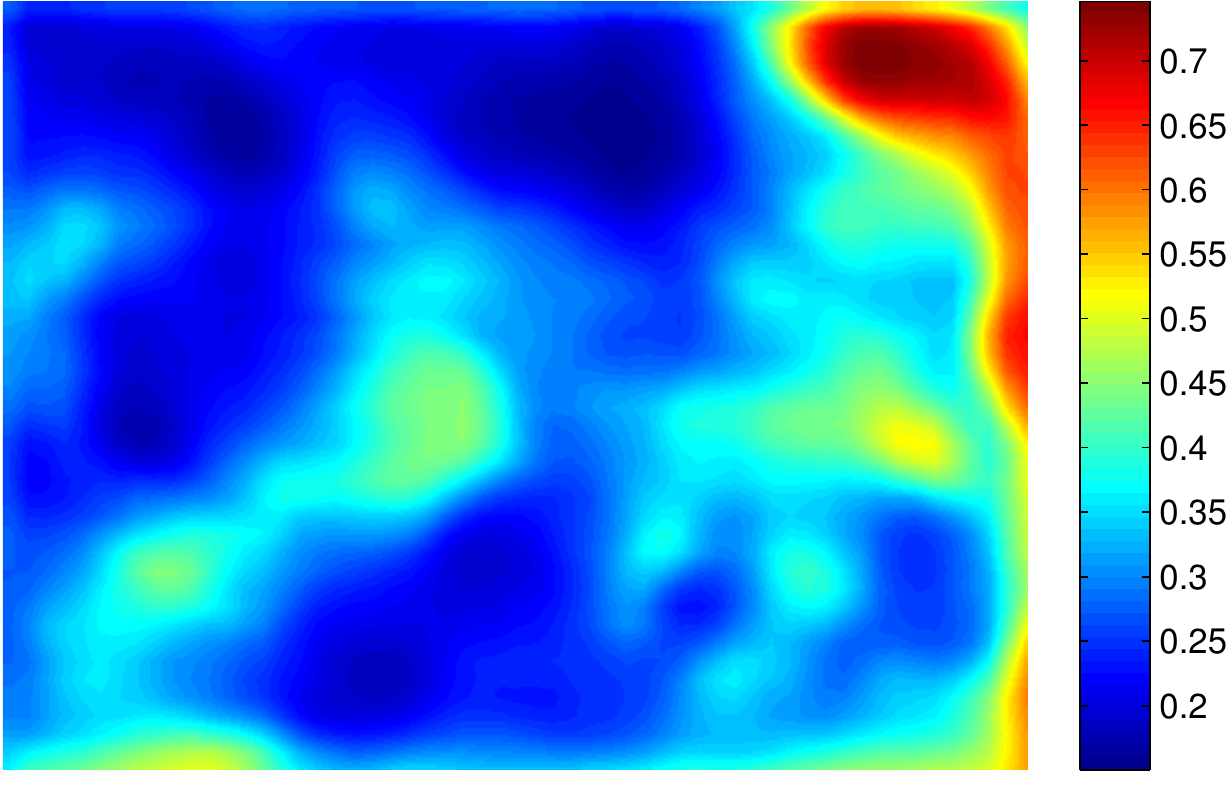} &
   \includegraphics[height=0.10\linewidth]{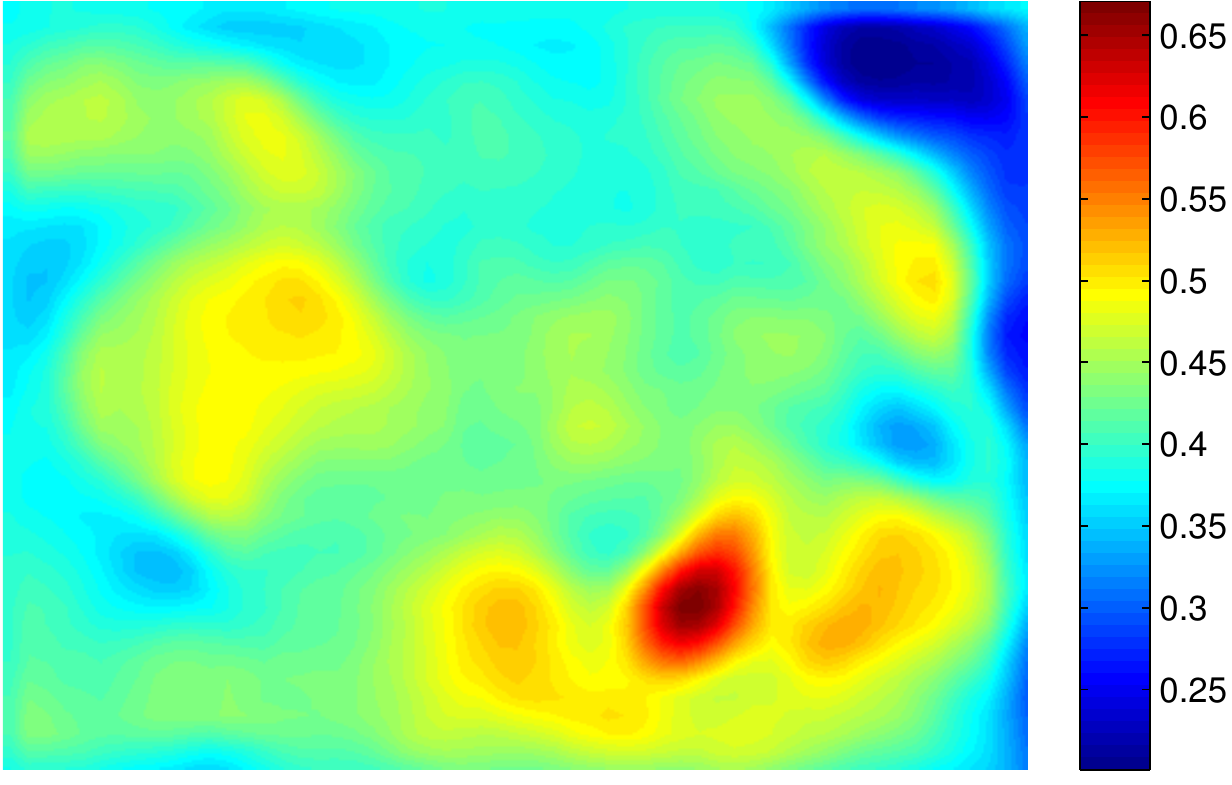} &
   \includegraphics[height=0.10\linewidth]{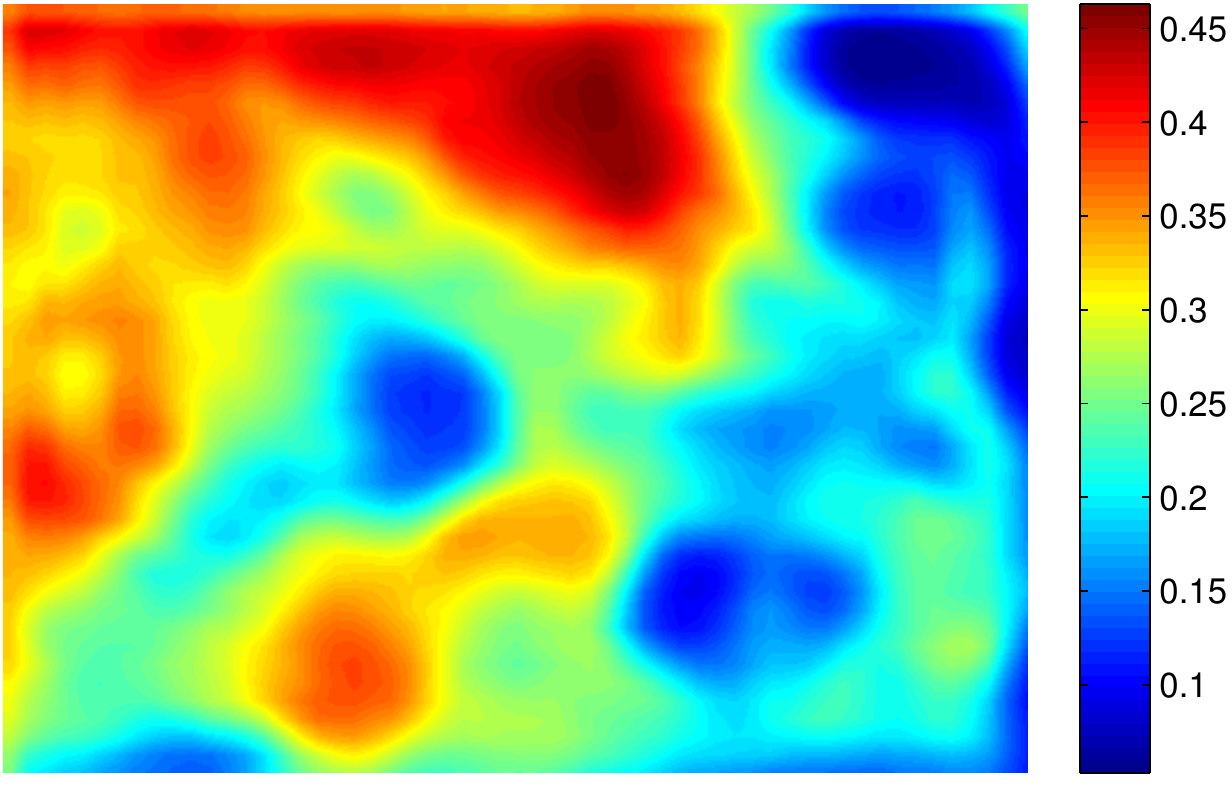} \\
   {\scriptsize (a) Image} & 
   {\scriptsize (b) Baseline} & 
   {\scriptsize (c) Our model} & 
   {\scriptsize (d) Scale-1 Attention} & 
   {\scriptsize (e) Scale-0.75 Attention} &
   {\scriptsize (f) Scale-0.5 Attention} \\
  \end{tabular}
  \caption{Results on PASCAL-Person-Part {\it validation} set. DeepLab-LargeFOV with one scale input is used as the baseline. Our model employs three scale inputs, attention model and extra supervision. Scale-1 attention captures small-scale parts, scale-0.75 attention catches middle-scale torsos and legs, while scale-0.5 attention focuses on large-scale legs and background. Bottom two rows show failure examples.}
  \label{fig:pascal_part_results}  
\end{figure*}

\textbf{Design choices:} For all the experiments reported in this work, our proposed attention model takes as input the convolutionalized $fc_7$ features \cite{simonyan2014very}, and employs a FCN consisting of two layers (the first layer has 512 filters with kernel size $\by{3}{3}$ and the second layer has $S$ filters with kernel size $\by{1}{1}$, where $S$ is the number of scales employed). We have experimented with different settings, including using only one layer for the attention model,
changing the kernel of the first layer to be $\by{1}{1}$, and varying the number of filters for the first layer. The performance does not vary too much; the degradation ranges from $0.1 \%$ to $0.4 \%$. Furthermore, we find that using $fc_8$ as features for the attention model results in worse performance (drops $\sim 0.5\%$) with similar results for $fc_6$ and $fc_7$. We also tried adding one more scale (four scales in total: $s \in \{1, 0.75, 0.5, 0.25\}$), however,
the performance drops by $0.5 \%$. We believe the score maps produced from scale $s=0.25$ were simply too small to be useful.

\textbf{Qualitative results:} We visualize the part segmentation results as well as the weight maps produced by the attention model in \figref{fig:pascal_part_results}. Merging the multi-scale features with the attention model yields not only better performance but also more interpretable weight maps. Specifically, scale-1 attention (\ie, the weight map learned by attention model for scale $s=1$) usually focuses on small-scale objects, scale-0.75 attention concentrates on middle-scale objects, and scale-0.5 attention usually puts large weight on large-scale objects or background, since it is easier to capture the largest scale objects or background contextual information when the image is shrunk to be half of the original resolution.

\textbf{Failure modes:} We show two failure examples in the bottom of \figref{fig:pascal_part_results}. The failure examples are due to the extremely difficult human poses or the confusion between cloth and person parts. The first problem may be resolved by acquiring more data, while the second one is challenging because person parts are usually covered by clothes.

\textbf{Supplementary materials:} In the supplementary materials, we apply our trained model to some videos from MPII Human Pose dataset \cite{andriluka14cvpr}. The model is not fine-tuned on the dataset, and the result is run frame-by-frame. As shown in the video, even for images from another dataset, our model is able to produce reasonably and visually good part segmentation results and it infers meaningful attention for different scales. Additionally, we provide more qualitative results
for all datasets in the supplementary materials.


\subsection{PASCAL VOC 2012}
\textbf{Dataset:} The PASCAL VOC 2012 segmentation benchmark \cite{everingham2014pascal} consists of 20 foreground object classes and one background class. Following the same experimental protocol \cite{chen2014semantic, dai2015boxsup, zheng2015conditional}, we augment the original training set from the annotations by \cite{hariharan2011semantic}. We report the results on the original PASCAL VOC 2012 validation set and test set.

\begin{table}
  \centering
  \rowcolors{2}{}{yelloworange!25}
  \addtolength{\tabcolsep}{2.5pt}
    \begin{tabular}{l c c}
      \toprule[0.2 em]
      \multicolumn{2}{l}{Baseline: DeepLab-LargeFOV} & 62.28  \\
      \toprule[0.2 em]
      {\bf Merging Method} & & w/ E-Supv \\
      \midrule \midrule
      {\it Scales = \{1, 0.5\}} & & \\
      Max-Pooling & 64.81 & 67.43 \\
      Average-Pooling & 64.86 & 67.79 \\
      Attention & 65.27 & 68.24 \\
      \midrule
      {\it Scales = \{1, 0.75, 0.5\}} & & \\
      Max-Pooling & 65.15 & 67.79 \\
      Average-Pooling & 63.92 & 67.98 \\
      Attention & 64.37 & {\bf 69.08} \\
      \bottomrule[0.1 em]
    \end{tabular}
    \caption{Results on PASCAL VOC 2012 {\it validation} set, pretrained with ImageNet. E-Supv: extra supervision.}
    \label{tab:deeplab_voc12_imagenet}
\end{table}

\textbf{Pretrained with ImageNet:} First, we experiment with the scenario where the underlying DeepLab-LargeFOV is only pretrained on ImageNet \cite{ILSVRC15}. Our reproduction of DeepLab-LargeFOV and DeepLab-MSc-LargeFOV yields performance of $62.28\%$ and $64.39\%$ on the validation set, respectively. They are similar to those ($62.25\%$ and $64.21\%$) reported in \cite{chen2014semantic}. We report results of the proposed methods on the validation set in \tabref{tab:deeplab_voc12_imagenet}. We observe similar experimental results as PASCAL-Person-Part dataset: (1) Using two input scales is better than single input scale. (2) Adding extra supervision is necessary to achieve better performance for merging three input scales, especially for average-pooling and the proposed attention model. (3) The best performance ($6.8\%$ improvement over the DeepLab-LargeFOV baseline) is obtained with three input scales, attention model, and extra supervision, and its performance is $4.69\%$ better than DeepLab-MSc-LargeFOV ($64.39\%$).

\begin{table}
  \centering
  \rowcolors{2}{}{yelloworange!25}
  \addtolength{\tabcolsep}{2.5pt}
  \begin{tabular} {l | c}
    \toprule[0.2 em]
    {\bf Method} & mIOU \\
    \midrule \midrule
    {\it Pretrained with ImageNet} & \\
    DeepLab-LargeFOV \cite{chen2014semantic} & 65.1 \\
    DeepLab-MSc-LargeFOV \cite{chen2014semantic} & 67.0 \\
    TTI\_zoomout\_v2 \cite{mostajabi2014feedforward} & 69.6 \\
    ParseNet \cite{liu2015parsenet} & 69.8 \\
    \midrule
    DeepLab-LargeFOV-{\bf AveragePooling} & 70.5 \\
    DeepLab-LargeFOV-{\bf MaxPooling} & 70.6 \\
    DeepLab-LargeFOV-{\bf Attention} & 71.5 \\
    \midrule \midrule
    {\it Pretrained with MS-COCO} & \\
    DeepLab-CRF-COCO-LargeFOV \cite{papandreou2015weakly} & 72.7 \\
    DeepLab-MSc-CRF-COCO-LargeFOV \cite{papandreou2015weakly} & 73.6 \\
    CRF-RNN \cite{zheng2015conditional} & 74.7 \\
    BoxSup \cite{dai2015boxsup} & 75.2 \\
    DPN \cite{liu2015semantic} & 77.5 \\
    Adelaide \cite{lin2015efficient} & 77.8 \\
    \midrule
    DeepLab-CRF-COCO-LargeFOV-{\bf Attention} & 75.1 \\
    DeepLab-CRF-COCO-LargeFOV-{\bf Attention}+ & 75.7 \\
    DeepLab-CRF-Attention-DT \cite{chen2015semantic} & 76.3 \\
    \bottomrule[0.1 em]
  \end{tabular}
  \caption{Labeling IOU on the PASCAL VOC 2012 test set.}
  \label{tab:voc2012}
\end{table}

We also report results on the test set for our best model in \tabref{tab:voc2012}. First, we observe that the attention model yields a $1\%$ improvement over average pooling, consistent with our results on the validation set. We then compare our models with DeepLab-LargeFOV and DeepLab-MSc-LargeFOV \cite{chen2014semantic}. We find that our proposed model improves $6.4\%$ over DeepLab-LargeFOV, and gives a $4.5\%$ boost over DeepLab-MSc\_LargeFOV. Finally, we compare our models with two other methods: ParseNet \cite{liu2015parsenet} and TTI\_zoomout\_v2 \cite{mostajabi2014feedforward}. ParseNet incorporates the image-level feature as global contextual information. We consider ParseNet as a special case to exploit multi-scale features, where the whole image is summarized by the image-level feature. TTI\_zoomout\_v2 also exploits features at different spatial scales. As shown in the table, our proposed model outperforms both of them. Note none of the methods discussed here employ a fully connected CRF \cite{KrahenbuhlK11}.

\begin{table}
  \centering
  \rowcolors{2}{}{yelloworange!25}
  \addtolength{\tabcolsep}{2.5pt}
    \begin{tabular}{l c c}
      \toprule[0.2 em]
      \multicolumn{2}{l}{Baseline: DeepLab-LargeFOV} & 67.58  \\
      \toprule[0.2 em]
      {\bf Merging Method} &  & w/ E-Supv \\
      \midrule
      \midrule
      {\it Scales = \{1, 0.5\}} & & \\
      Max-Pooling & 69.15 & 70.01 \\
      Average-Pooling & 69.22 & 70.44 \\
      Attention & 69.90 & 70.76 \\
      \midrule
      {\it Scales = \{1, 0.75, 0.5\}} & & \\
      Max-Pooling & 69.70 & 70.06 \\
      Average-Pooling & 68.82 & 70.55 \\
      Attention & 69.47 & {\bf 71.42} \\
      \bottomrule[0.1 em]
    \end{tabular}
    \caption{Results on PASCAL VOC 2012 {\it validation} set, pretrained with MS-COCO. E-Supv: extra supervision.}
    \label{tab:deeplab_voc12}
\end{table}

\textbf{Pretrained with MS-COCO:} Second, we experiment with the scenario where the underlying baseline, DeepLab-LargeFOV, has been pretrained on the MS-COCO 2014 dataset \cite{lin2014microsoft}. The goal is to test if we can still observe any improvement with such a strong baseline. As shown in \tabref{tab:deeplab_voc12}, we again observe similar experimental results, and our best model still outperforms the DeepLab-LargeFOV baseline by $3.84\%$. We also report the
best model on the {\it test} set in the bottom of \tabref{tab:voc2012}. For a fair comparison with the reported DeepLab variants on the test set, we employ a fully connected CRF \cite{KrahenbuhlK11} as post processing. As shown in the table, our model attains the performance of $75.1\%$, outperforming DeepLab-CRF-LargeFOV and DeepLab-MSc-CRF-LaregeFOV by $2.4\%$, and $1.5\%$, respectively. Motivated by \cite{lin2015efficient}, incorporating data augmentation by randomly scaling input images (from 0.6 to 1.4) during training brings extra $0.6\%$ improvement in our model.

Note our models do not outperform current best models \cite{lin2015efficient, liu2015semantic}, which employ joint training of CRF (\eg, with the {\it spatial} pairwise term) and FCNs \cite{chen2014learning}. However, we believe our proposed method (\eg, attention model for scales) could be complementary to them. We emphasize that our models are trained end-to-end with one pass to exploit multi-scale features, instead of multiple training steps. Recently, \cite{chen2015semantic} has been shown that further improvement can be attained by combining our proposed model and a discriminatively trained domain transform \cite{GastalOliveira2011DomainTransform}.

\begin{figure*}
  \centering
  \scalebox{0.95}{
  \begin{tabular}{c c c c c c}
   \includegraphics[height=0.1\linewidth]{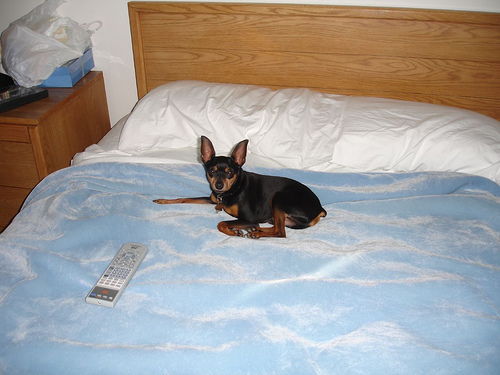} &
   \includegraphics[height=0.1\linewidth]{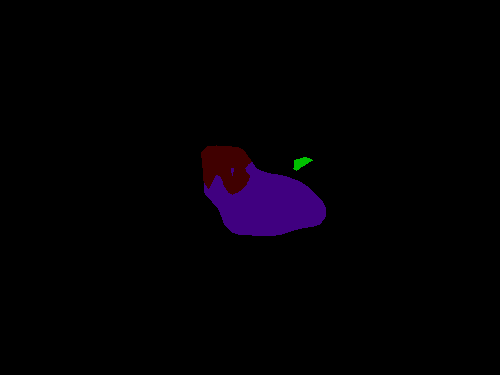} &
   \includegraphics[height=0.1\linewidth]{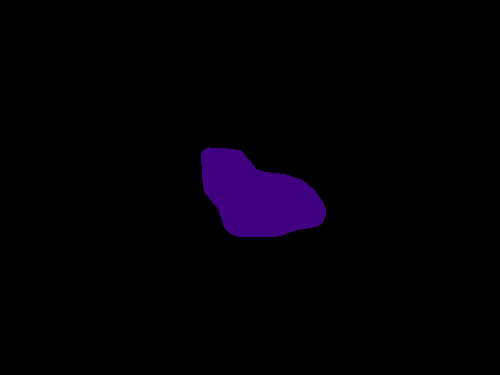} &
   \includegraphics[height=0.1\linewidth]{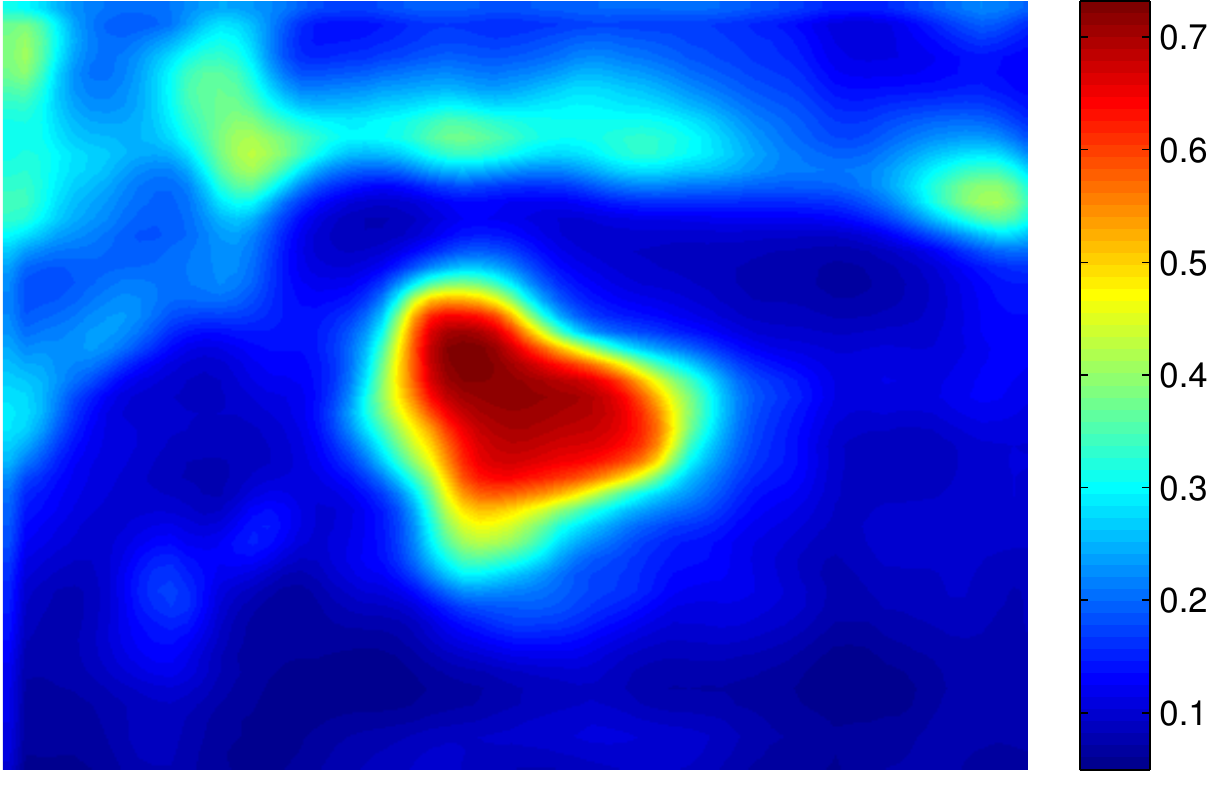} &
   \includegraphics[height=0.1\linewidth]{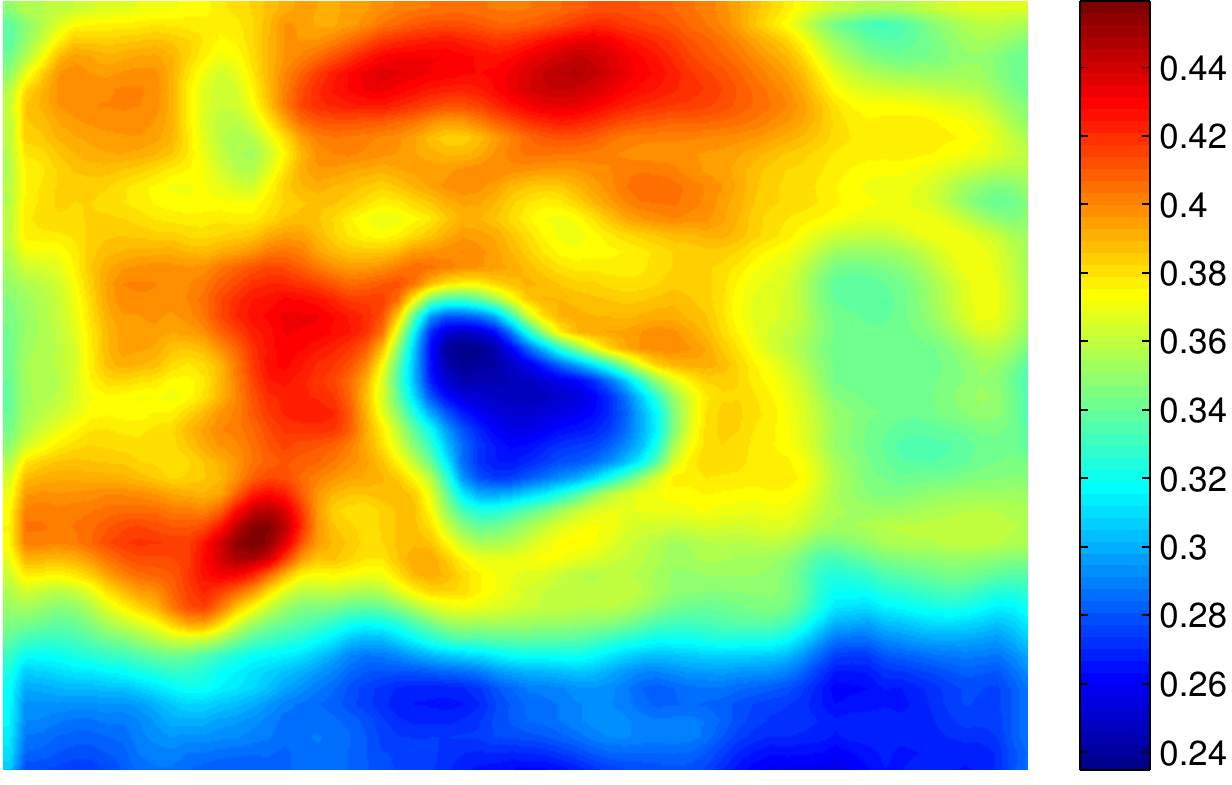} &
   \includegraphics[height=0.1\linewidth]{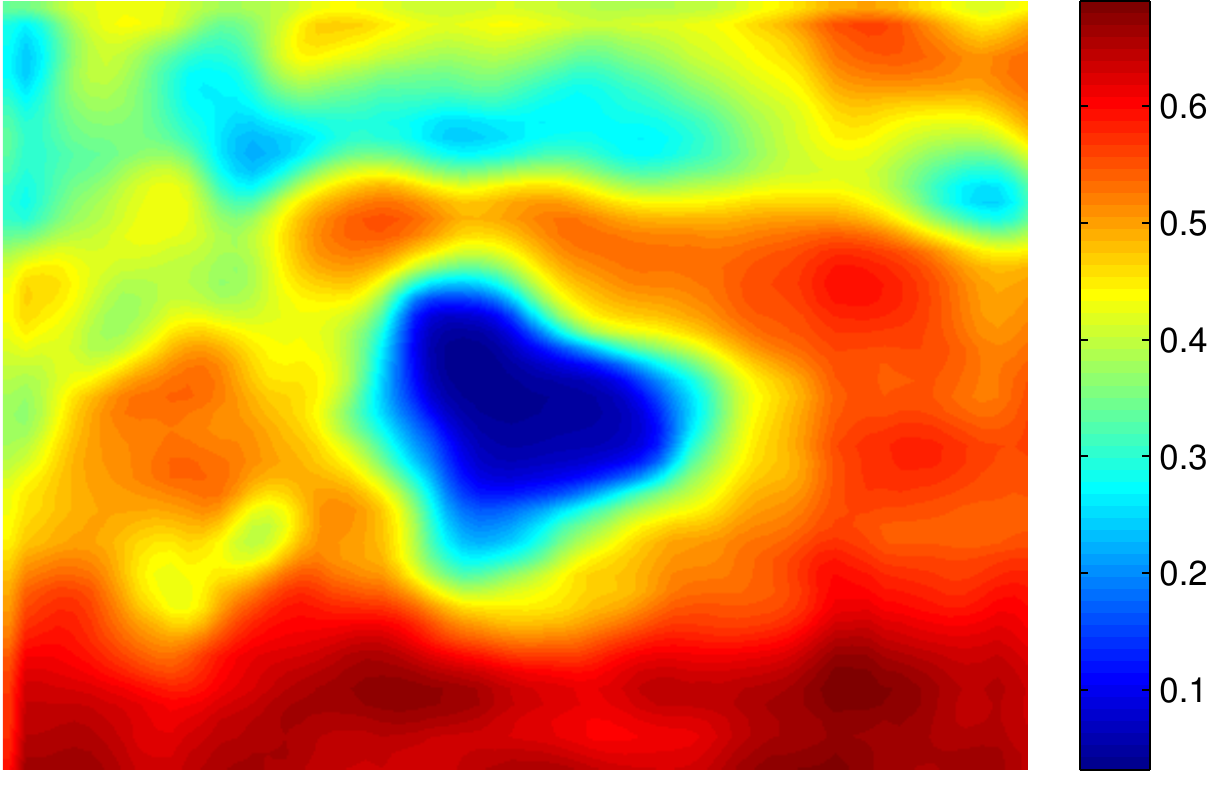} \\
   \includegraphics[height=0.1\linewidth]{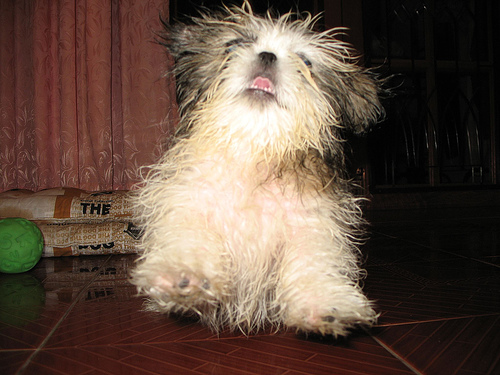} &
   \includegraphics[height=0.1\linewidth]{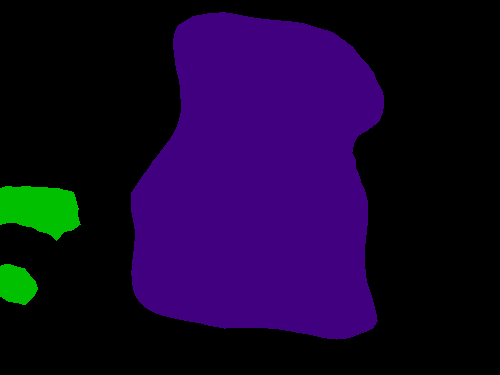} &
   \includegraphics[height=0.1\linewidth]{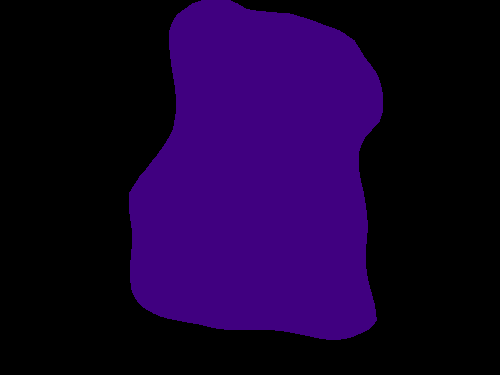} &
   \includegraphics[height=0.1\linewidth]{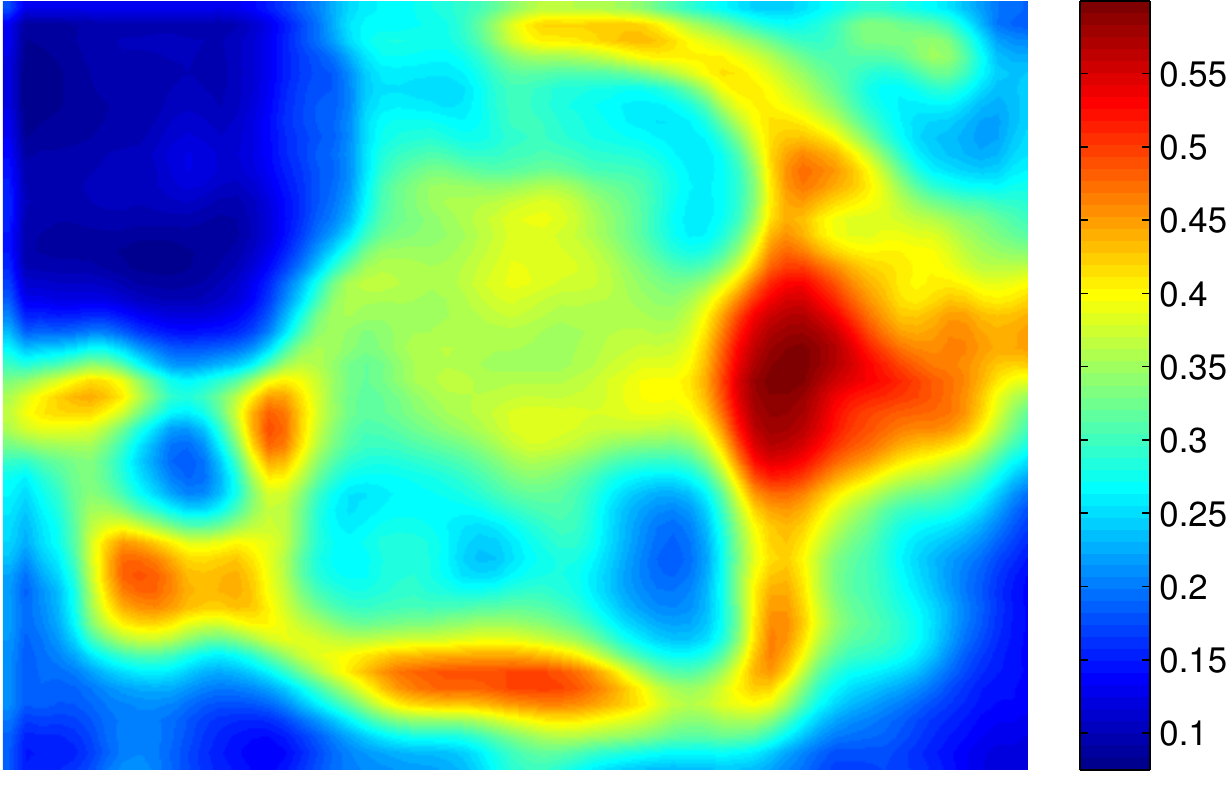} &
   \includegraphics[height=0.1\linewidth]{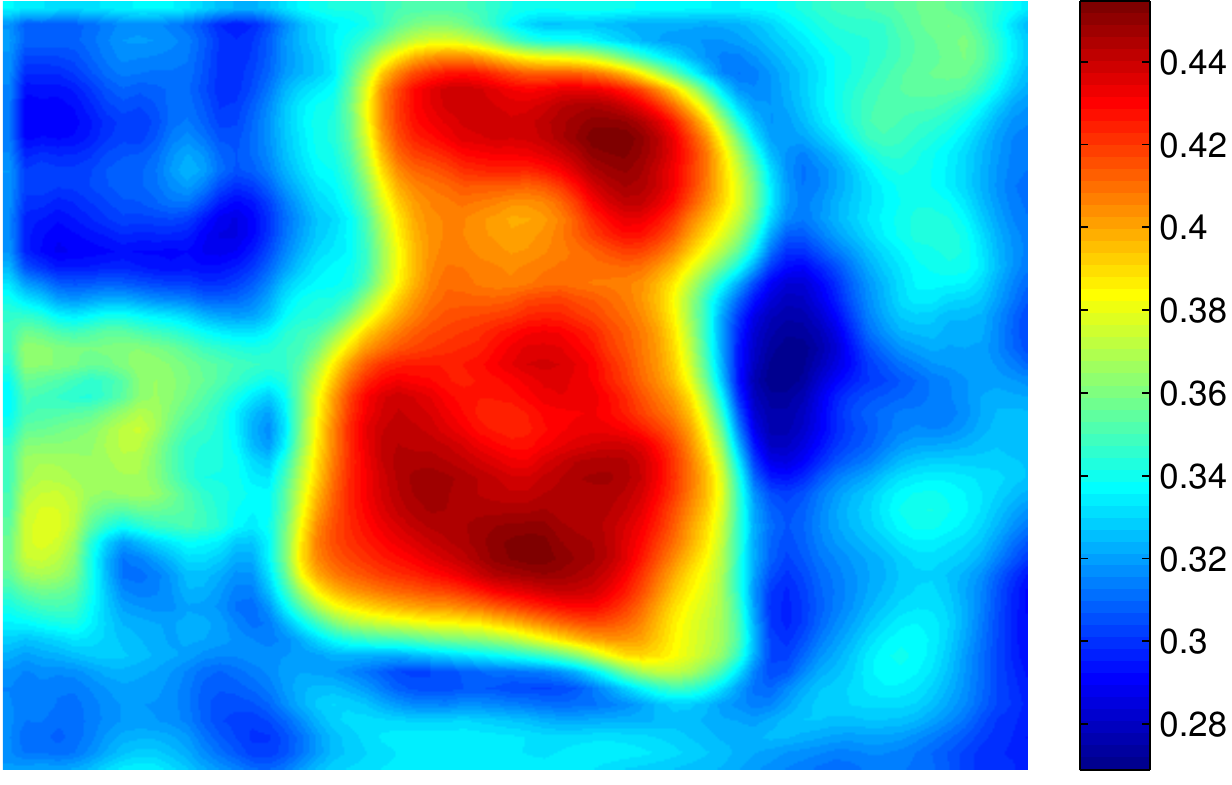} &
   \includegraphics[height=0.1\linewidth]{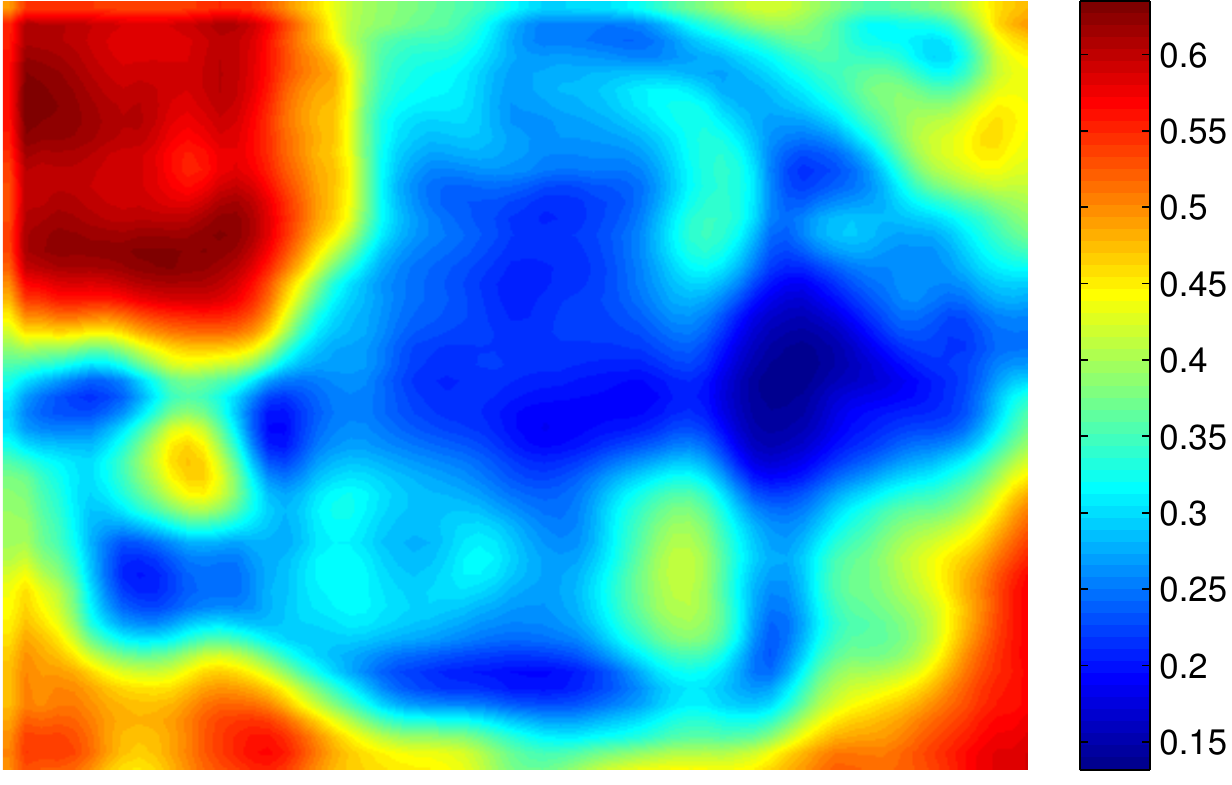} \\
   \includegraphics[height=0.1\linewidth]{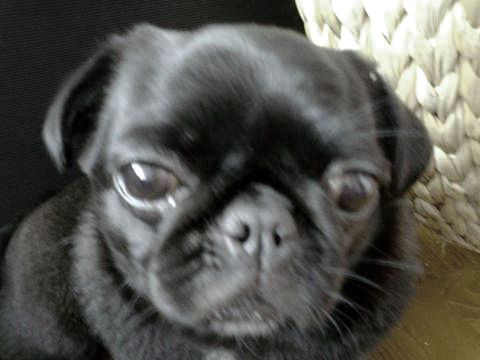} &
   \includegraphics[height=0.1\linewidth]{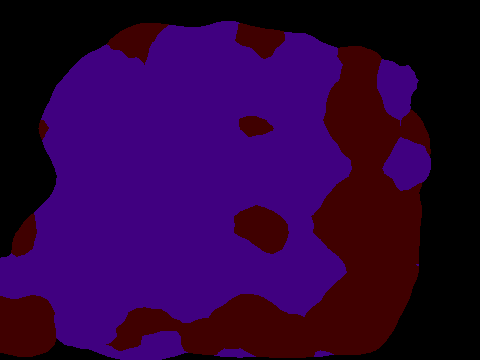} &
   \includegraphics[height=0.1\linewidth]{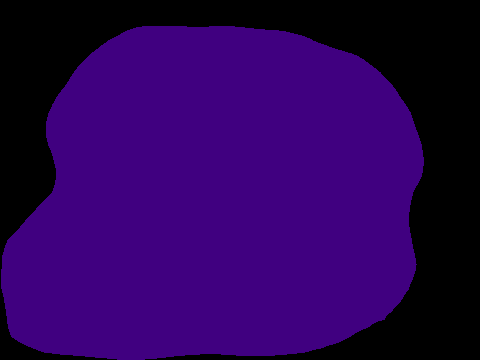} &
   \includegraphics[height=0.1\linewidth]{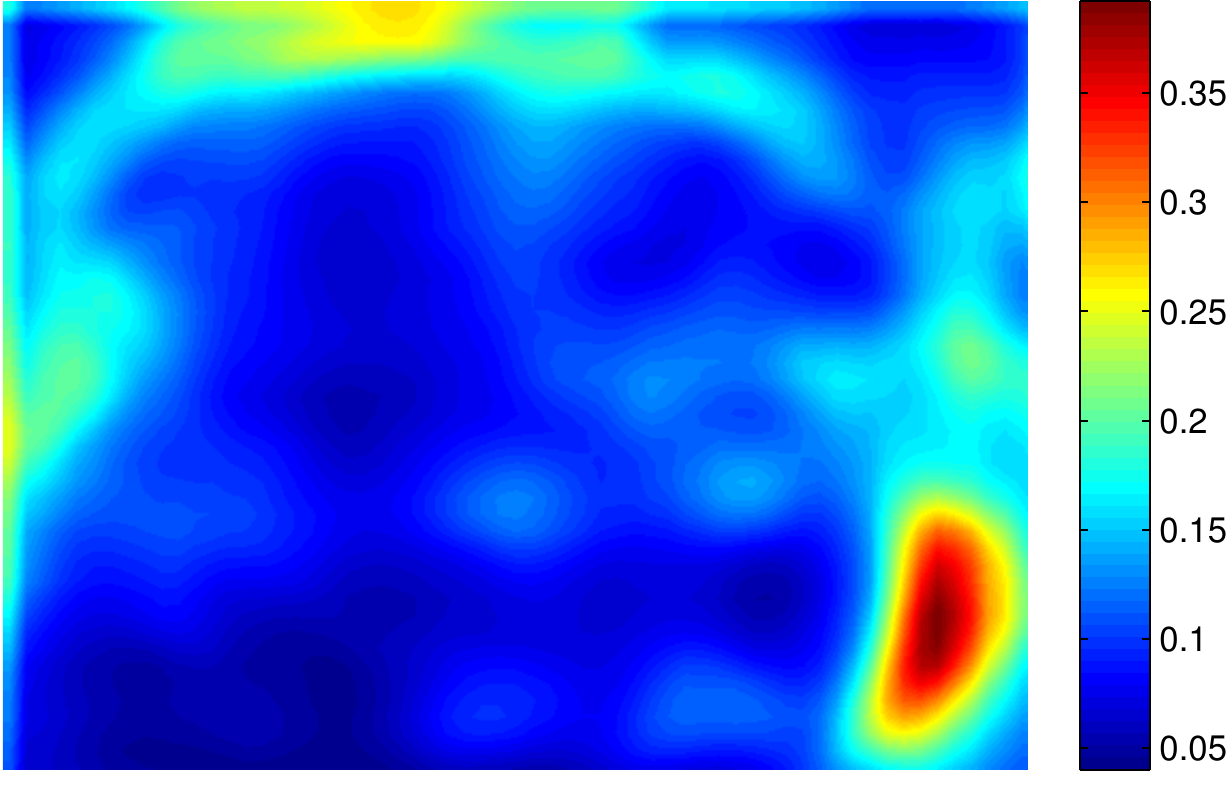} &
   \includegraphics[height=0.1\linewidth]{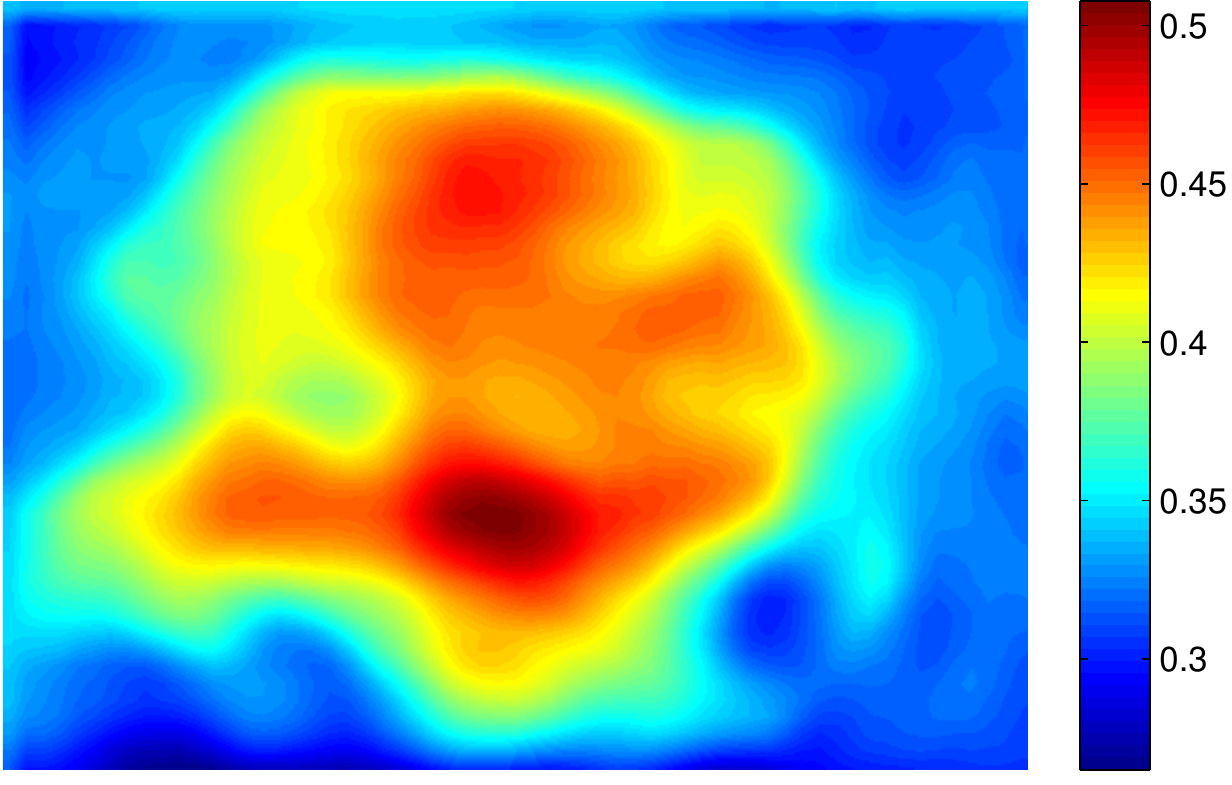} &
   \includegraphics[height=0.1\linewidth]{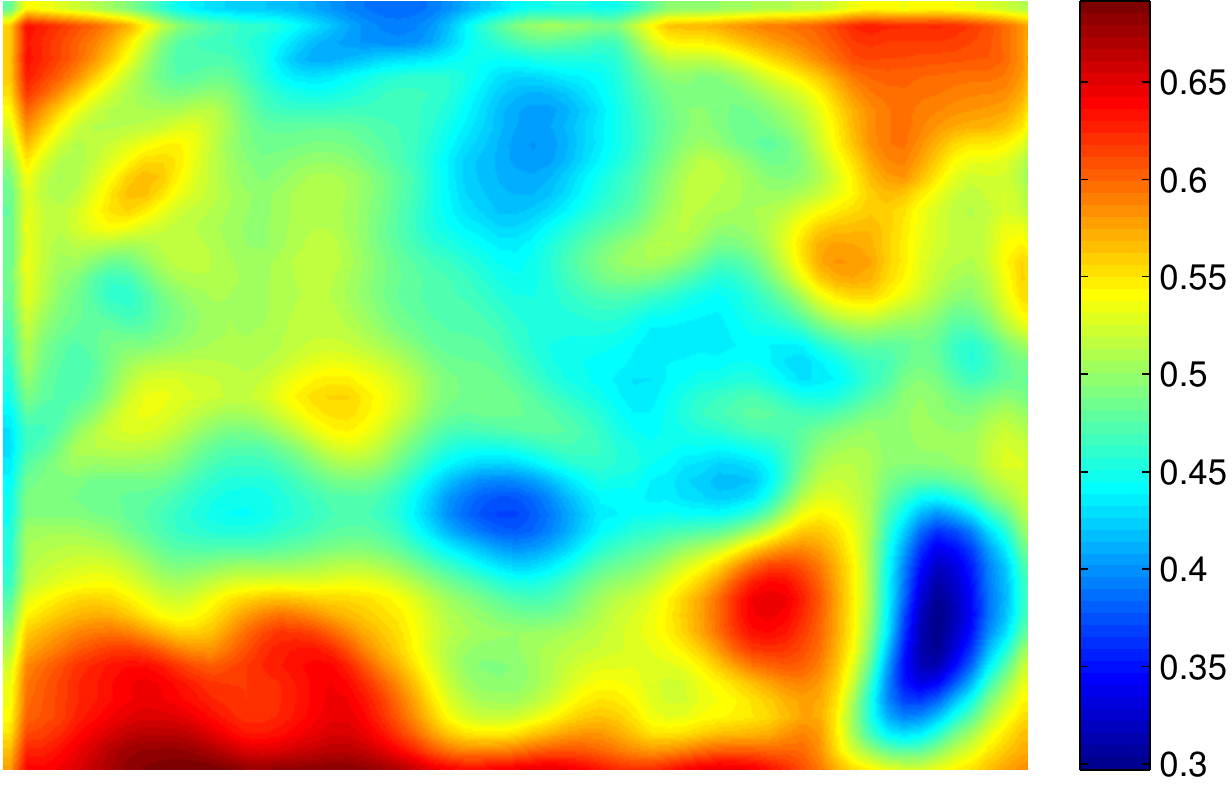} \\
   \includegraphics[height=0.1\linewidth]{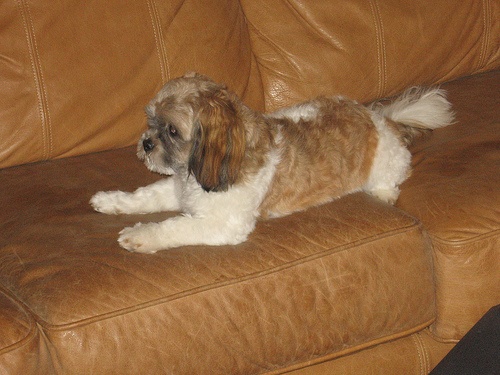} &
   \includegraphics[height=0.1\linewidth]{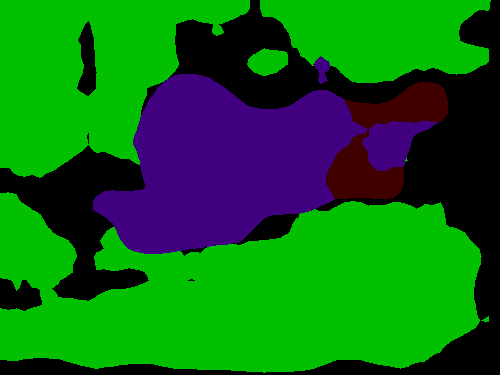} &
   \includegraphics[height=0.1\linewidth]{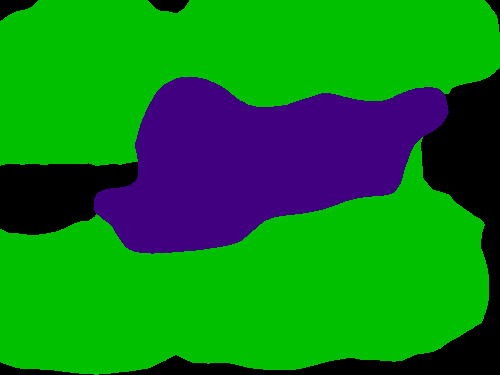} &
   \includegraphics[height=0.1\linewidth]{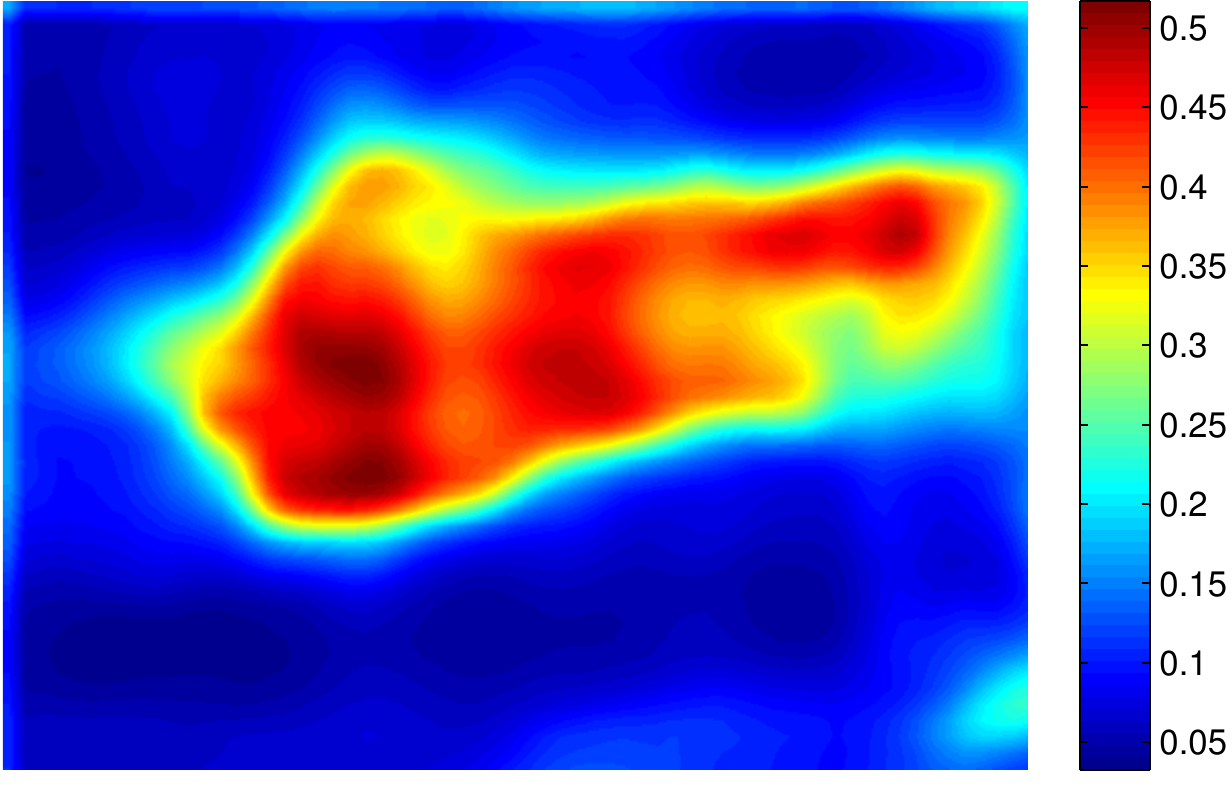} &
   \includegraphics[height=0.1\linewidth]{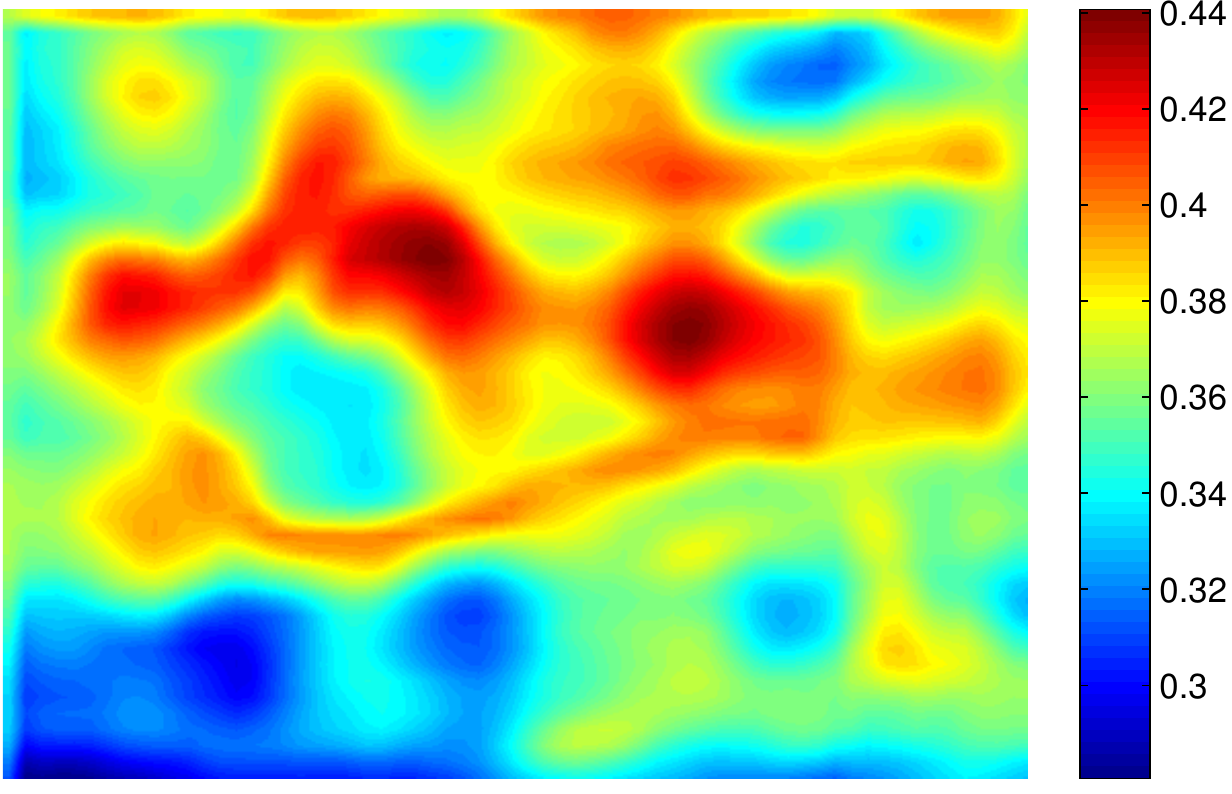} &
   \includegraphics[height=0.1\linewidth]{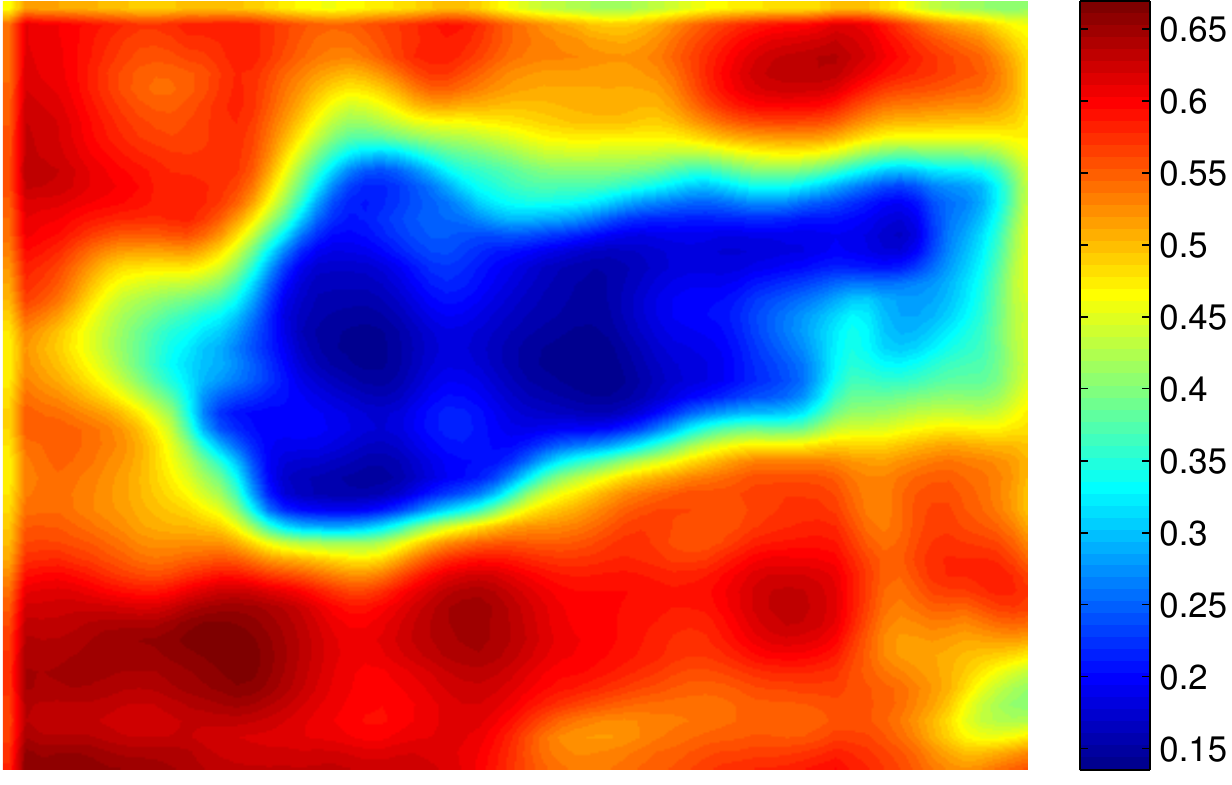} \\
   {\scriptsize (a) Image} & 
   {\scriptsize (b) Baseline} & 
   {\scriptsize (c) Our model} & 
   {\scriptsize (d) Scale-1 Attention} & 
   {\scriptsize (e) Scale-0.75 Attention} &
   {\scriptsize (f) Scale-0.5 Attention} \\
  \end{tabular}
  }
  \caption{Results on PASCAL VOC 2012 {\it validation} set. DeepLab-LargeFOV with one scale input is used as baseline. Our model employs three scale inputs, attention model and extra supervision. Scale-1 attention captures small-scale dogs (dark blue label), scale-0.75 attention concentrates on middle-scale dogs and part of sofa (light green label), while scale-0.5 attention catches largest-scale dogs and sofa.} 
  \label{fig:pascal_voc12_results}  
\end{figure*}

\begin{figure*}[!ht]
  \centering
  \scalebox{0.95}{
  \begin{tabular}{c c c c c c}
   \includegraphics[height=0.14\linewidth]{fig/coco/img/COCO_val2014_000000003109.jpg} &
   \includegraphics[height=0.14\linewidth]{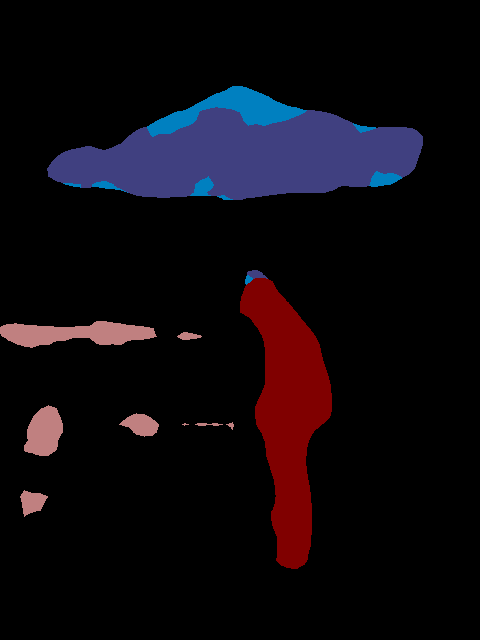} &
   \includegraphics[height=0.14\linewidth]{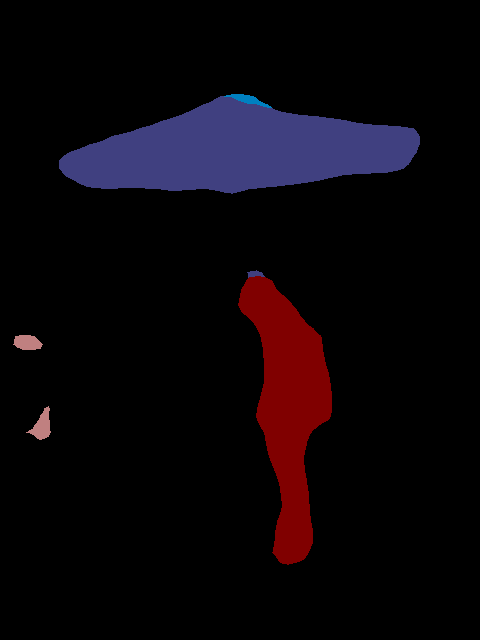} &
   \includegraphics[height=0.14\linewidth]{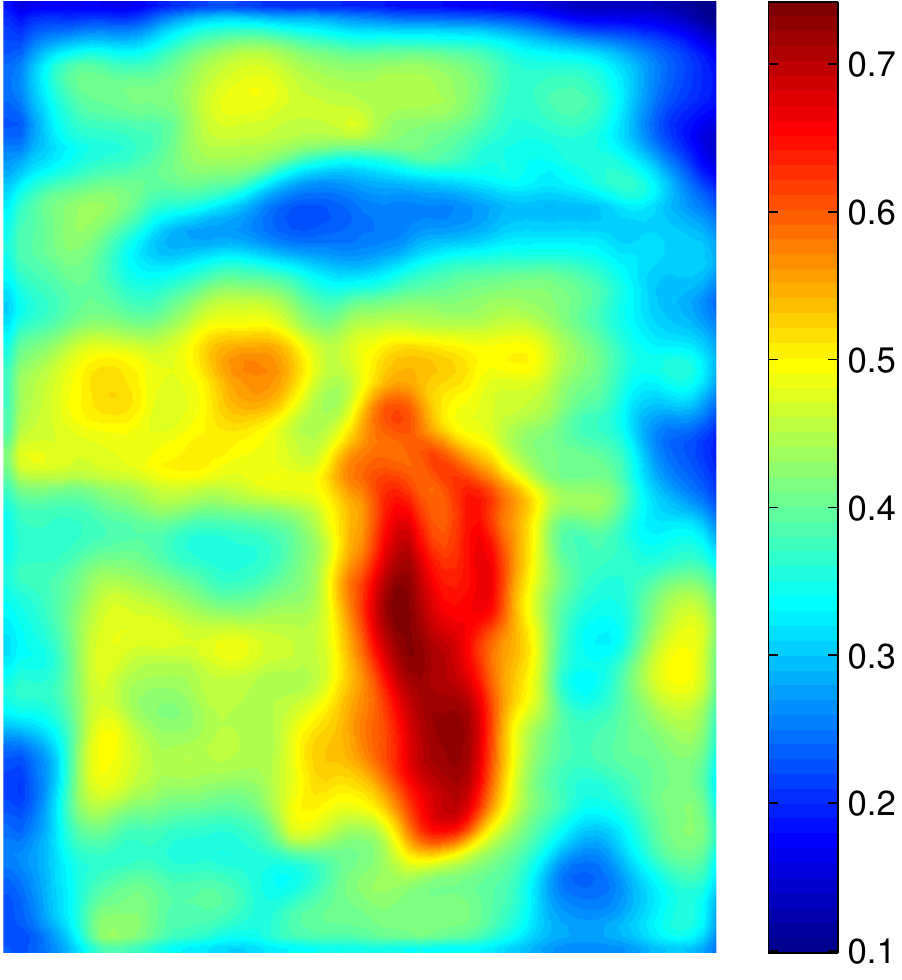} &
   \includegraphics[height=0.14\linewidth]{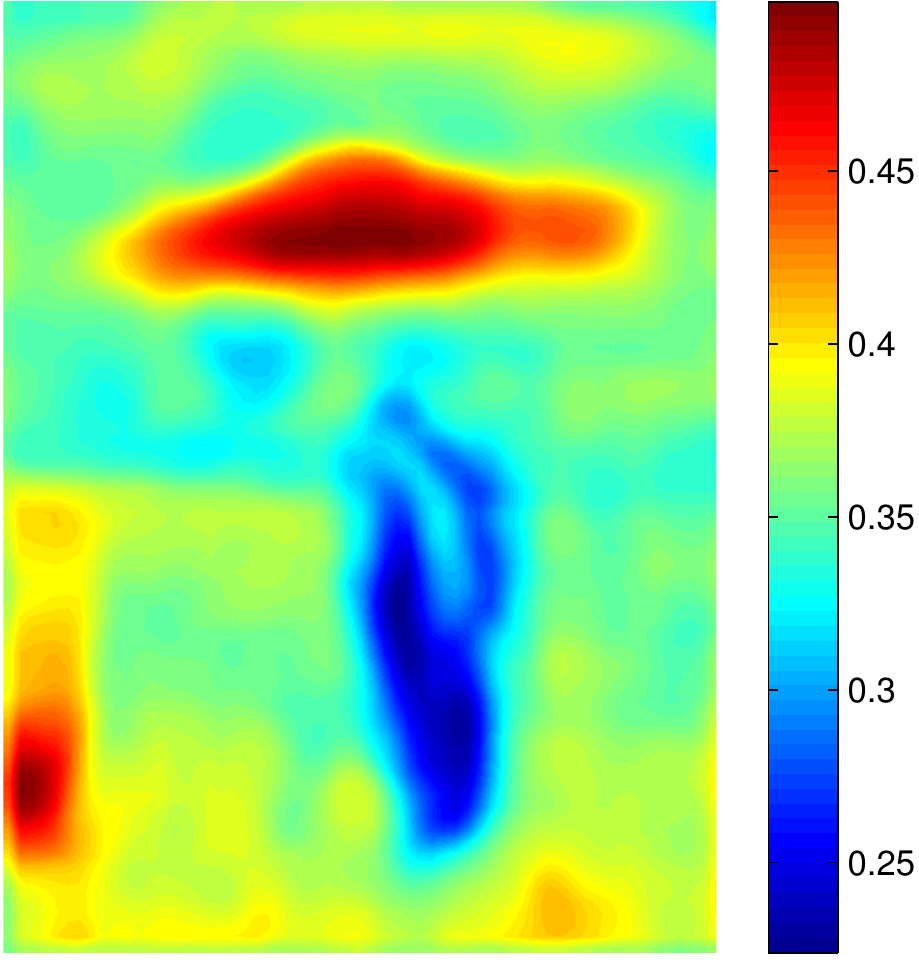} &
   \includegraphics[height=0.14\linewidth]{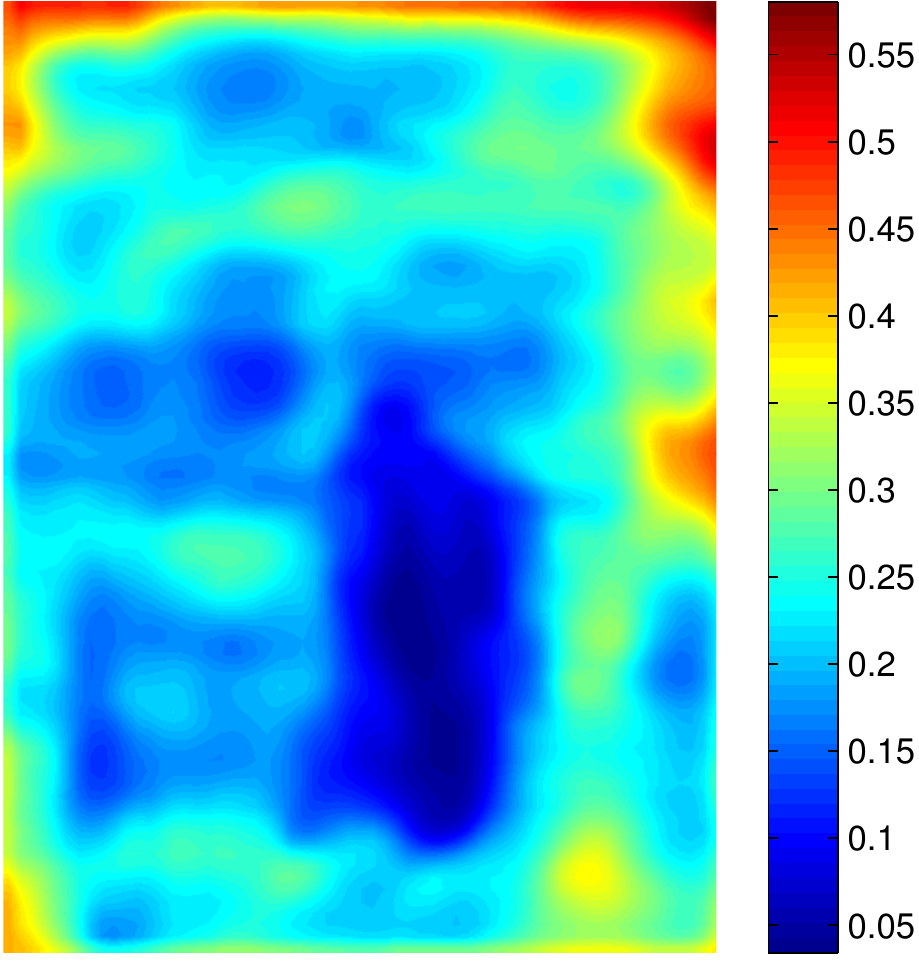} \\
   \includegraphics[height=0.095\linewidth]{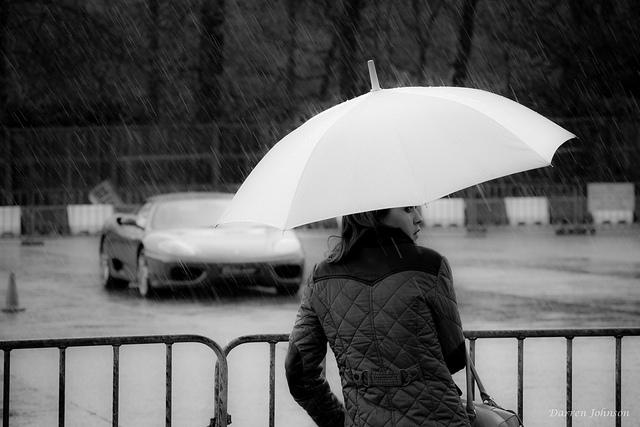} &
   \includegraphics[height=0.095\linewidth]{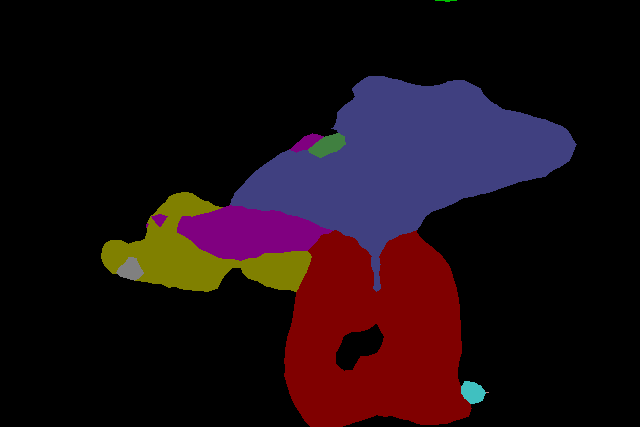} &
   \includegraphics[height=0.095\linewidth]{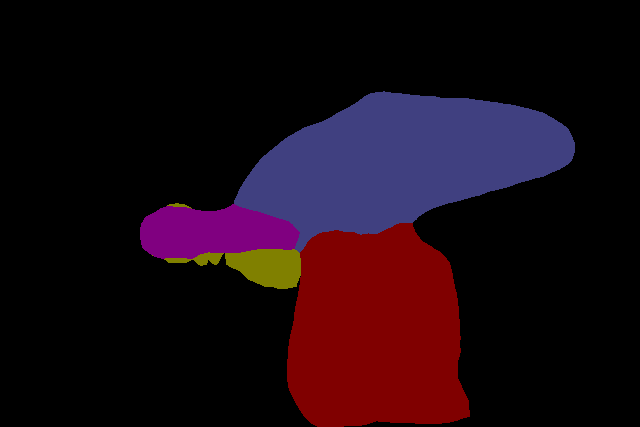} &
   \includegraphics[height=0.095\linewidth]{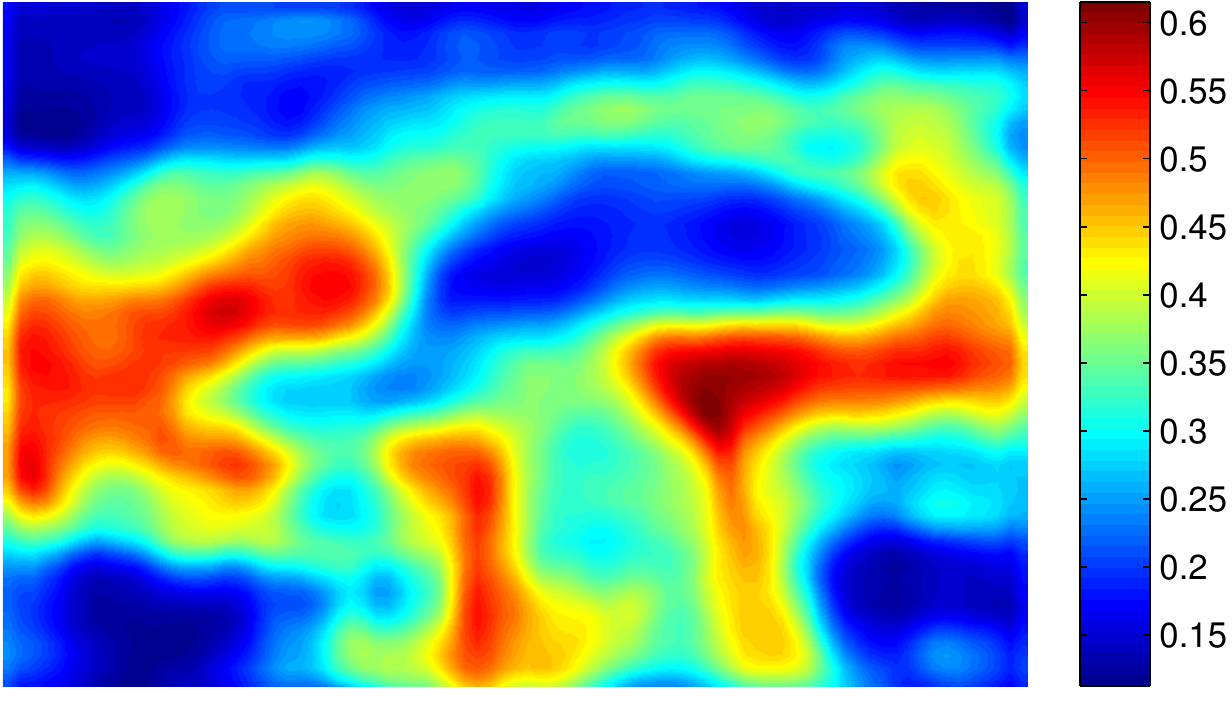} &
   \includegraphics[height=0.095\linewidth]{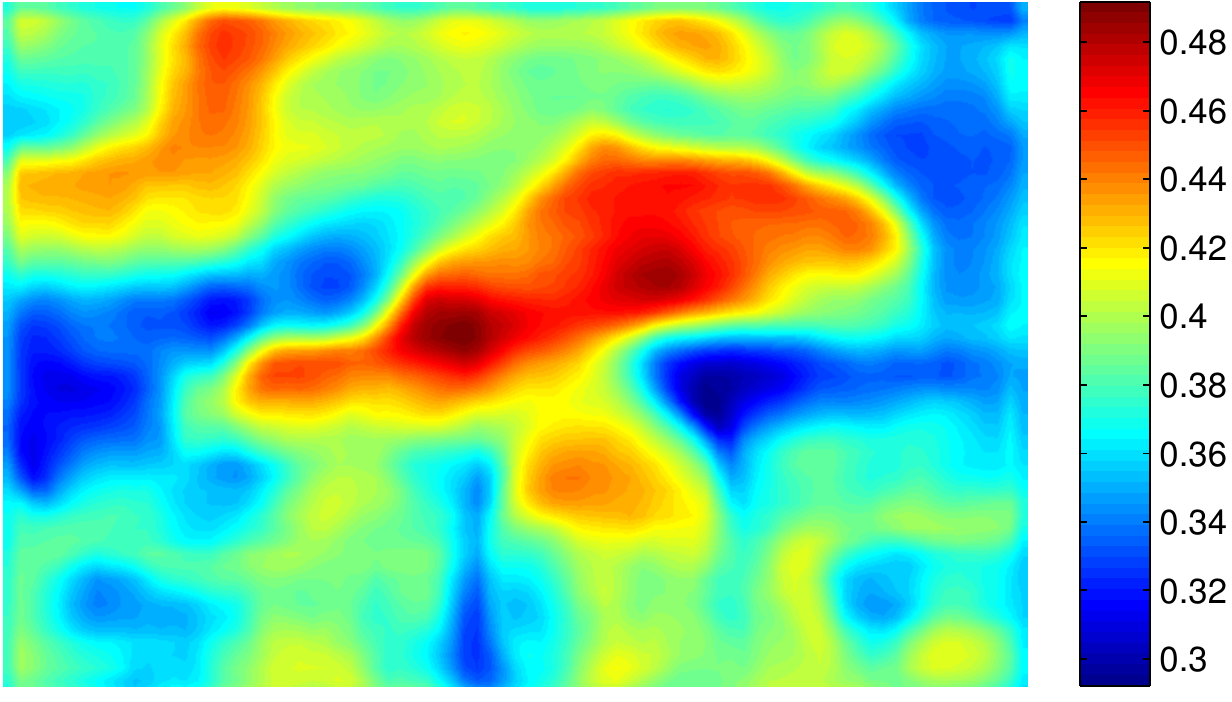} &
   \includegraphics[height=0.095\linewidth]{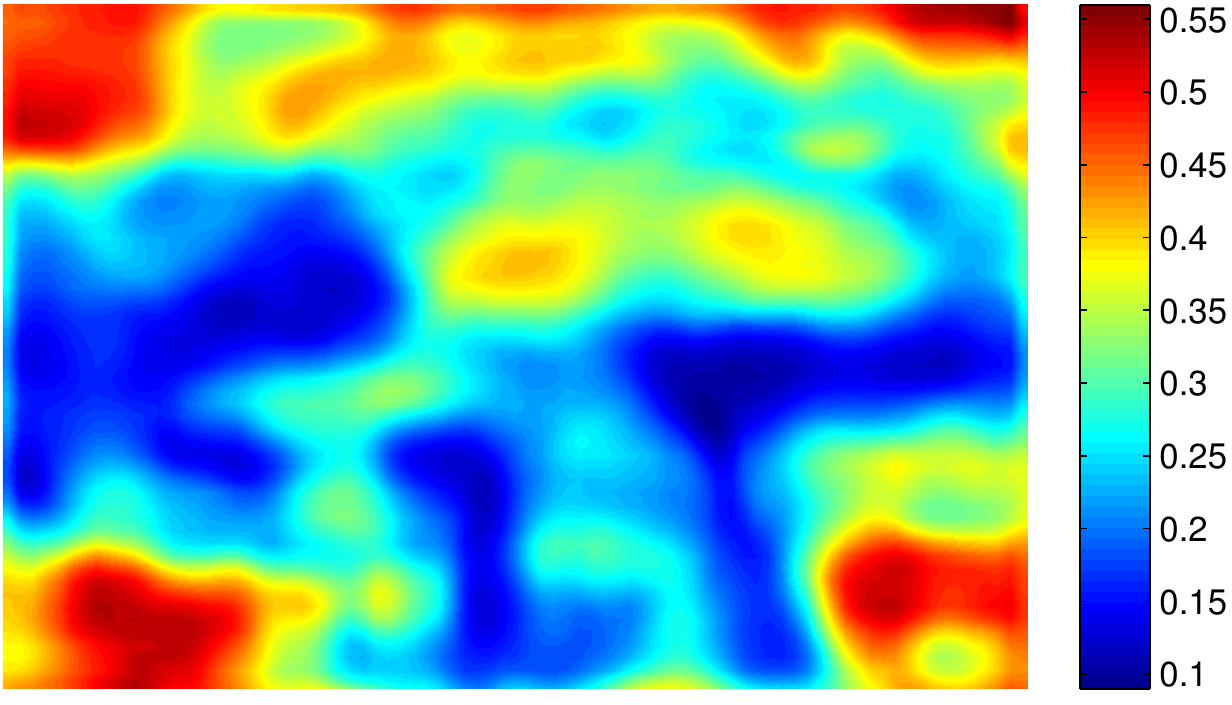} \\
   \includegraphics[height=0.095\linewidth]{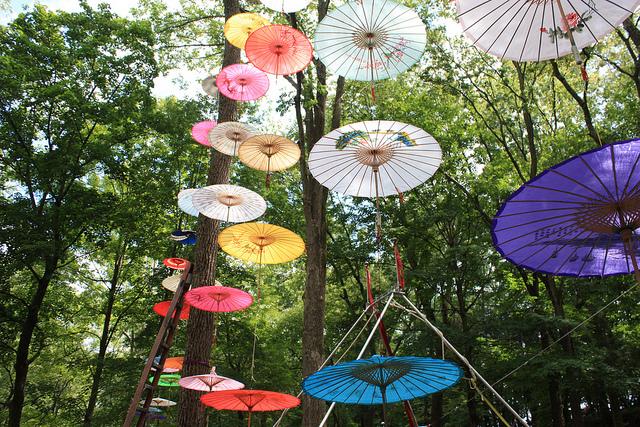} &
   \includegraphics[height=0.095\linewidth]{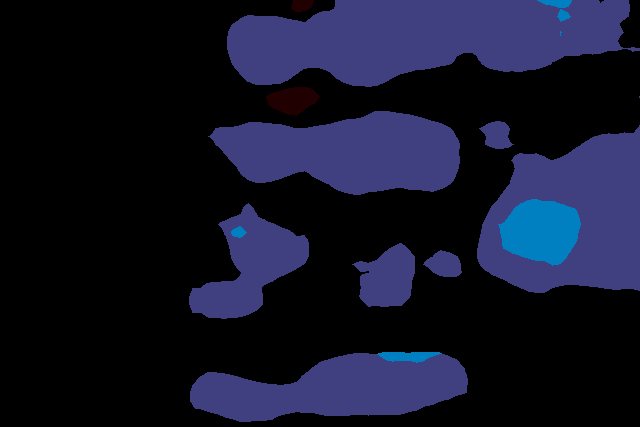} &
   \includegraphics[height=0.095\linewidth]{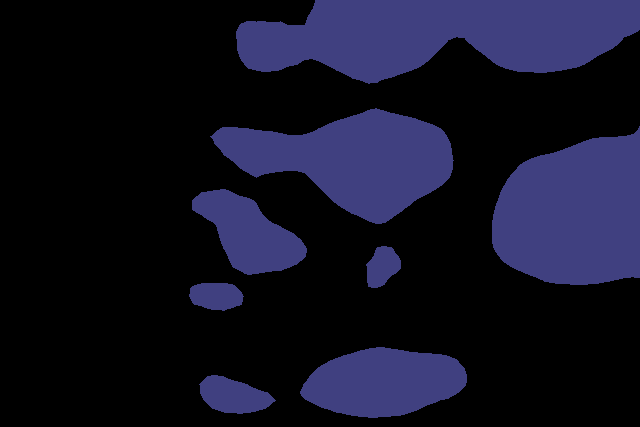} &
   \includegraphics[height=0.095\linewidth]{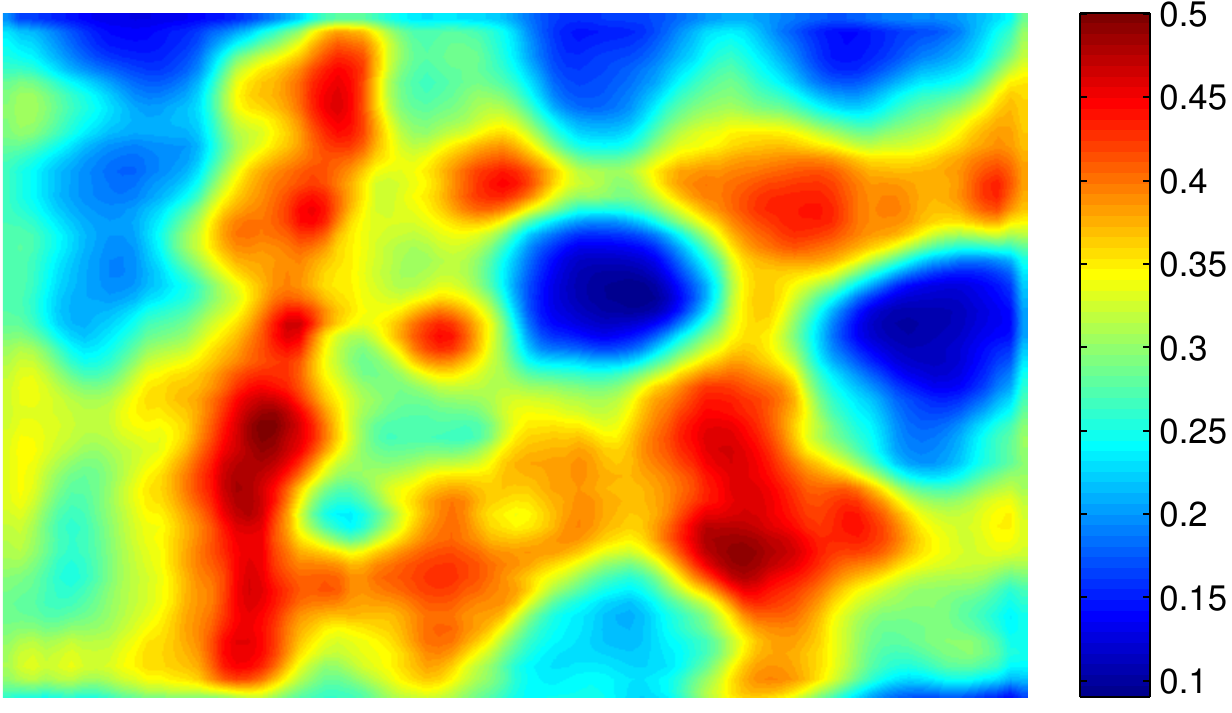} &
   \includegraphics[height=0.095\linewidth]{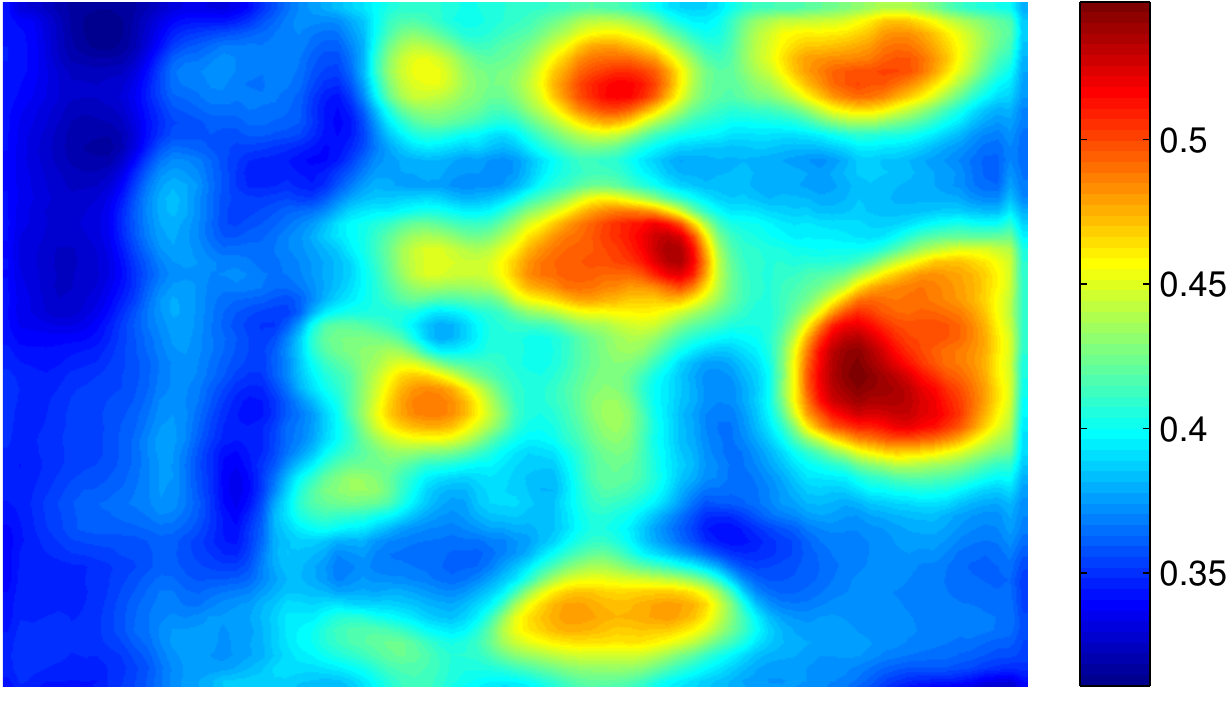} &
   \includegraphics[height=0.095\linewidth]{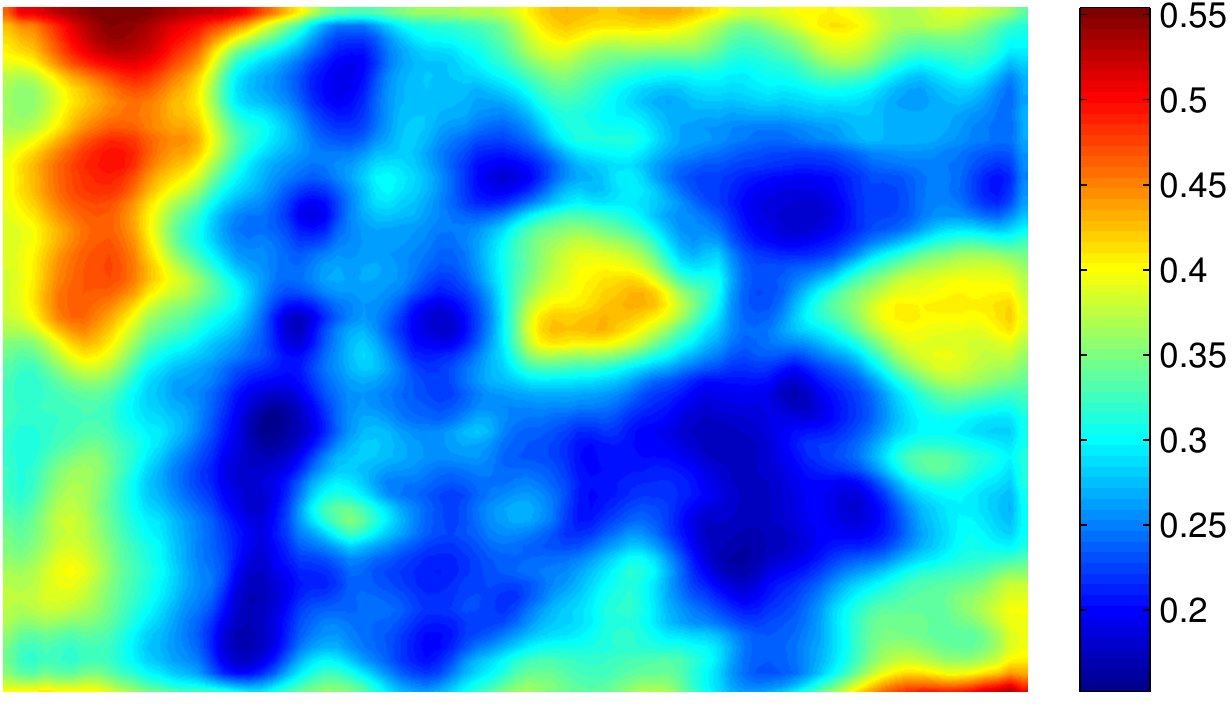} \\
   \includegraphics[height=0.095\linewidth]{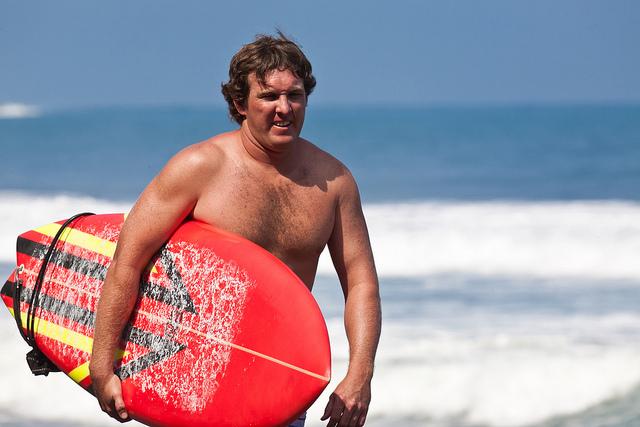} &
   \includegraphics[height=0.095\linewidth]{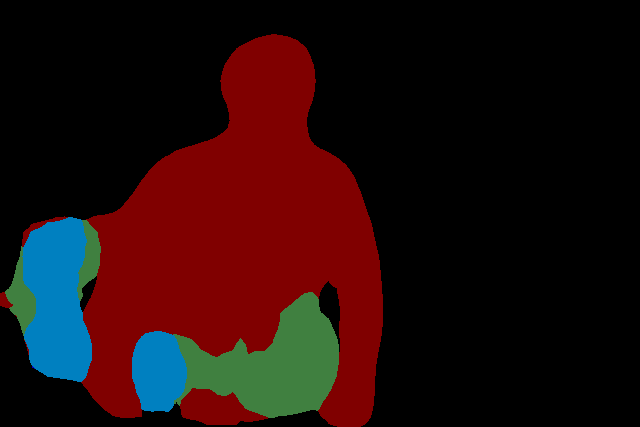} &
   \includegraphics[height=0.095\linewidth]{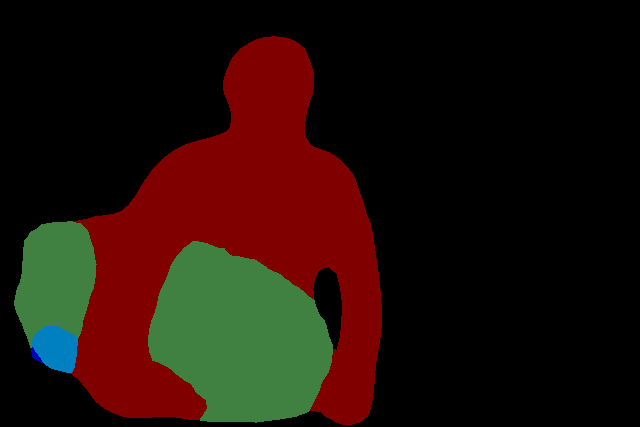} &
   \includegraphics[height=0.095\linewidth]{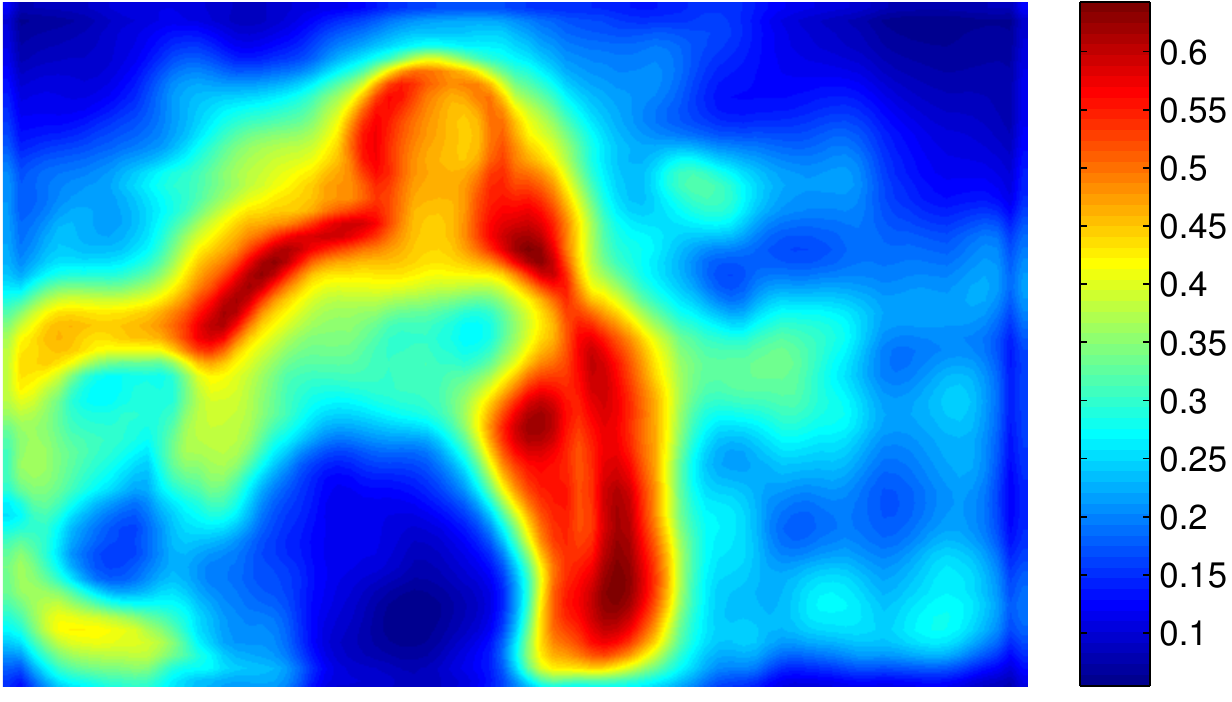} &
   \includegraphics[height=0.095\linewidth]{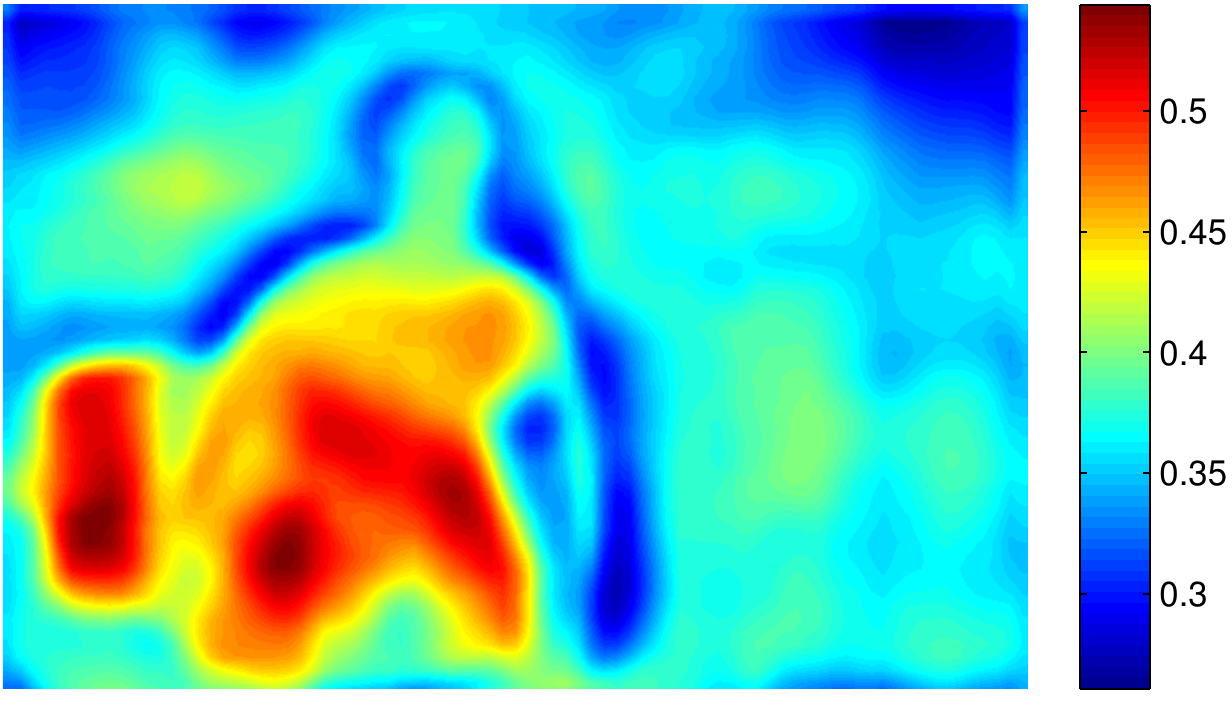} &
   \includegraphics[height=0.095\linewidth]{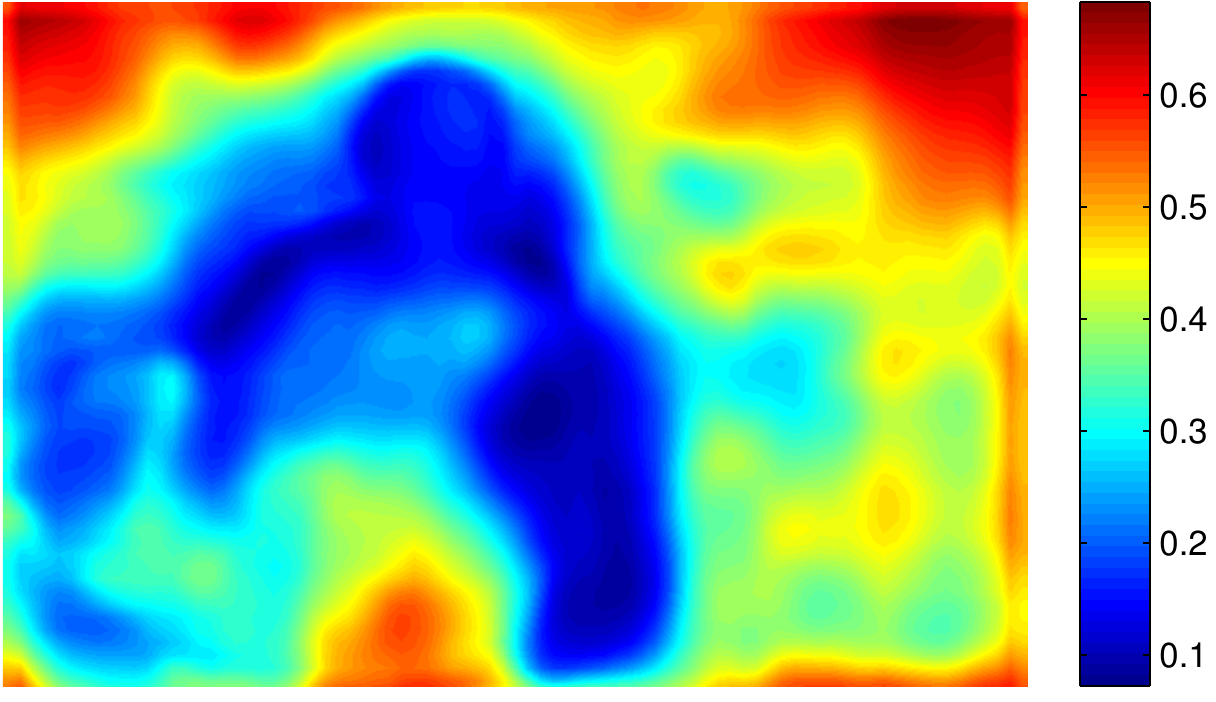} \\
   \includegraphics[height=0.115\linewidth]{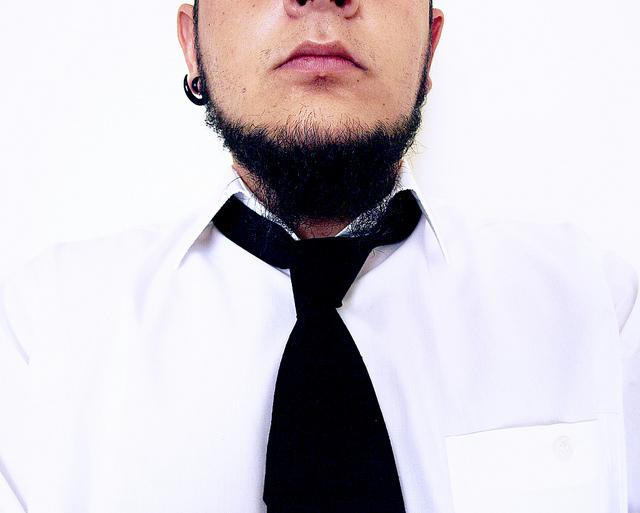} &
   \includegraphics[height=0.115\linewidth]{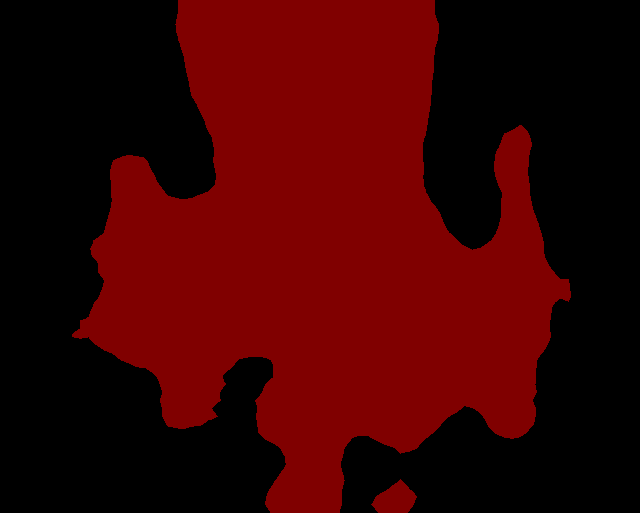} &
   \includegraphics[height=0.115\linewidth]{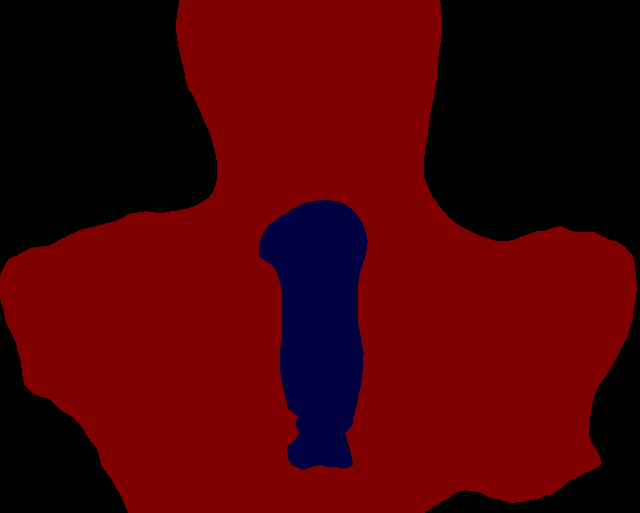} &
   \includegraphics[height=0.115\linewidth]{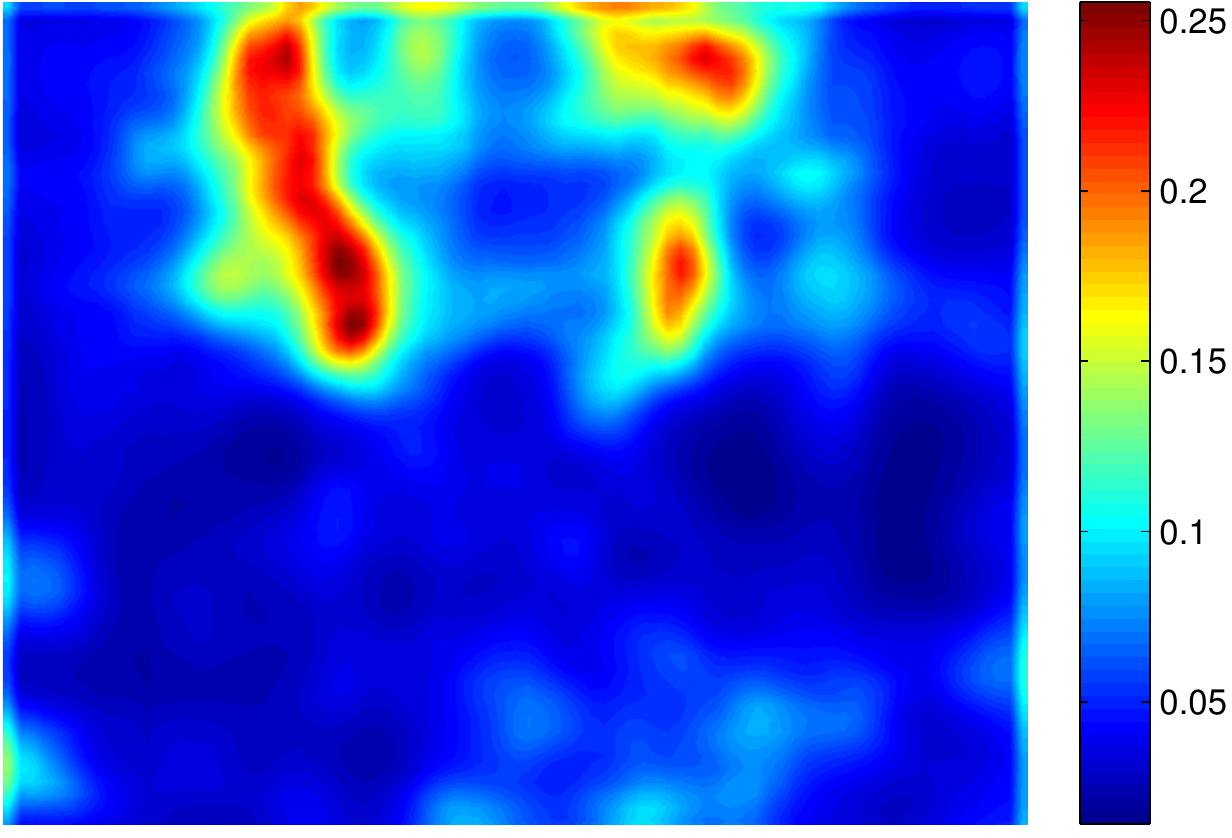} &
   \includegraphics[height=0.115\linewidth]{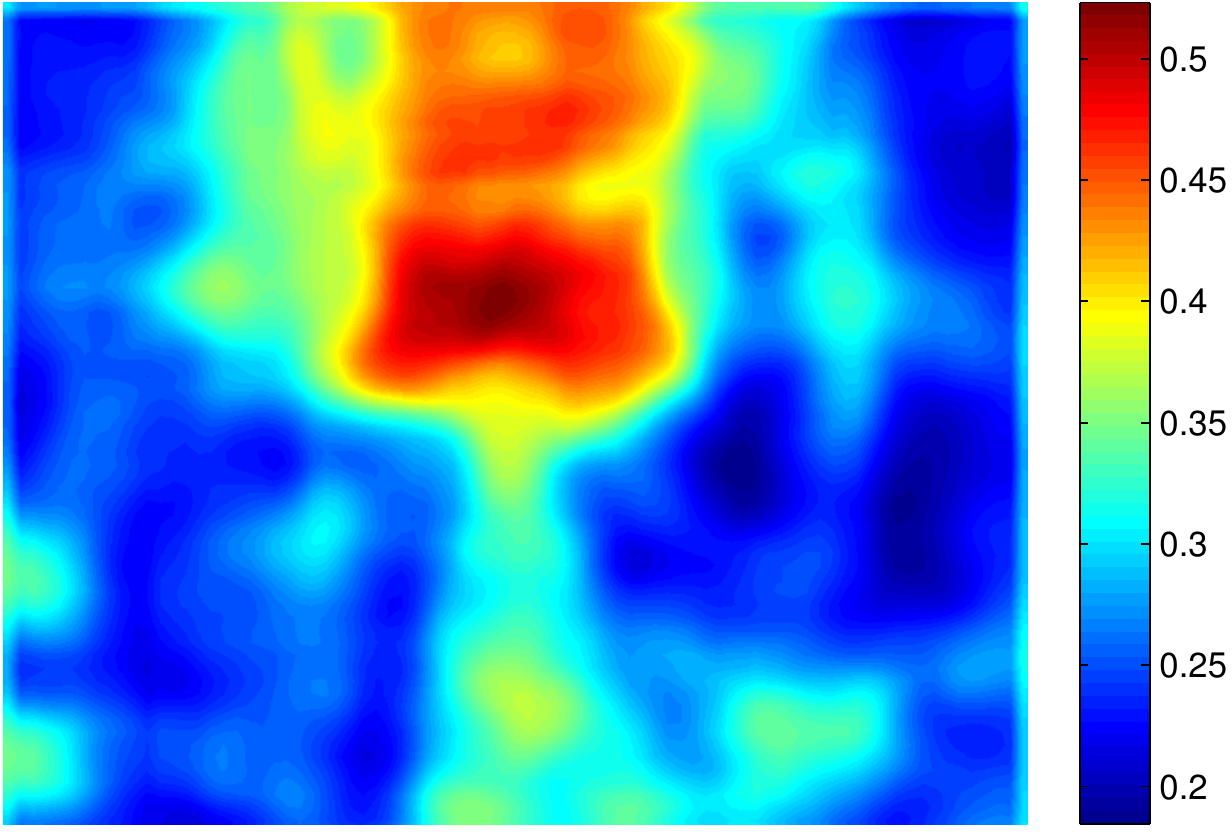} &
   \includegraphics[height=0.115\linewidth]{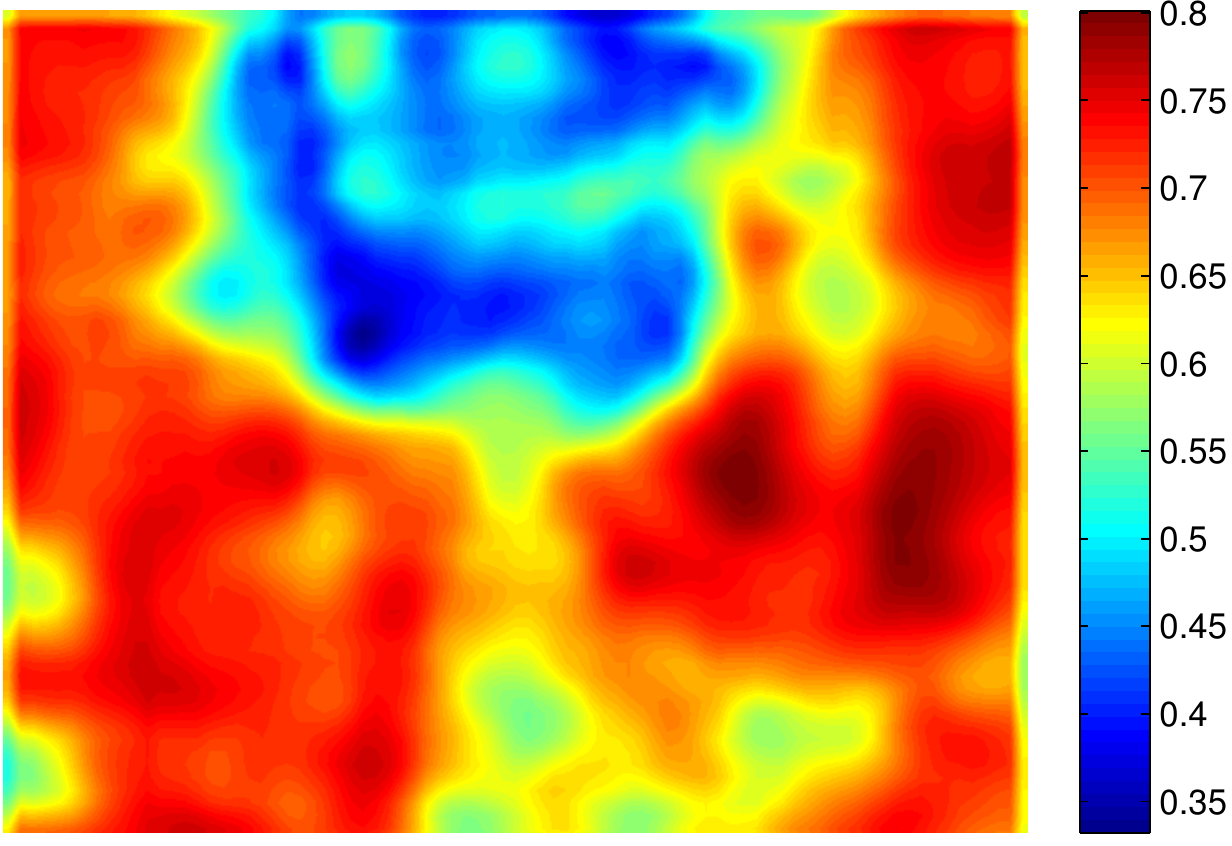} \\ 
   {\scriptsize (a) Image} & 
   {\scriptsize (b) Baseline} & 
   {\scriptsize (c) Our model} & 
   {\scriptsize (d) Scale-1 Attention} & 
   {\scriptsize (e) Scale-0.75 Attention} &
   {\scriptsize (f) Scale-0.5 Attention} \\
  \end{tabular}
  }
  \caption{Results on subset of MS-COCO 2014 {\it validation} set. DeepLab-LargeFOV with one scale input is used as baseline. Our model employs three scale inputs, attention model and extra supervision. Scale-1 attention captures small-scale person (dark red label) and umbrella (violet label). Scale-0.75 attention concentrates on middle-scale umbrella and head, while scale-0.5 attention catches large-scale person torso.}
  \label{fig:pascal_coco_results}  
\end{figure*}

\subsection{Subset of MS-COCO}
\textbf{Dataset:} The MS-COCO 2014 dataset \cite{lin2014microsoft} contains 80 foreground object classes and one background class. The training set has about 80K images, and 40K images for validation. We randomly select 10K images from the training set and 1,500 images from the validation set (the resulting training and validation sets have same sizes as those we used for PASCAL VOC 2012). The goal is to demonstrate our model on another challenging dataset.

\begin{table}
  \centering
  \rowcolors{2}{}{yelloworange!25}
  \addtolength{\tabcolsep}{2.5pt}
    \begin{tabular}{l c c}
      \toprule[0.2 em]
      \multicolumn{2}{l}{Baseline: DeepLab-LargeFOV} & 31.22 \\
      \toprule[0.2 em]
      {\bf Merging Method} & & w/ E-Supv \\
      \midrule \midrule
      {\it Scales = \{1, 0.5\}} & & \\
      Max-Pooling & 32.95 & 34.70 \\
      Average-Pooling & 33.69 & 35.14 \\
      Attention & 34.03 & 35.41 \\
      \midrule
      {\it Scales = \{1, 0.75, 0.5\}} & & \\
      Max-Pooling & 33.58 & 35.08 \\
      Average-Pooling & 33.74 & 35.72 \\
      Attention & 33.42 & {\bf 35.78} \\
      \bottomrule[0.1 em]
    \end{tabular}
    \caption{Results on the subset of MS-COCO {\it validation} set with DeepLab-LargeFOV as the baseline. E-Supv: extra supervision.}
    \label{tab:deeplab_coco}
\end{table}

\textbf{Improvement over DeepLab:} In addition to observing similar results as before, we find that the DeepLab-LargeFOV baseline achieves a low mean IOU $31.22 \%$ in \tabref{tab:deeplab_coco} due to the difficulty of MS-COCO dataset (\eg, large object scale variance and more object classes). However, employing multi-scale inputs, attention model, and extra supervision can still bring $4.6\%$ improvement over the DeepLab-LargeFOV baseline, and $4.17\%$ over DeepLab-MSc-LargeFOV ($31.61\%$). We find that the results of employing average-pooling and the attention model as merging methods are very similar. We hypothesize that many small object classes (\eg, fork, mouse, and toothbrush) with extremely low prediction accuracy reduce the improvement. This challenging problem (\ie, segment small objects and handle imbalanced classes) is considered as future work. On the other hand, we show the performance for the {\it person} class in \tabref{tab:deeplab_coco_person} because it occurs most frequently and appears with different scales (see Fig. 5(a), and Fig. 13(b) in \cite{lin2014microsoft}) in this dataset. As shown in the table, the improvement from the proposed methods becomes more noticeable in this case, and we observe the same results as before. The qualitative results are shown in \figref{fig:pascal_coco_results}.

\begin{table}
  \centering
  \rowcolors{2}{}{yelloworange!25}
  \addtolength{\tabcolsep}{2.5pt}
    \begin{tabular}{l c c}
      \toprule[0.2 em]
      \multicolumn{2}{l}{Baseline: DeepLab-LargeFOV} & 68.76 \\
      \toprule[0.2 em]
      {\bf Merging Method} & & w/ E-Supv \\
      \midrule \midrule
      {\it Scales = \{1, 0.5\}} & & \\
      Max-Pooling & 70.07 & 71.06 \\
      Average-Pooling & 70.38 & 71.60 \\
      Attention & 70.66 & 72.20 \\
      \midrule
      {\it Scales = \{1, 0.75, 0.5\}} & & \\
      Max-Pooling & 69.97 & 71.43 \\
      Average-Pooling & 69.69 & 71.70 \\
      Attention & 70.14 & {\bf 72.72} \\
      \bottomrule[0.1 em]
    \end{tabular}
    \caption{{\bf Person} class IOU on subset of MS-COCO {\it validation} set with DeepLab-LargeFOV as baseline. E-Supv: extra supervision.}
    \label{tab:deeplab_coco_person}
\end{table}

\section{Conclusion}
For semantic segmentation, this paper adapts a state-of-the-art model (\ie, DeepLab-LargeFOV) to exploit multi-scale inputs. Experiments on three datasets have shown that: (1) Using multi-scale inputs yields better performance than a single scale input. (2) Merging the multi-scale features with the proposed attention model not only improves the performance over average- or max-pooling baselines, but also allows us to diagnostically visualize the importance of features at different positions and scales. (3) Excellent performance can be obtained by adding extra supervision to the final output of networks for each scale. 



\paragraph{Acknowledgments} 
This work wast partly supported by ARO 62250-CS and NIH Grant 5R01EY022247-03. We thank Xiao-Chen Lian for valuable discussions. We also thank Sam Hallman and Haonan Yu for the proofreading.

\appendix
\section*{Supplementary Material}
We include as appendix: (1) more qualitative results on PASCAL-Person-Part, PASCAL VOC 2012, and subset of MS-COCO 2014 datasets.

\section{More qualitative results}
\begin{figure*}
  \centering
  \begin{tabular}{c c c c c c}
   \multicolumn{6}{c}{\includegraphics[width=0.5\linewidth]{fig/voc10_part/legend.pdf}} \\
   \includegraphics[height=0.09\linewidth]{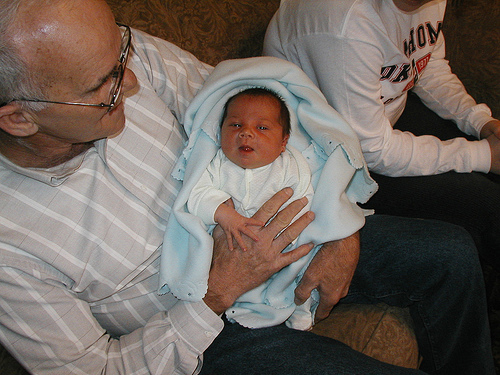} &
   \includegraphics[height=0.09\linewidth]{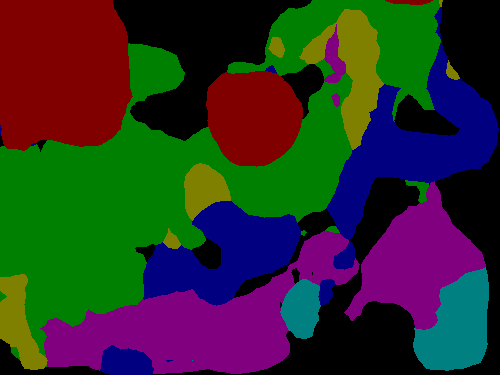} &
   \includegraphics[height=0.09\linewidth]{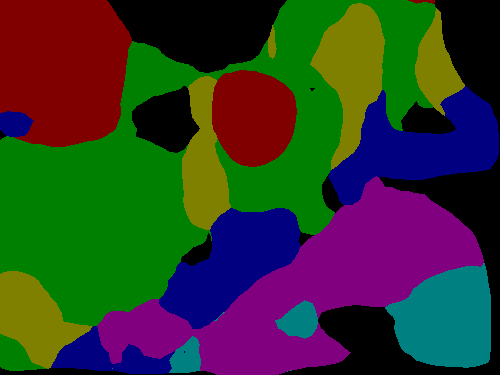} &
   \includegraphics[height=0.09\linewidth]{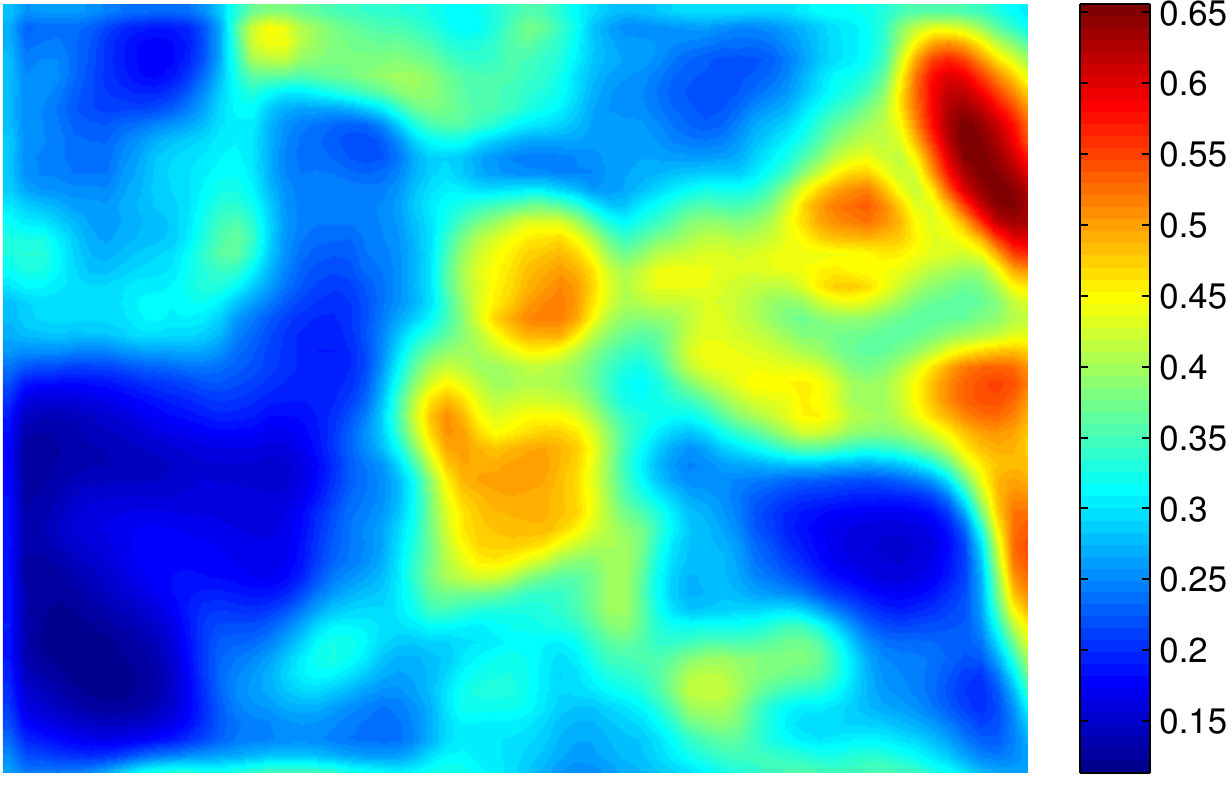} &
   \includegraphics[height=0.09\linewidth]{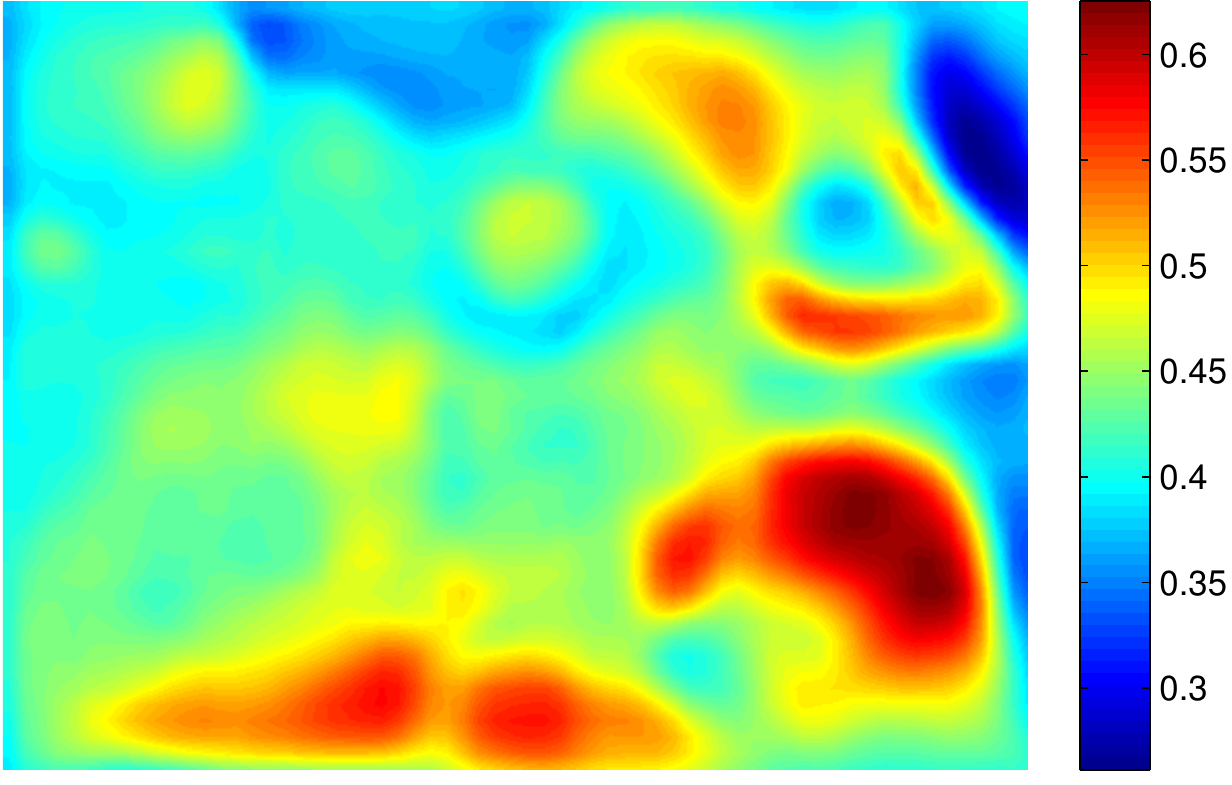} &
   \includegraphics[height=0.09\linewidth]{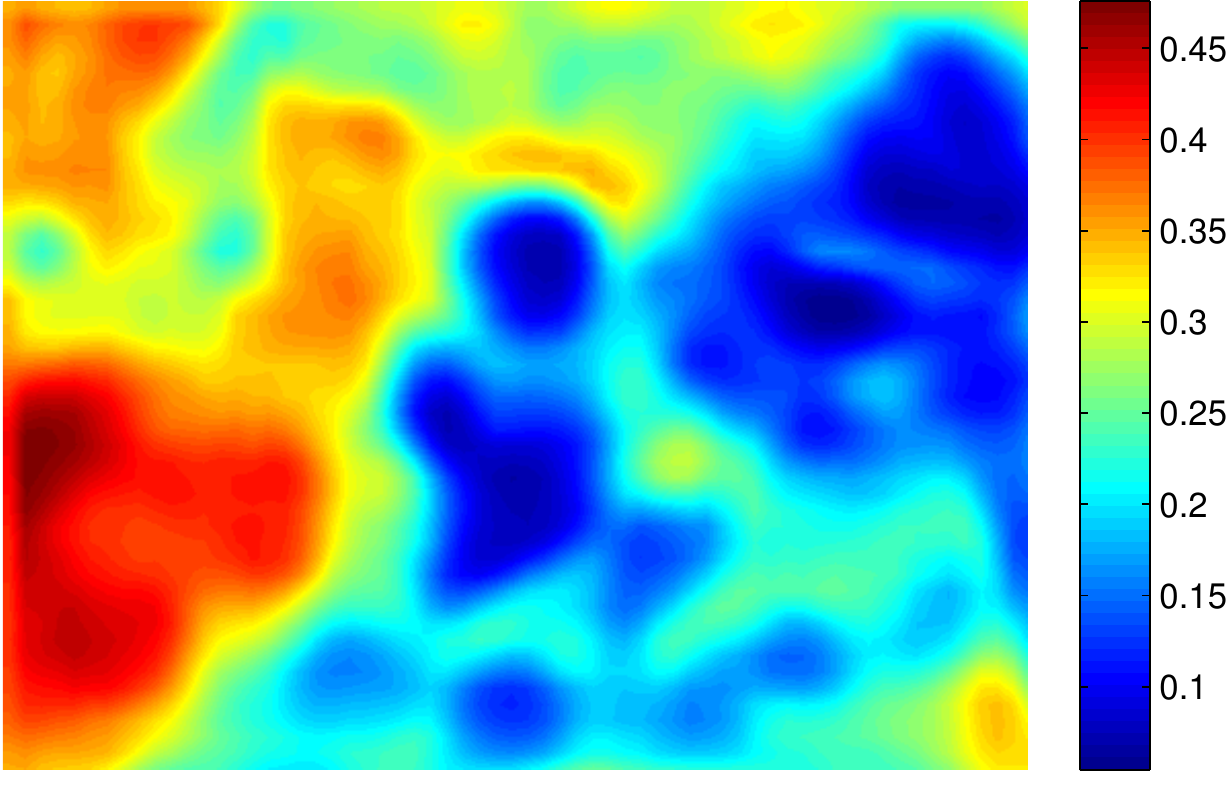} \\
   \includegraphics[height=0.13\linewidth]{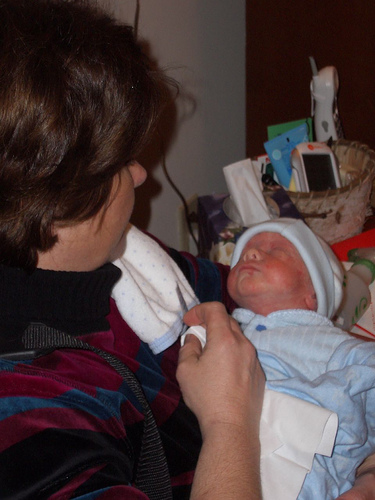} &
   \includegraphics[height=0.13\linewidth]{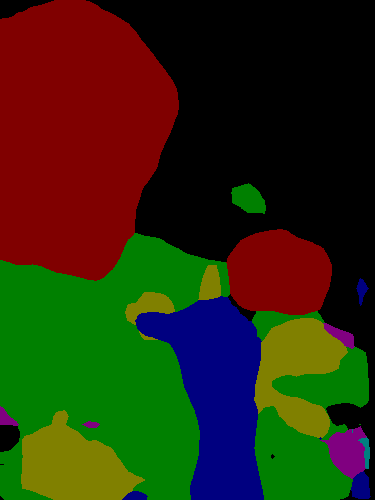} &
   \includegraphics[height=0.13\linewidth]{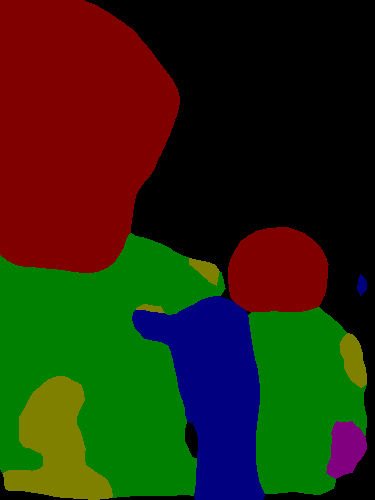} &
   \includegraphics[height=0.13\linewidth]{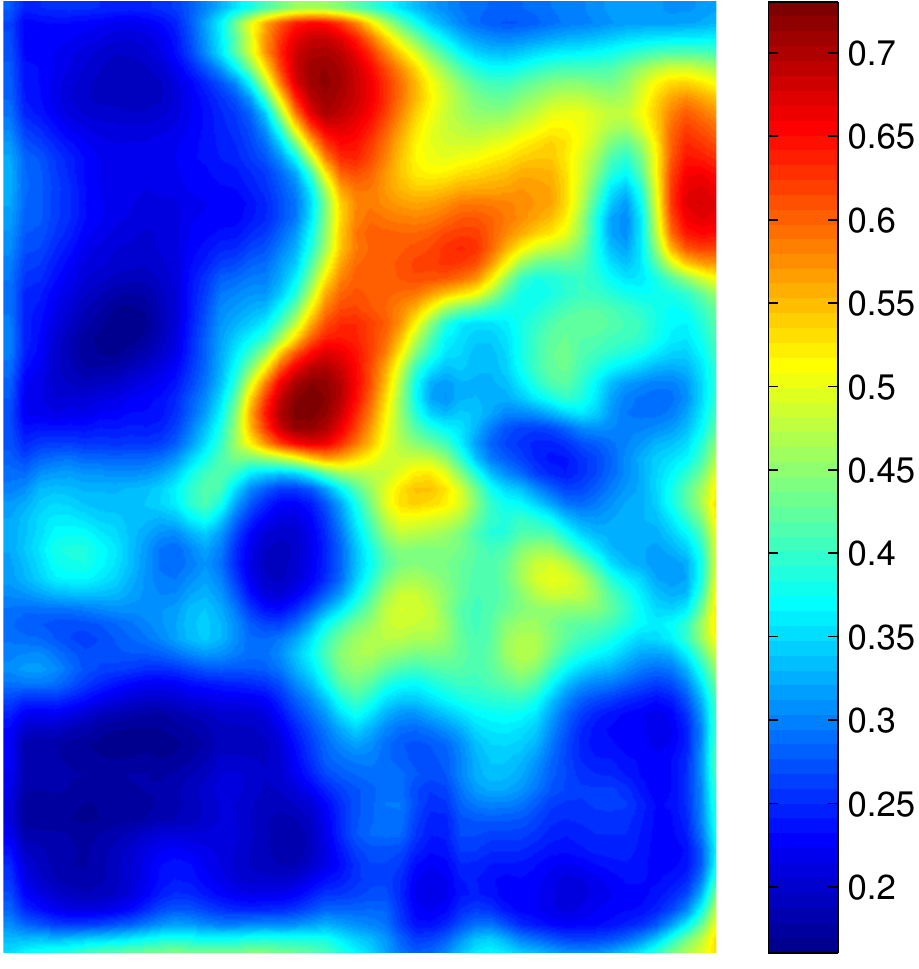} &
   \includegraphics[height=0.13\linewidth]{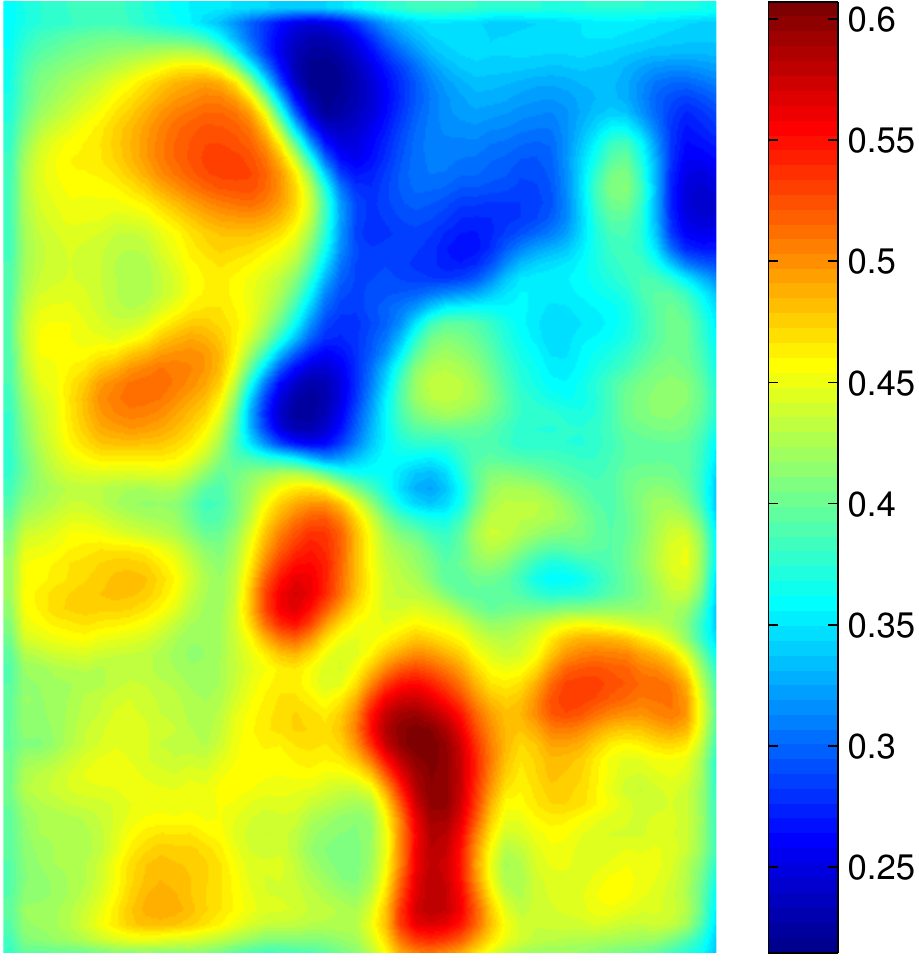} &
   \includegraphics[height=0.13\linewidth]{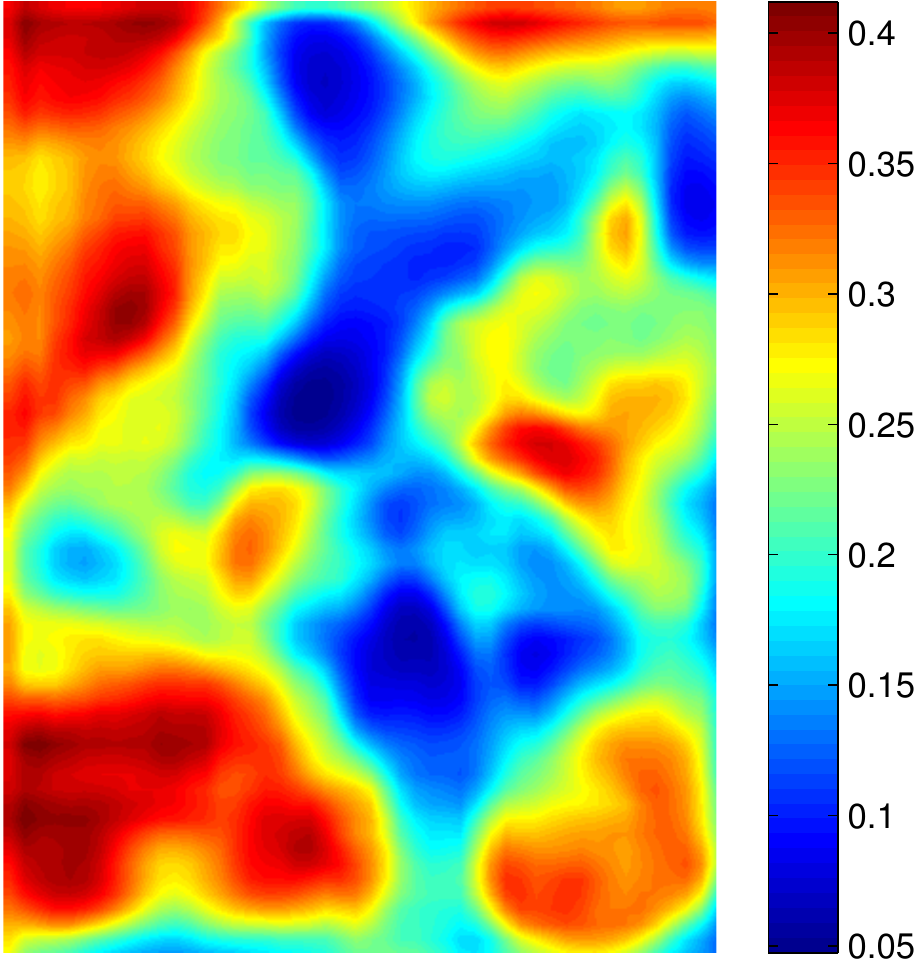} \\
   \includegraphics[height=0.09\linewidth]{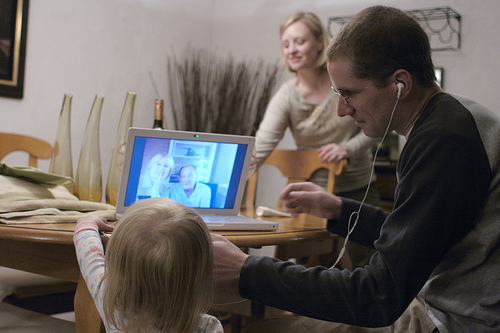} &
   \includegraphics[height=0.09\linewidth]{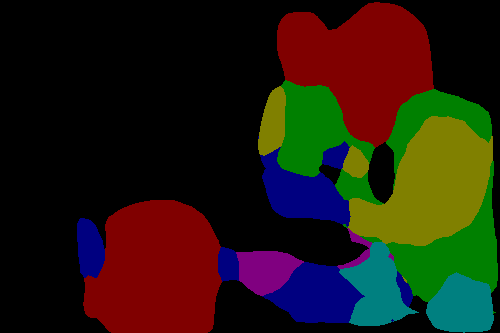} &
   \includegraphics[height=0.09\linewidth]{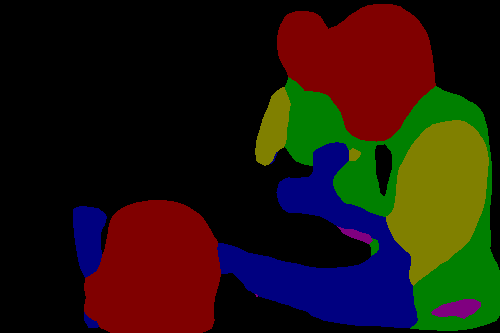} &
   \includegraphics[height=0.09\linewidth]{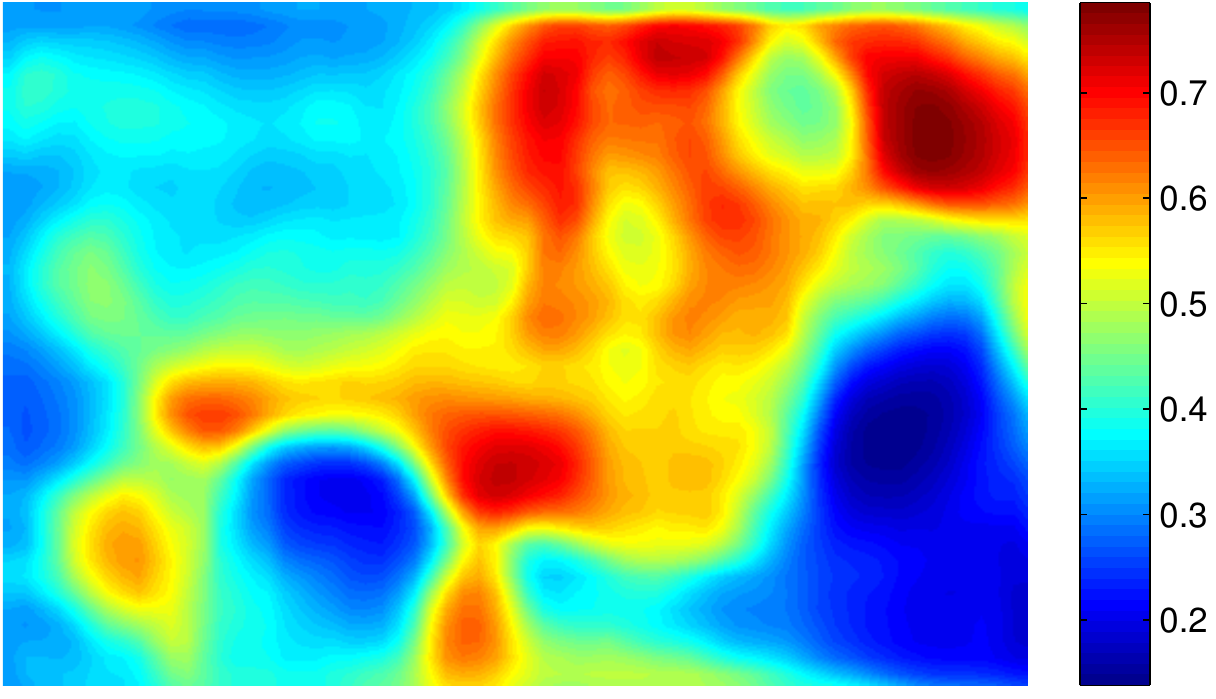} &
   \includegraphics[height=0.09\linewidth]{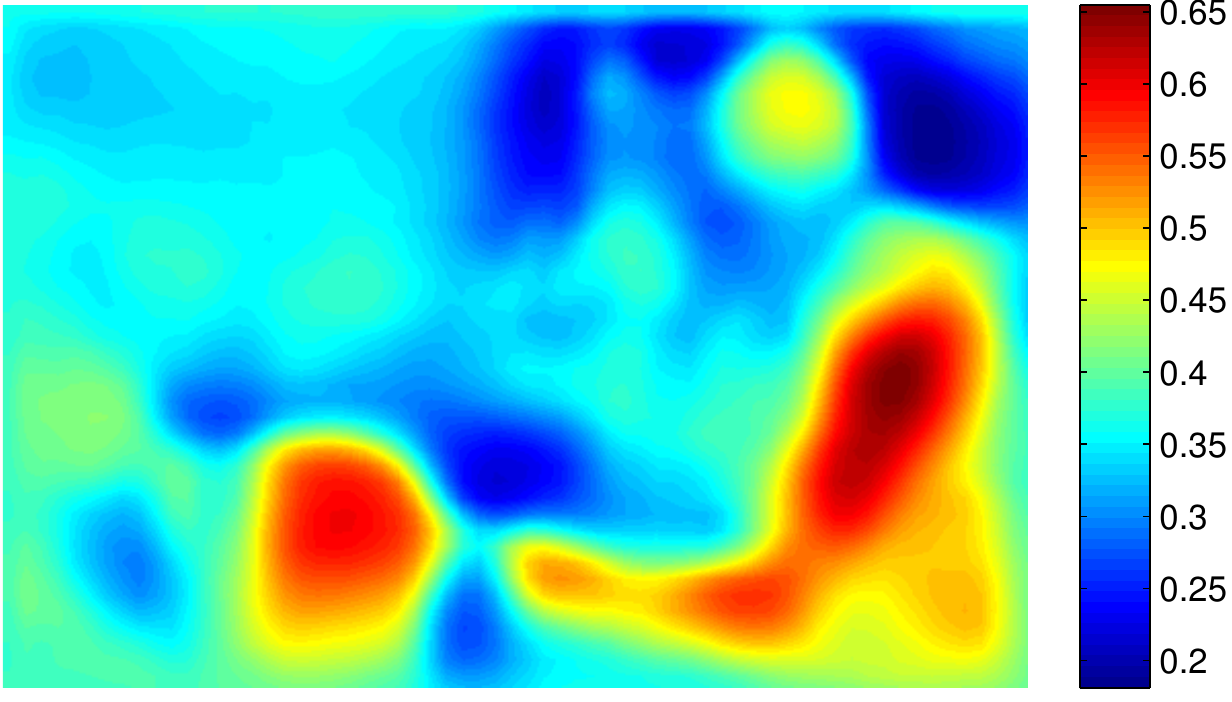} &
   \includegraphics[height=0.09\linewidth]{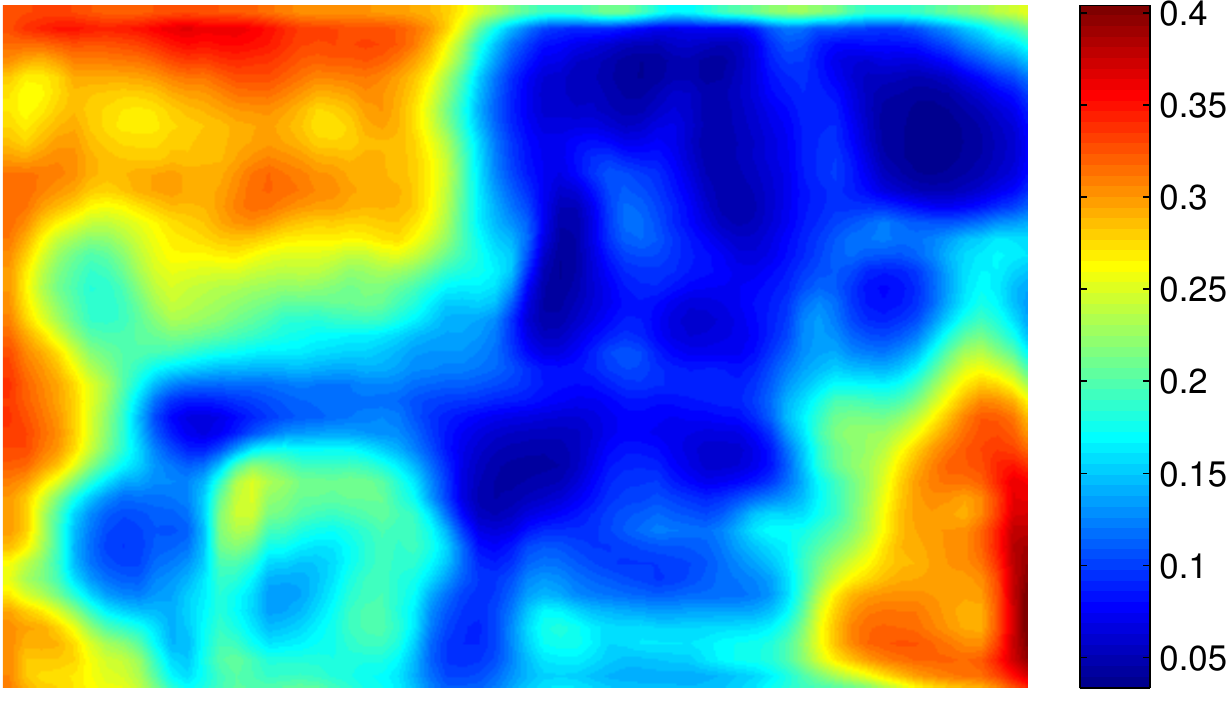} \\
   \includegraphics[height=0.13\linewidth]{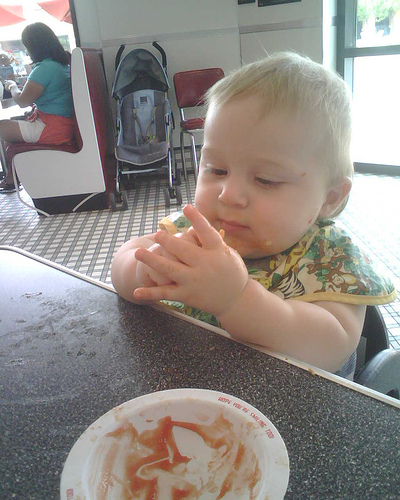} &
   \includegraphics[height=0.13\linewidth]{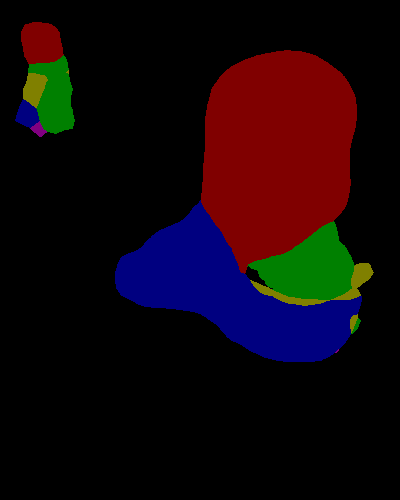} &
   \includegraphics[height=0.13\linewidth]{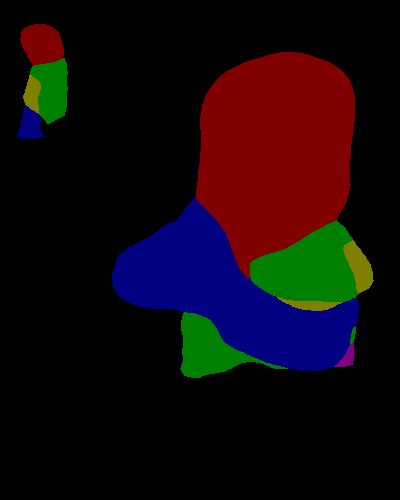} &
   \includegraphics[height=0.13\linewidth]{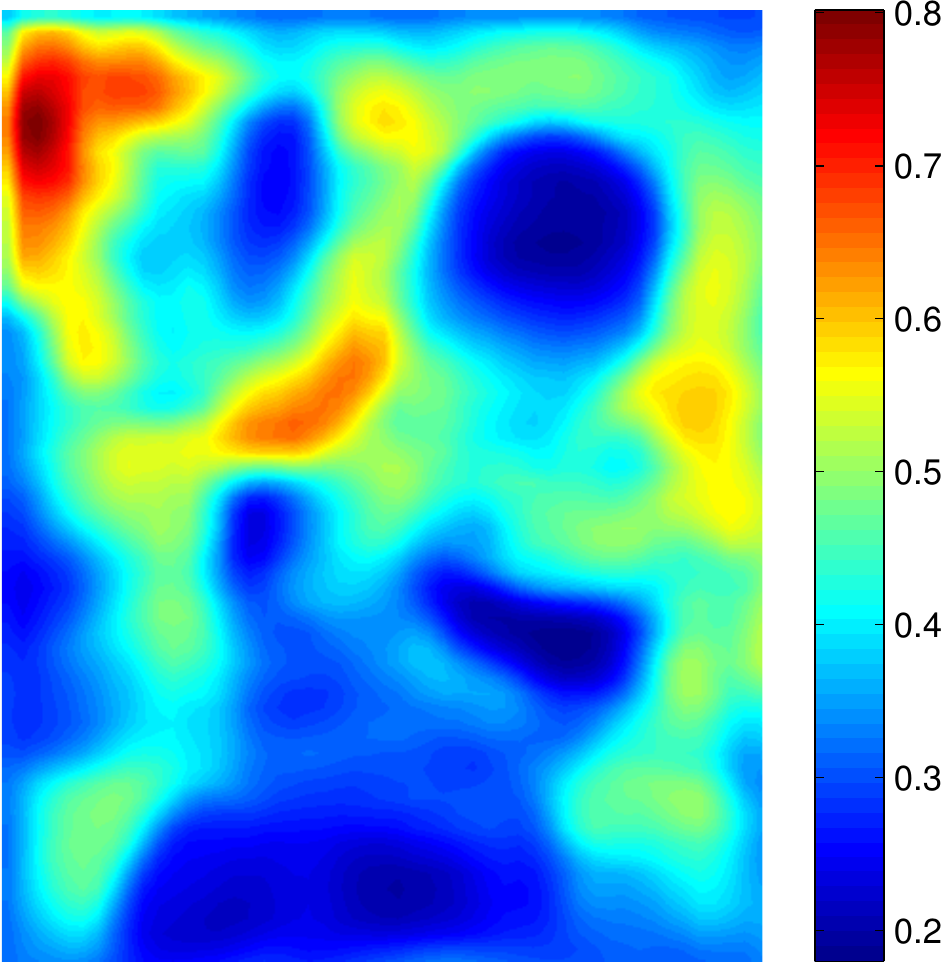} &
   \includegraphics[height=0.13\linewidth]{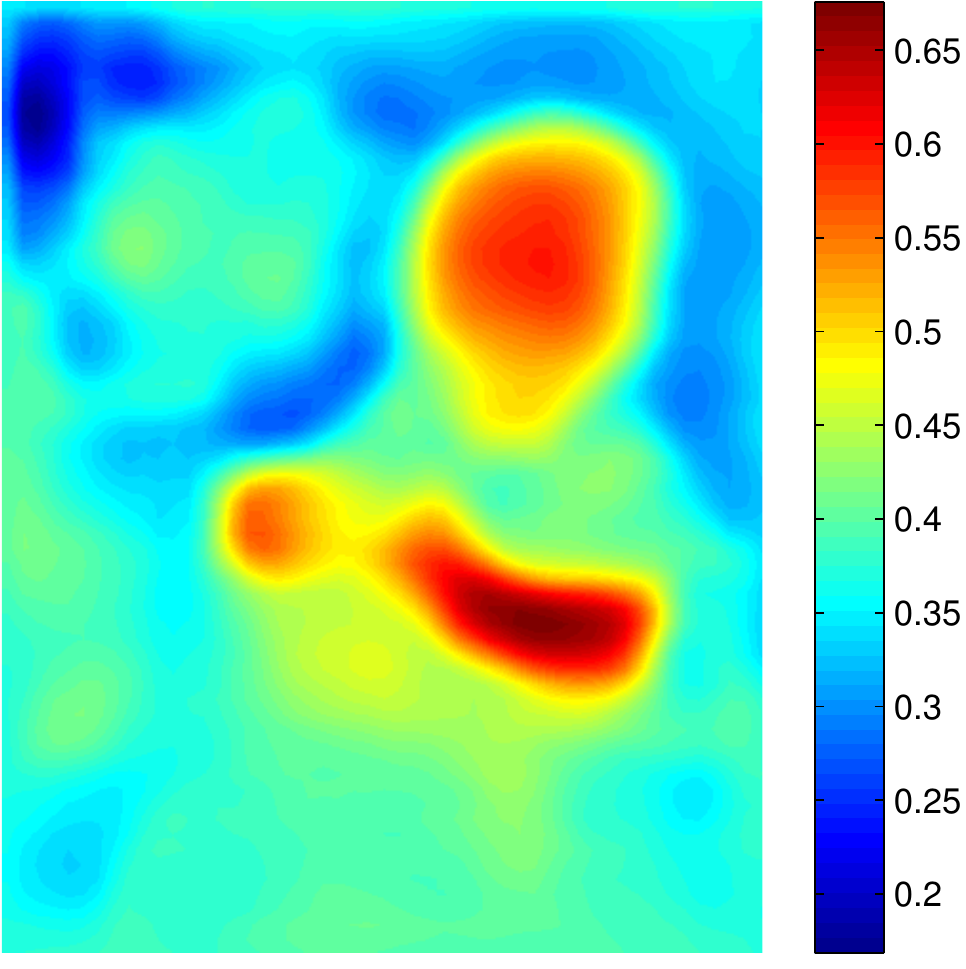} &
   \includegraphics[height=0.13\linewidth]{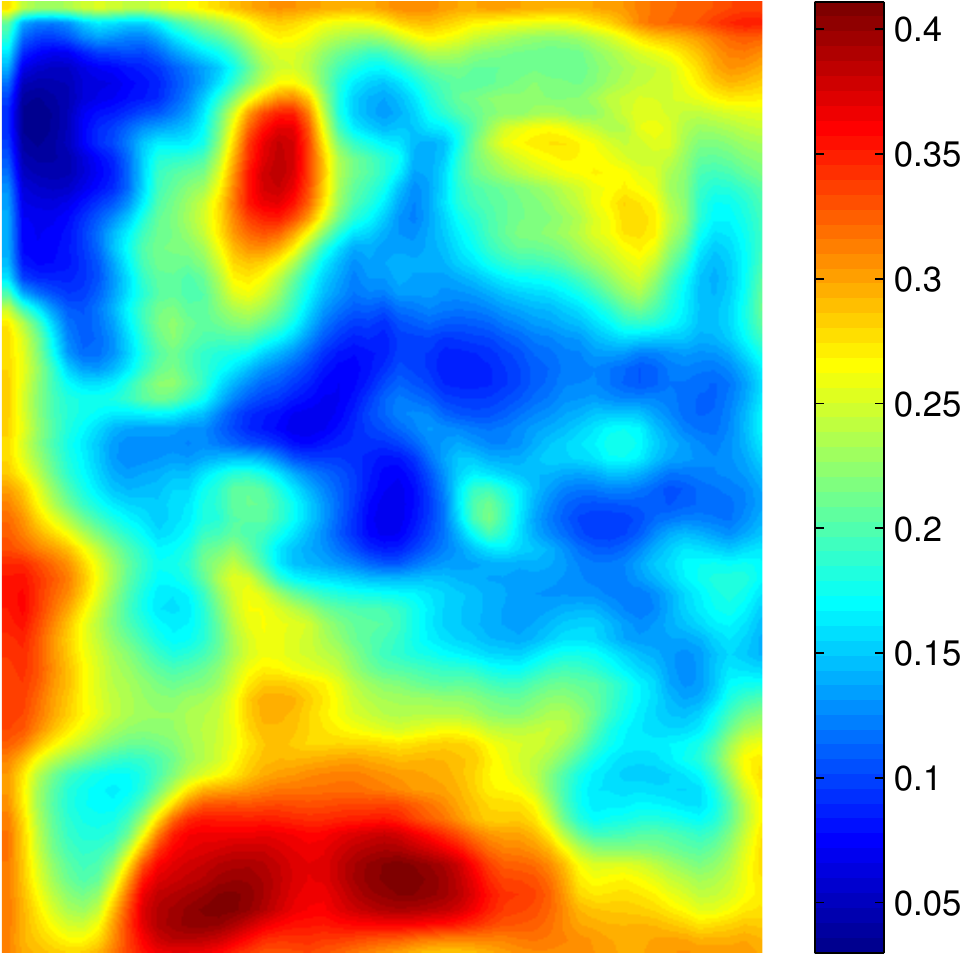} \\
   \includegraphics[height=0.13\linewidth]{fig/voc10_part/img/2008_003344.jpg} &
   \includegraphics[height=0.13\linewidth]{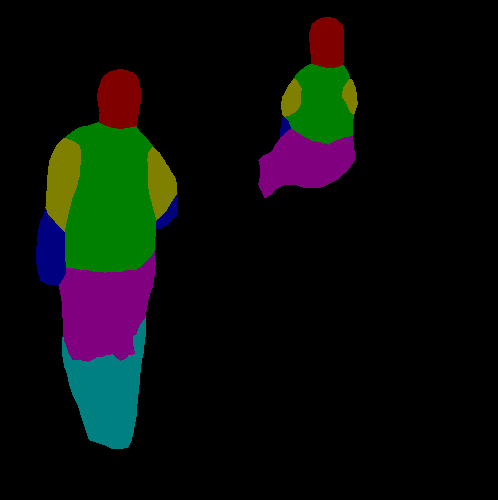} &
   \includegraphics[height=0.13\linewidth]{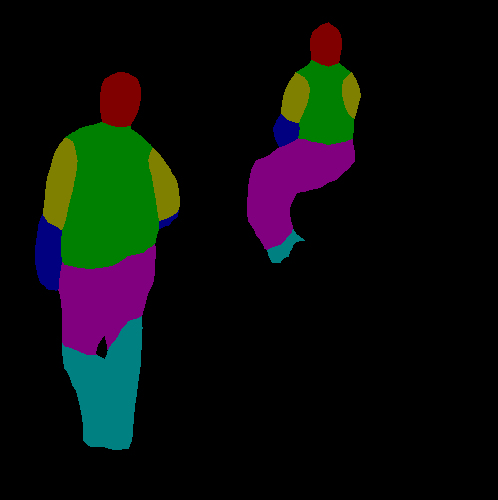} &
   \includegraphics[height=0.13\linewidth]{fig/voc10_part/att1/2008_003344.pdf} &
   \includegraphics[height=0.13\linewidth]{fig/voc10_part/att2/2008_003344.pdf} &
   \includegraphics[height=0.13\linewidth]{fig/voc10_part/att3/2008_003344.pdf} \\
   \includegraphics[height=0.11\linewidth]{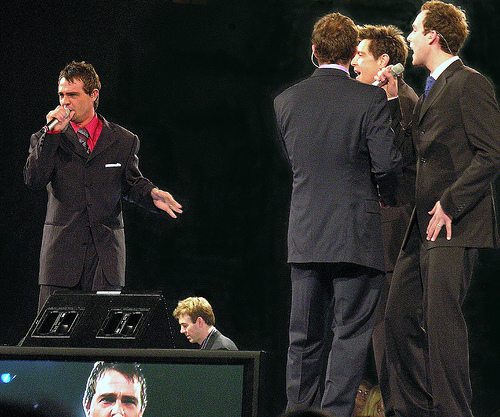} &
   \includegraphics[height=0.11\linewidth]{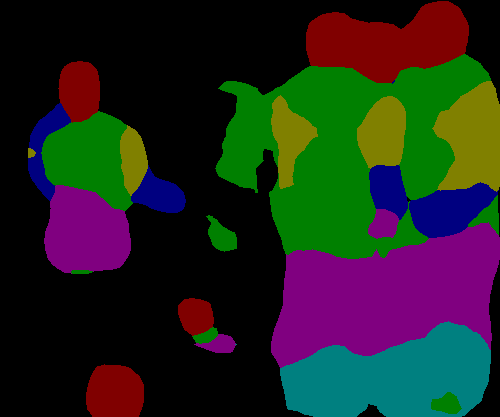} &
   \includegraphics[height=0.11\linewidth]{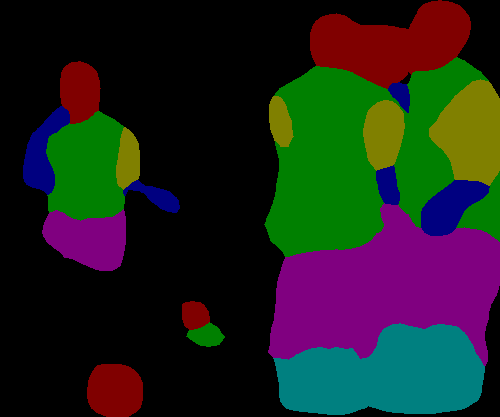} &
   \includegraphics[height=0.11\linewidth]{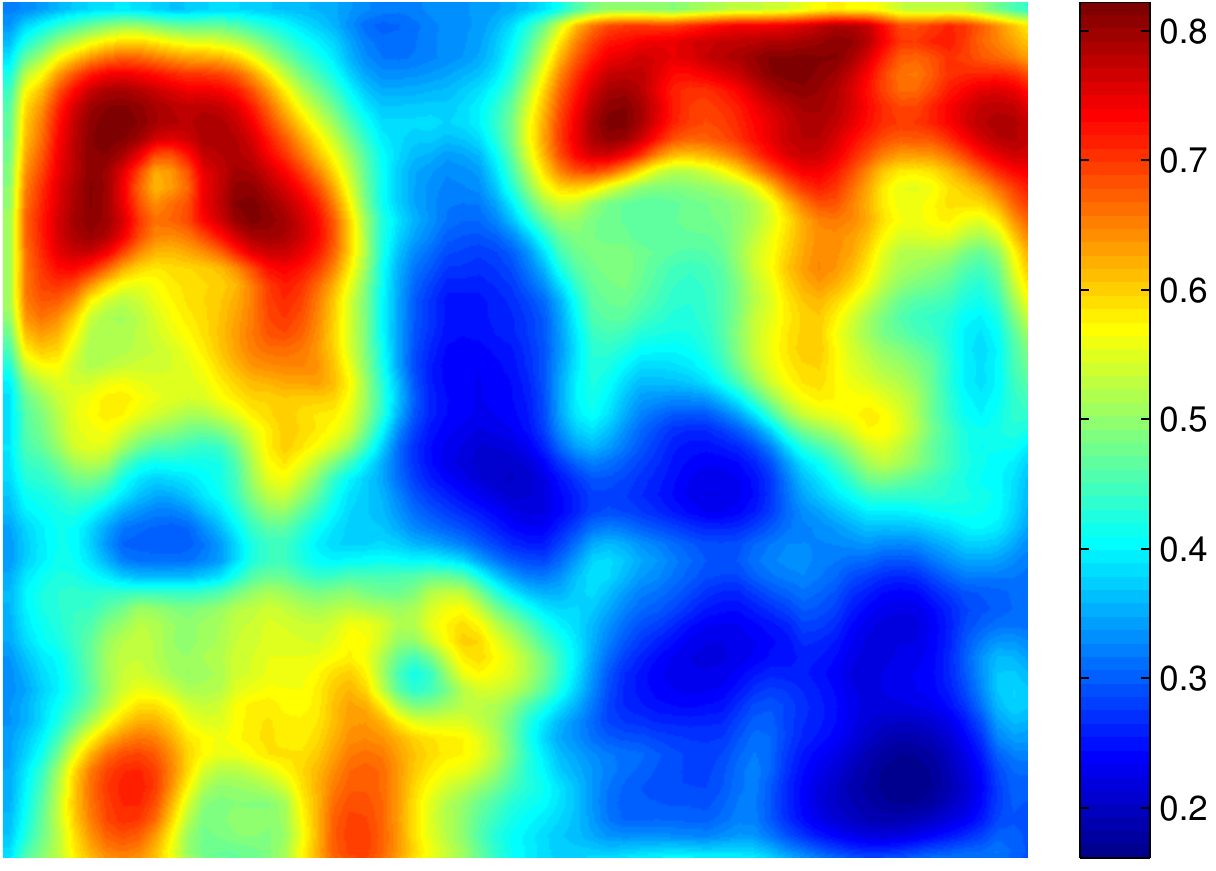} &
   \includegraphics[height=0.11\linewidth]{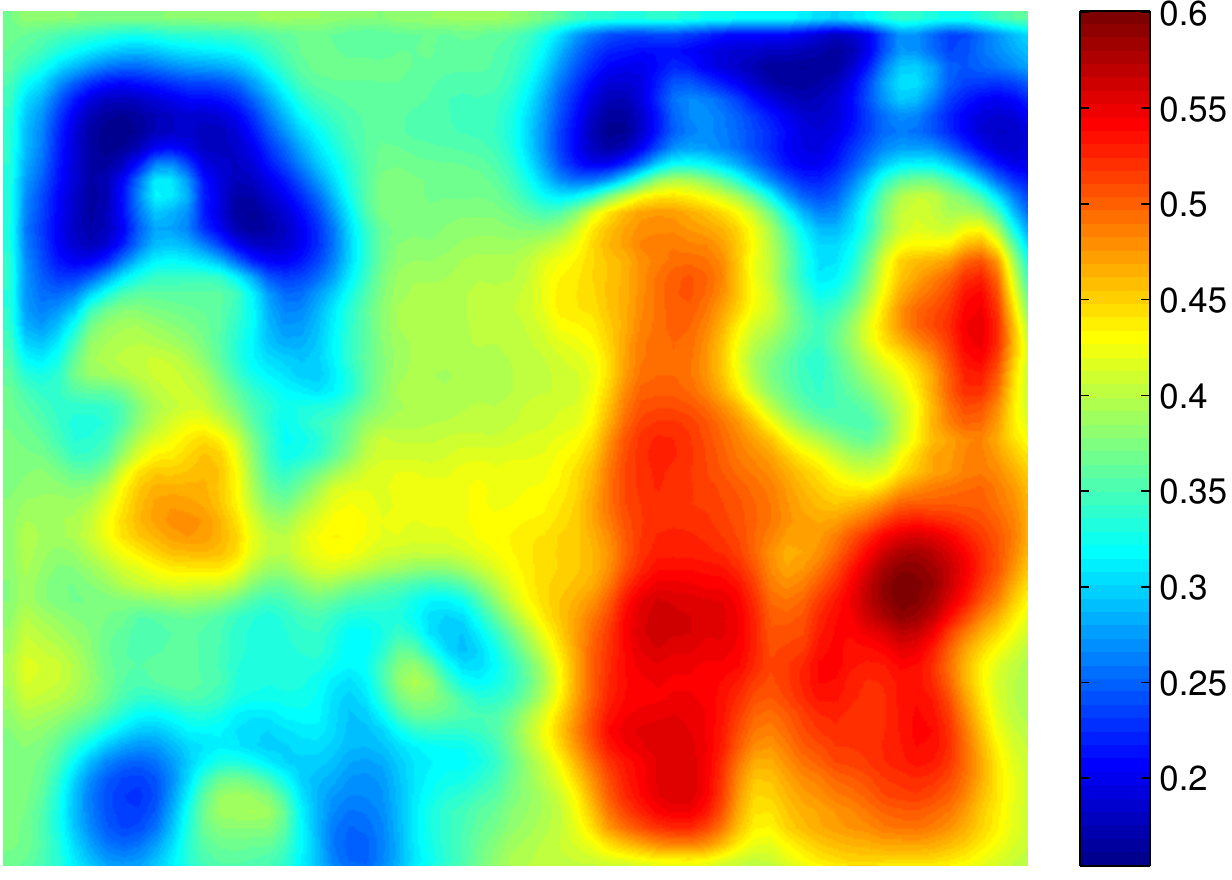} &
   \includegraphics[height=0.11\linewidth]{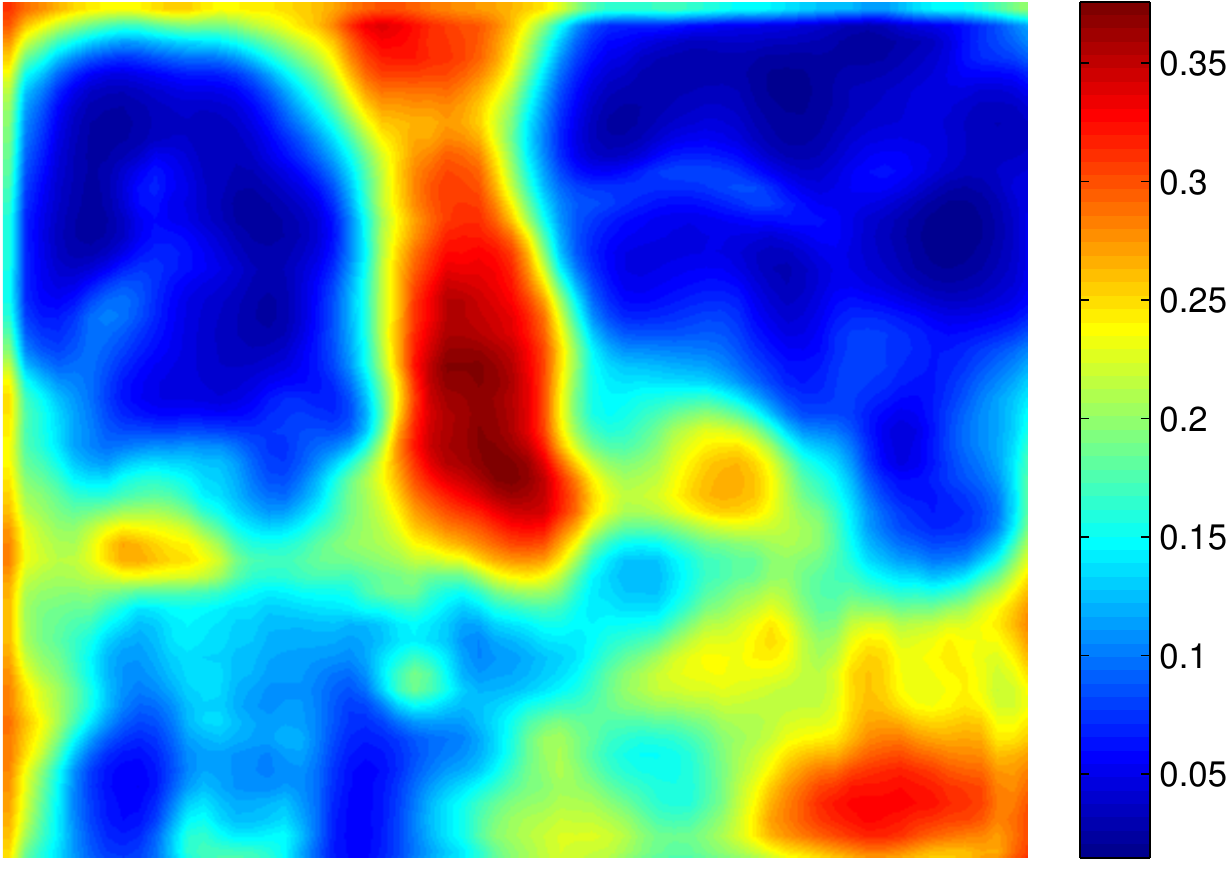} \\
   \includegraphics[height=0.1\linewidth]{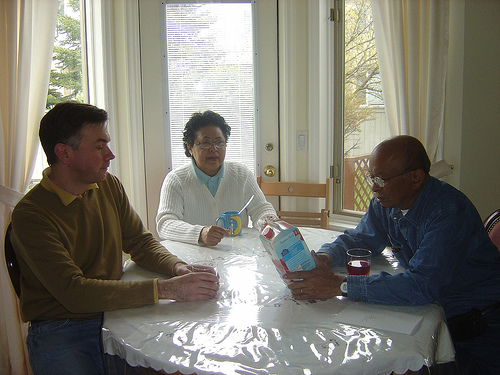} &
   \includegraphics[height=0.1\linewidth]{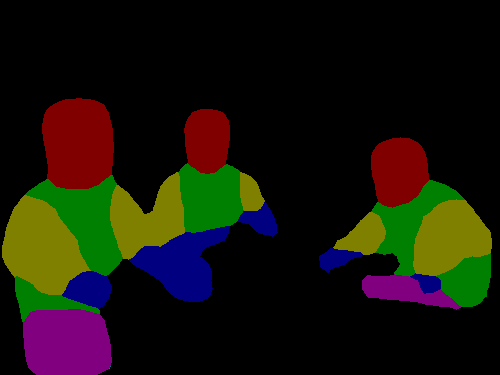} &
   \includegraphics[height=0.1\linewidth]{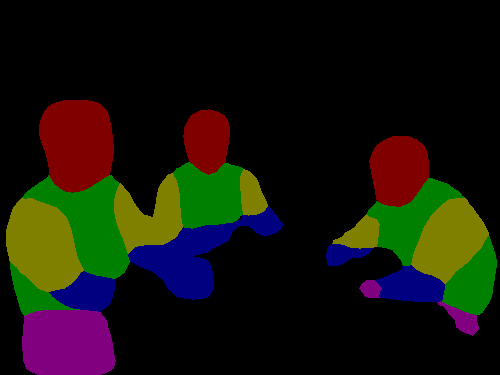} &
   \includegraphics[height=0.1\linewidth]{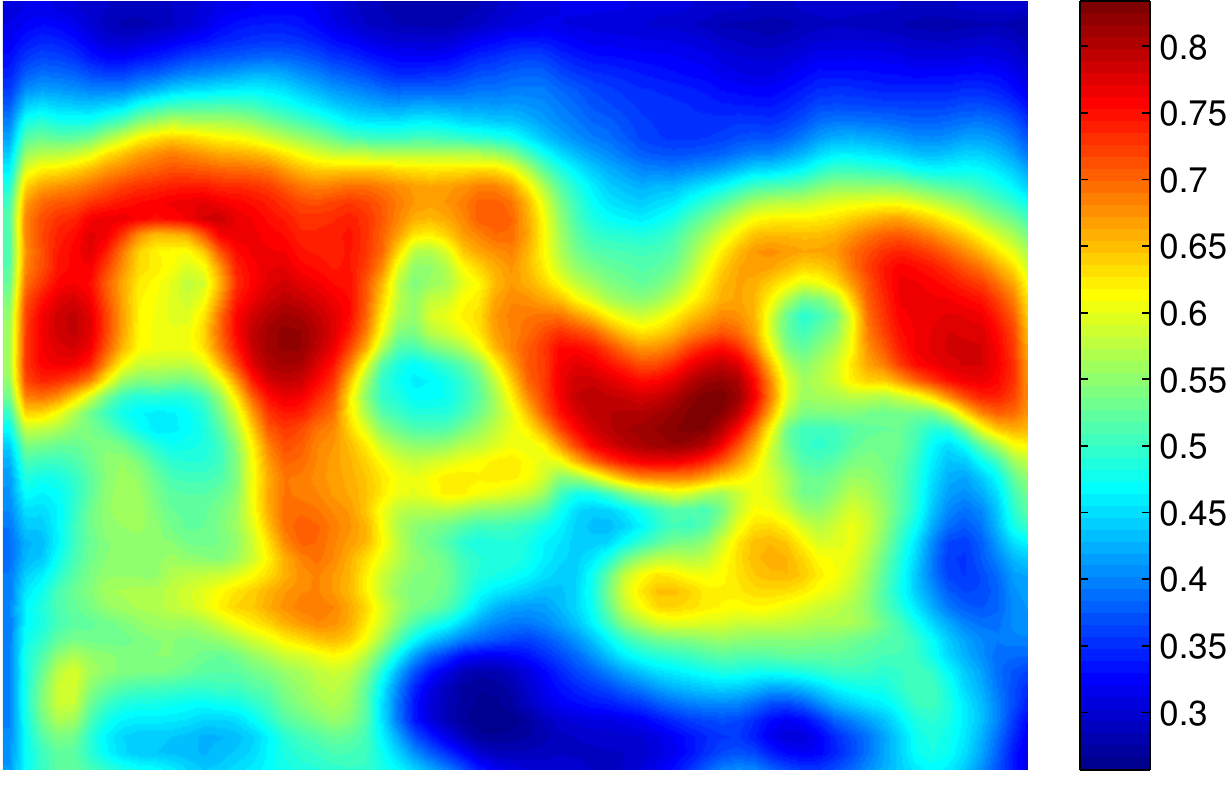} &
   \includegraphics[height=0.1\linewidth]{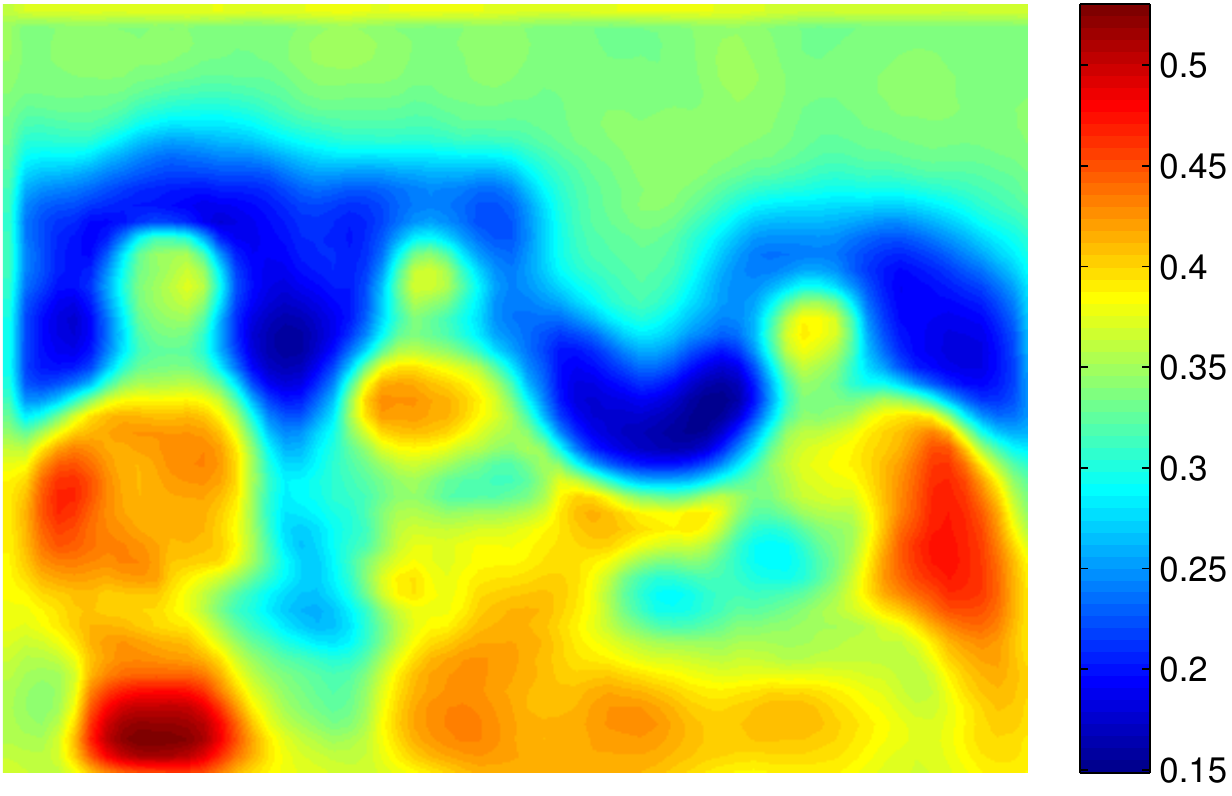} &
   \includegraphics[height=0.1\linewidth]{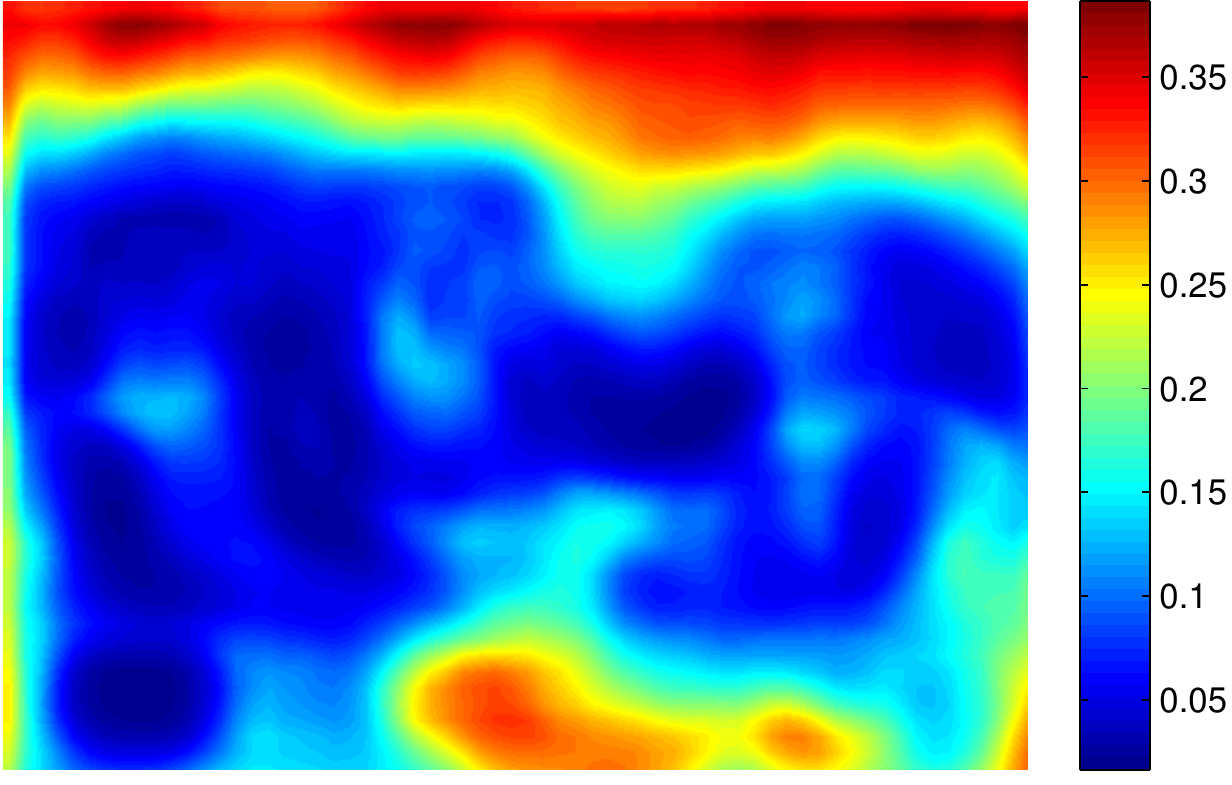} \\
   \includegraphics[height=0.1\linewidth]{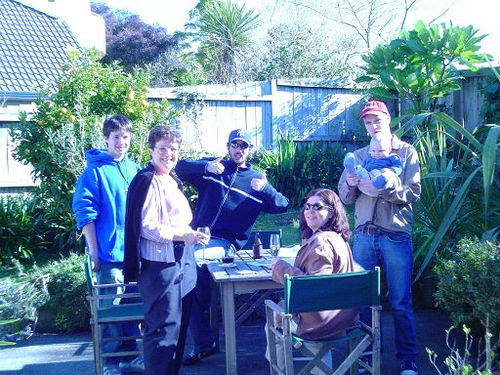} &
   \includegraphics[height=0.1\linewidth]{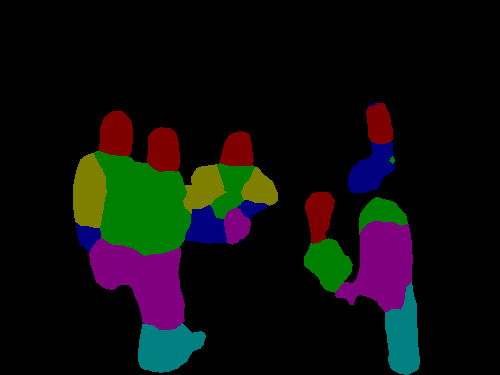} &
   \includegraphics[height=0.1\linewidth]{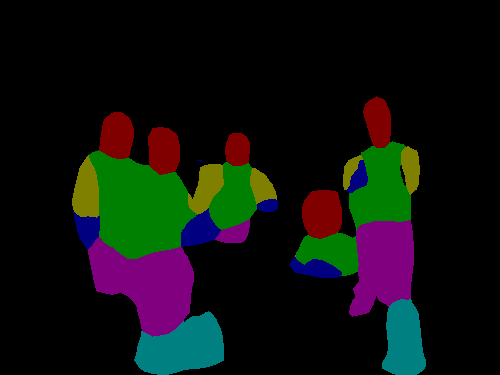} &
   \includegraphics[height=0.1\linewidth]{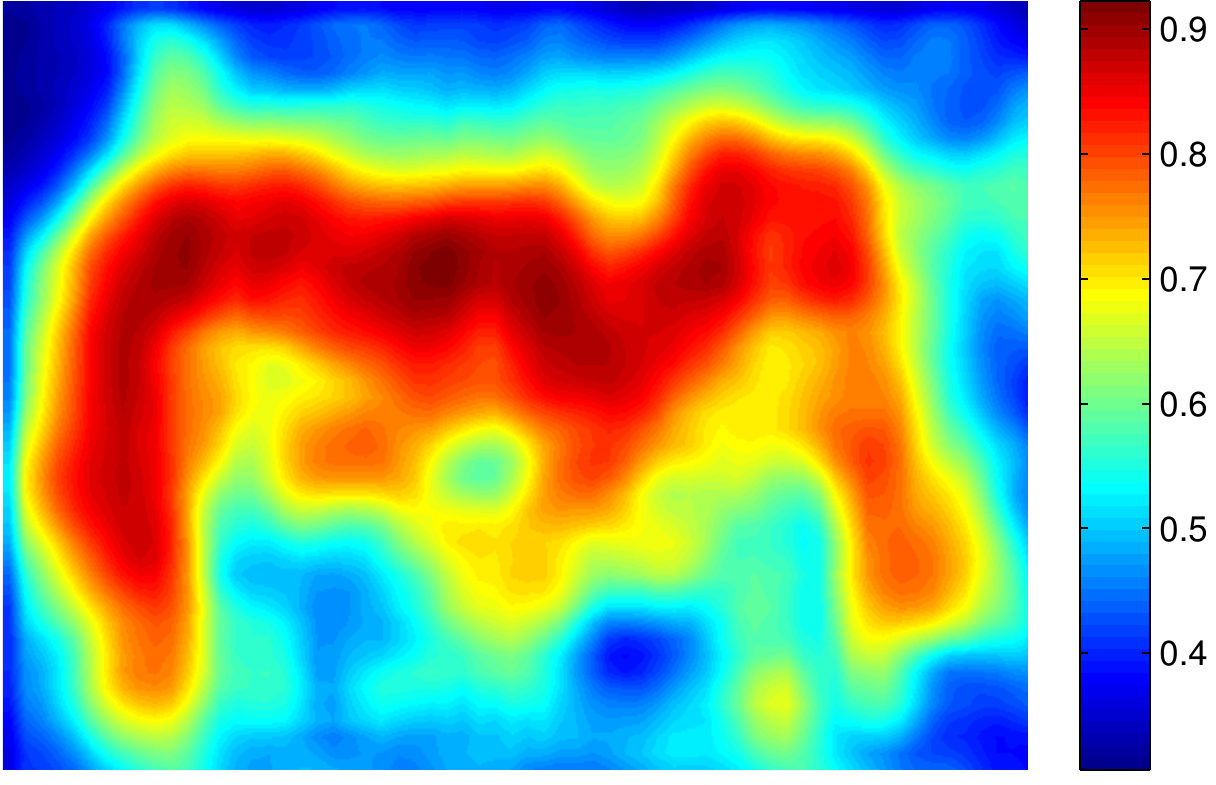} &
   \includegraphics[height=0.1\linewidth]{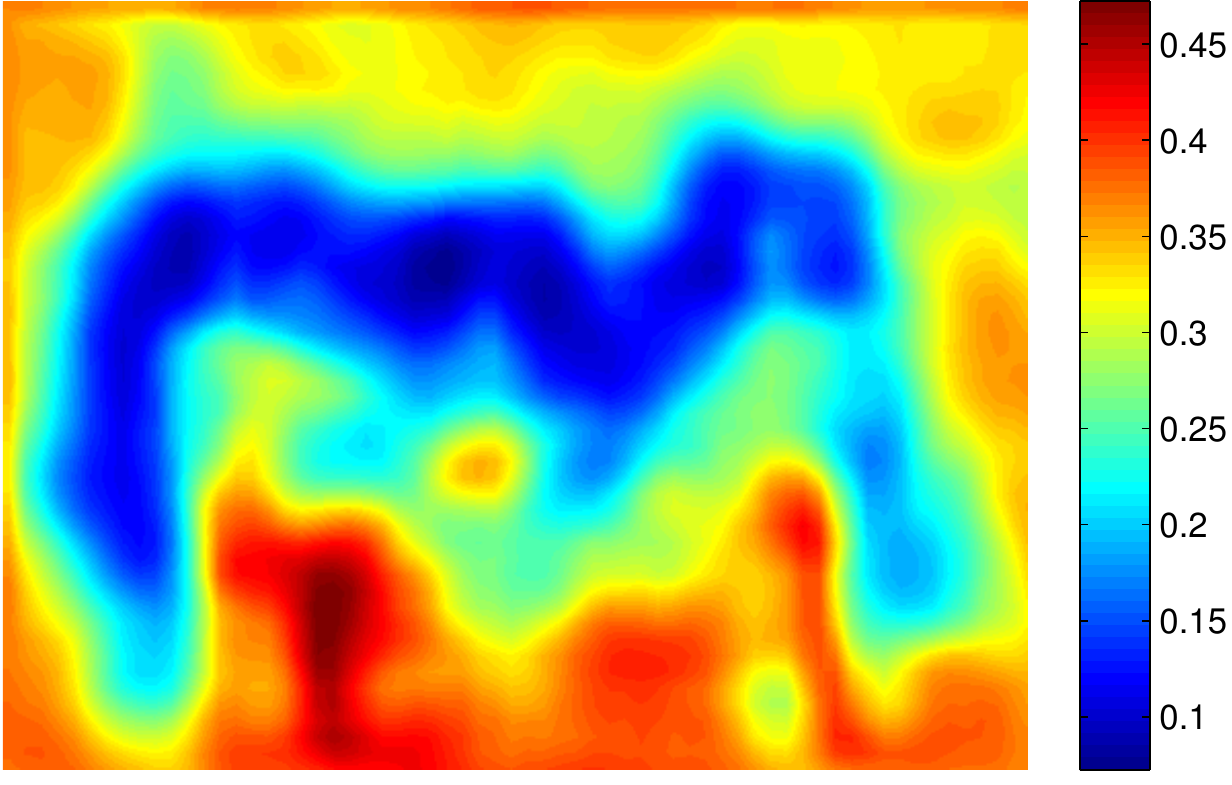} &
   \includegraphics[height=0.1\linewidth]{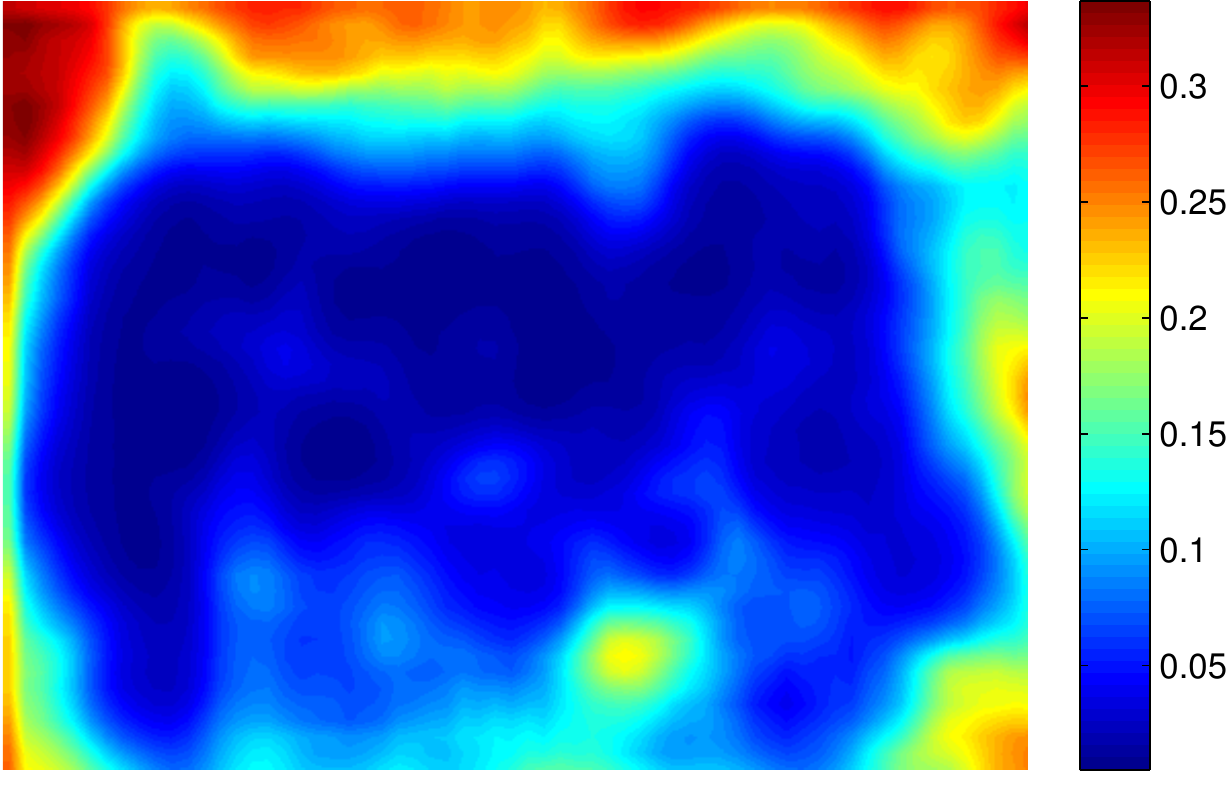} \\
   \includegraphics[height=0.1\linewidth]{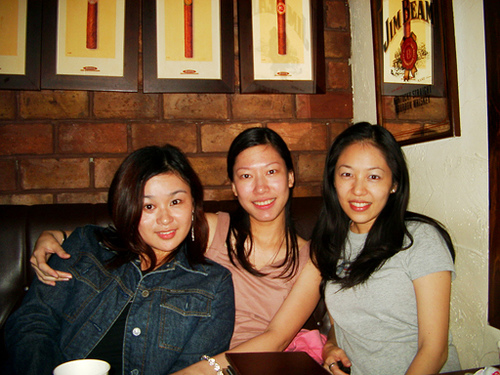} &
   \includegraphics[height=0.1\linewidth]{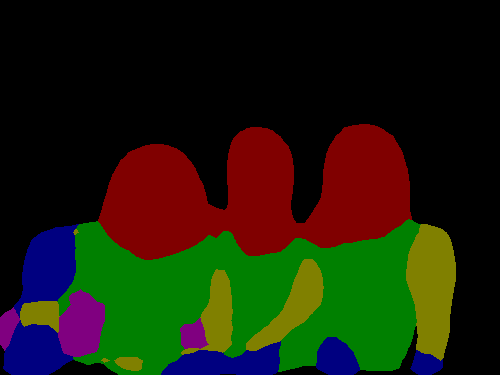} &
   \includegraphics[height=0.1\linewidth]{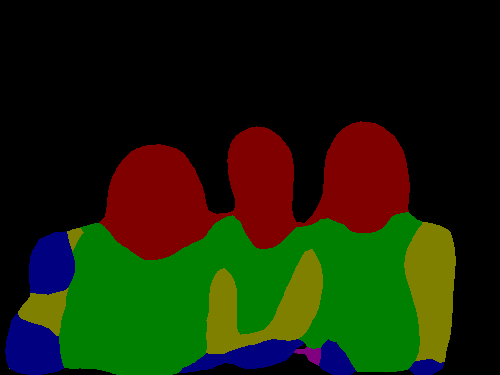} &
   \includegraphics[height=0.1\linewidth]{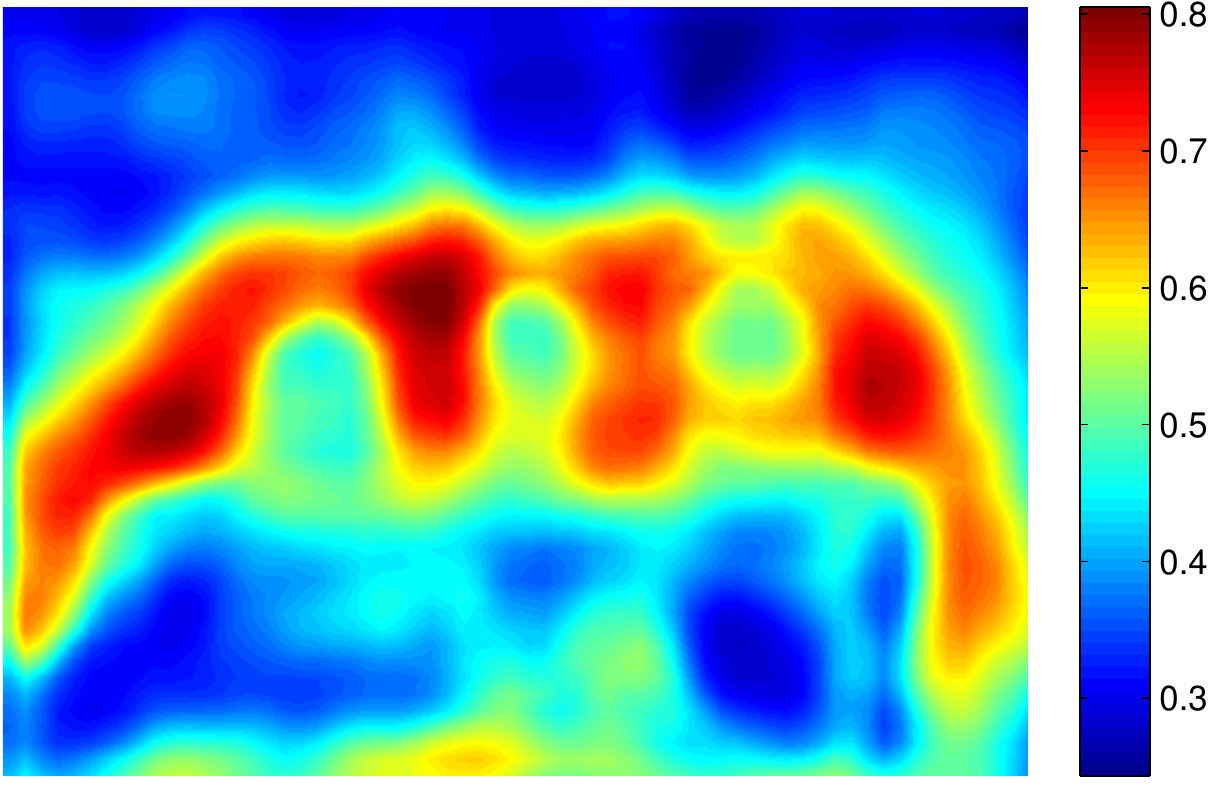} &
   \includegraphics[height=0.1\linewidth]{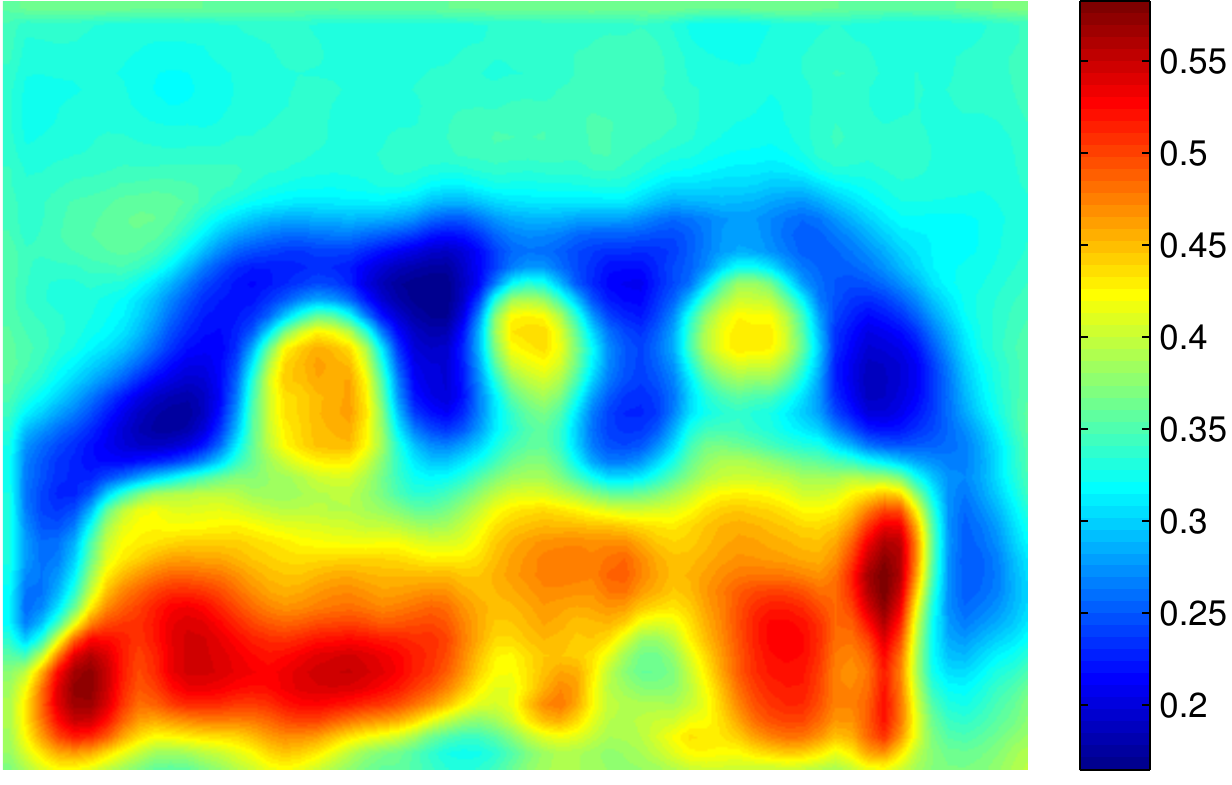} &
   \includegraphics[height=0.1\linewidth]{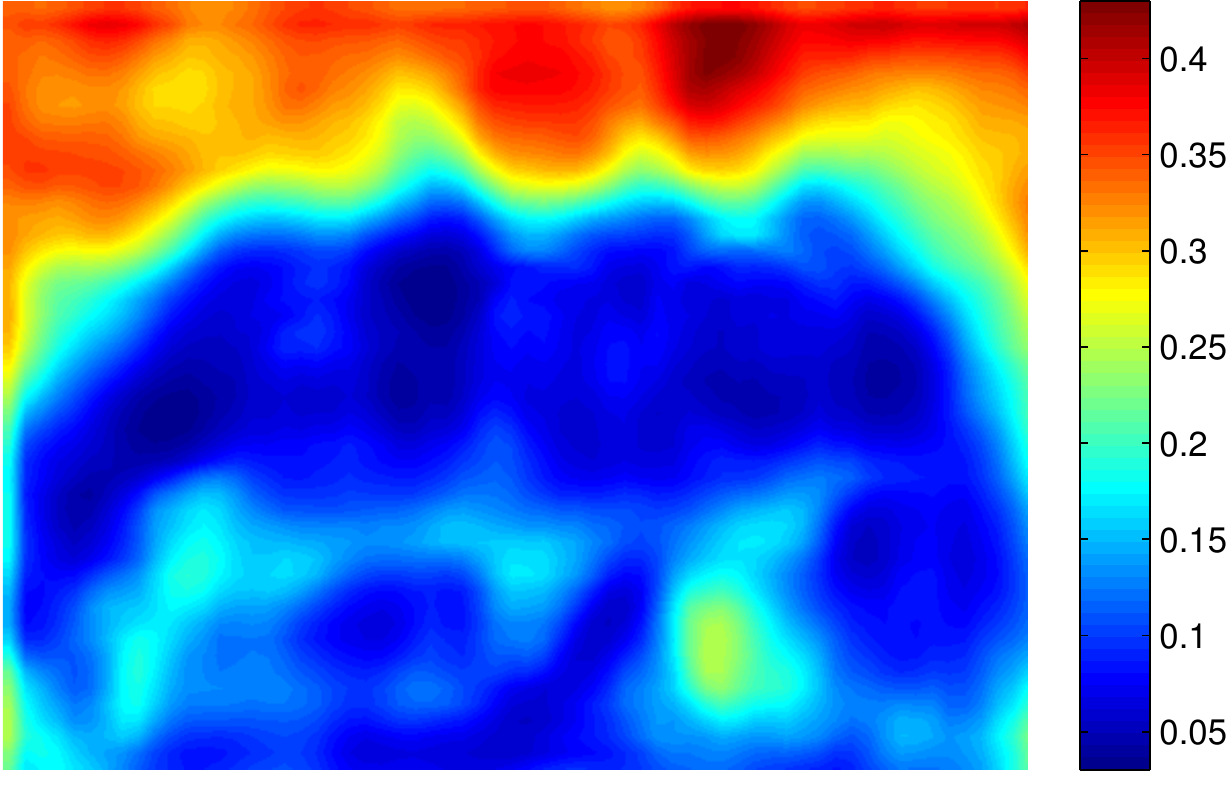} \\
   \includegraphics[height=0.087\linewidth]{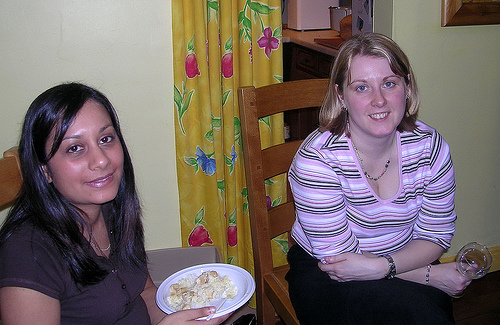} &
   \includegraphics[height=0.087\linewidth]{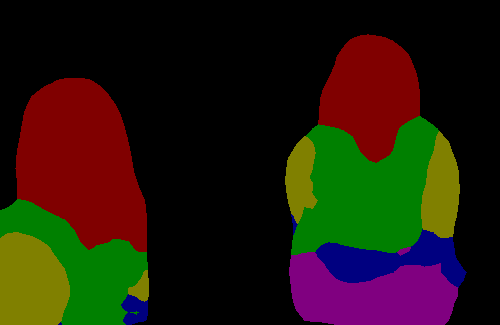} &
   \includegraphics[height=0.087\linewidth]{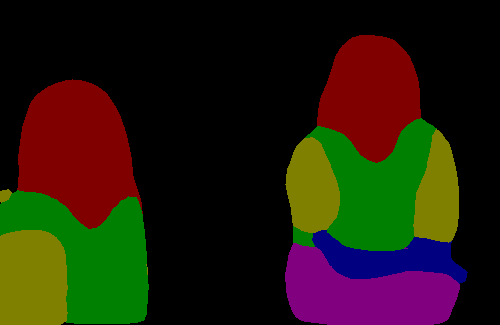} &
   \includegraphics[height=0.087\linewidth]{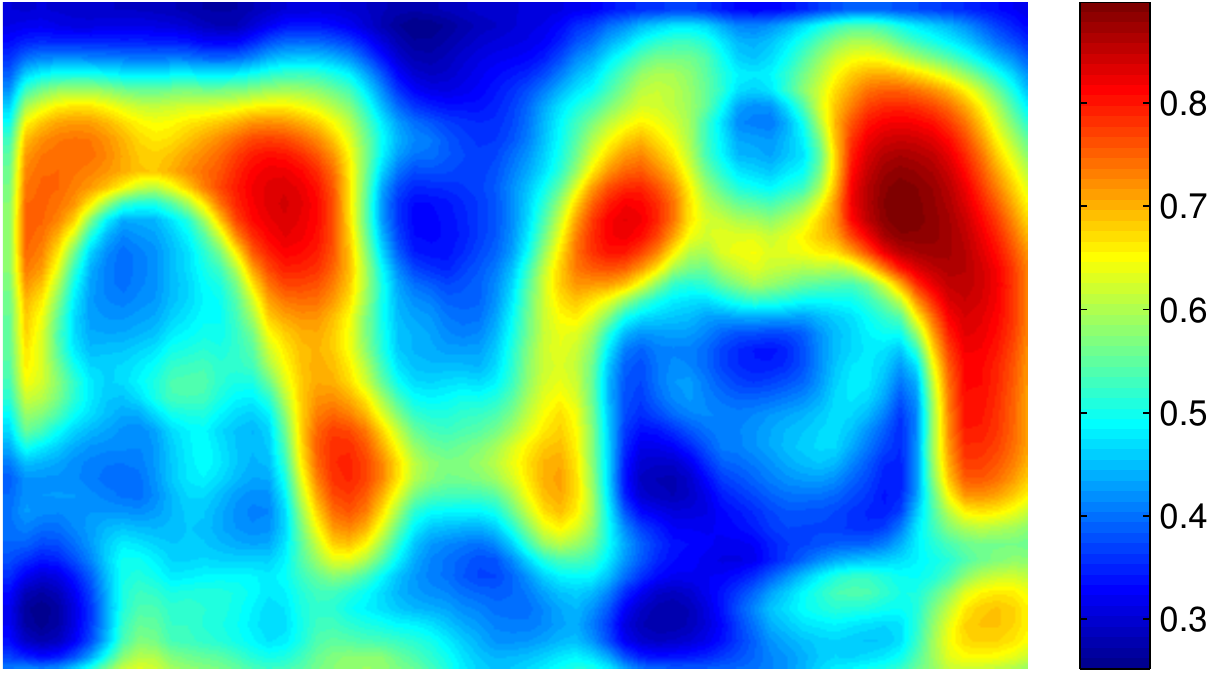} &
   \includegraphics[height=0.087\linewidth]{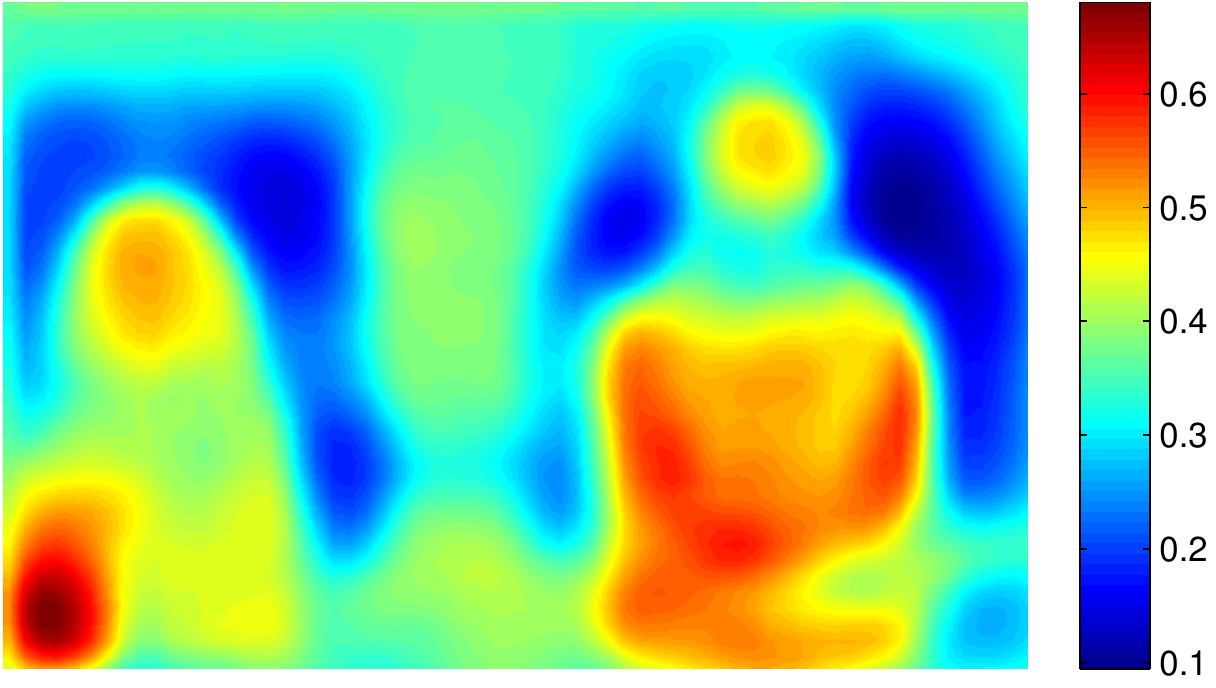} &
   \includegraphics[height=0.087\linewidth]{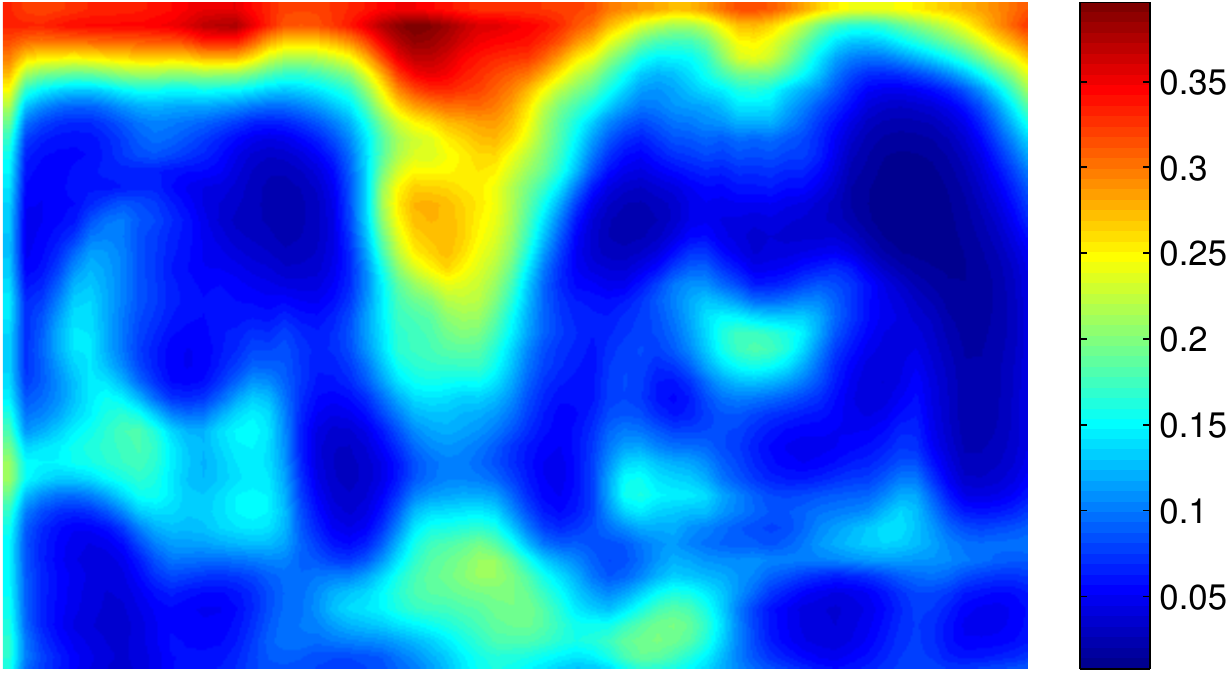} \\
   {\scriptsize (a) Image} & 
   {\scriptsize (b) Baseline} & 
   {\scriptsize (c) Our model} & 
   {\scriptsize (d) Scale-1 Attention} & 
   {\scriptsize (e) Scale-0.75 Attention} &
   {\scriptsize (f) Scale-0.5 Attention} \\
  \end{tabular}
  \caption{Qualitative segmentation results on PASCAL-Person-Part validation set.}
  \label{fig:pascal_part_results_app}  
\end{figure*}

\begin{figure*}
  \centering
  \begin{tabular}{c c c c c c}
   \includegraphics[height=0.09\linewidth]{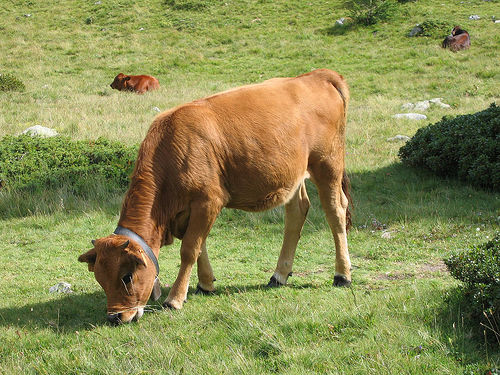} &
   \includegraphics[height=0.09\linewidth]{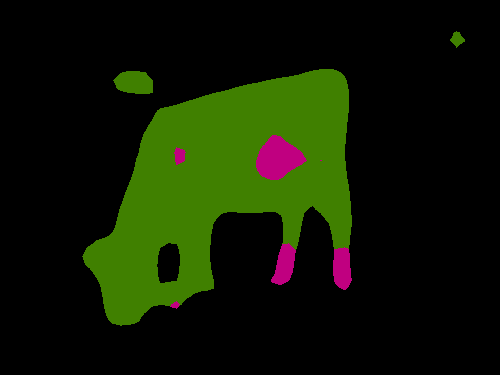} &
   \includegraphics[height=0.09\linewidth]{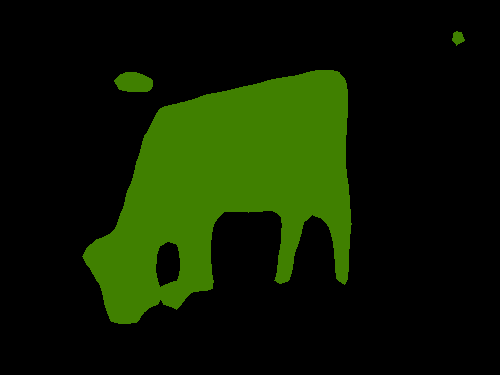} &
   \includegraphics[height=0.09\linewidth]{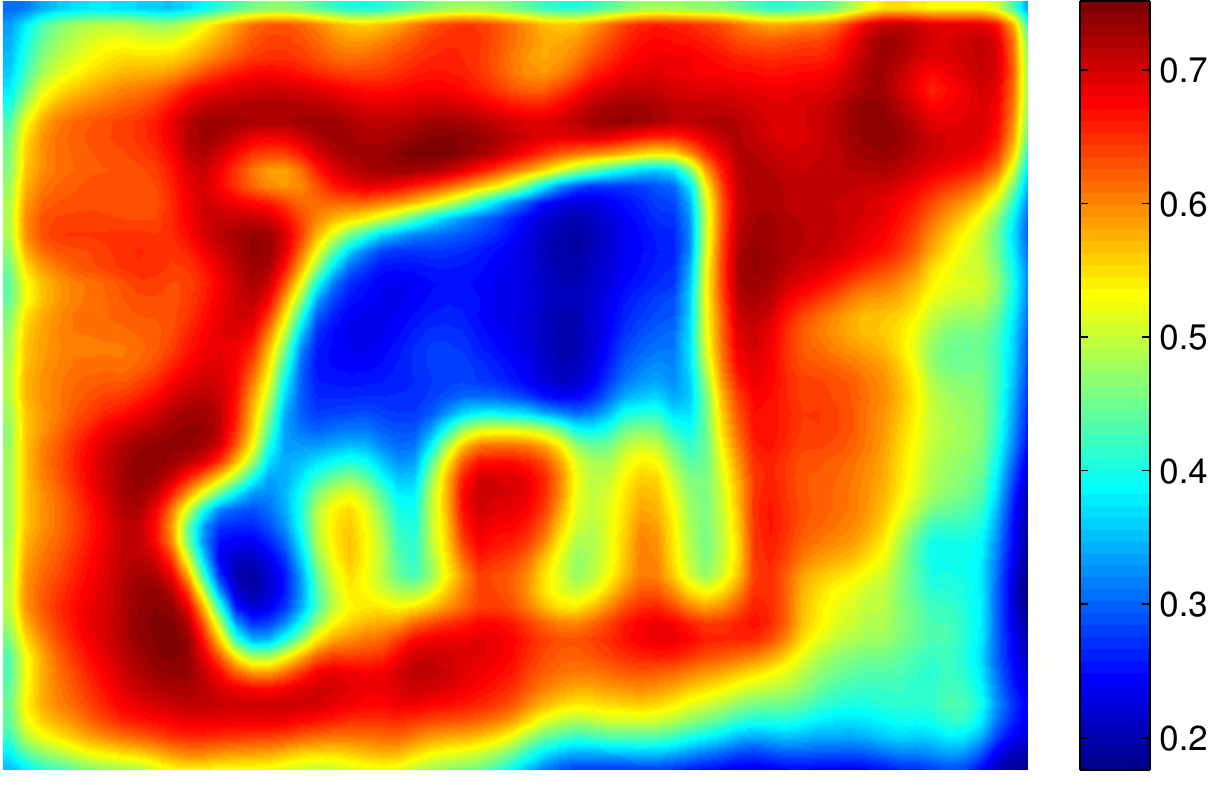} &
   \includegraphics[height=0.09\linewidth]{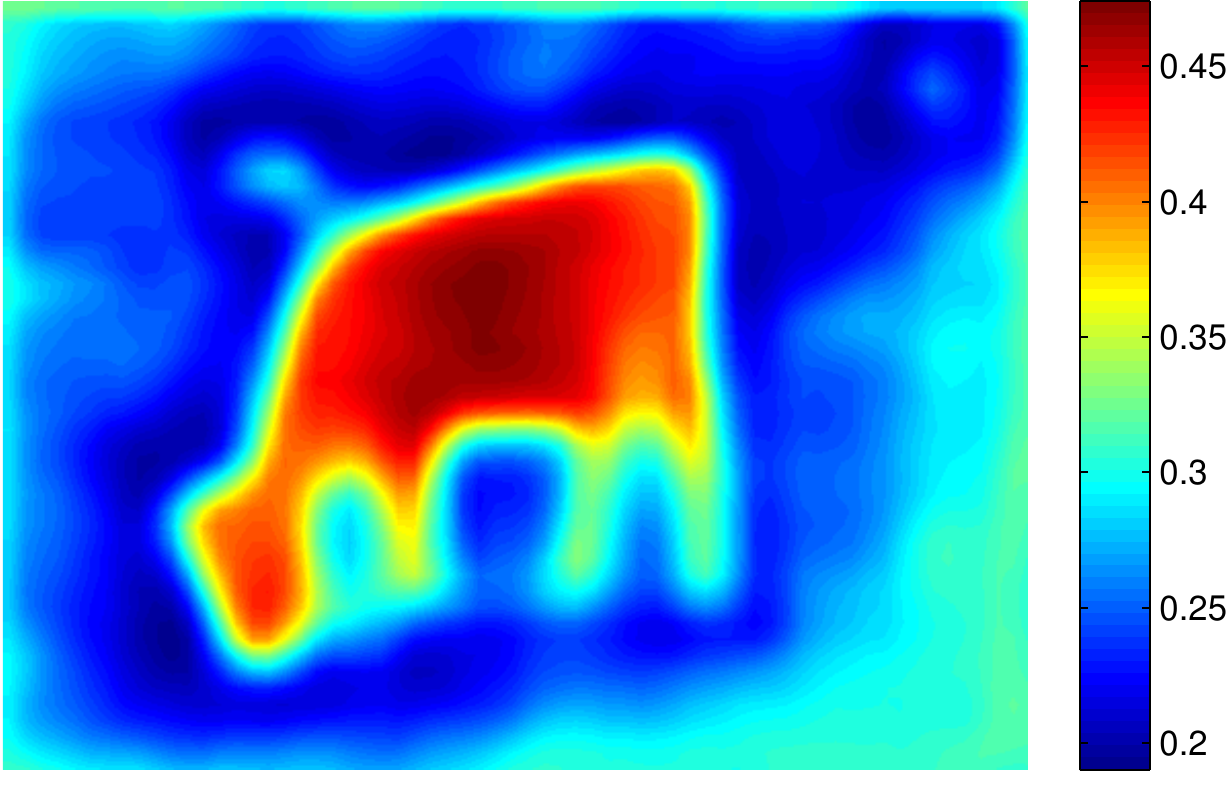} &
   \includegraphics[height=0.09\linewidth]{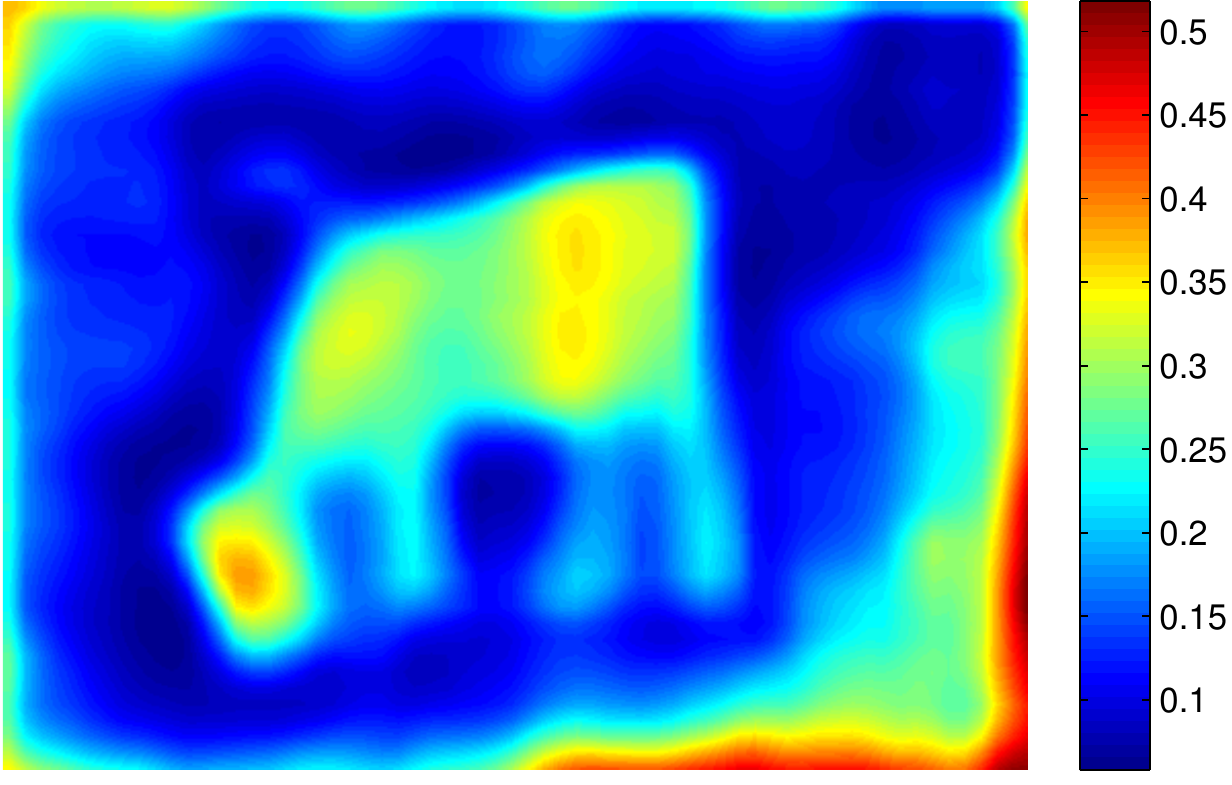} \\
   \includegraphics[height=0.08\linewidth]{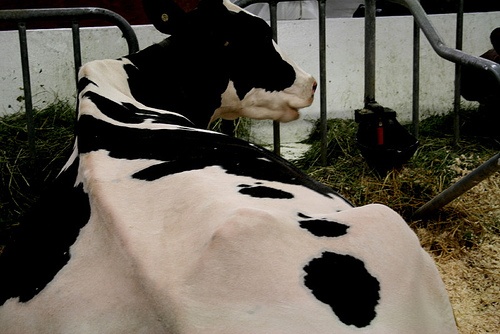} &
   \includegraphics[height=0.08\linewidth]{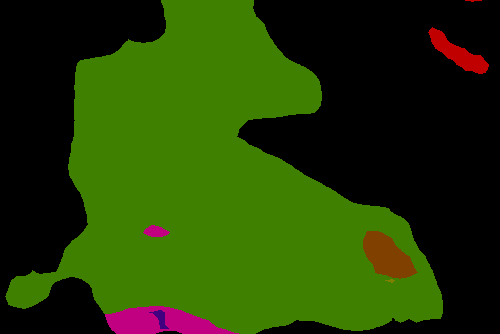} &
   \includegraphics[height=0.08\linewidth]{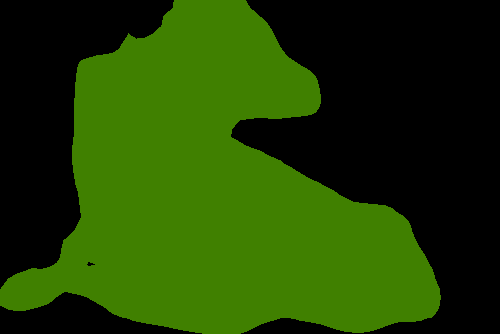} &
   \includegraphics[height=0.08\linewidth]{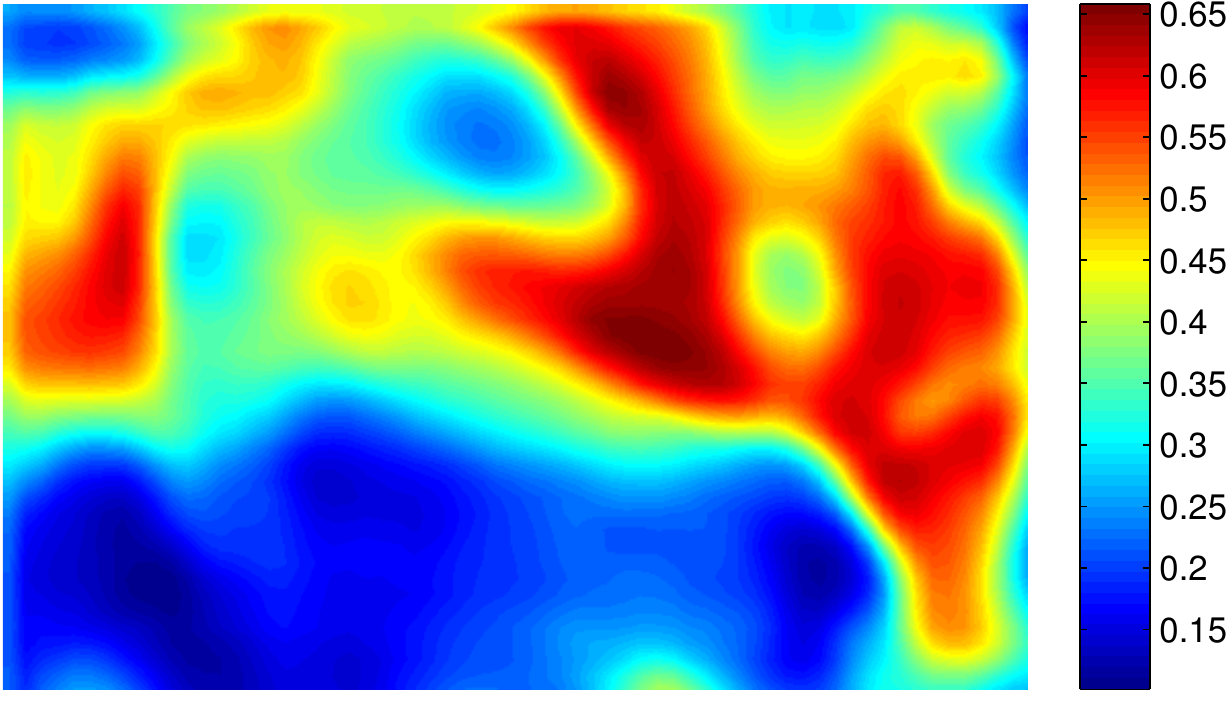} &
   \includegraphics[height=0.08\linewidth]{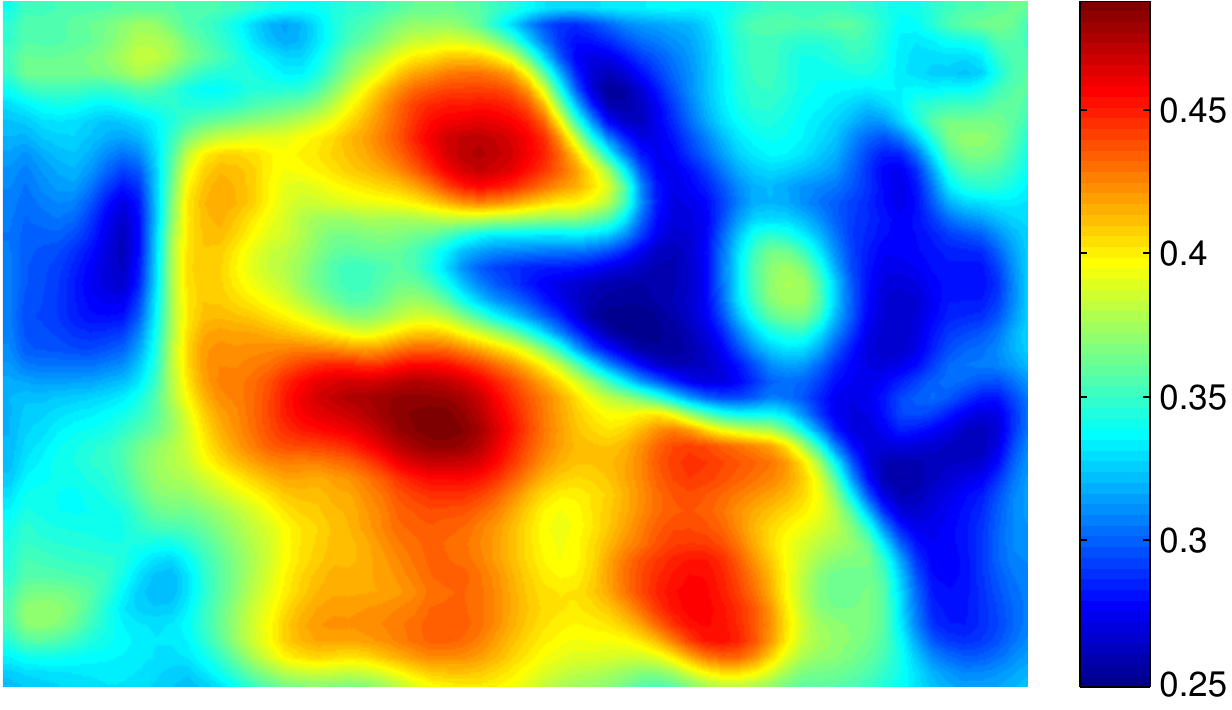} &
   \includegraphics[height=0.08\linewidth]{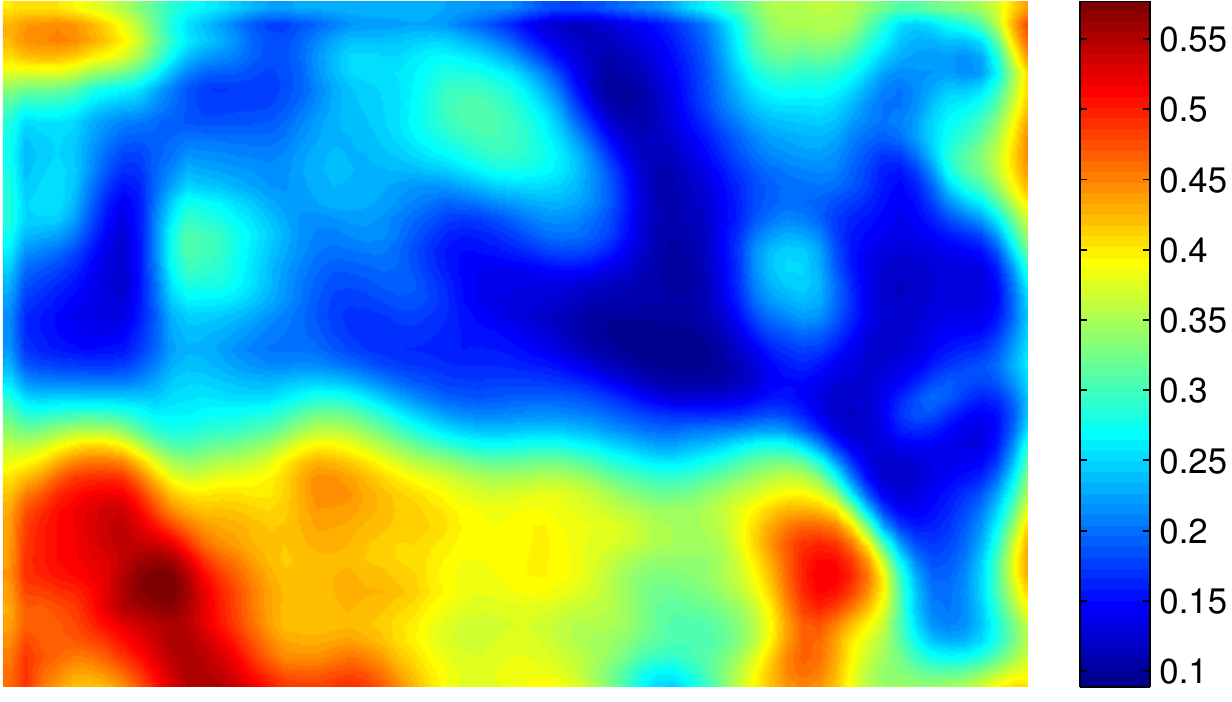} \\
   \includegraphics[height=0.12\linewidth]{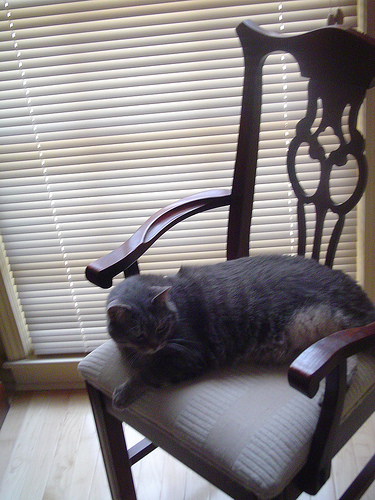} &
   \includegraphics[height=0.12\linewidth]{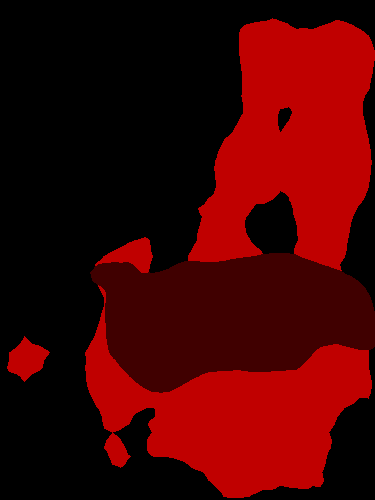} &
   \includegraphics[height=0.12\linewidth]{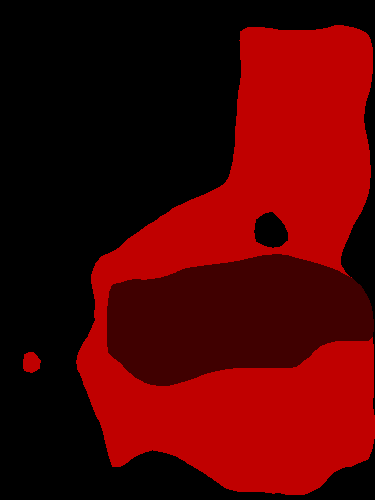} &
   \includegraphics[height=0.12\linewidth]{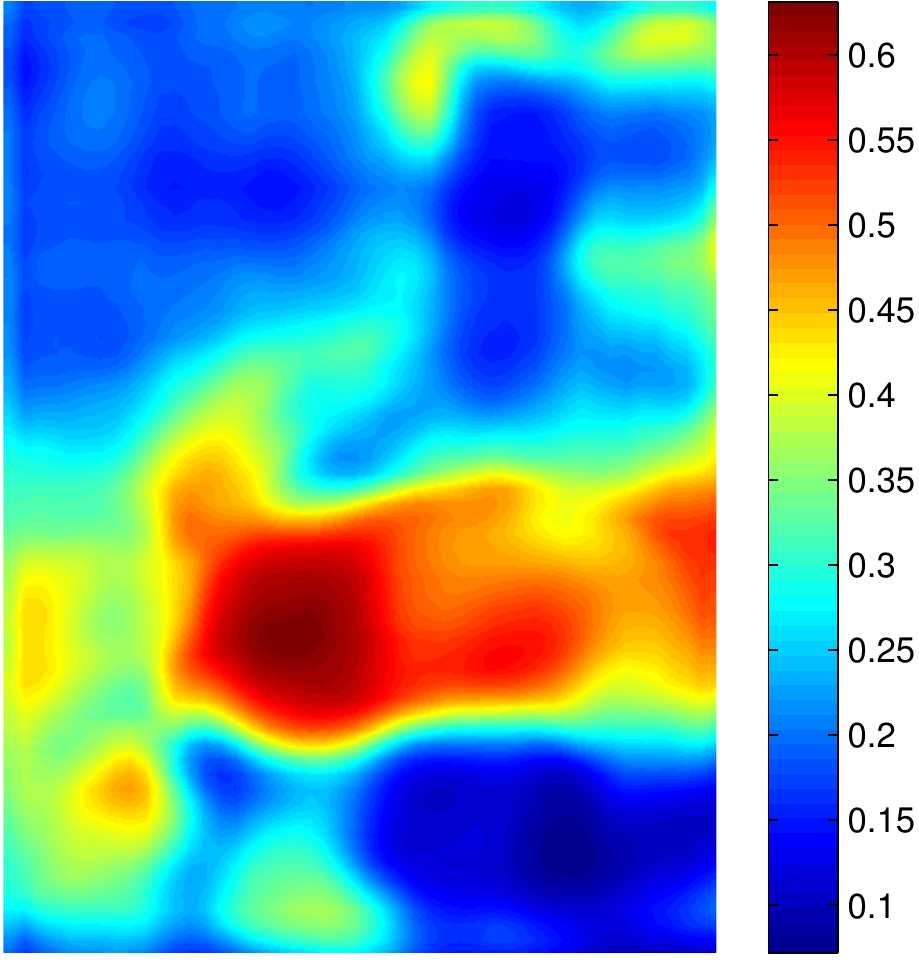} &
   \includegraphics[height=0.12\linewidth]{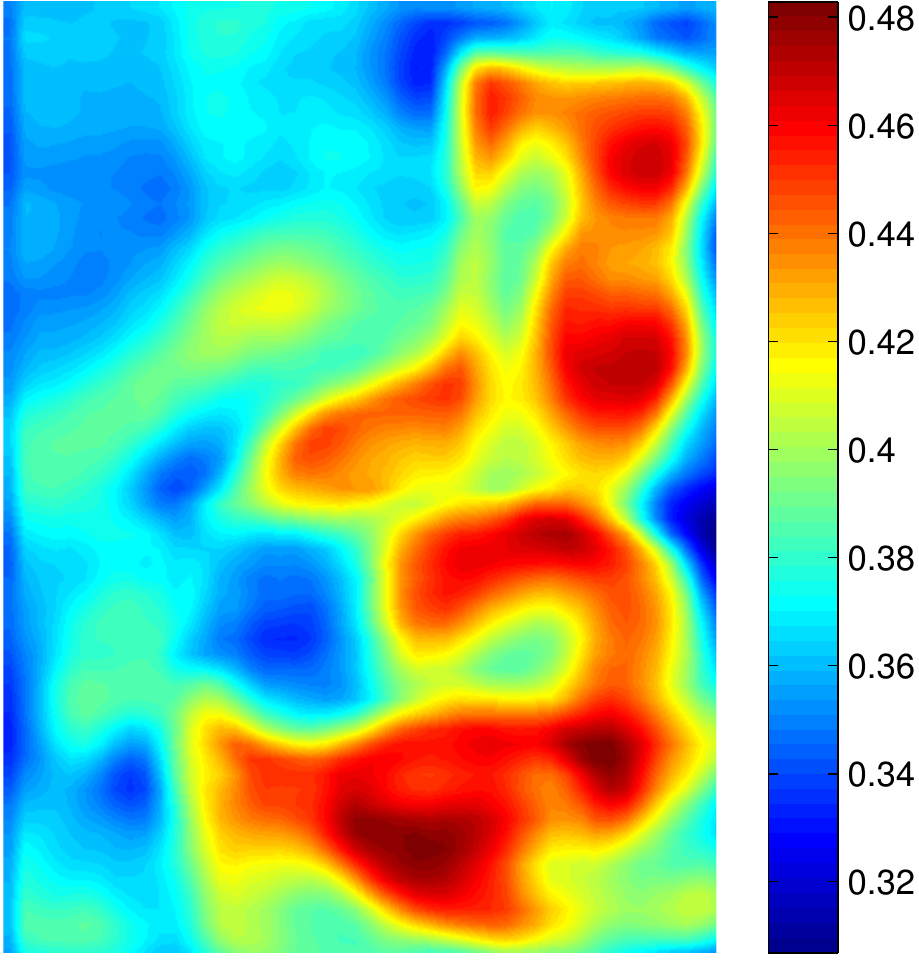} &
   \includegraphics[height=0.12\linewidth]{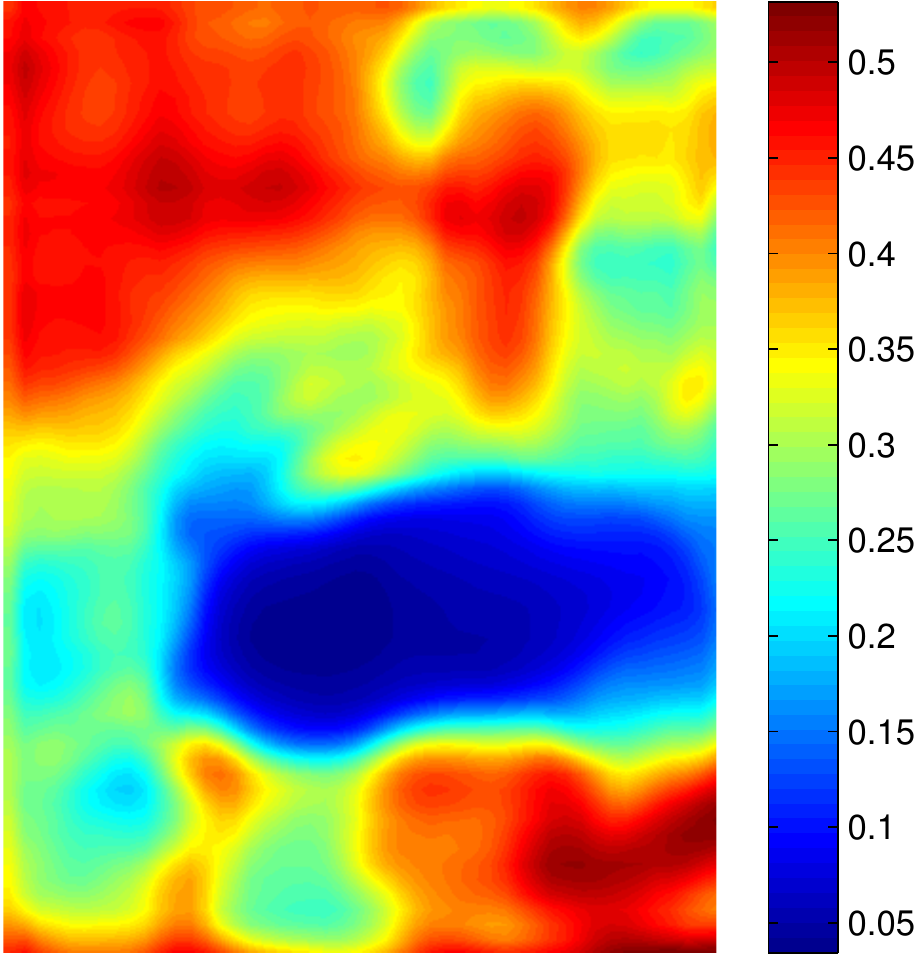} \\
   \includegraphics[height=0.092\linewidth]{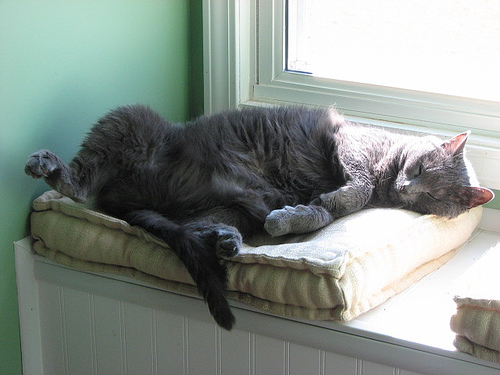} &
   \includegraphics[height=0.092\linewidth]{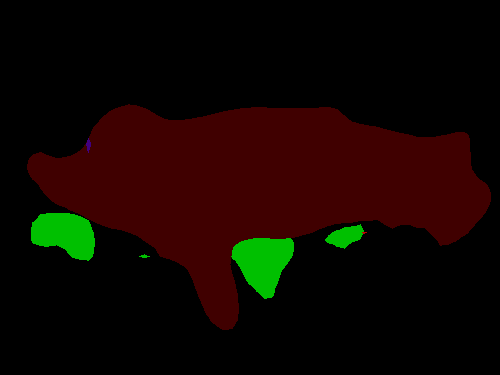} &
   \includegraphics[height=0.092\linewidth]{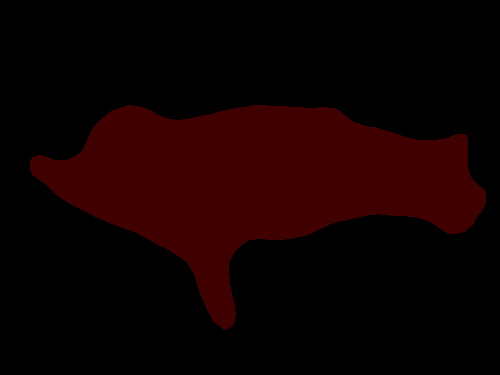} &
   \includegraphics[height=0.092\linewidth]{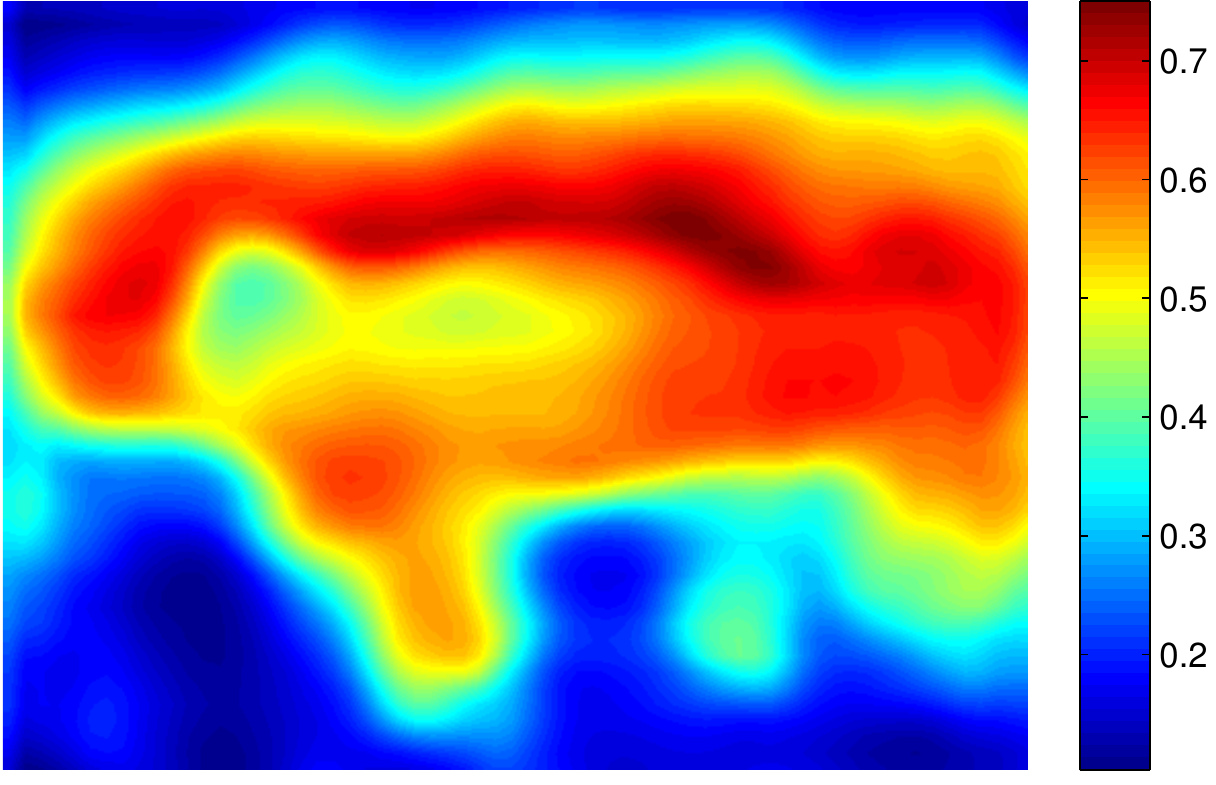} &
   \includegraphics[height=0.092\linewidth]{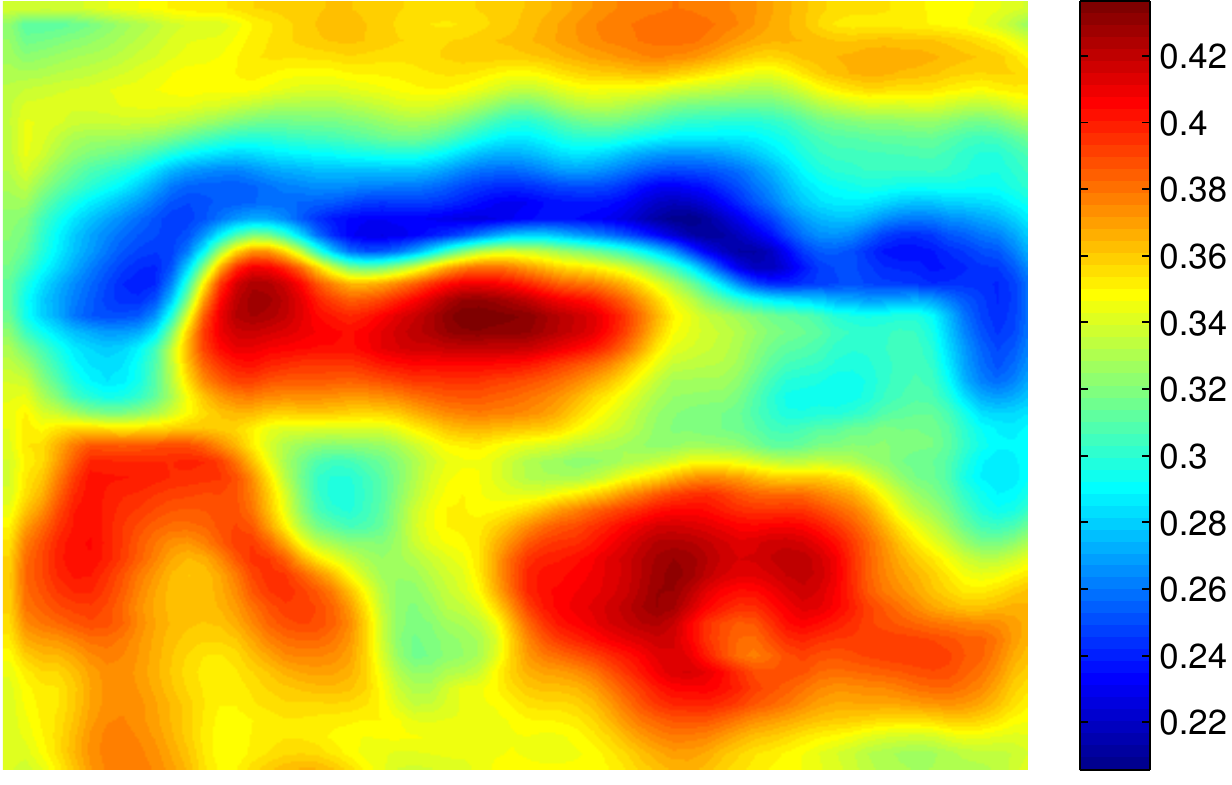} &
   \includegraphics[height=0.092\linewidth]{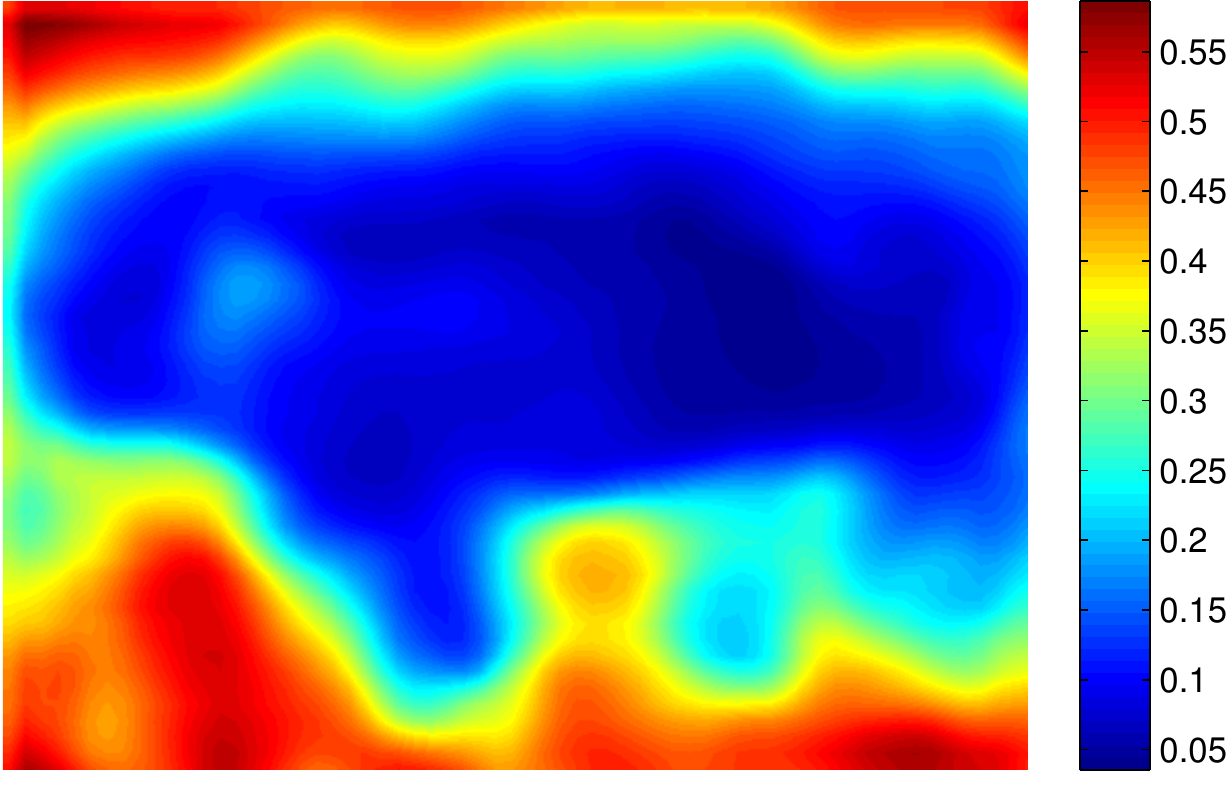} \\
   \includegraphics[height=0.09\linewidth]{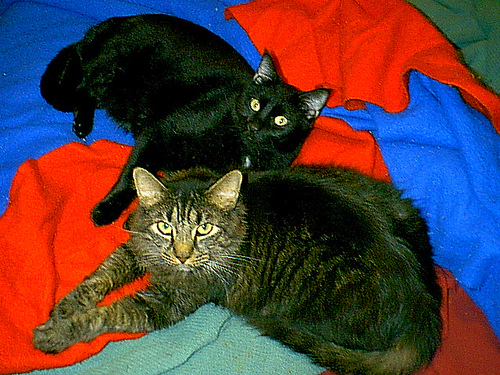} &
   \includegraphics[height=0.09\linewidth]{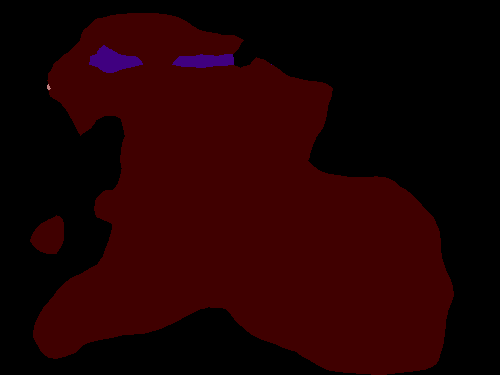} &
   \includegraphics[height=0.09\linewidth]{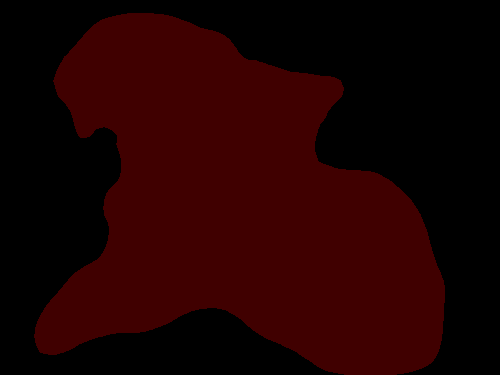} &
   \includegraphics[height=0.09\linewidth]{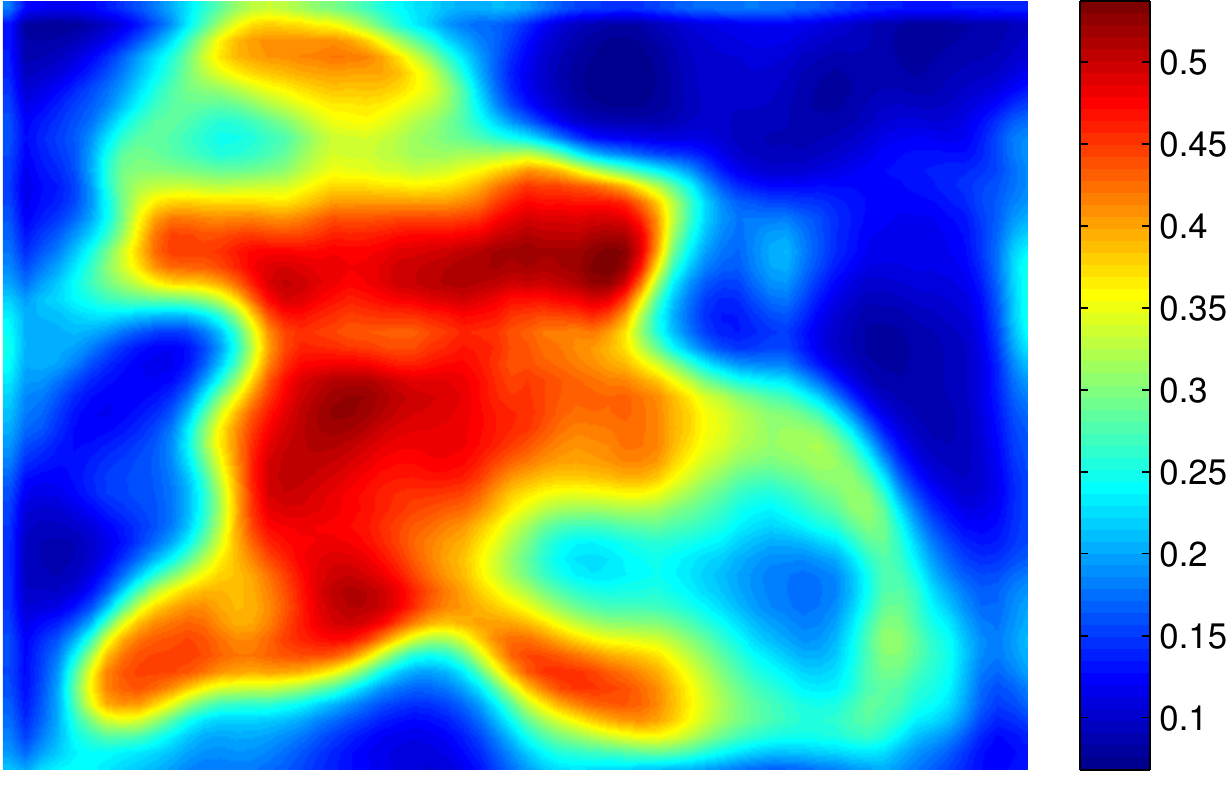} &
   \includegraphics[height=0.09\linewidth]{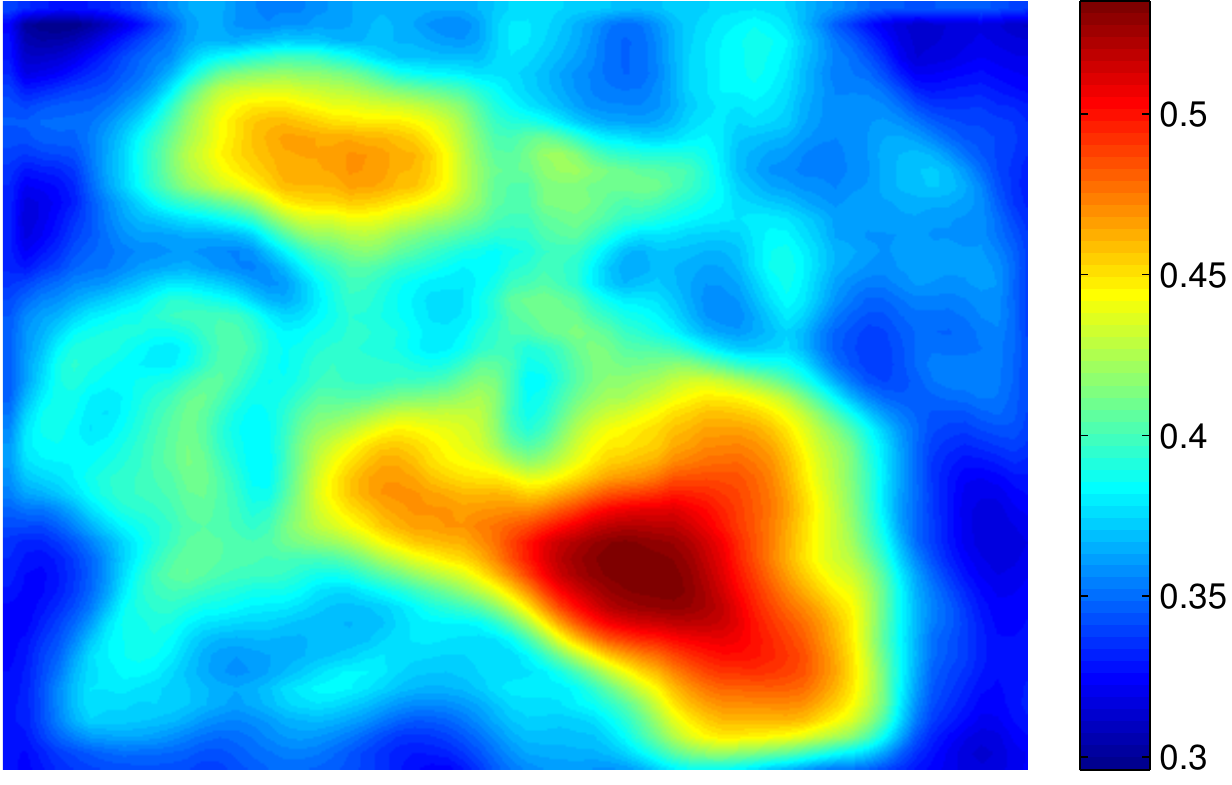} &
   \includegraphics[height=0.09\linewidth]{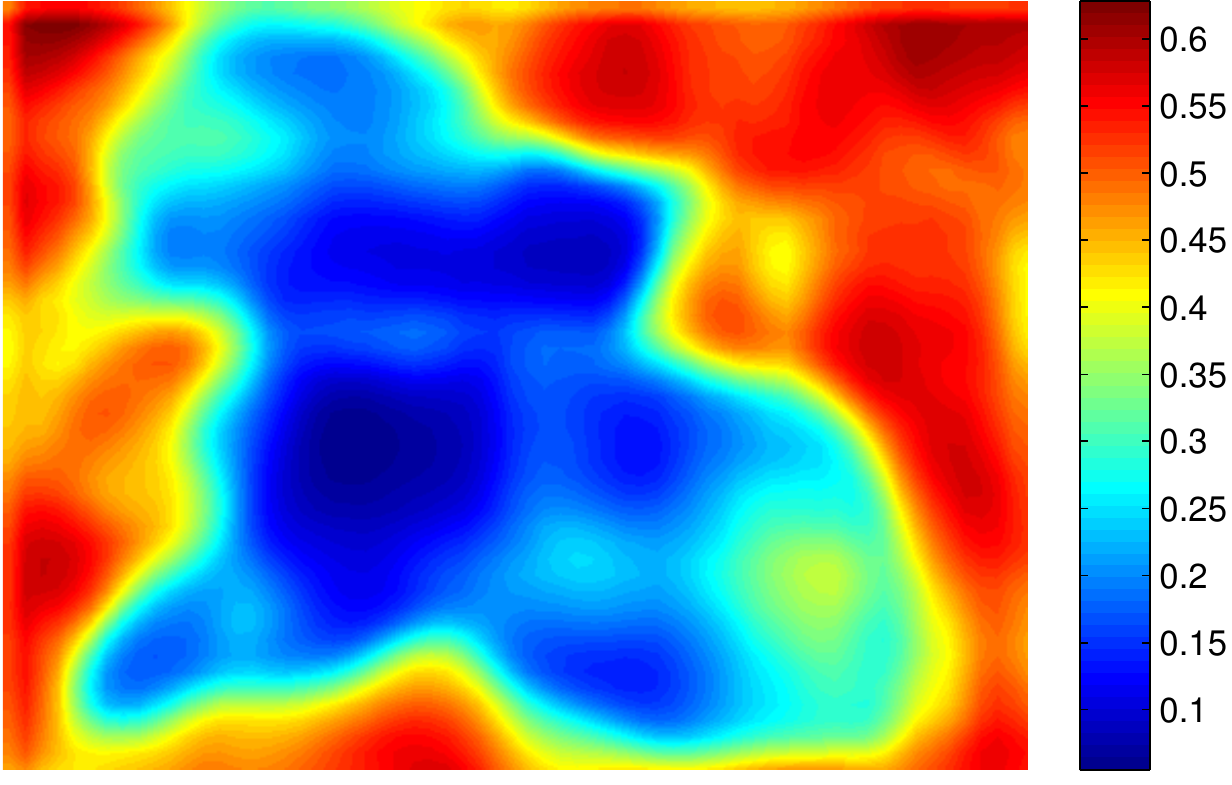} \\
   \includegraphics[height=0.09\linewidth]{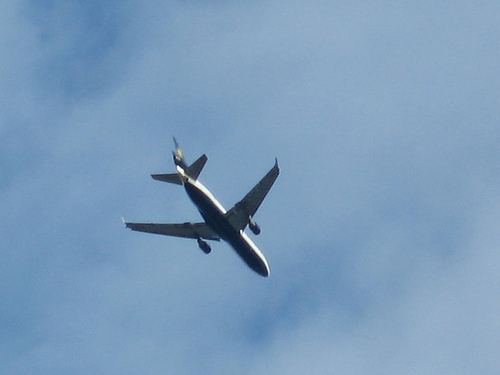} &
   \includegraphics[height=0.09\linewidth]{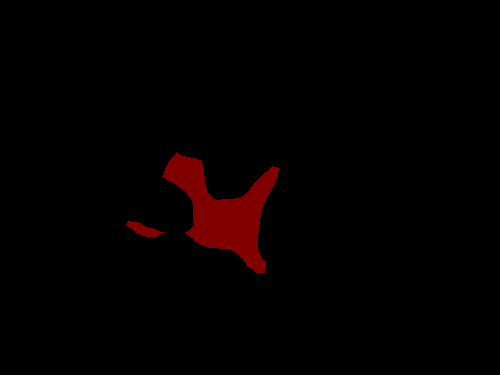} &
   \includegraphics[height=0.09\linewidth]{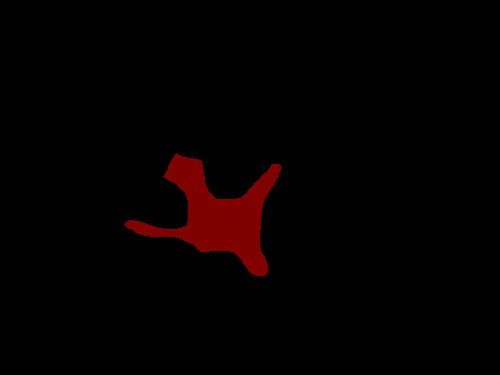} &
   \includegraphics[height=0.09\linewidth]{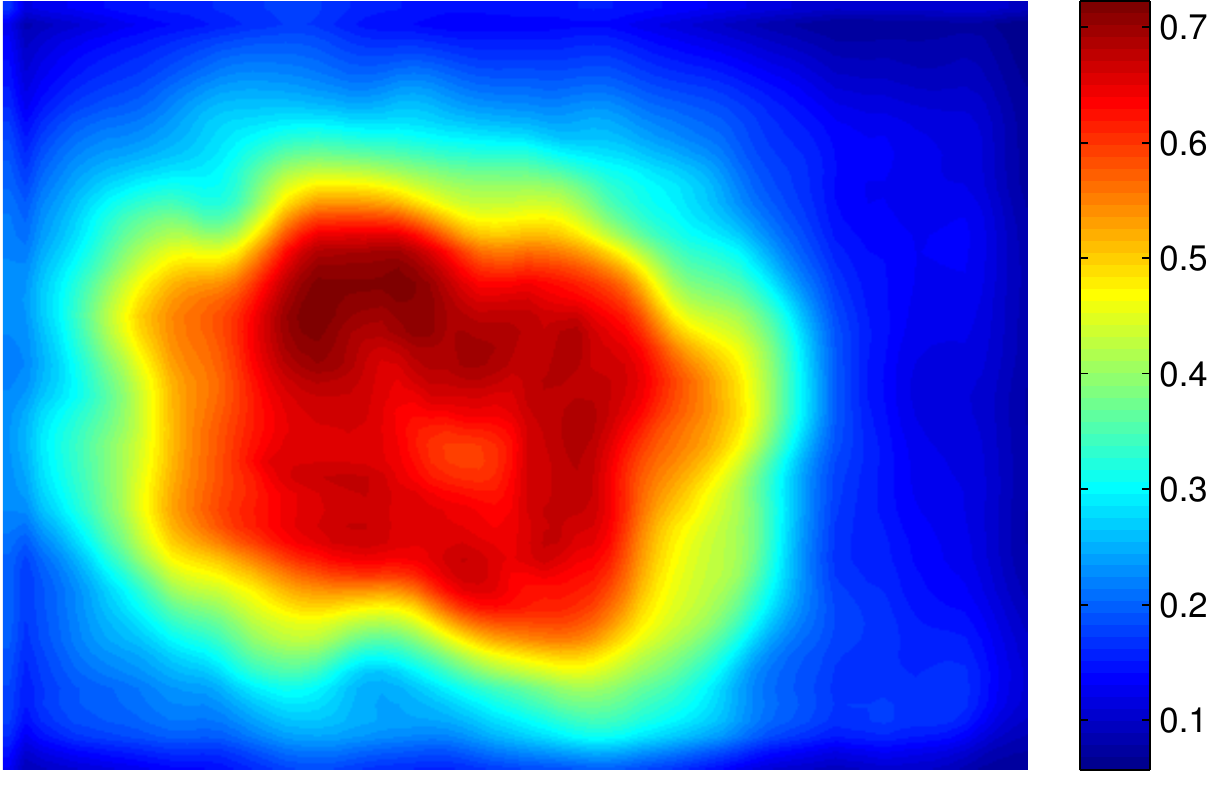} &
   \includegraphics[height=0.09\linewidth]{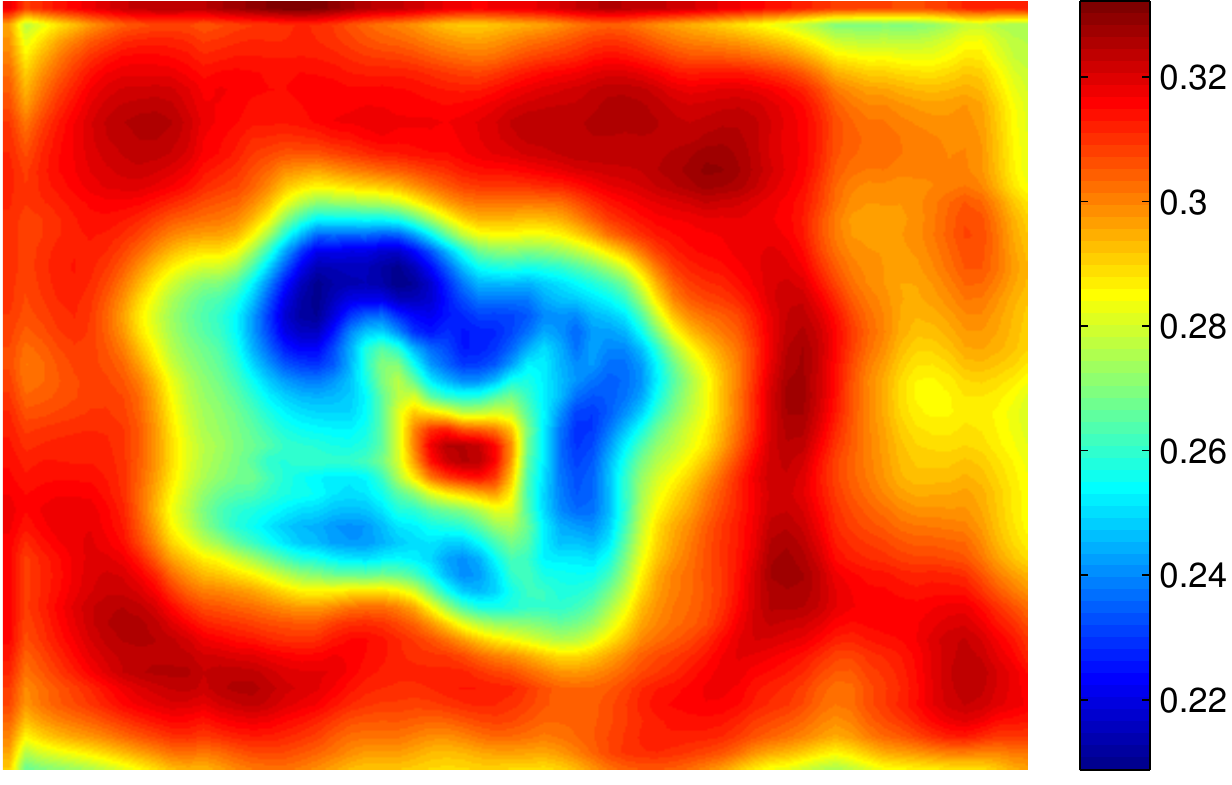} &
   \includegraphics[height=0.09\linewidth]{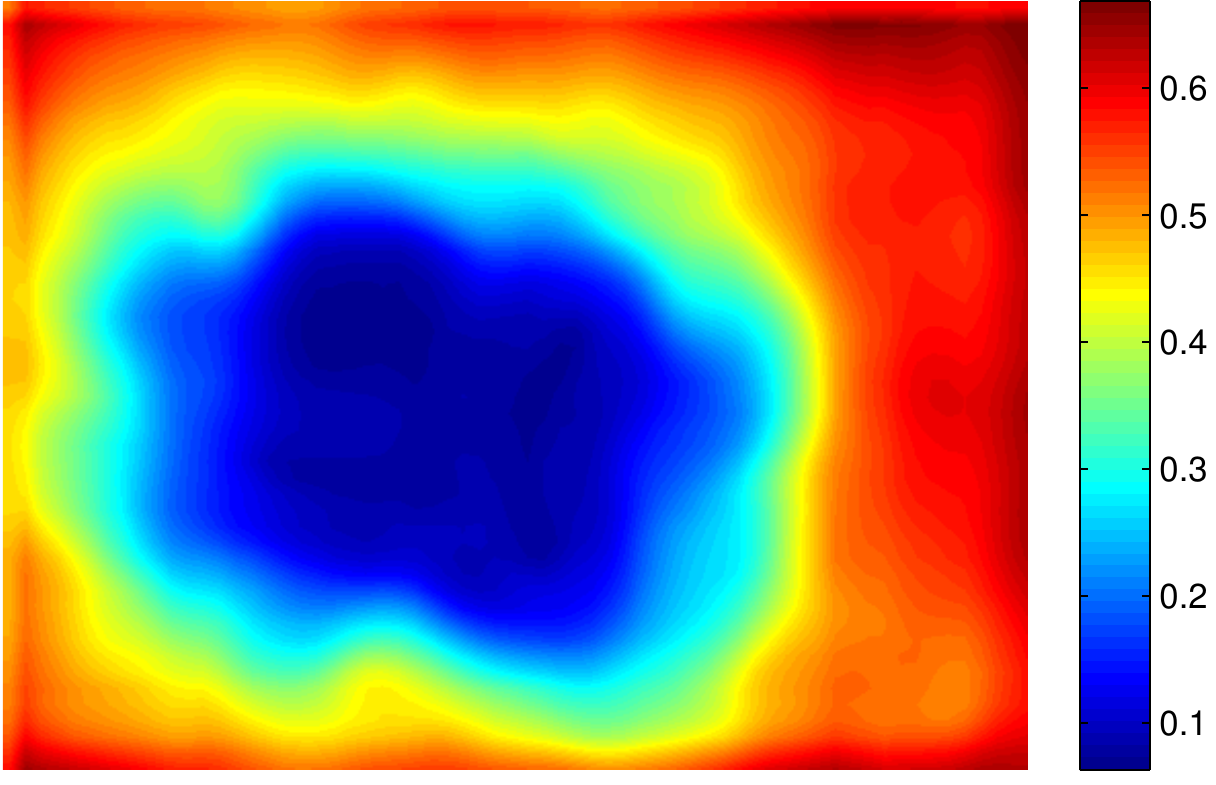} \\
   \includegraphics[height=0.092\linewidth]{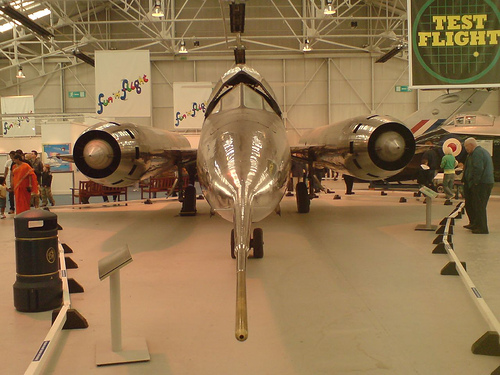} &
   \includegraphics[height=0.092\linewidth]{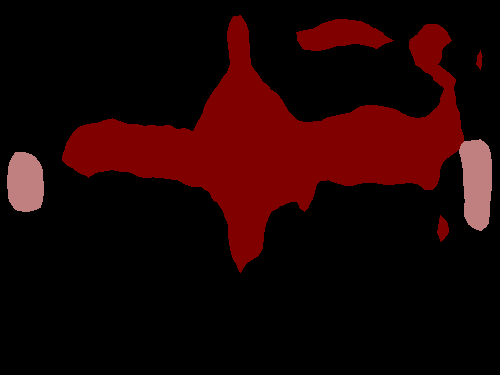} &
   \includegraphics[height=0.092\linewidth]{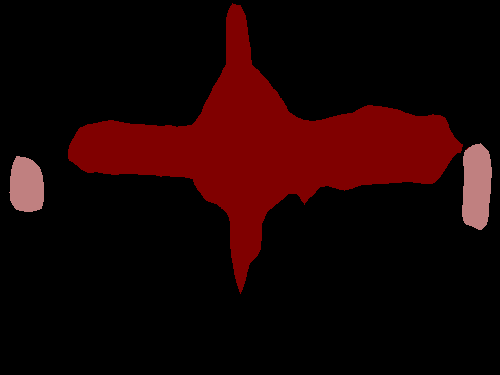} &
   \includegraphics[height=0.092\linewidth]{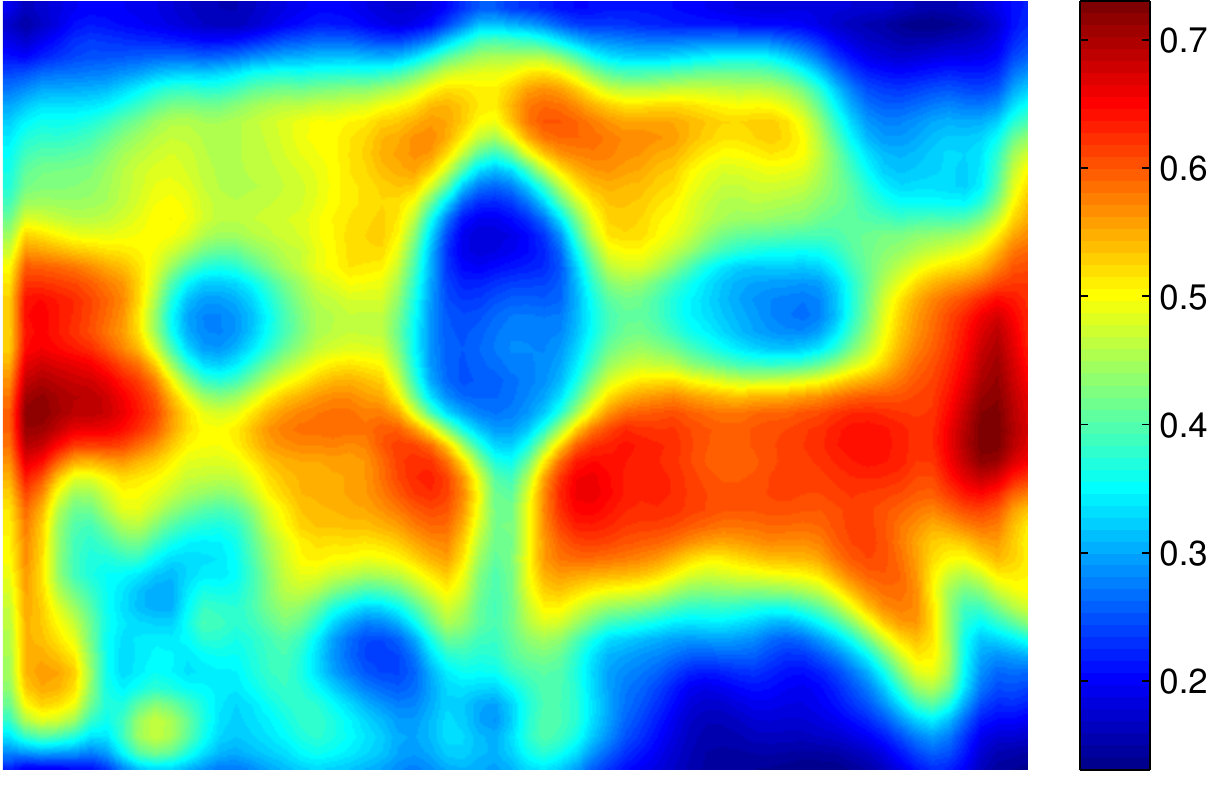} &
   \includegraphics[height=0.092\linewidth]{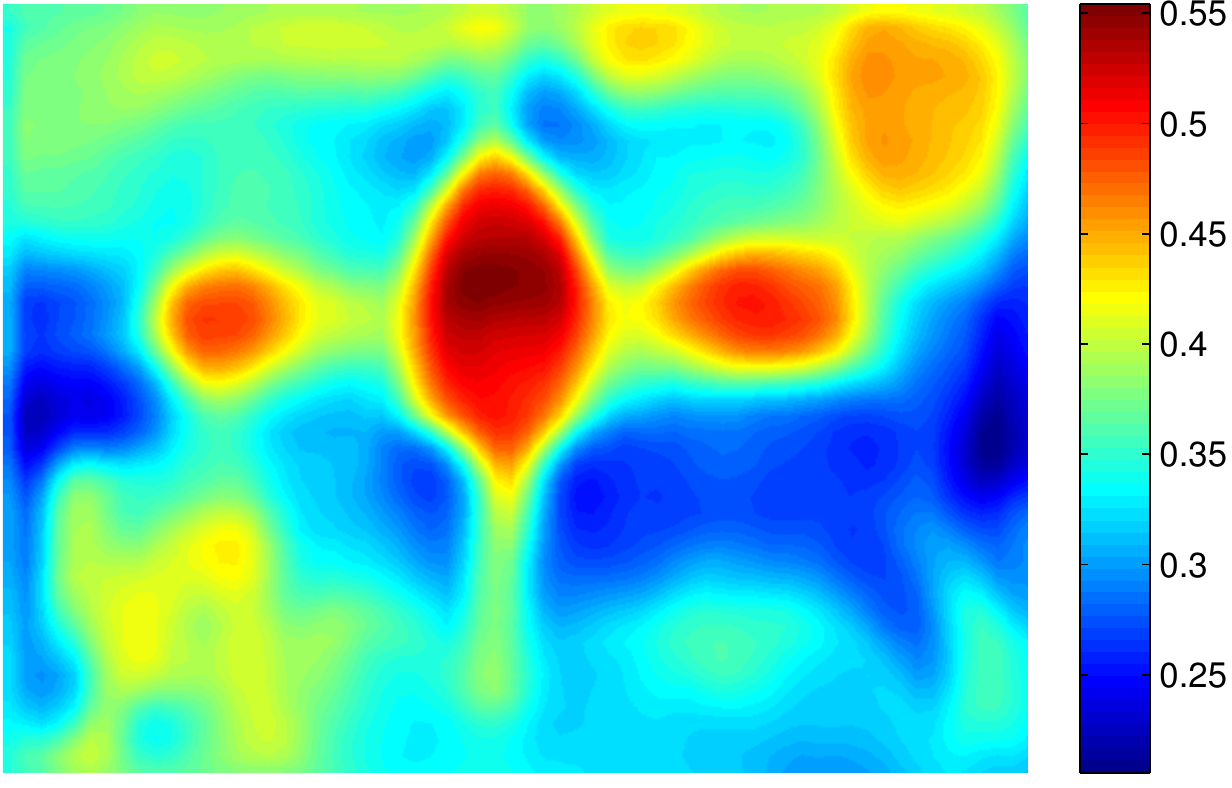} &
   \includegraphics[height=0.092\linewidth]{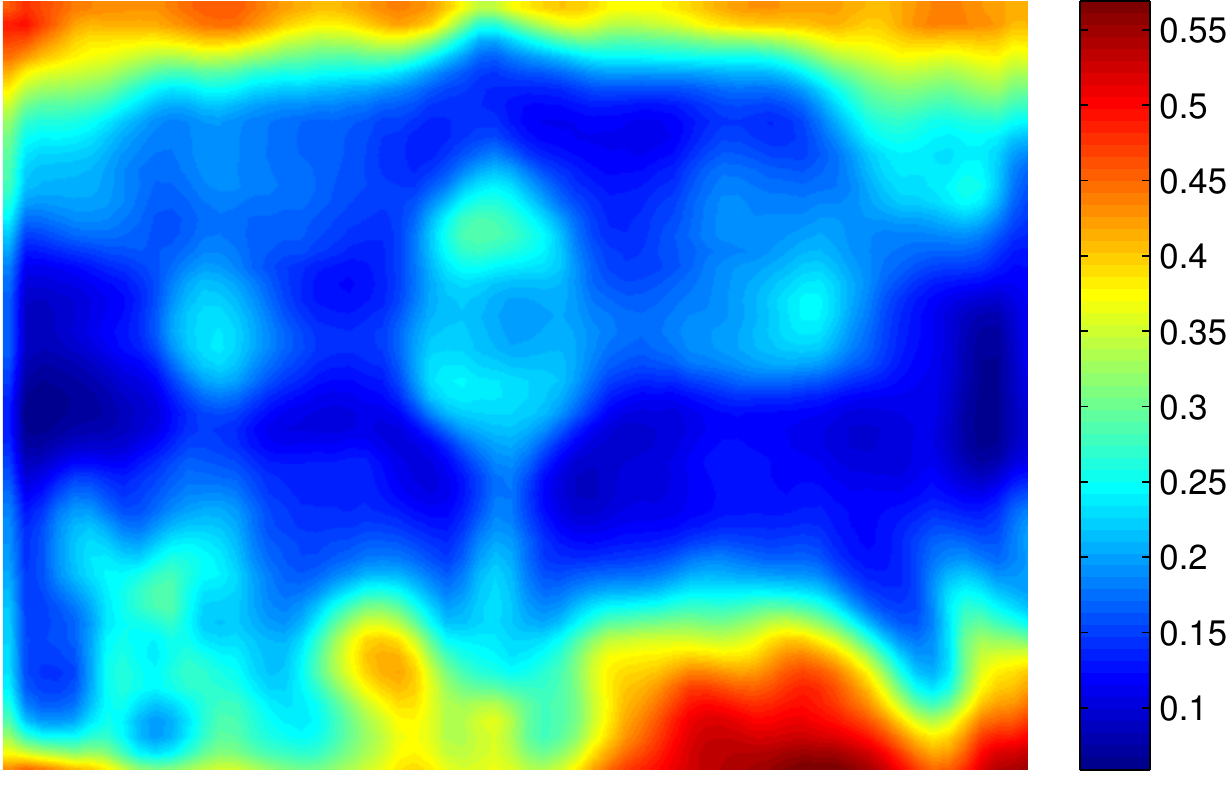} \\
   \includegraphics[height=0.122\linewidth]{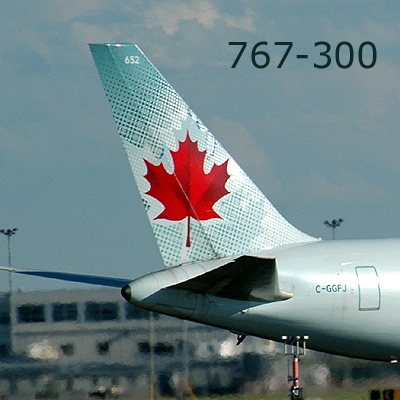} &
   \includegraphics[height=0.122\linewidth]{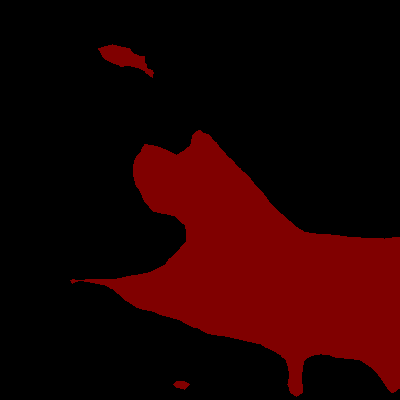} &
   \includegraphics[height=0.122\linewidth]{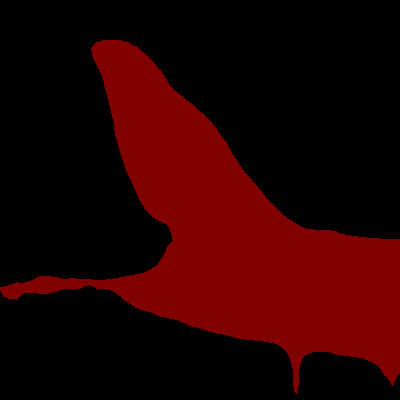} &
   \includegraphics[height=0.122\linewidth]{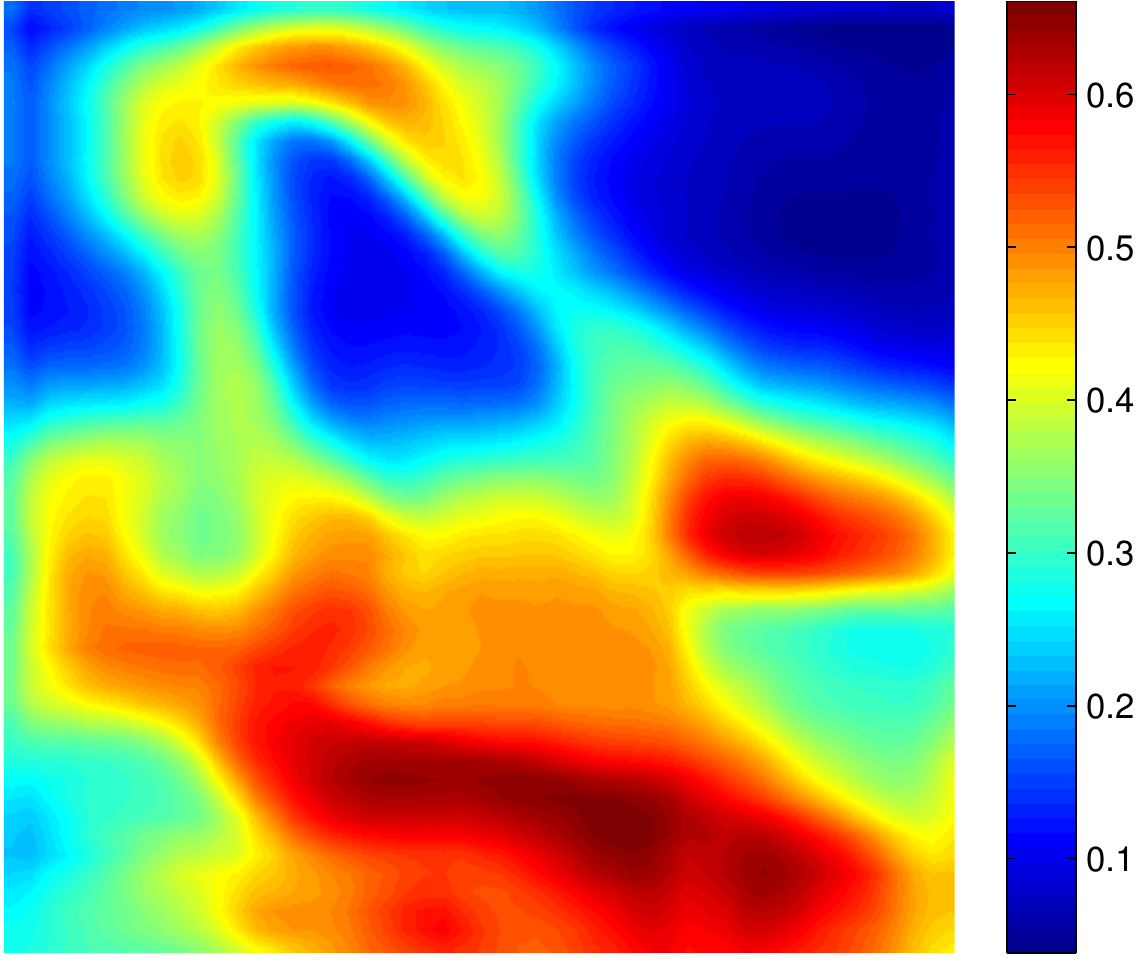} &
   \includegraphics[height=0.122\linewidth]{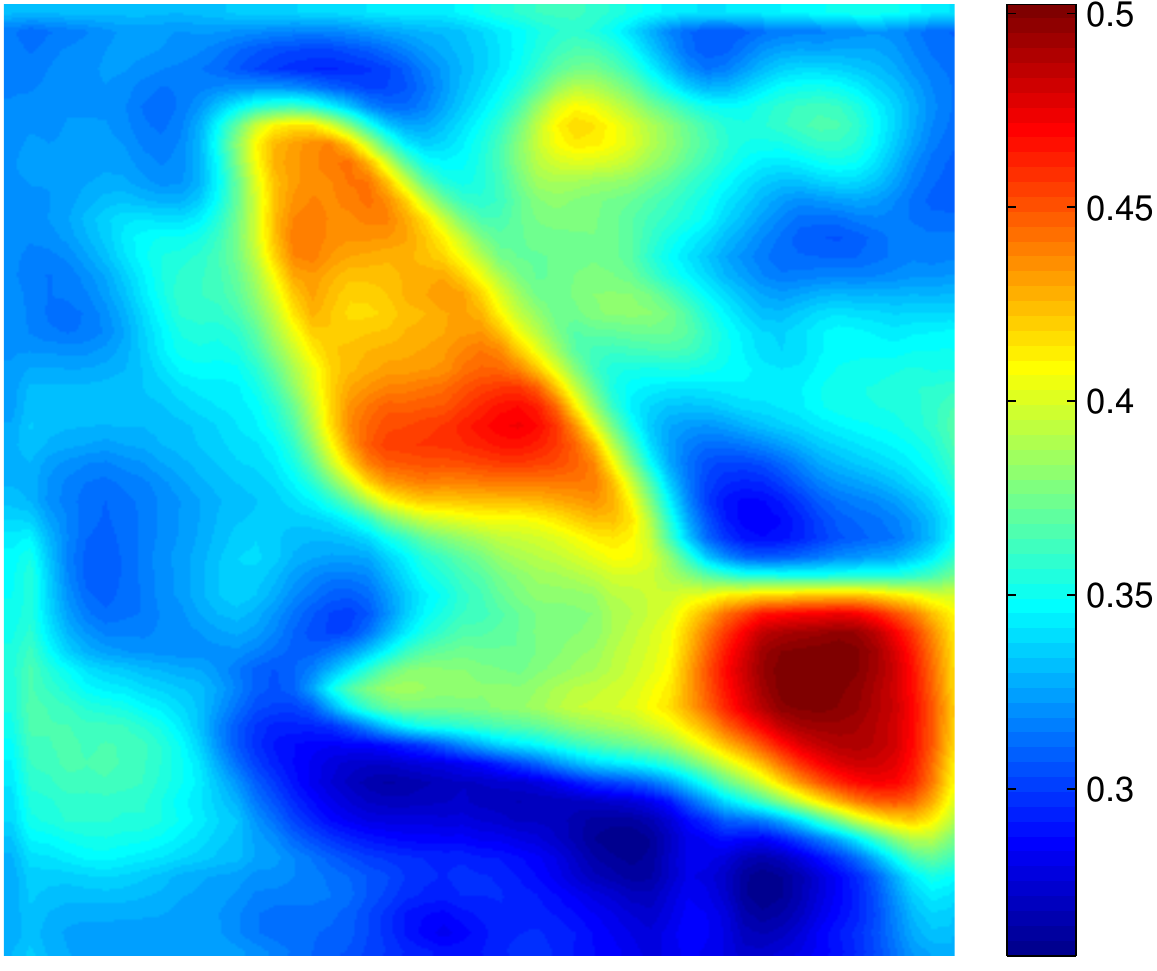} &
   \includegraphics[height=0.122\linewidth]{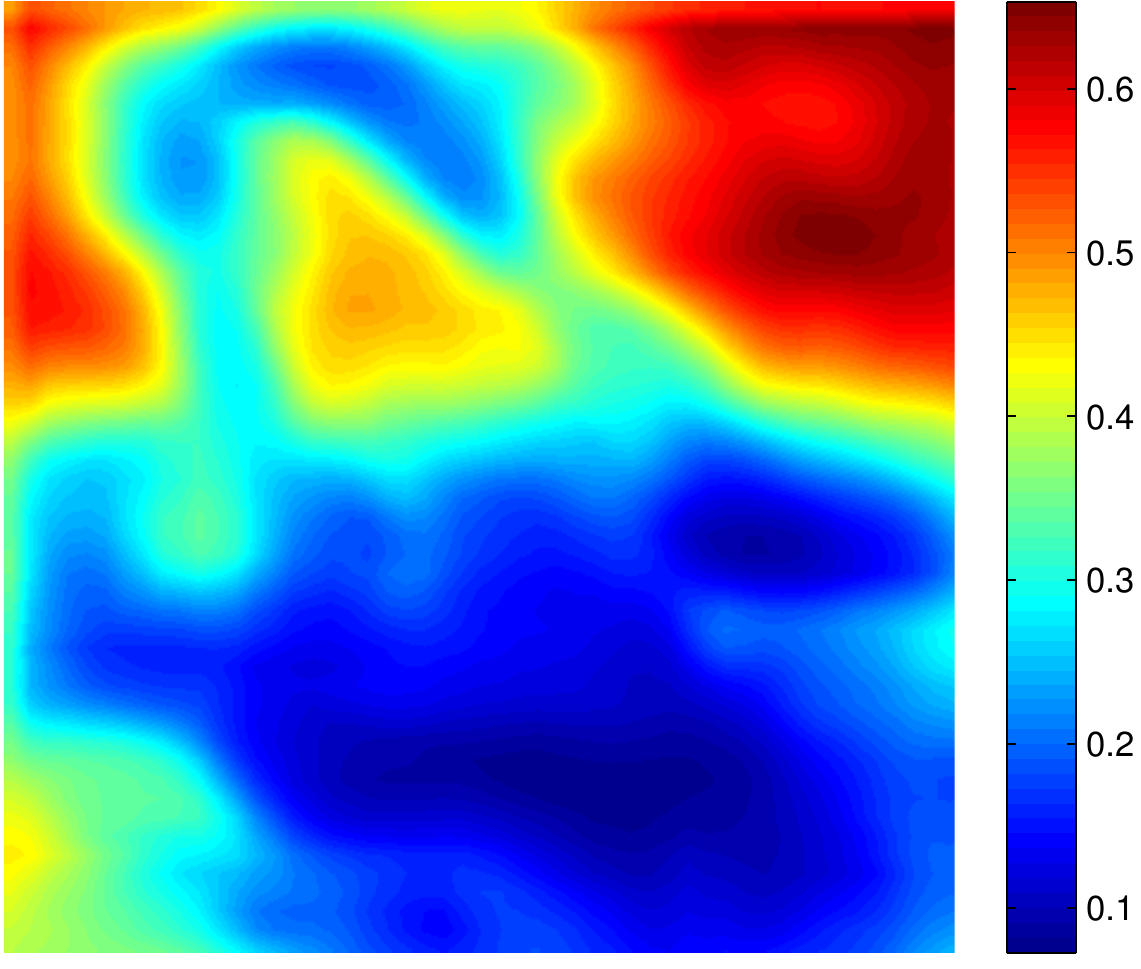} \\
   \includegraphics[height=0.085\linewidth]{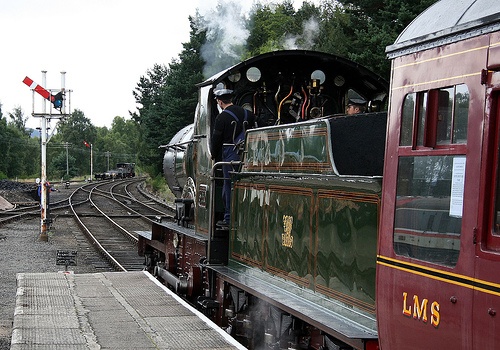} &
   \includegraphics[height=0.085\linewidth]{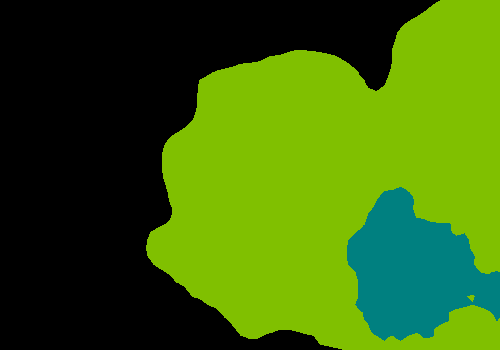} &
   \includegraphics[height=0.085\linewidth]{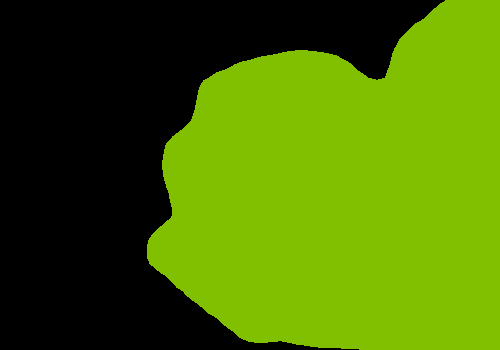} &
   \includegraphics[height=0.085\linewidth]{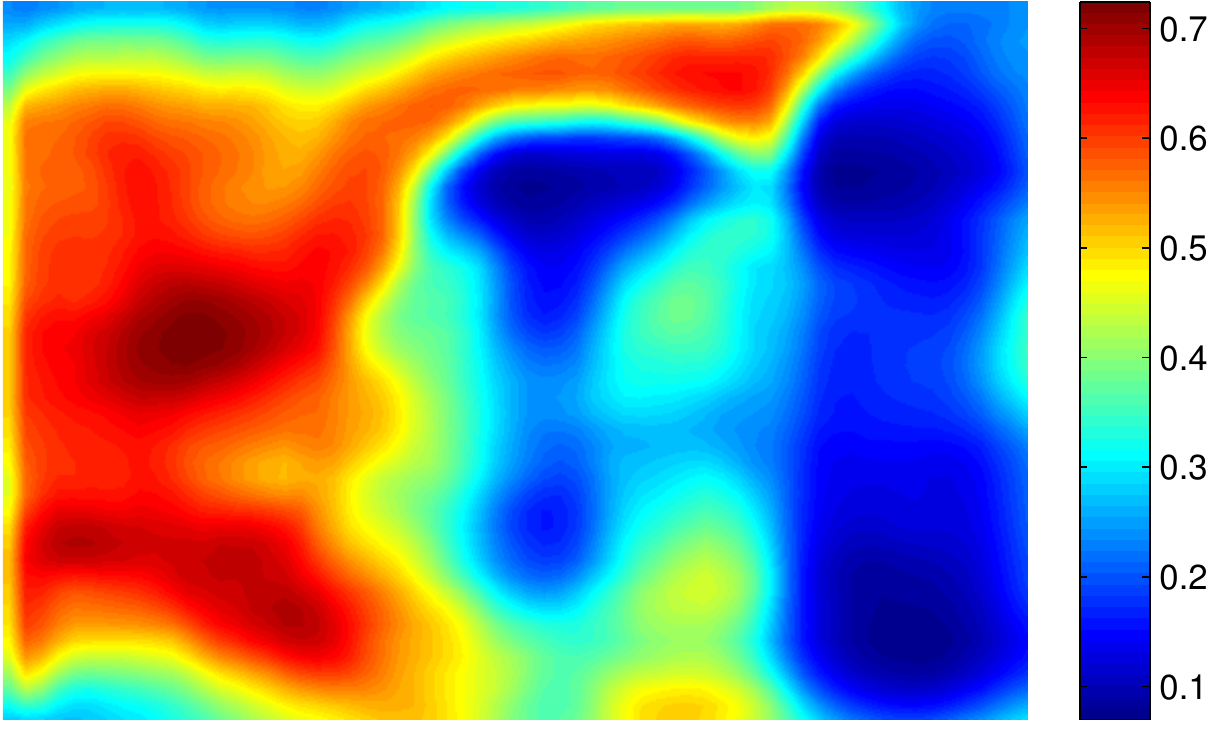} &
   \includegraphics[height=0.085\linewidth]{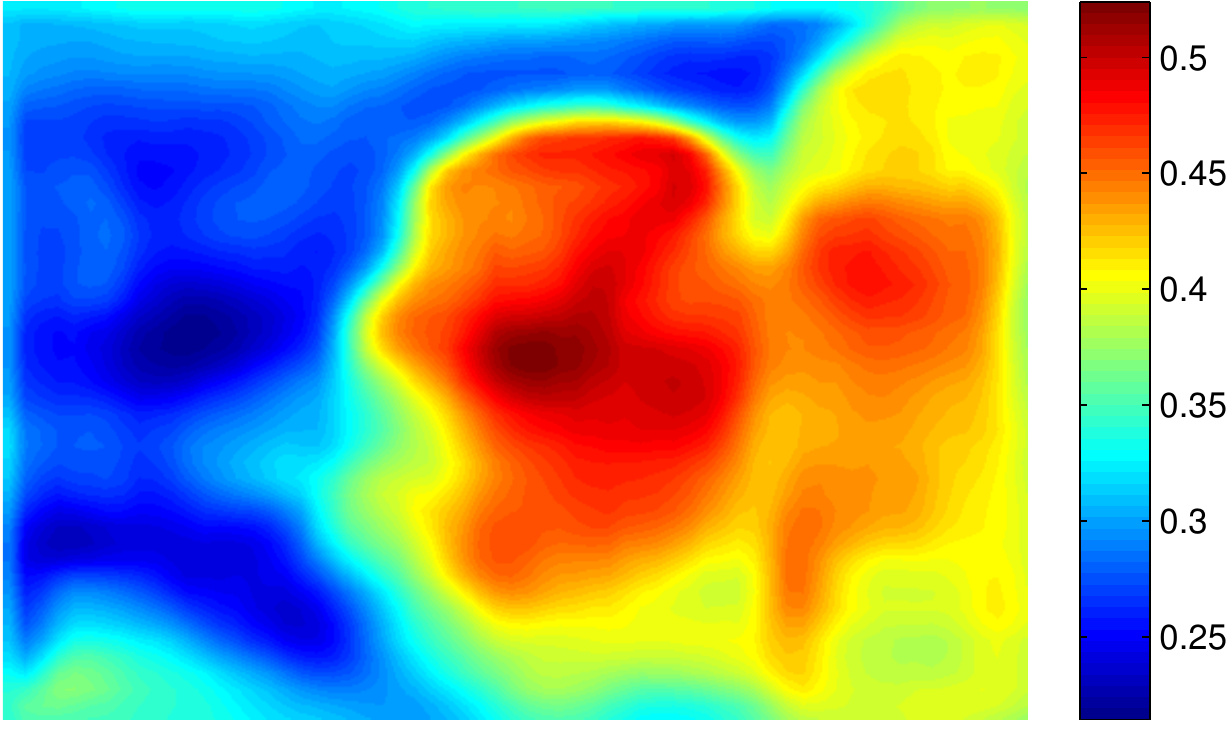} &
   \includegraphics[height=0.085\linewidth]{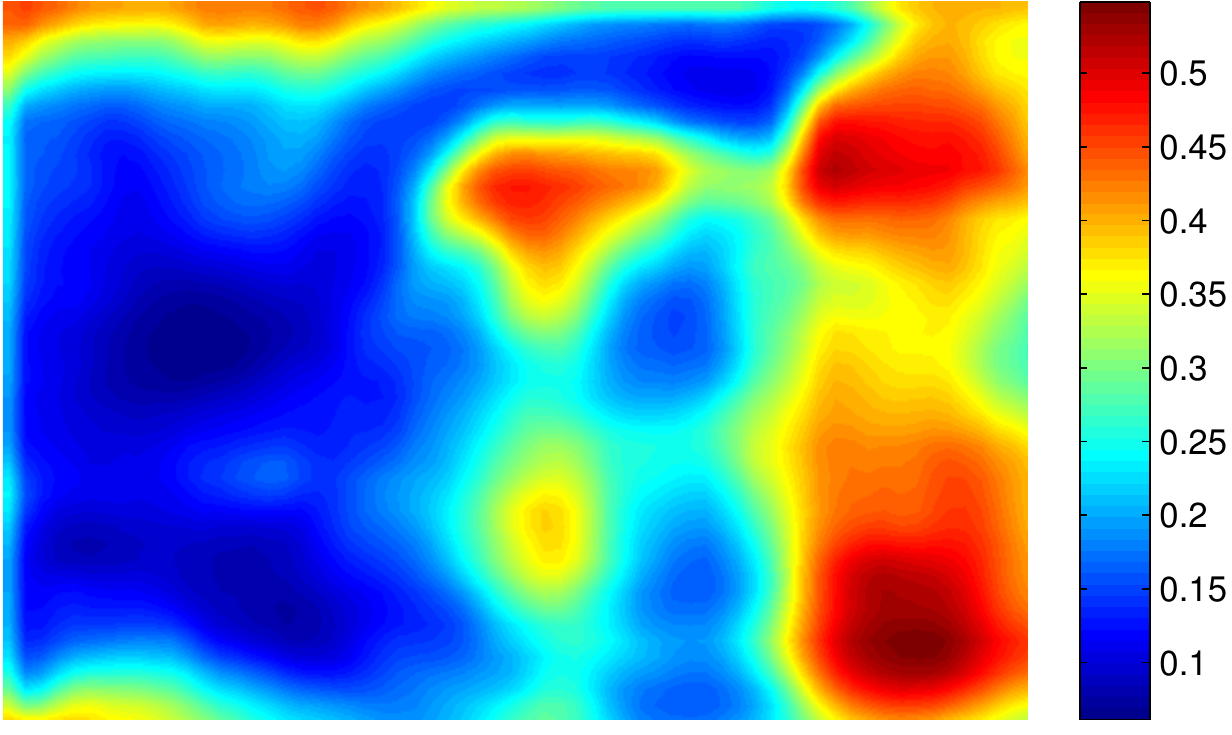} \\
   \includegraphics[height=0.09\linewidth]{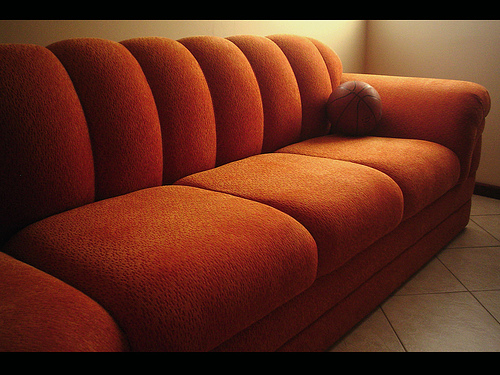} &
   \includegraphics[height=0.09\linewidth]{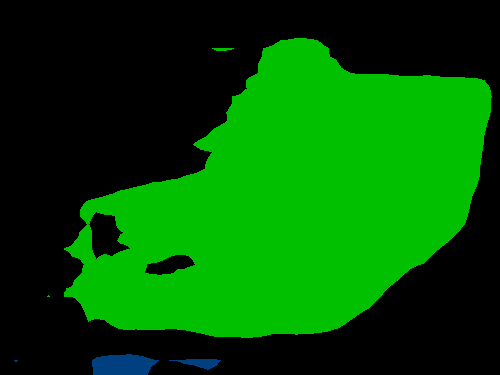} &
   \includegraphics[height=0.09\linewidth]{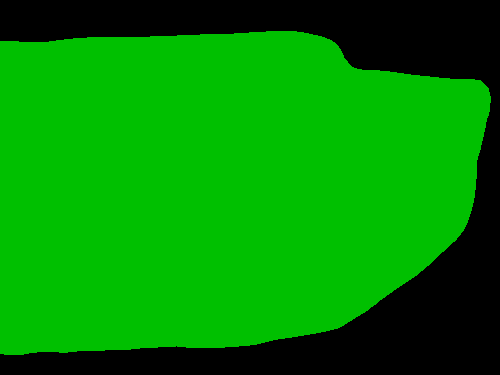} &
   \includegraphics[height=0.09\linewidth]{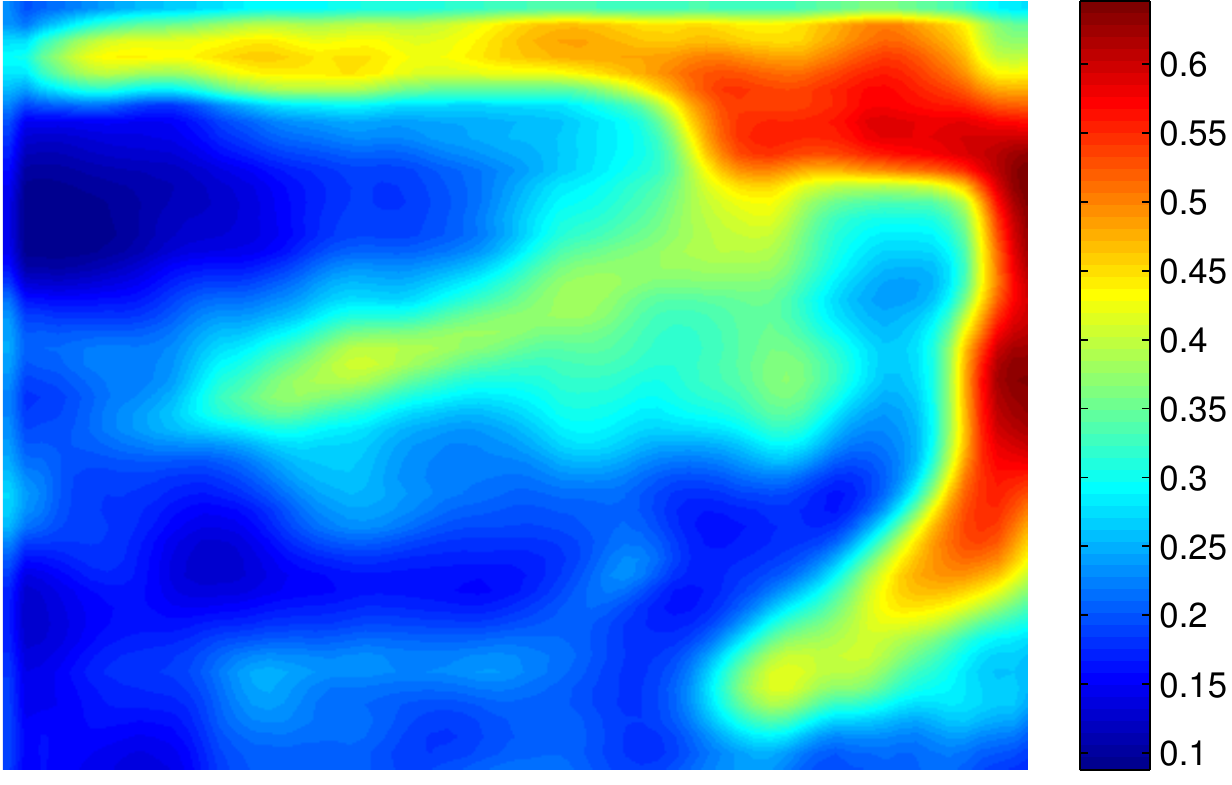} &
   \includegraphics[height=0.09\linewidth]{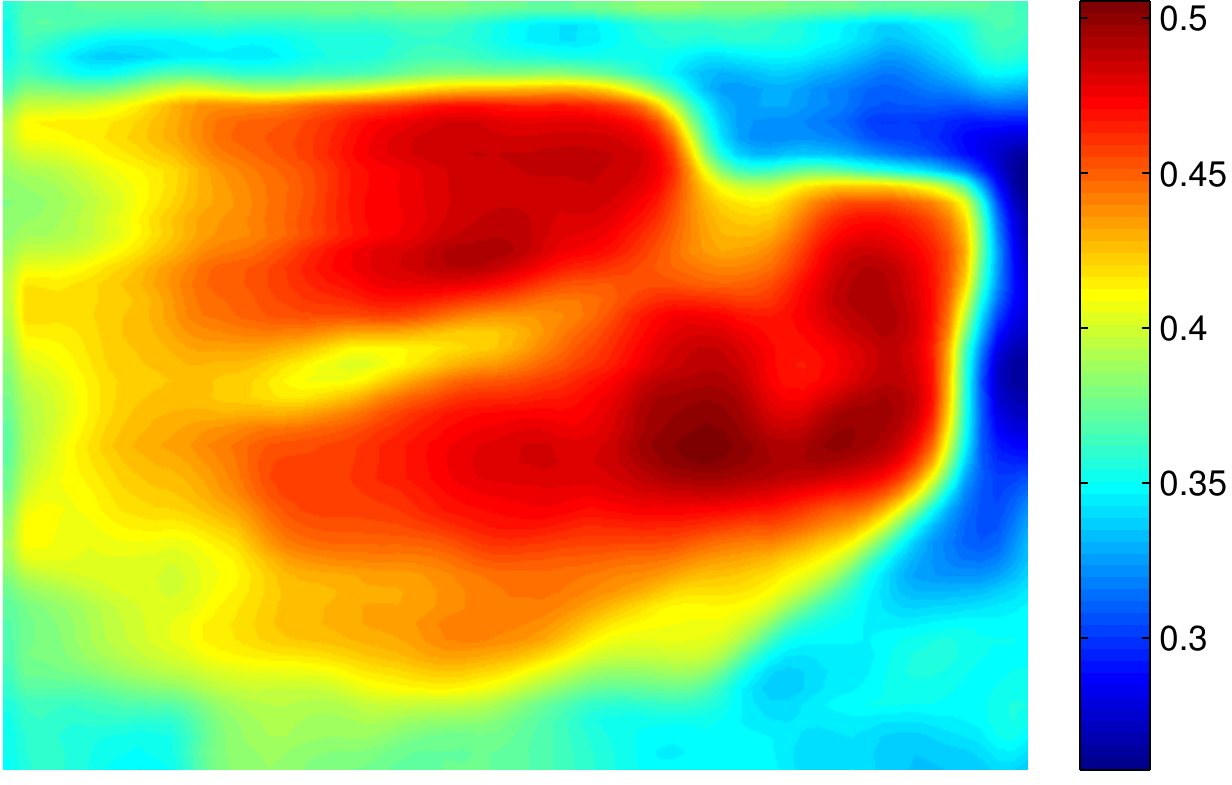} &
   \includegraphics[height=0.09\linewidth]{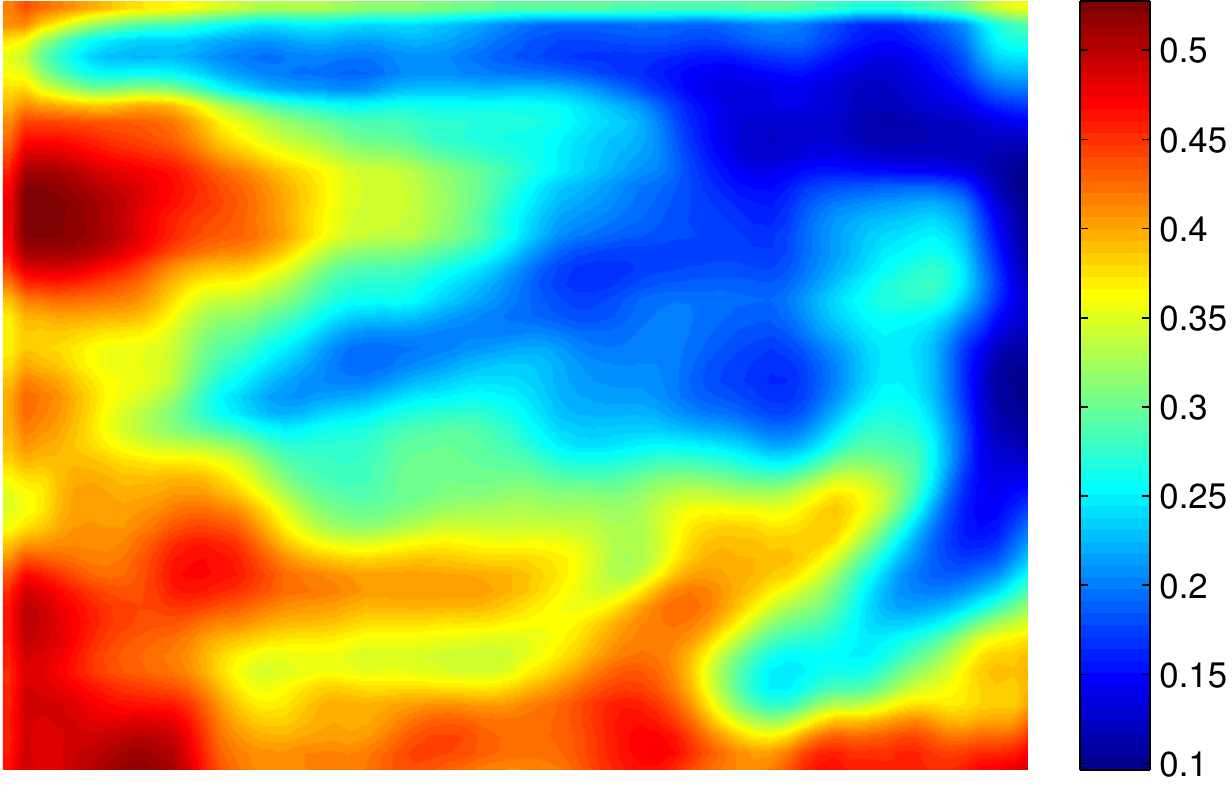} \\
   \includegraphics[height=0.13\linewidth]{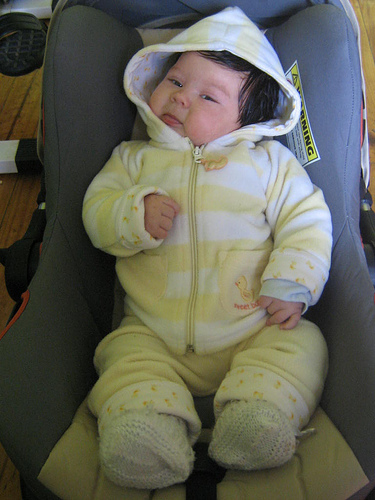} &
   \includegraphics[height=0.13\linewidth]{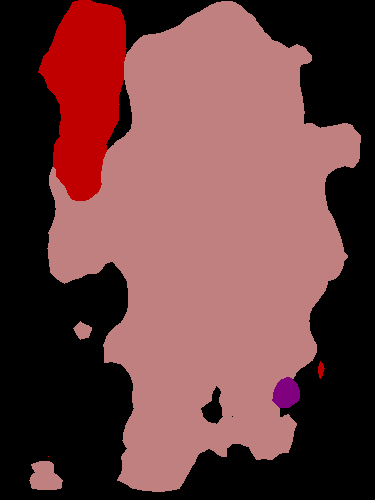} &
   \includegraphics[height=0.13\linewidth]{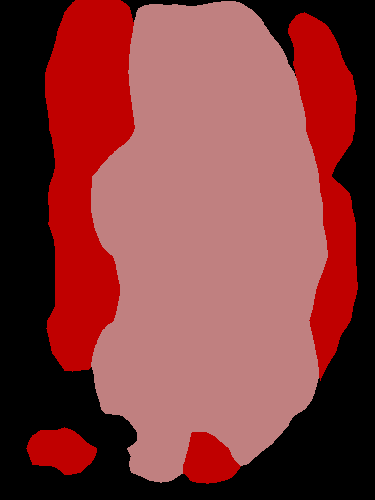} &
   \includegraphics[height=0.13\linewidth]{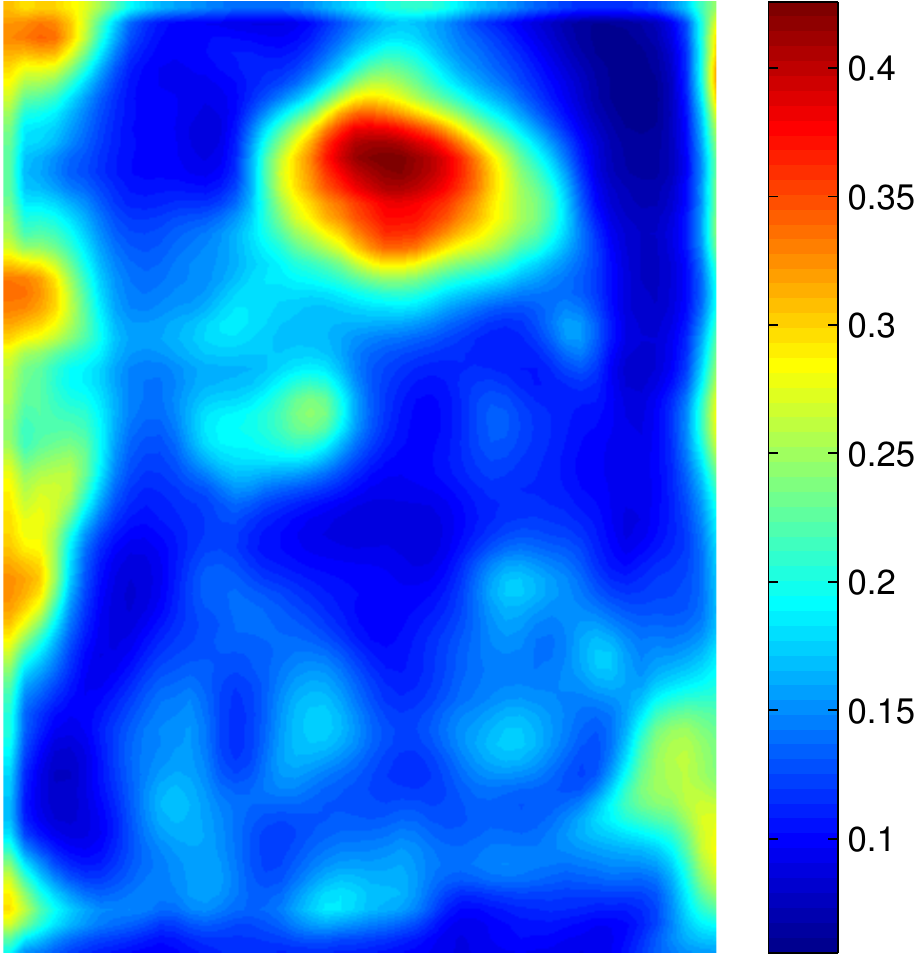} &
   \includegraphics[height=0.13\linewidth]{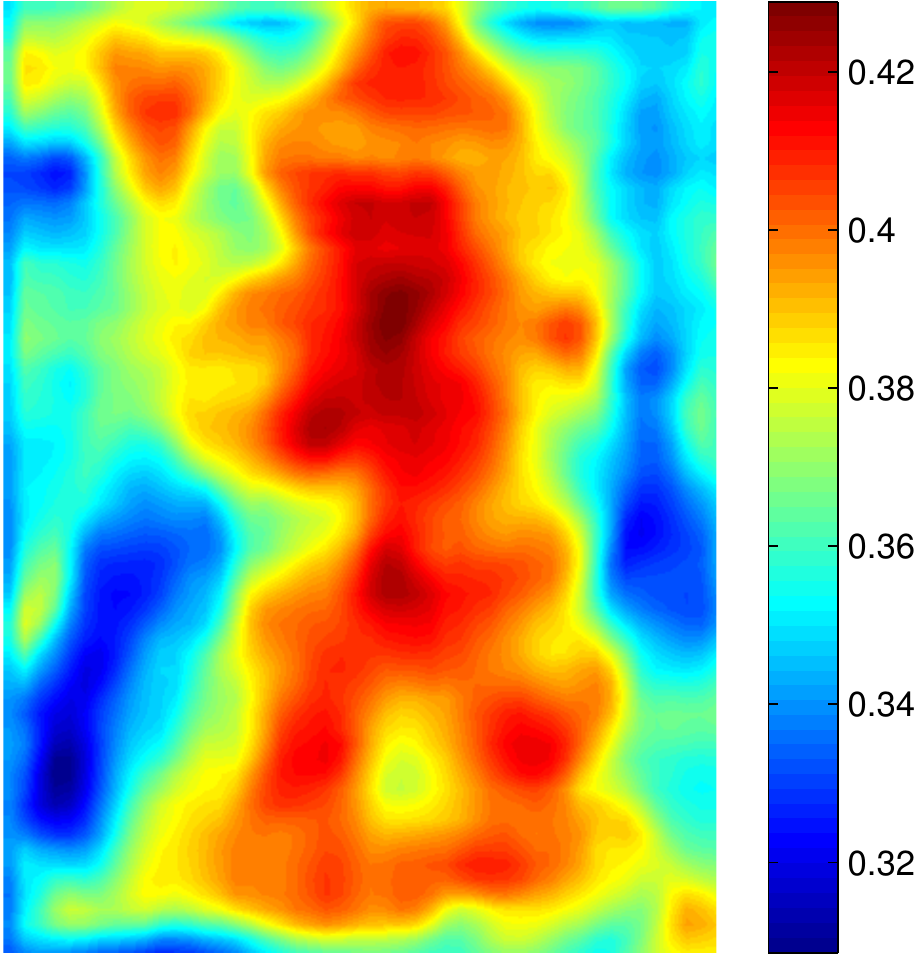} &
   \includegraphics[height=0.13\linewidth]{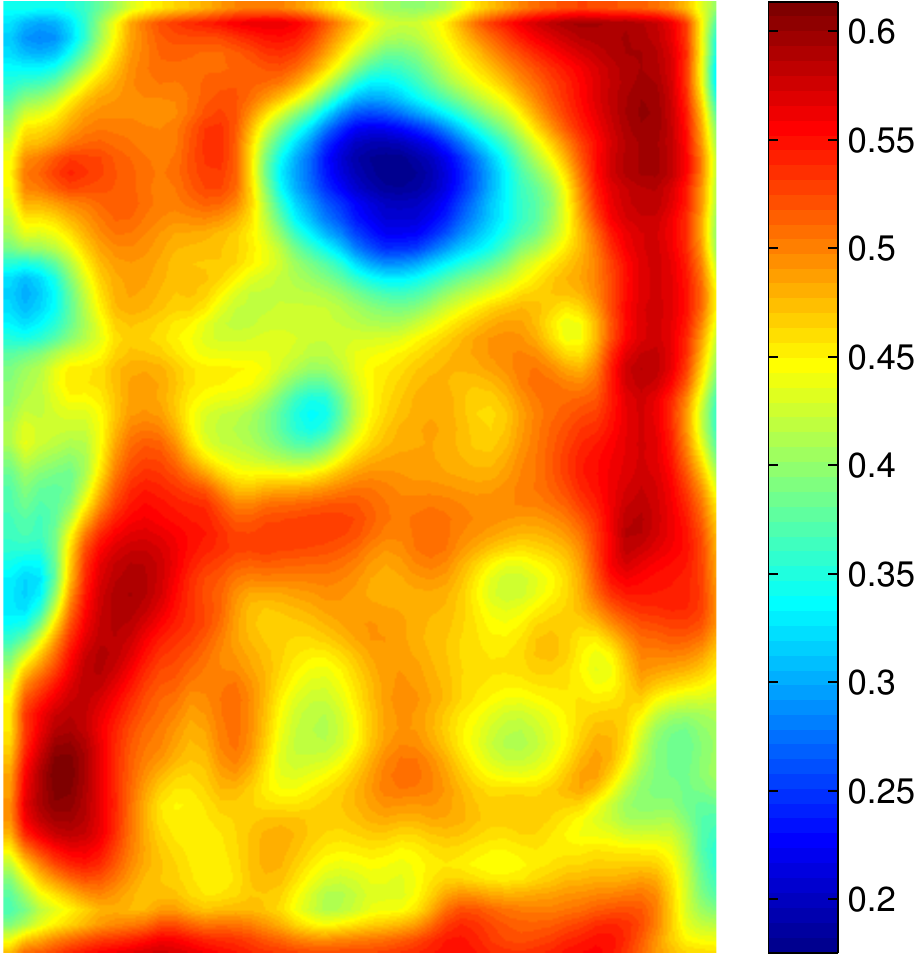} \\
   \includegraphics[height=0.11\linewidth]{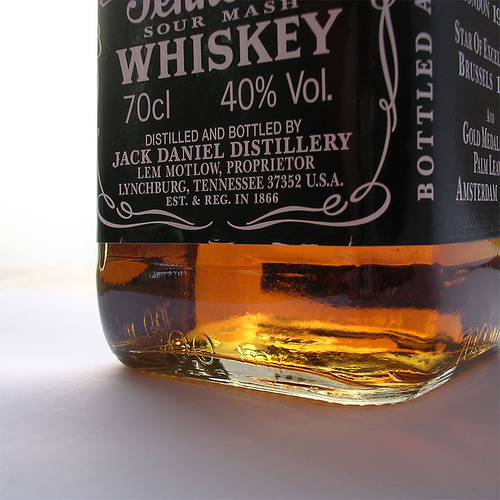} &
   \includegraphics[height=0.11\linewidth]{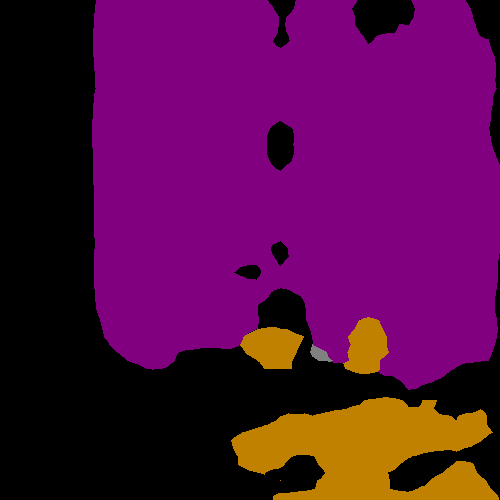} &
   \includegraphics[height=0.11\linewidth]{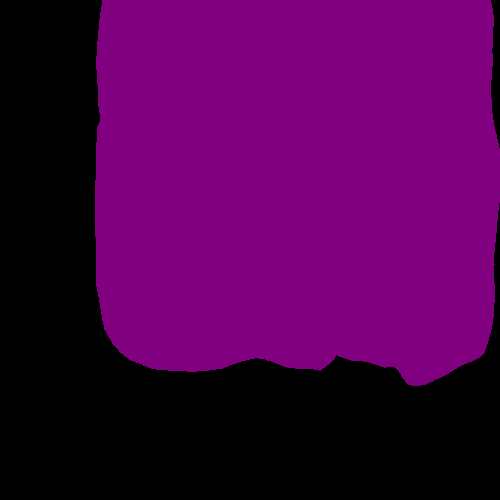} &
   \includegraphics[height=0.11\linewidth]{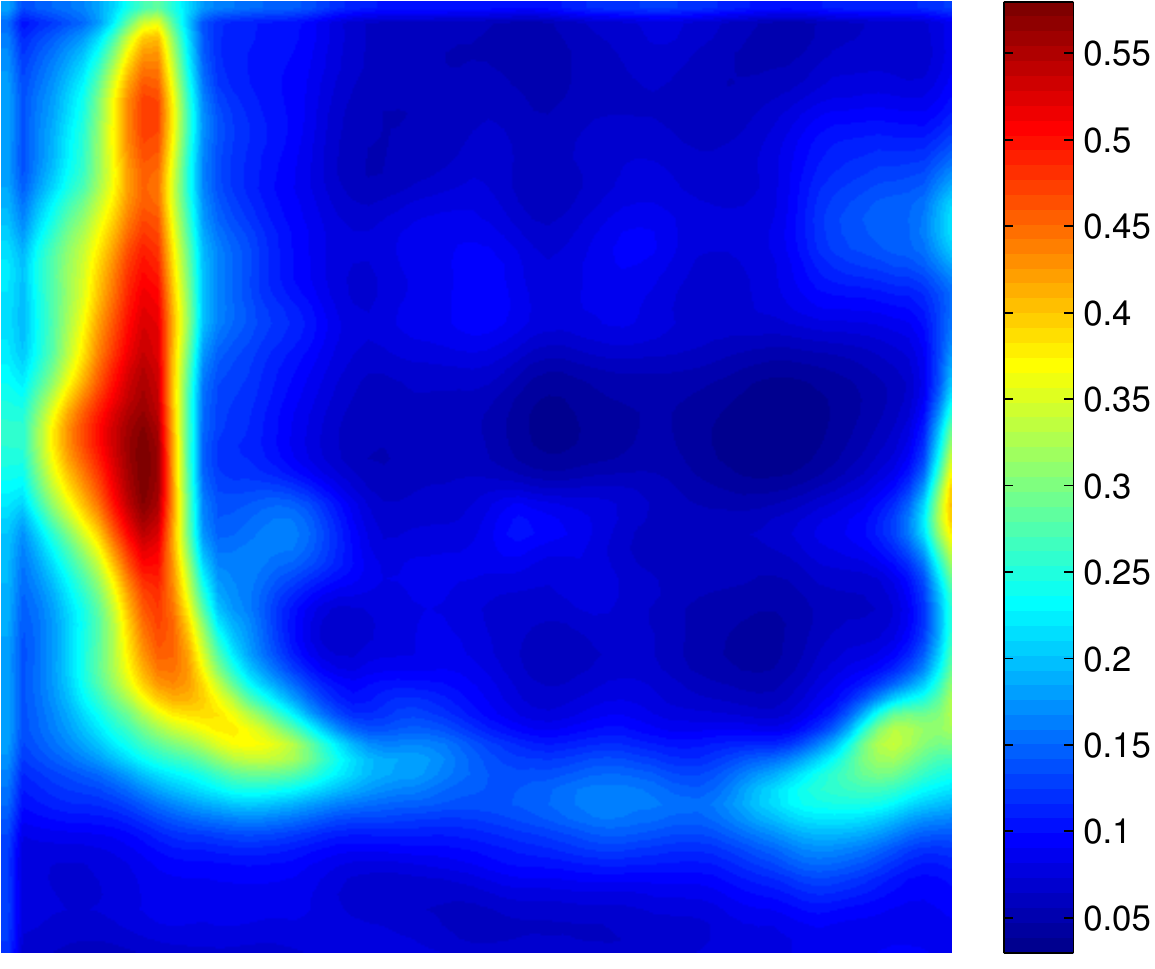} &
   \includegraphics[height=0.11\linewidth]{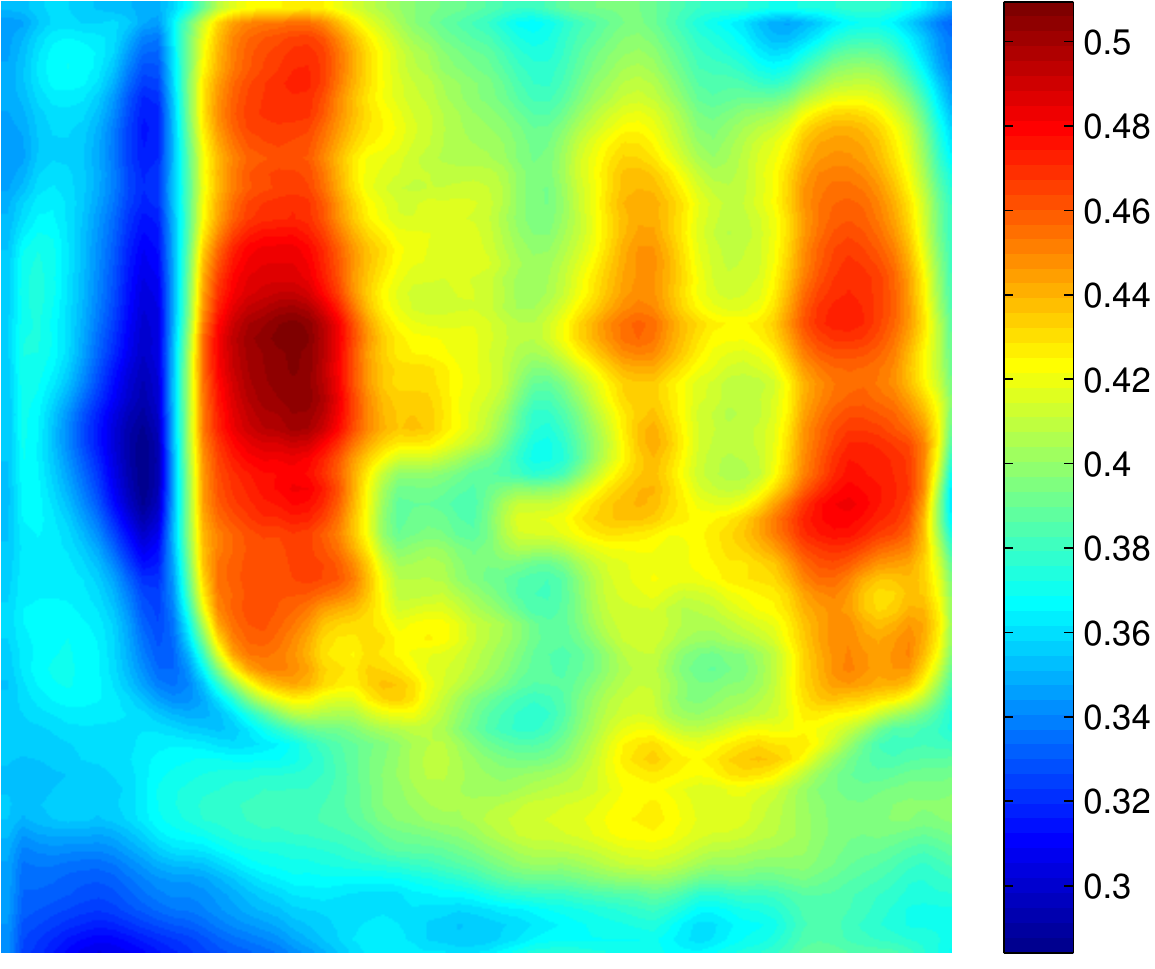} &
   \includegraphics[height=0.11\linewidth]{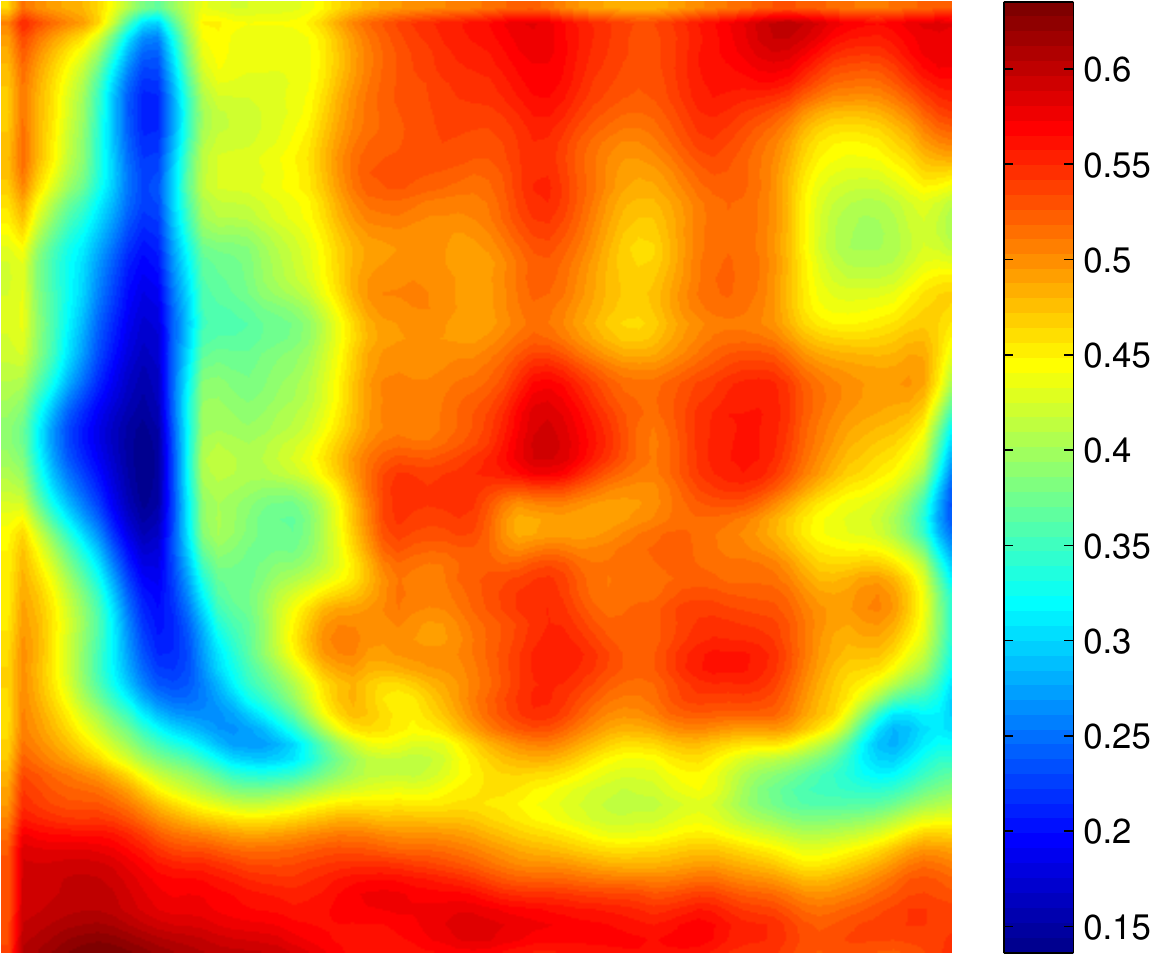} \\
   {\scriptsize (a) Image} & 
   {\scriptsize (b) Baseline} & 
   {\scriptsize (c) Our model} & 
   {\scriptsize (d) Scale-1 Attention} & 
   {\scriptsize (e) Scale-0.75 Attention} &
   {\scriptsize (f) Scale-0.5 Attention} \\
  \end{tabular}
  \caption{Qualitative segmentation results on PASCAL VOC 2012 validation set.} 
  \label{fig:pascal_voc12_results_app}  
\end{figure*}

\begin{figure*}
  \centering
  \begin{tabular}{c c c c c c}
   \includegraphics[height=0.1\linewidth]{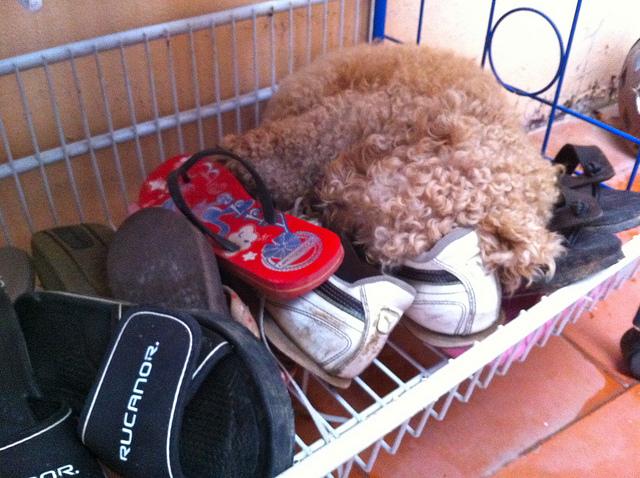} &
   \includegraphics[height=0.1\linewidth]{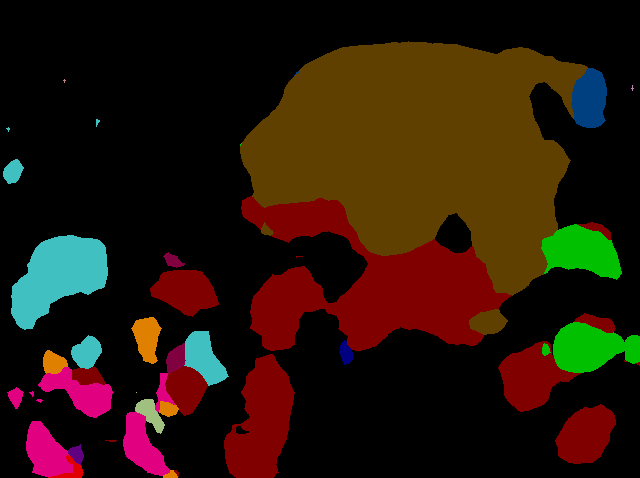} &
   \includegraphics[height=0.1\linewidth]{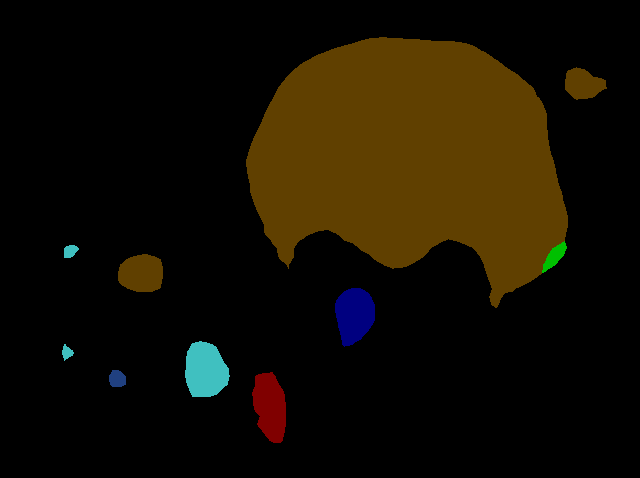} &
   \includegraphics[height=0.1\linewidth]{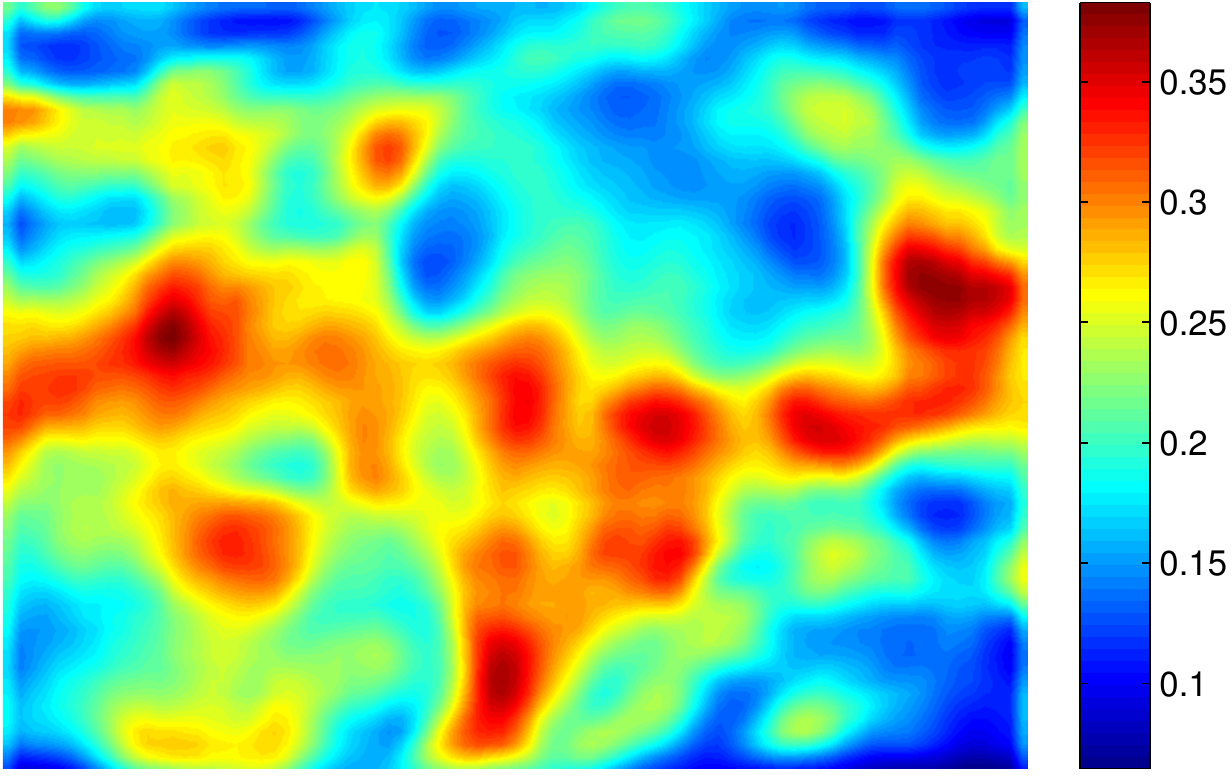} &
   \includegraphics[height=0.1\linewidth]{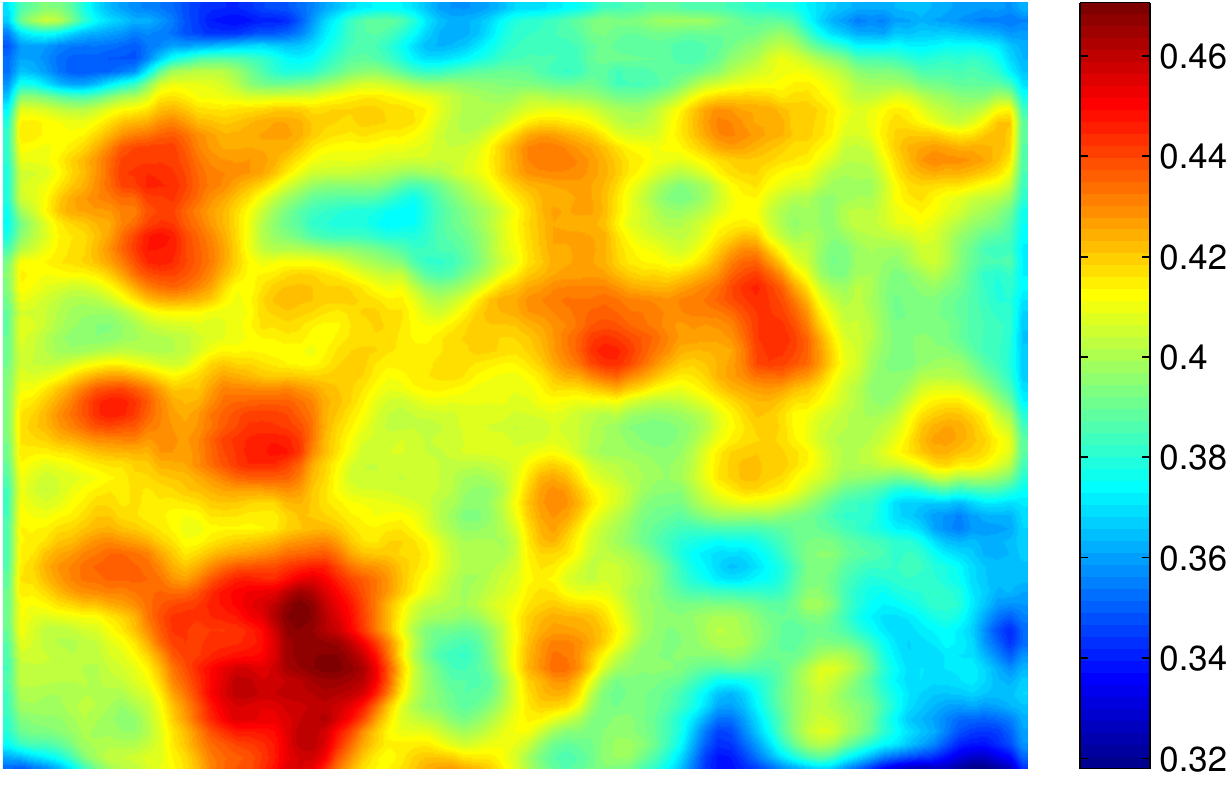} &
   \includegraphics[height=0.1\linewidth]{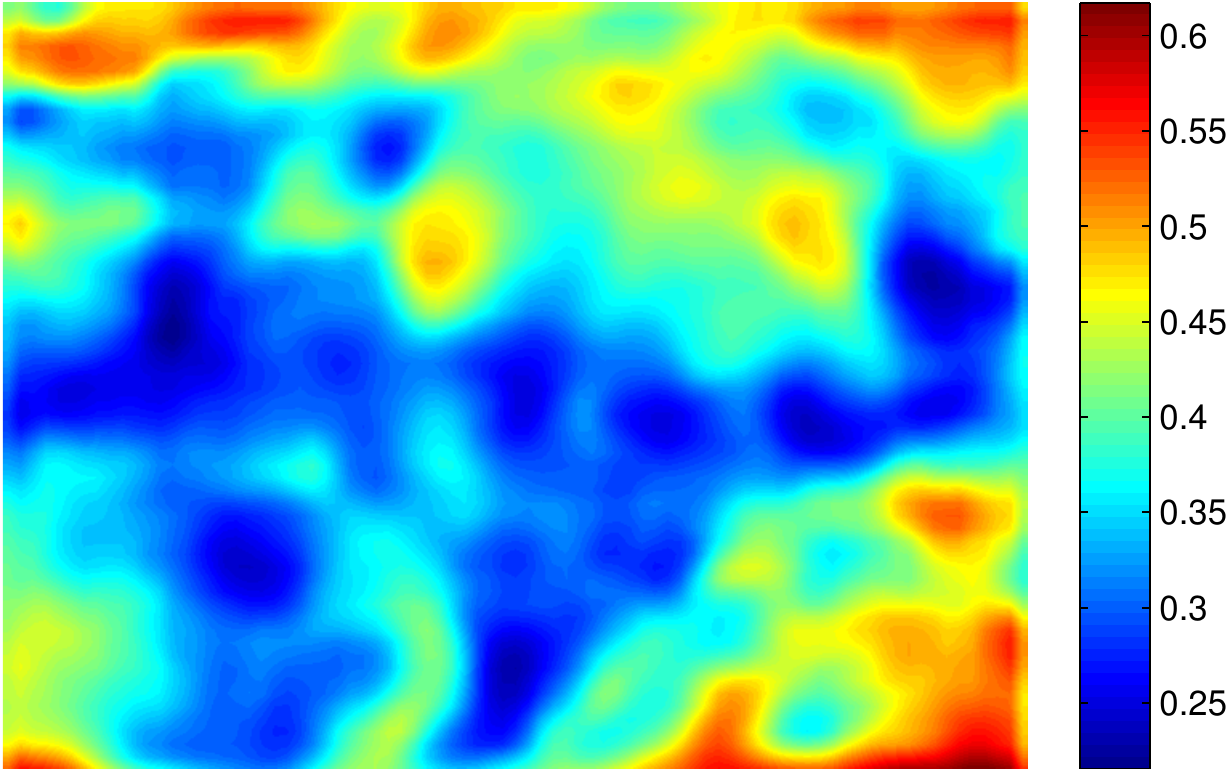} \\
   \includegraphics[height=0.087\linewidth]{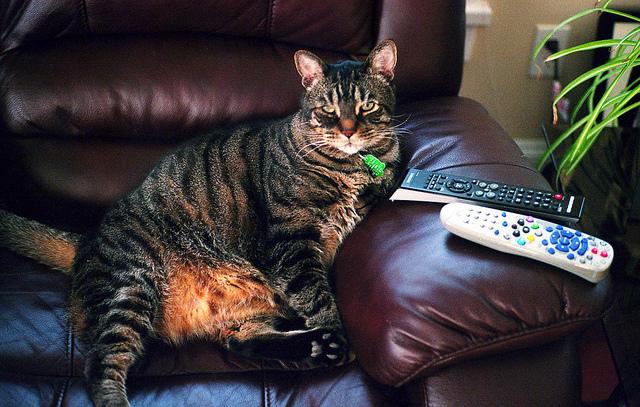} &
   \includegraphics[height=0.087\linewidth]{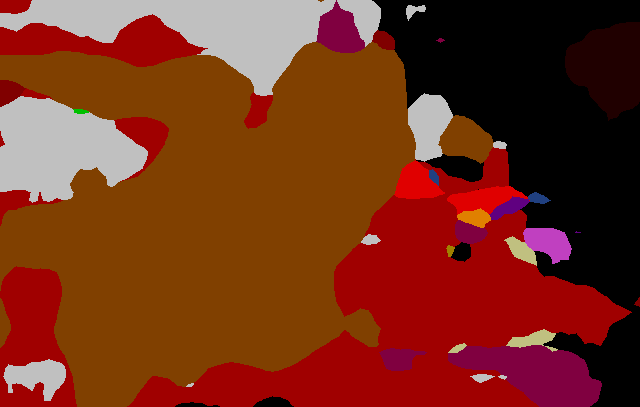} &
   \includegraphics[height=0.087\linewidth]{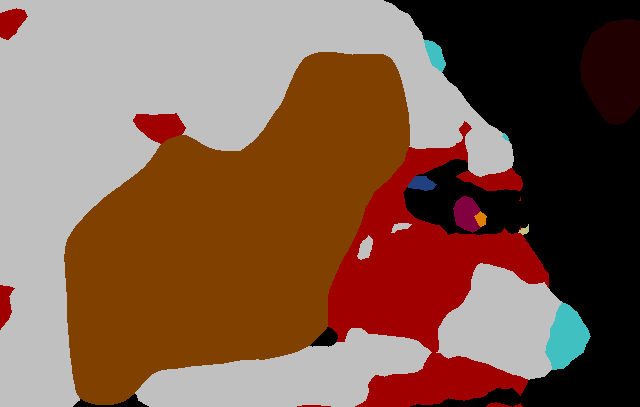} &
   \includegraphics[height=0.087\linewidth]{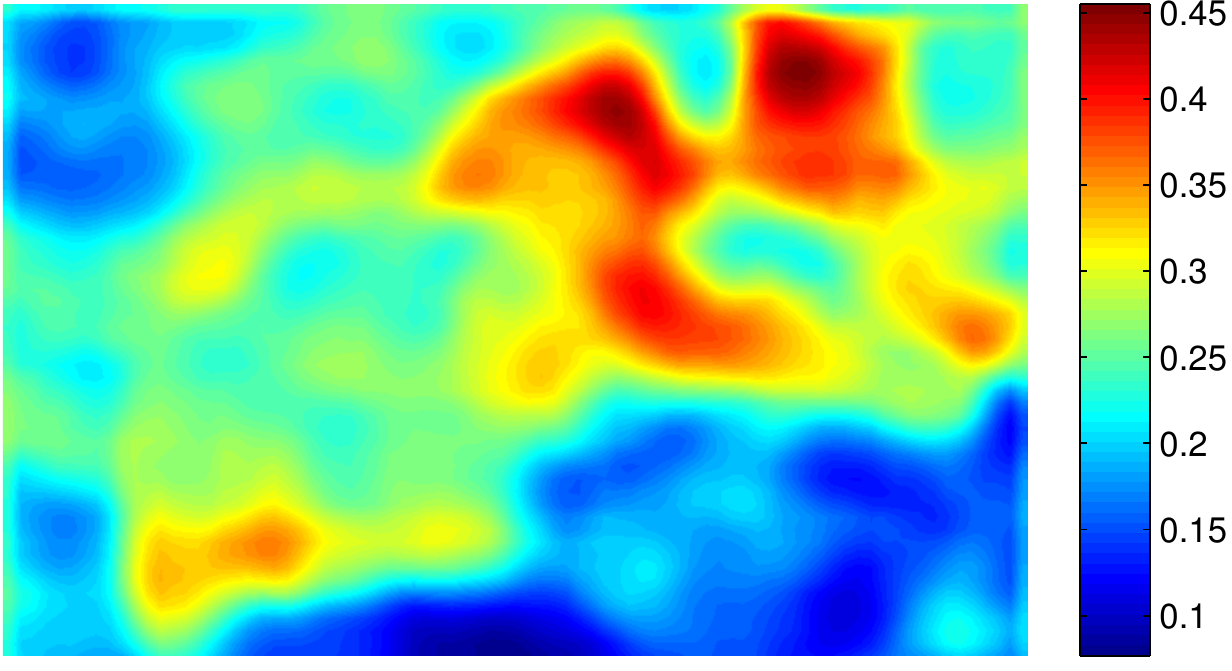} &
   \includegraphics[height=0.087\linewidth]{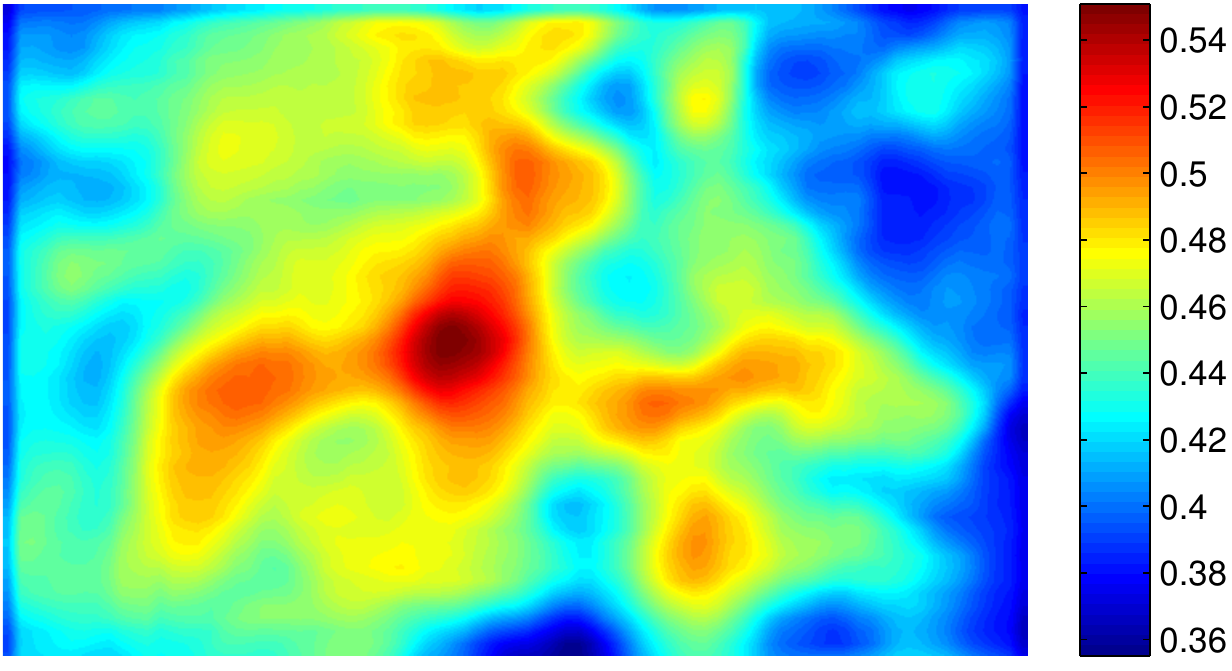} &
   \includegraphics[height=0.087\linewidth]{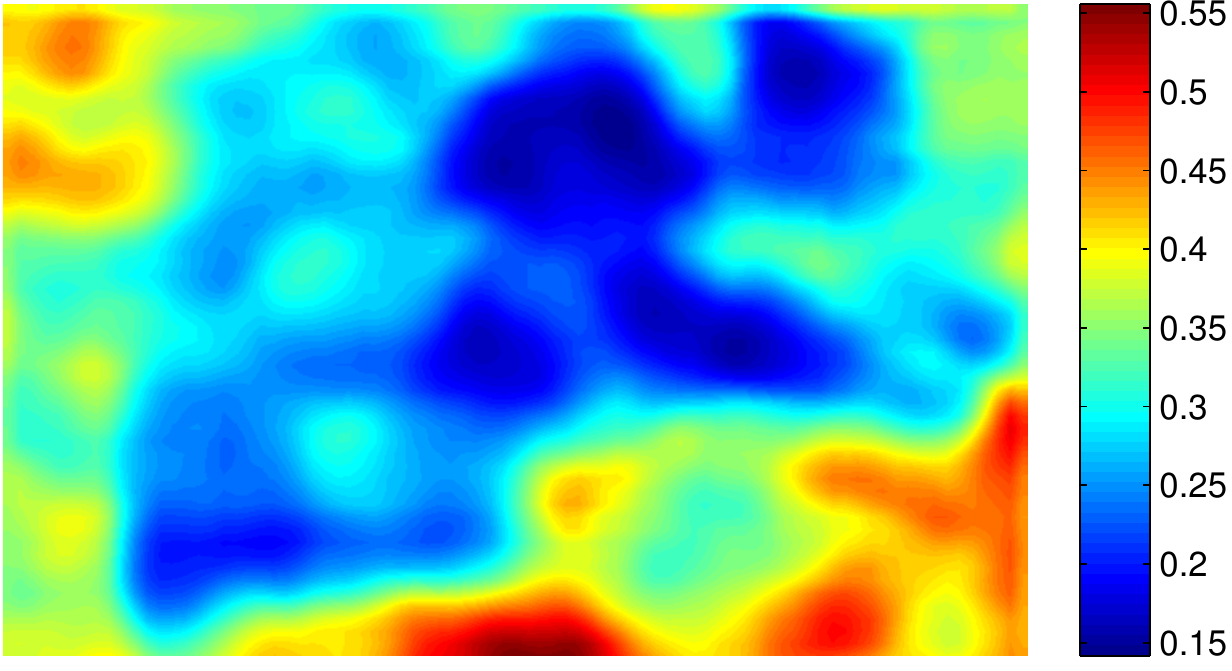} \\
   \includegraphics[height=0.14\linewidth]{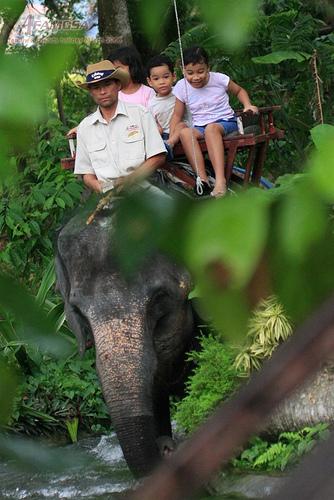} &
   \includegraphics[height=0.14\linewidth]{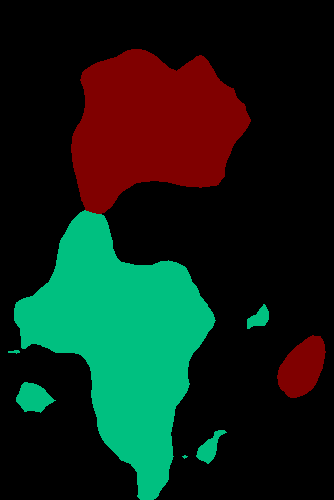} &
   \includegraphics[height=0.14\linewidth]{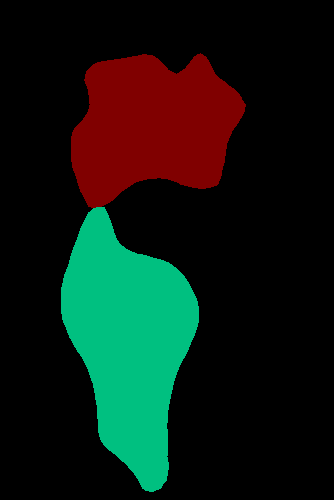} &
   \includegraphics[height=0.14\linewidth]{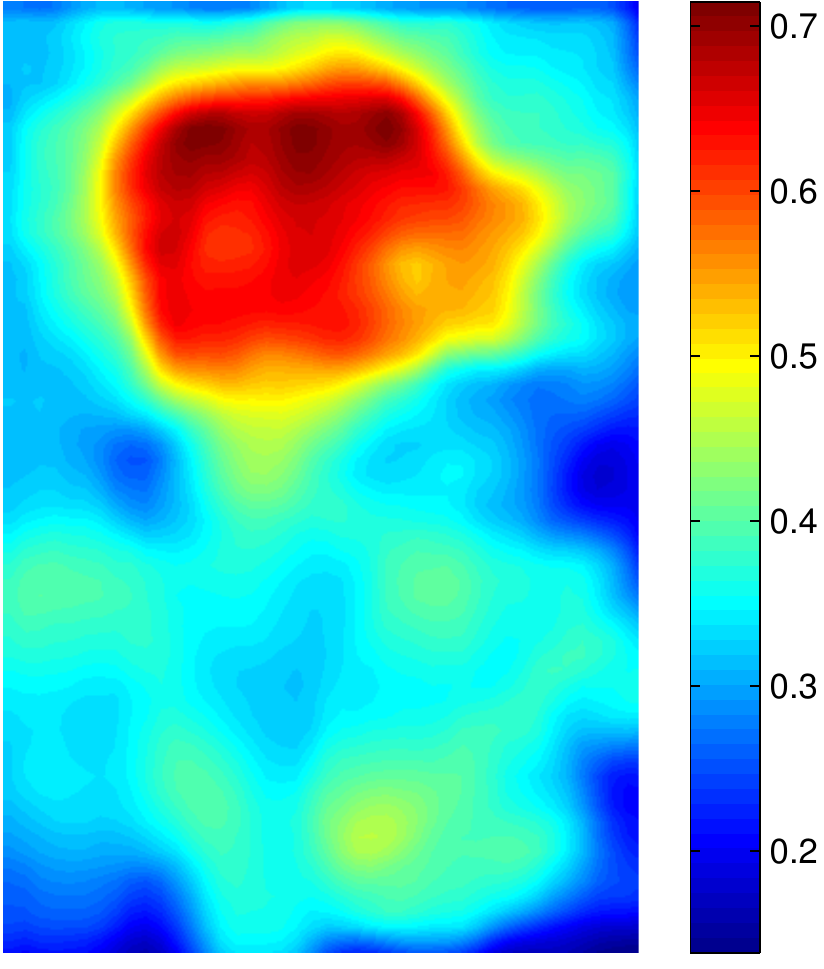} &
   \includegraphics[height=0.14\linewidth]{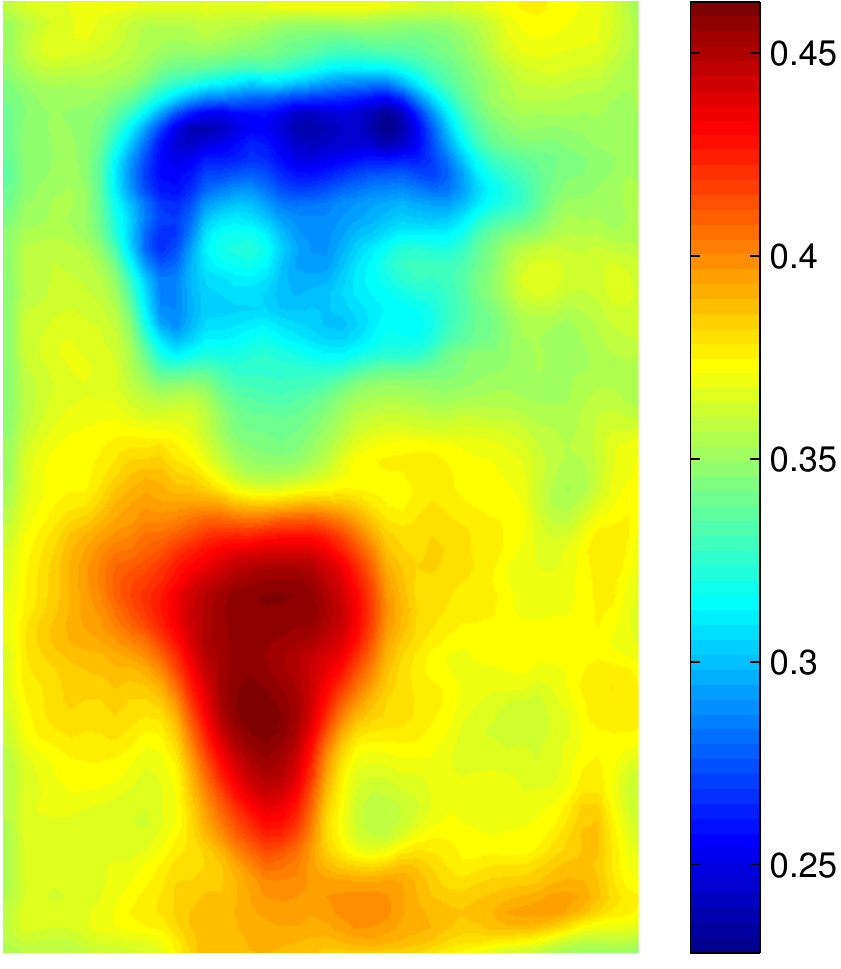} &
   \includegraphics[height=0.14\linewidth]{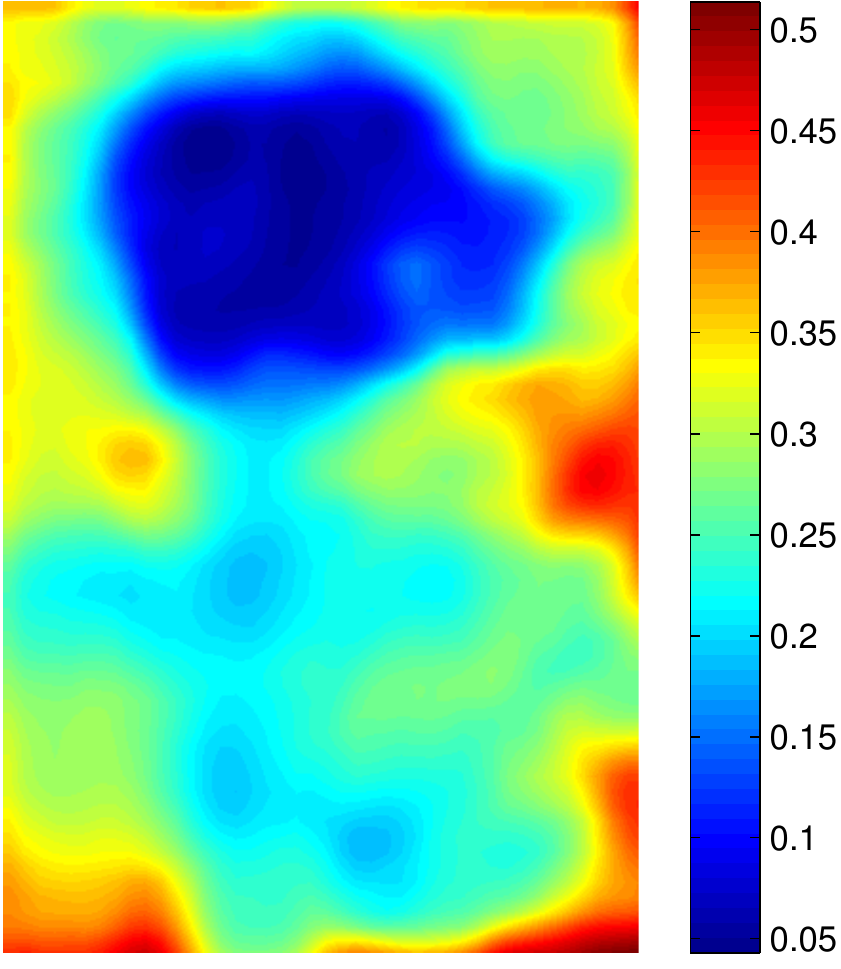} \\
   \includegraphics[height=0.1\linewidth]{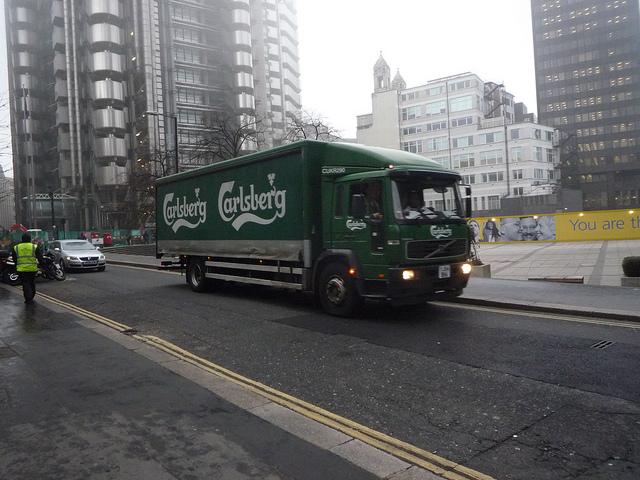} &
   \includegraphics[height=0.1\linewidth]{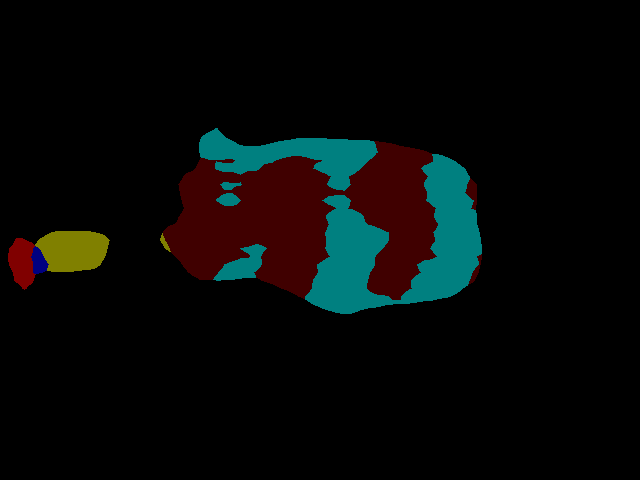} &
   \includegraphics[height=0.1\linewidth]{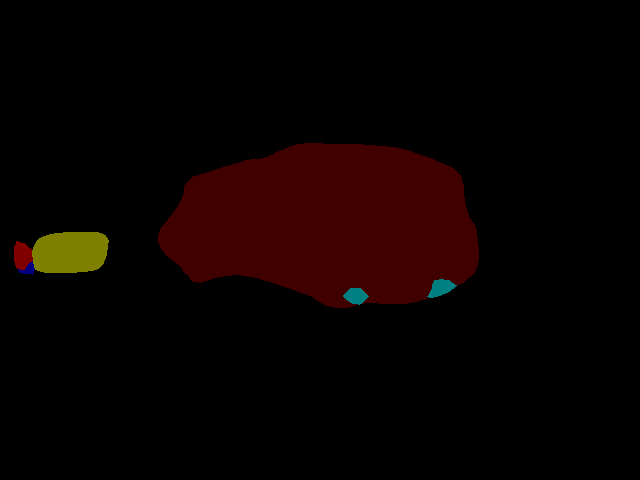} &
   \includegraphics[height=0.1\linewidth]{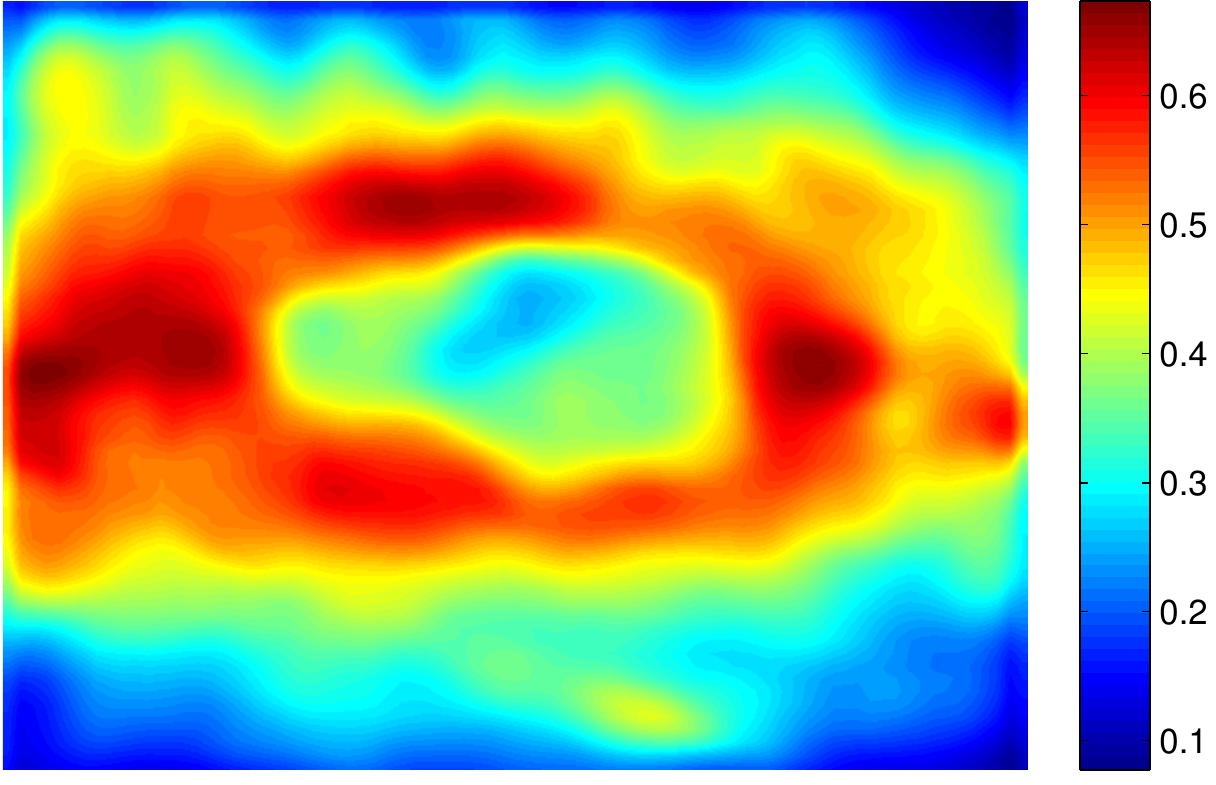} &
   \includegraphics[height=0.1\linewidth]{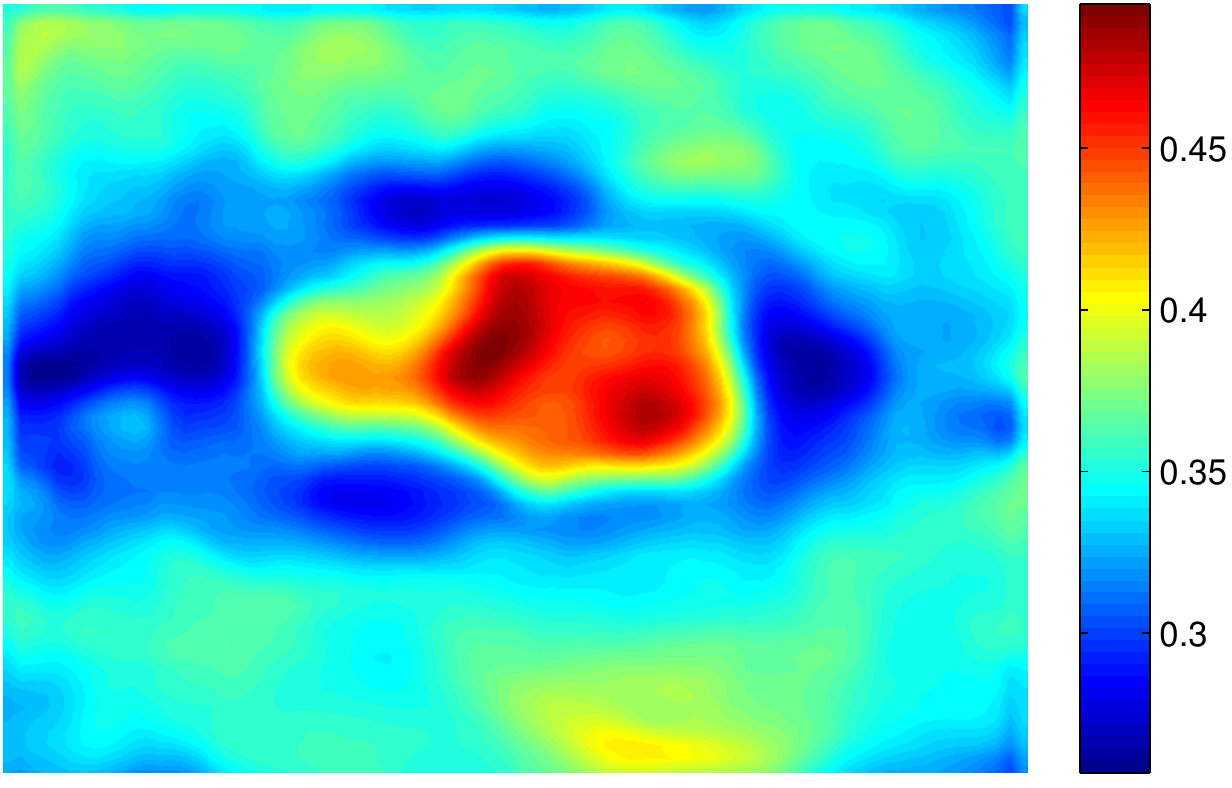} &
   \includegraphics[height=0.1\linewidth]{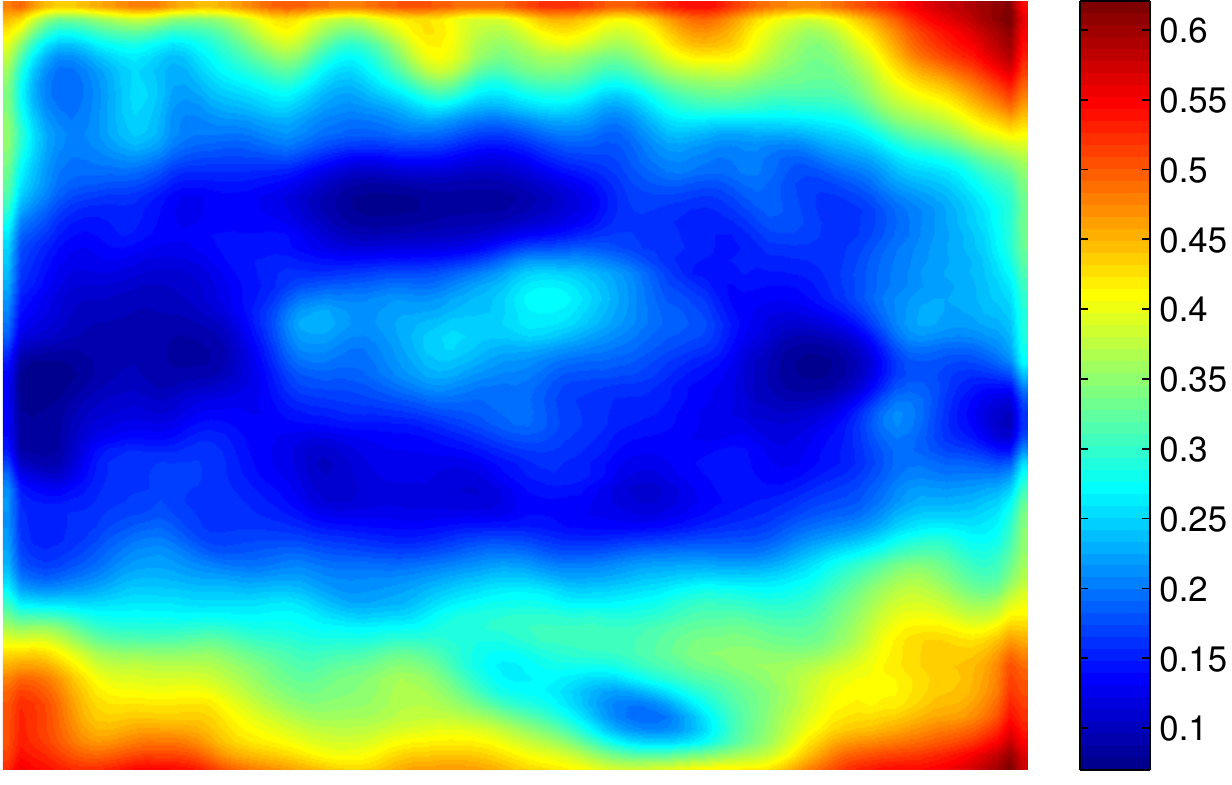} \\
   \includegraphics[height=0.12\linewidth]{fig/coco/img/COCO_val2014_000000001089.jpg} &
   \includegraphics[height=0.12\linewidth]{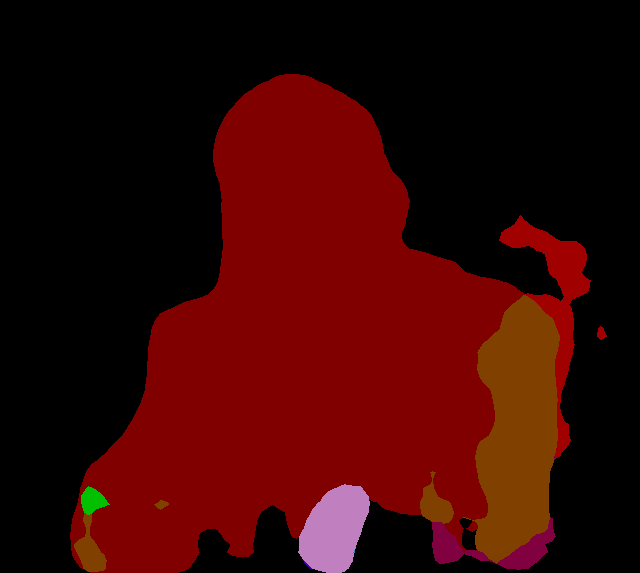} &
   \includegraphics[height=0.12\linewidth]{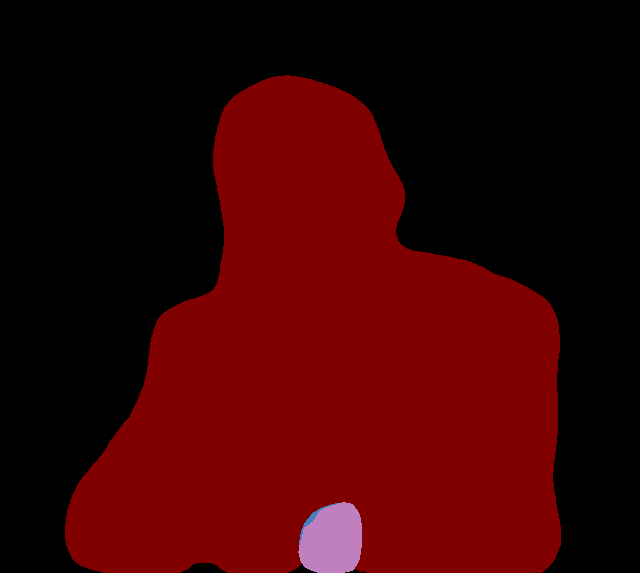} &
   \includegraphics[height=0.12\linewidth]{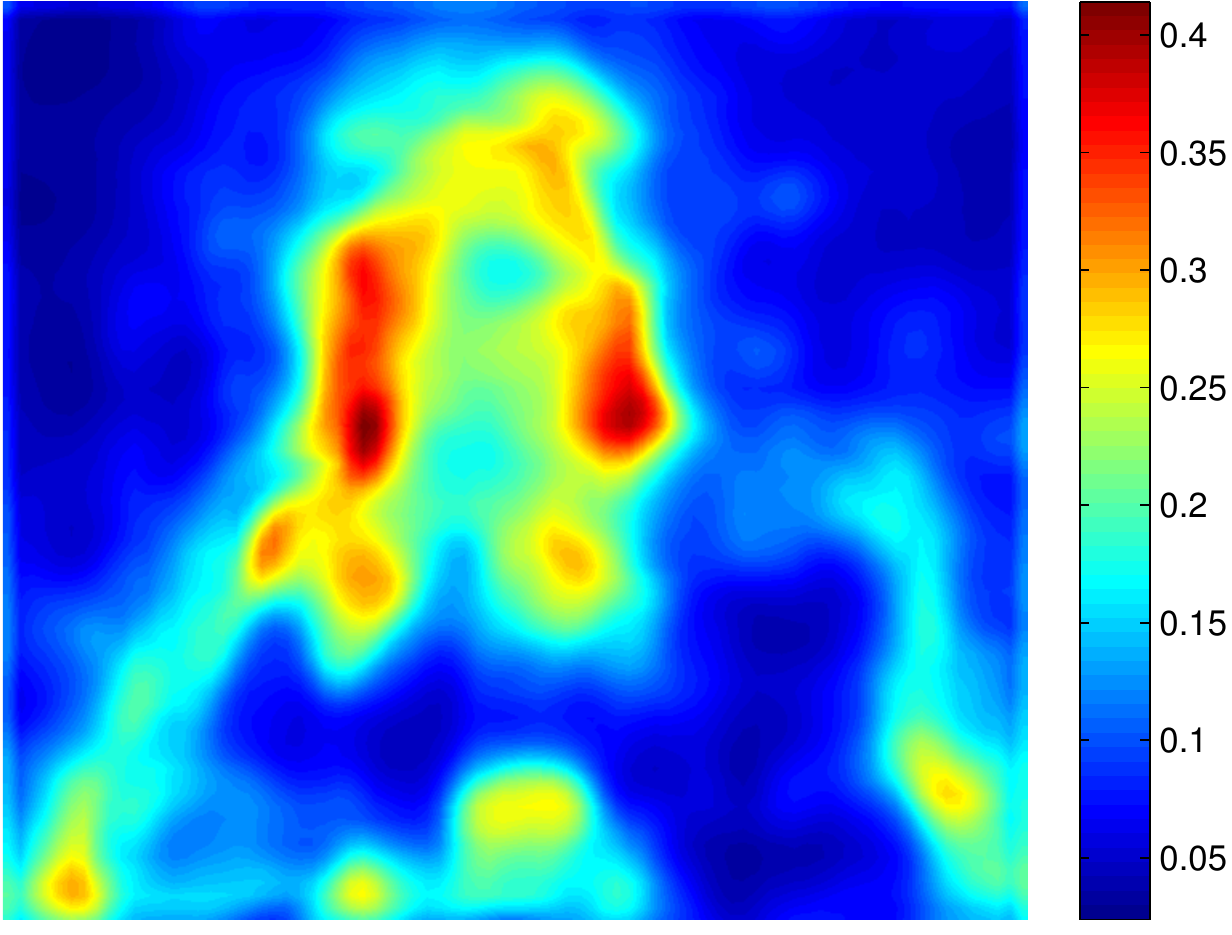} &
   \includegraphics[height=0.12\linewidth]{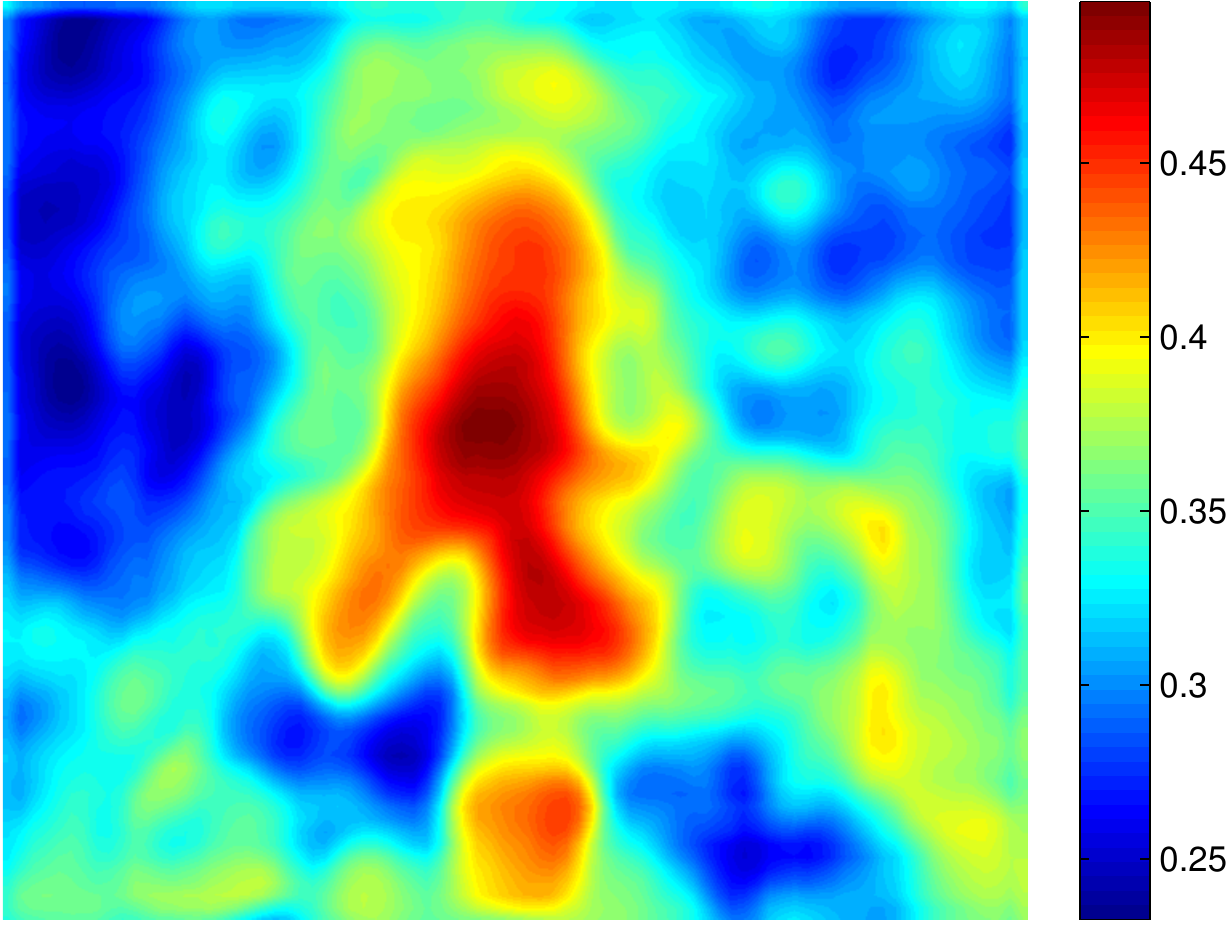} &
   \includegraphics[height=0.12\linewidth]{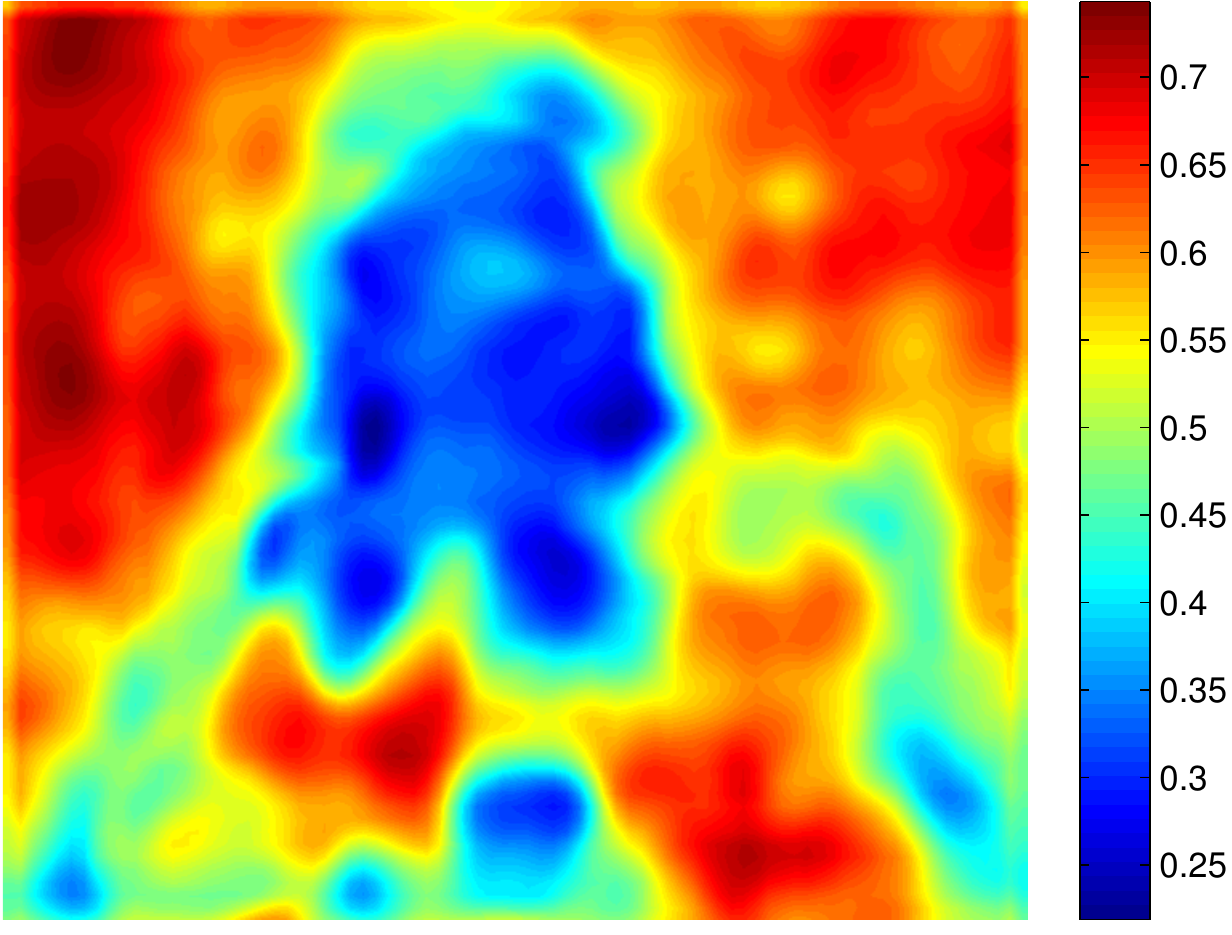} \\
   \includegraphics[height=0.13\linewidth]{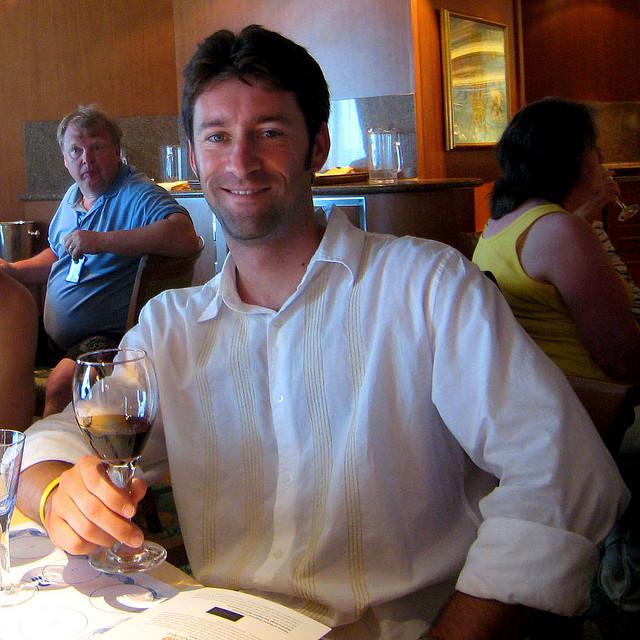} &
   \includegraphics[height=0.13\linewidth]{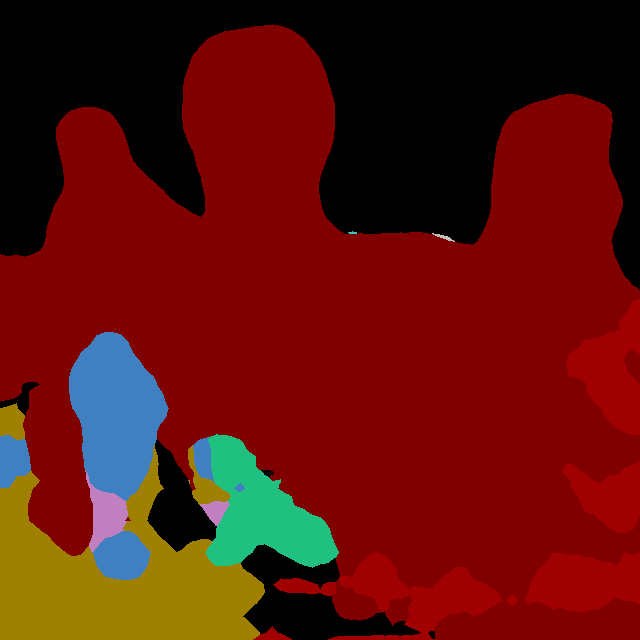} &
   \includegraphics[height=0.13\linewidth]{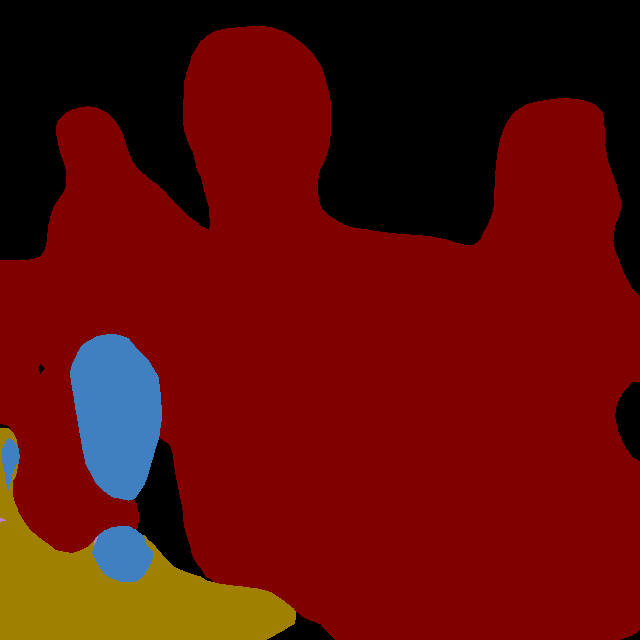} &
   \includegraphics[height=0.13\linewidth]{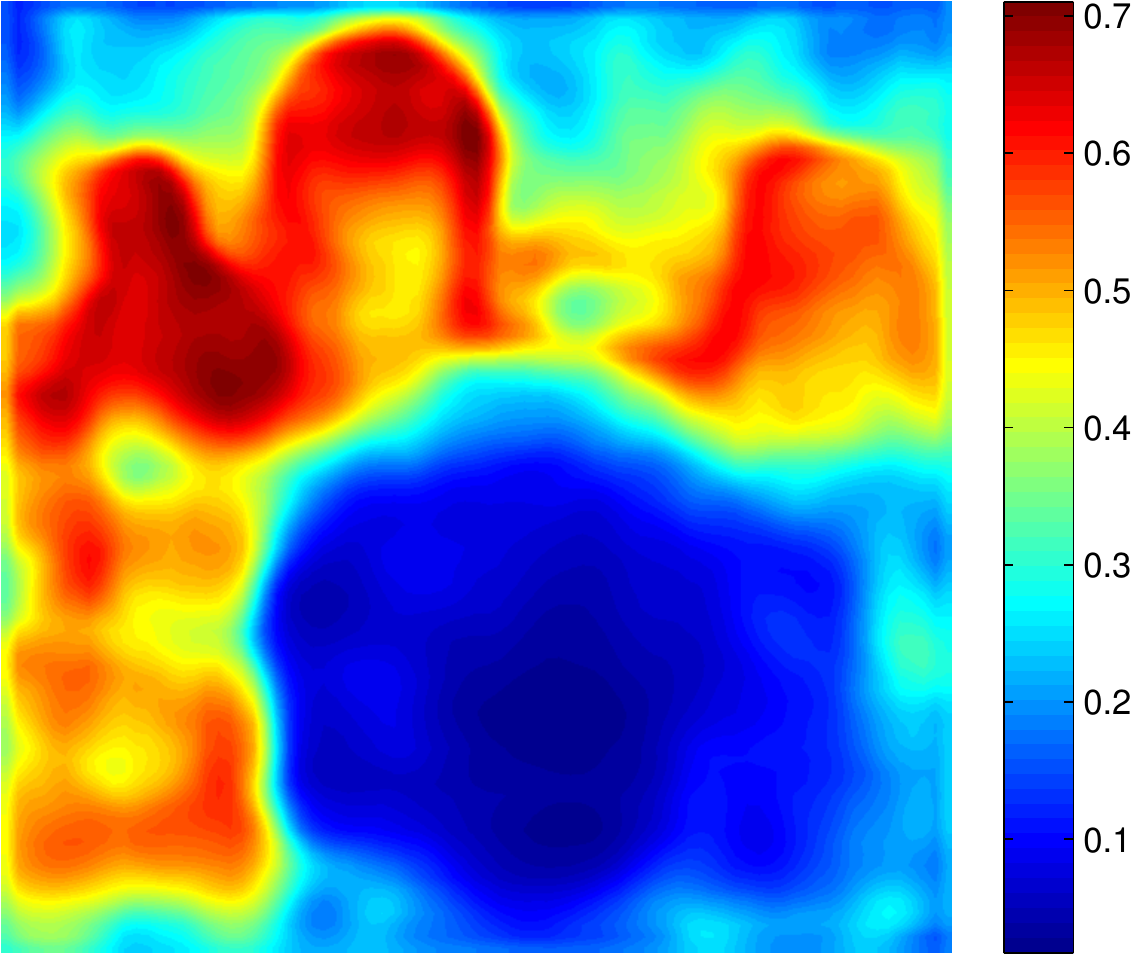} &
   \includegraphics[height=0.13\linewidth]{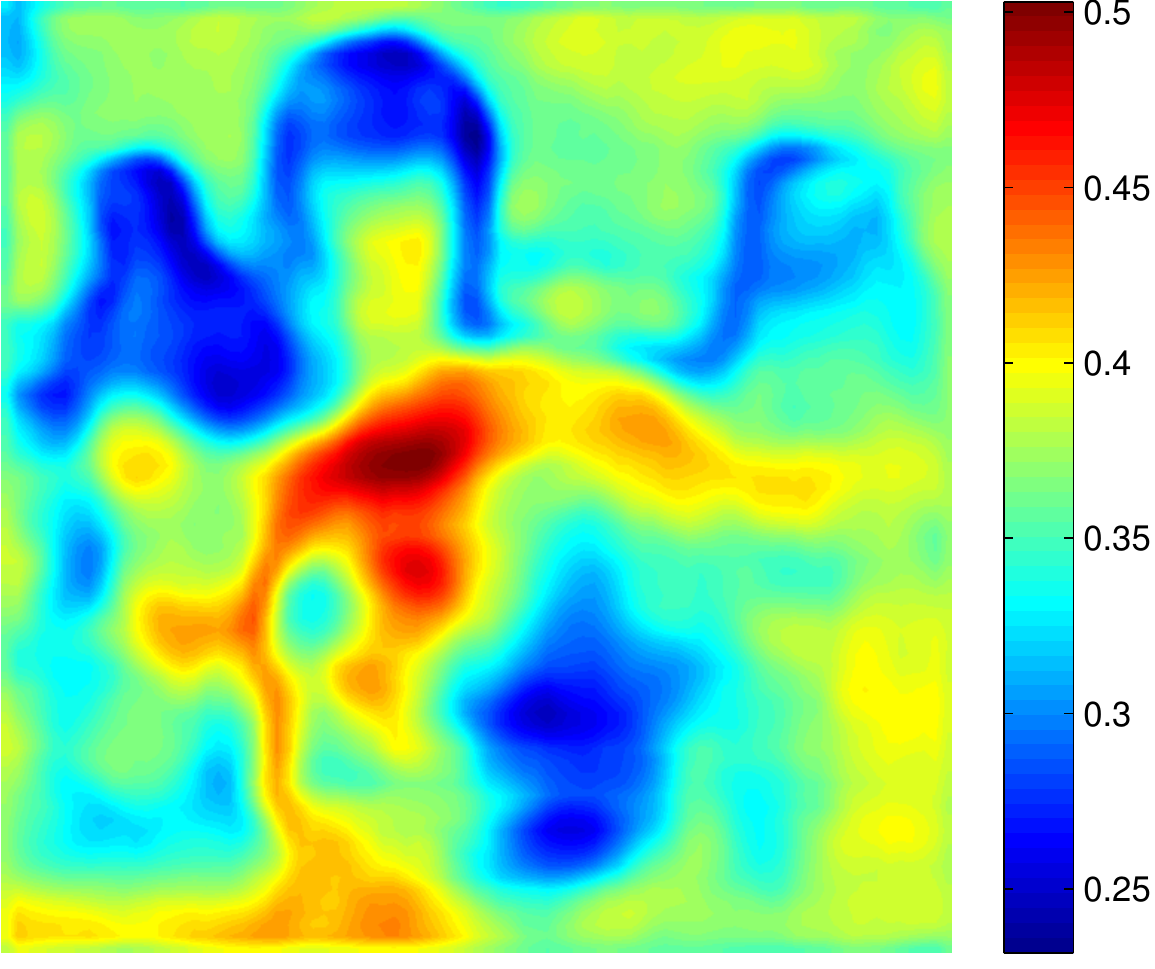} &
   \includegraphics[height=0.13\linewidth]{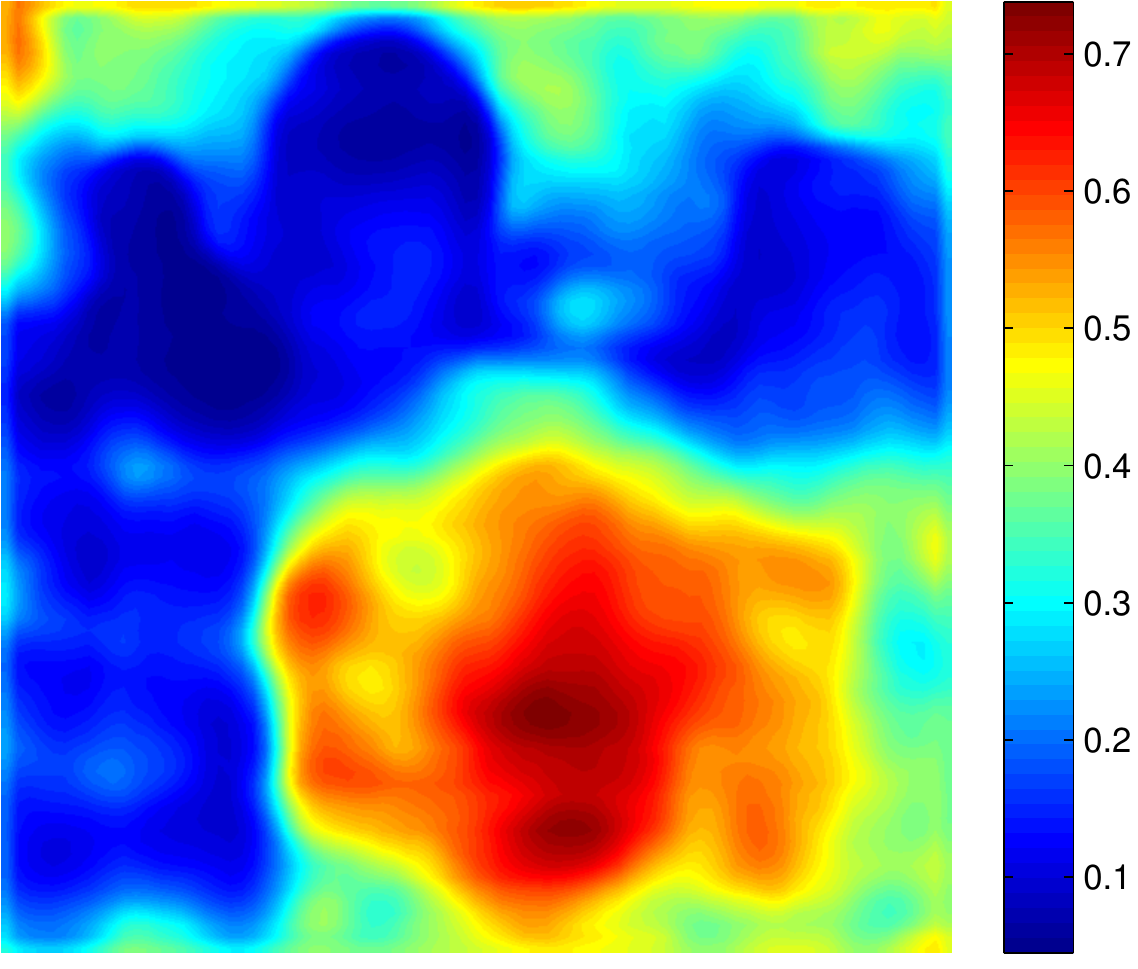} \\
   \includegraphics[height=0.087\linewidth]{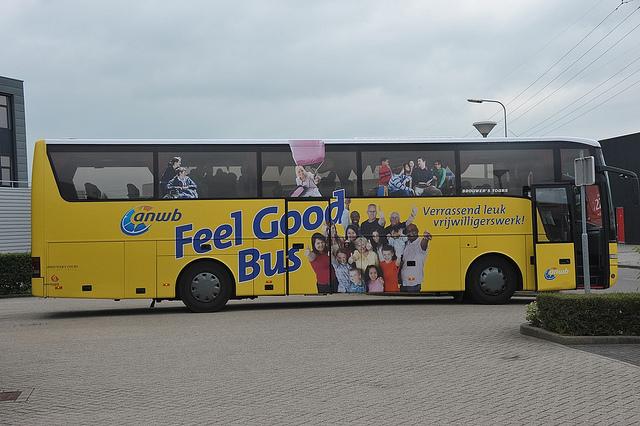} &
   \includegraphics[height=0.087\linewidth]{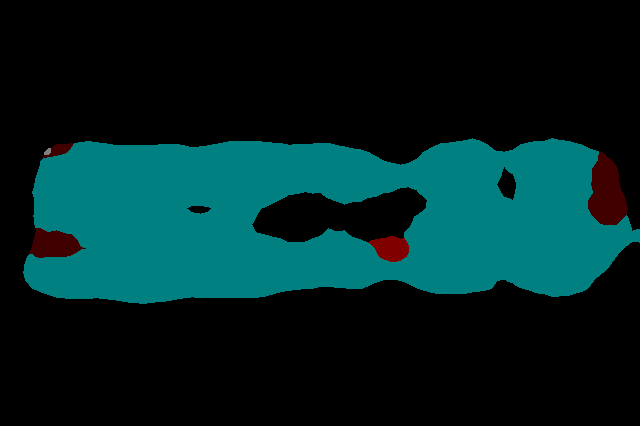} &
   \includegraphics[height=0.087\linewidth]{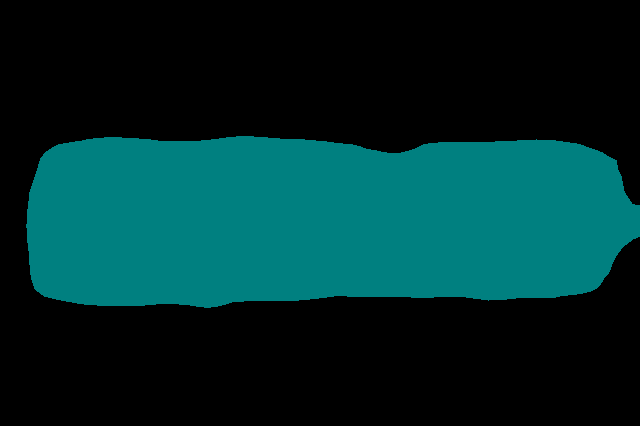} &
   \includegraphics[height=0.087\linewidth]{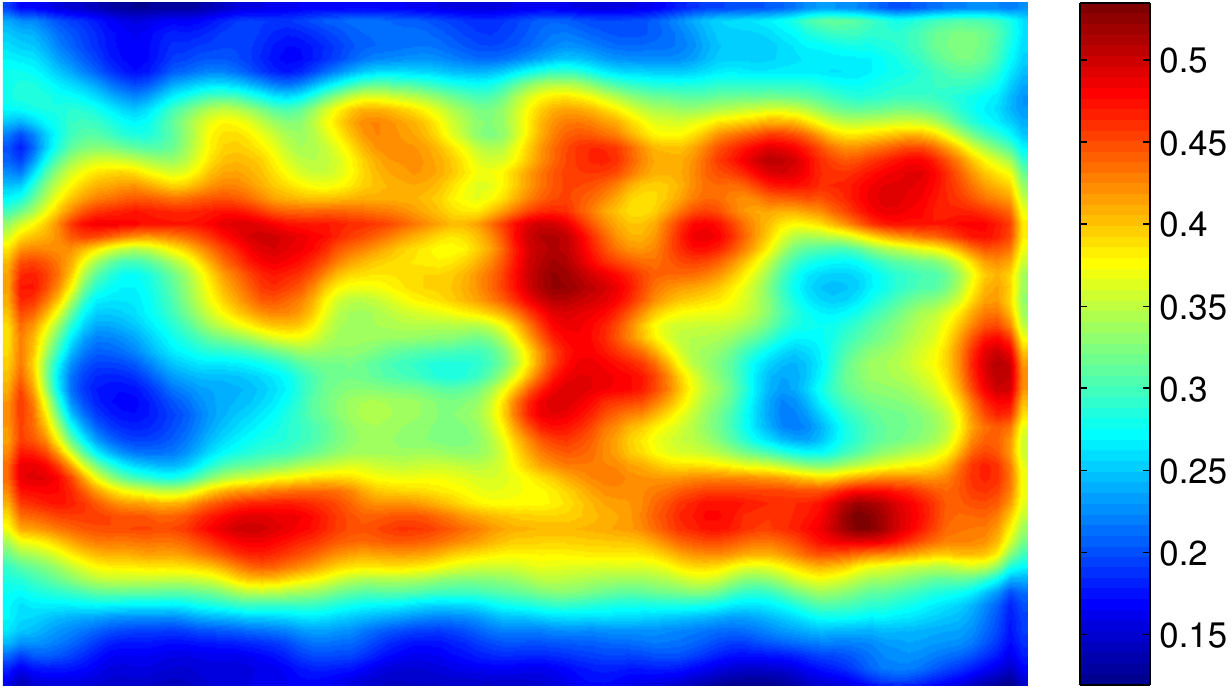} &
   \includegraphics[height=0.087\linewidth]{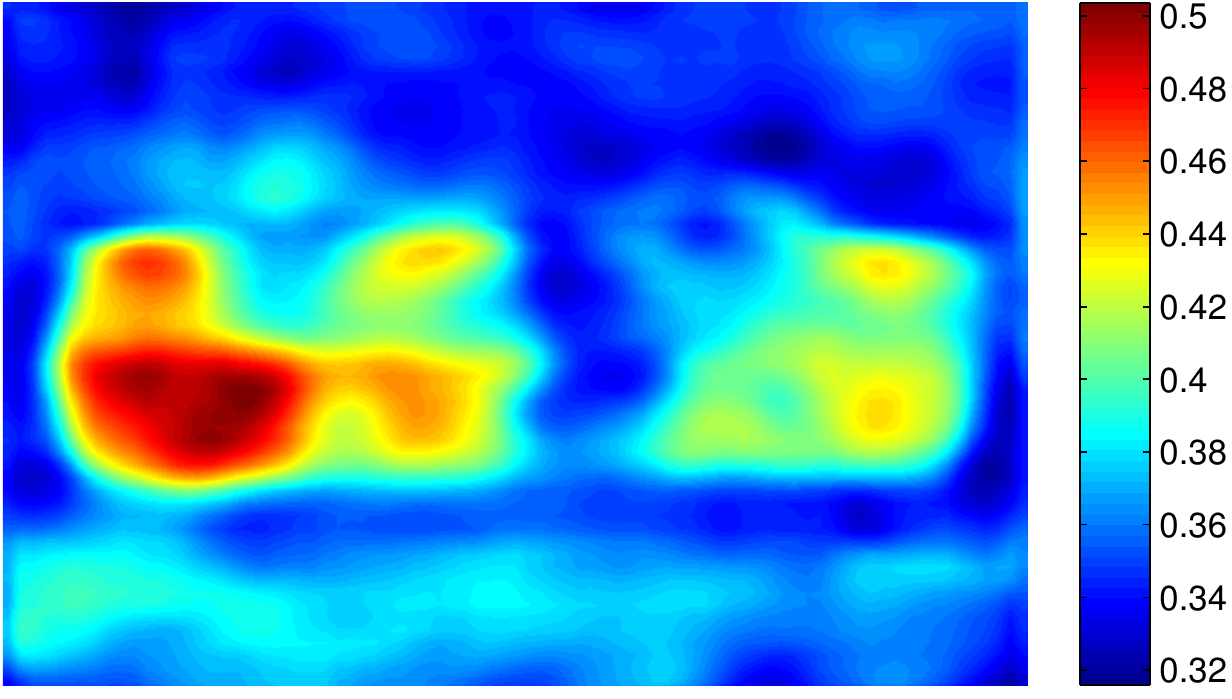} &
   \includegraphics[height=0.087\linewidth]{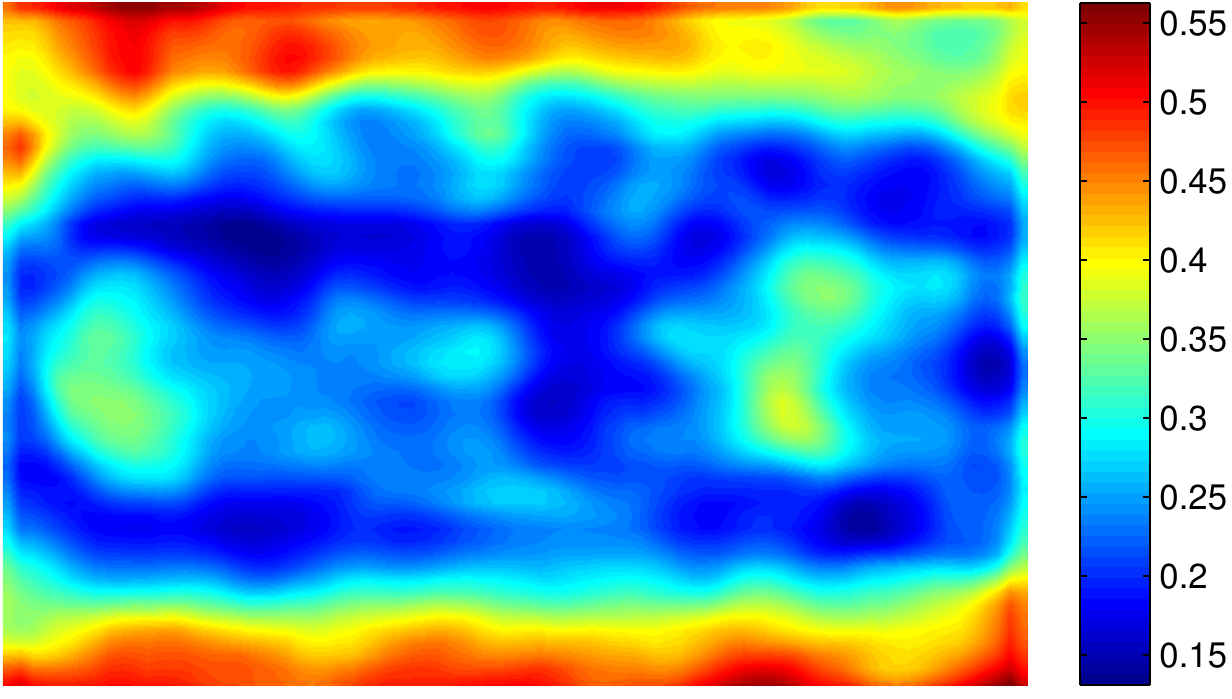} \\
   \includegraphics[height=0.13\linewidth]{fig/coco/img/COCO_val2014_000000012748.jpg} &
   \includegraphics[height=0.13\linewidth]{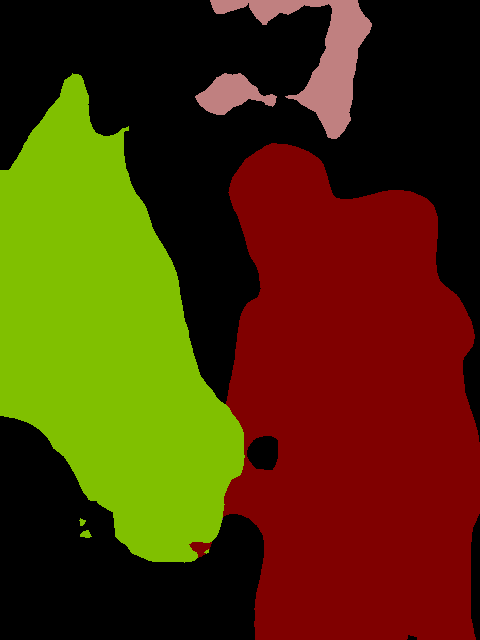} &
   \includegraphics[height=0.13\linewidth]{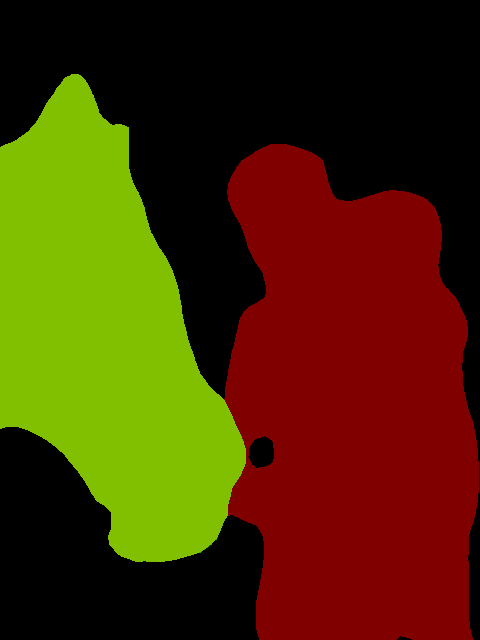} &
   \includegraphics[height=0.13\linewidth]{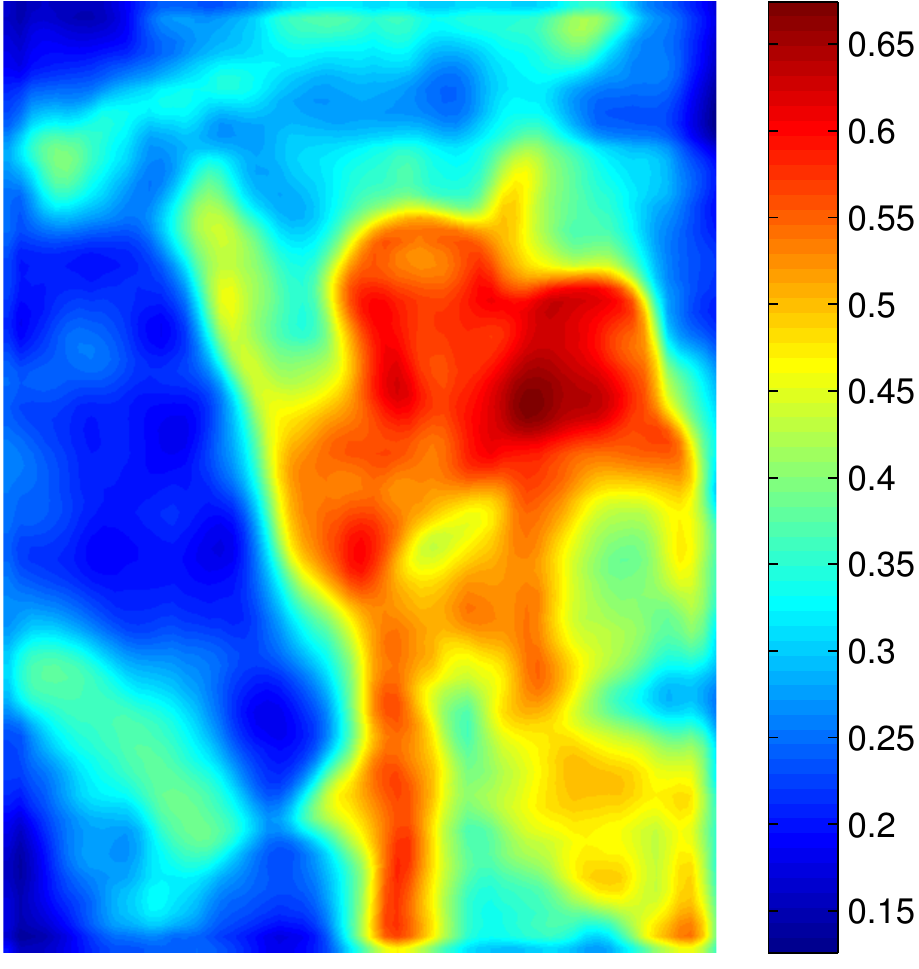} &
   \includegraphics[height=0.13\linewidth]{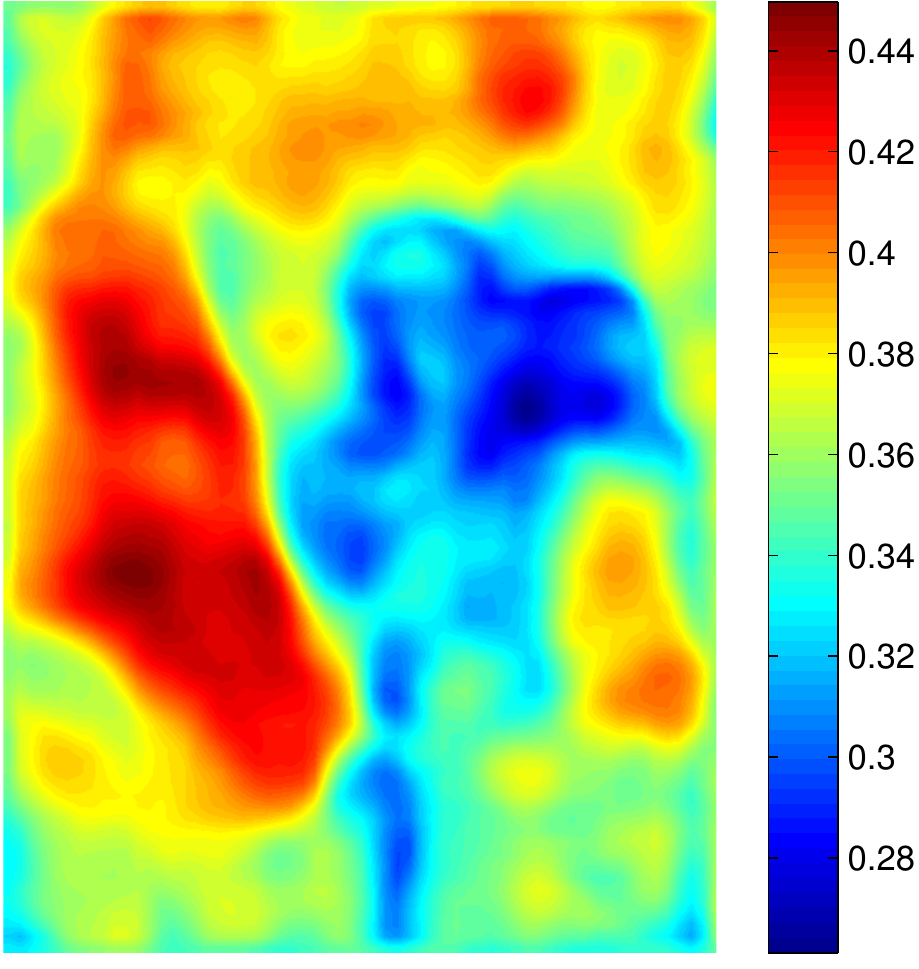} &
   \includegraphics[height=0.13\linewidth]{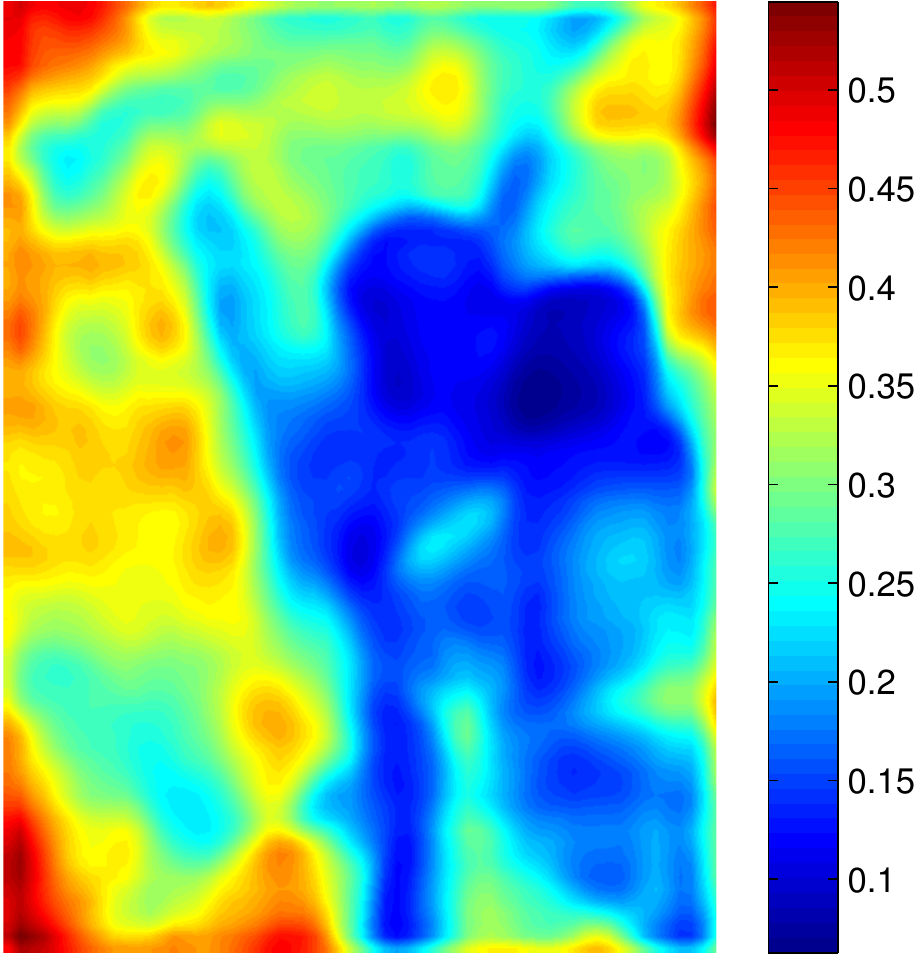} \\
   \includegraphics[height=0.1\linewidth]{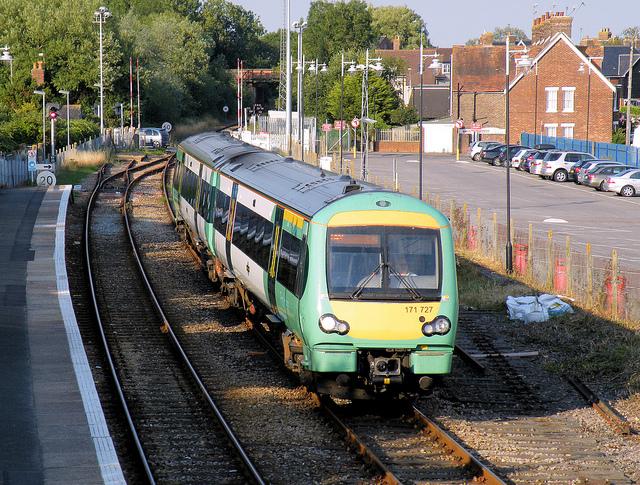} &
   \includegraphics[height=0.1\linewidth]{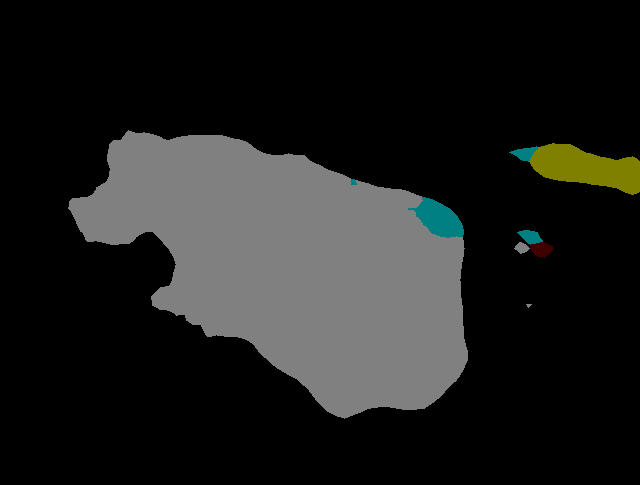} &
   \includegraphics[height=0.1\linewidth]{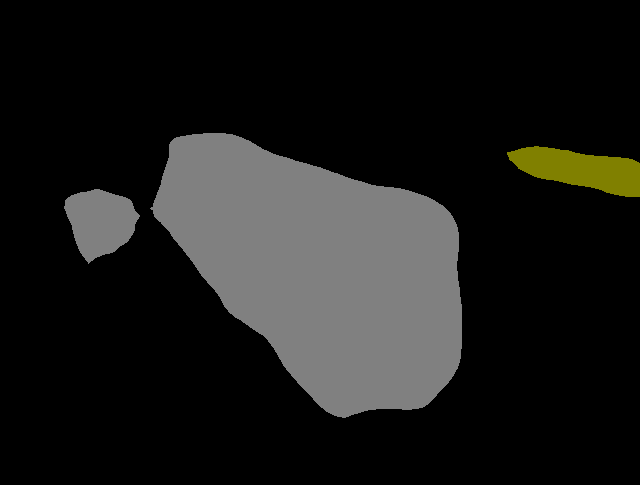} &
   \includegraphics[height=0.1\linewidth]{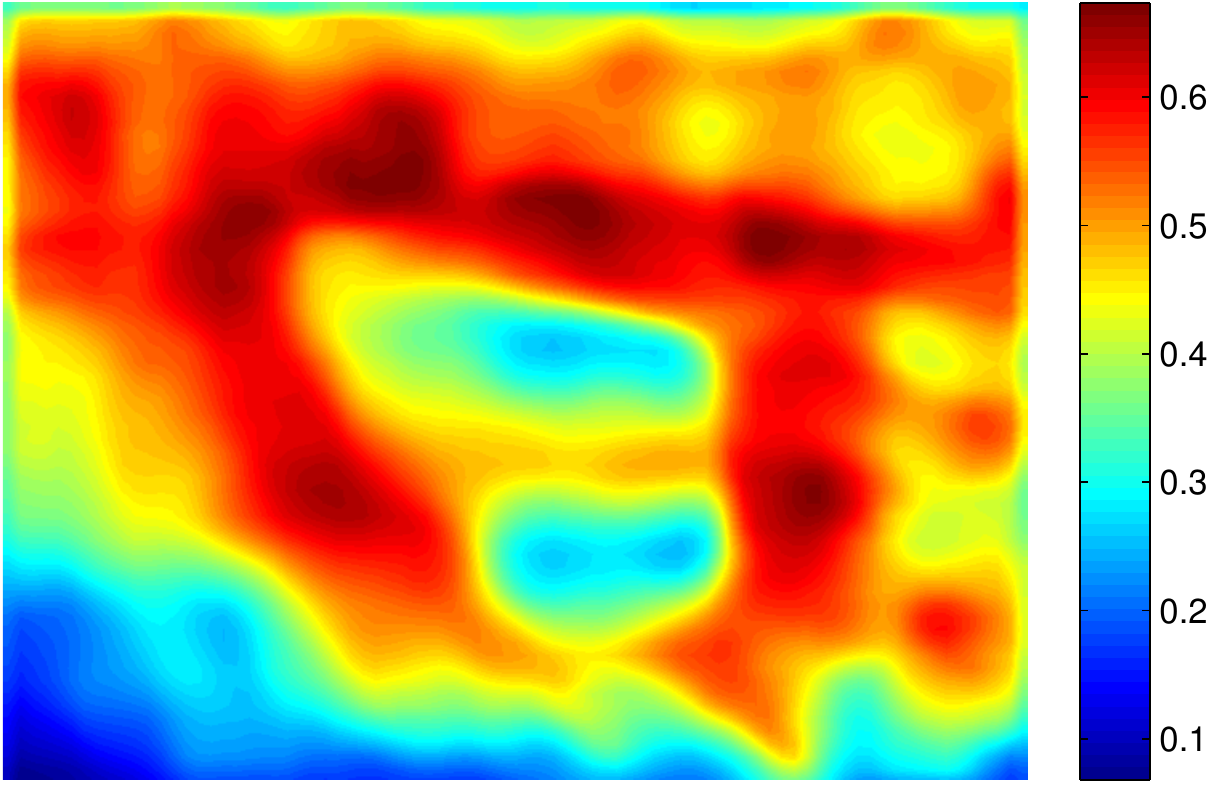} &
   \includegraphics[height=0.1\linewidth]{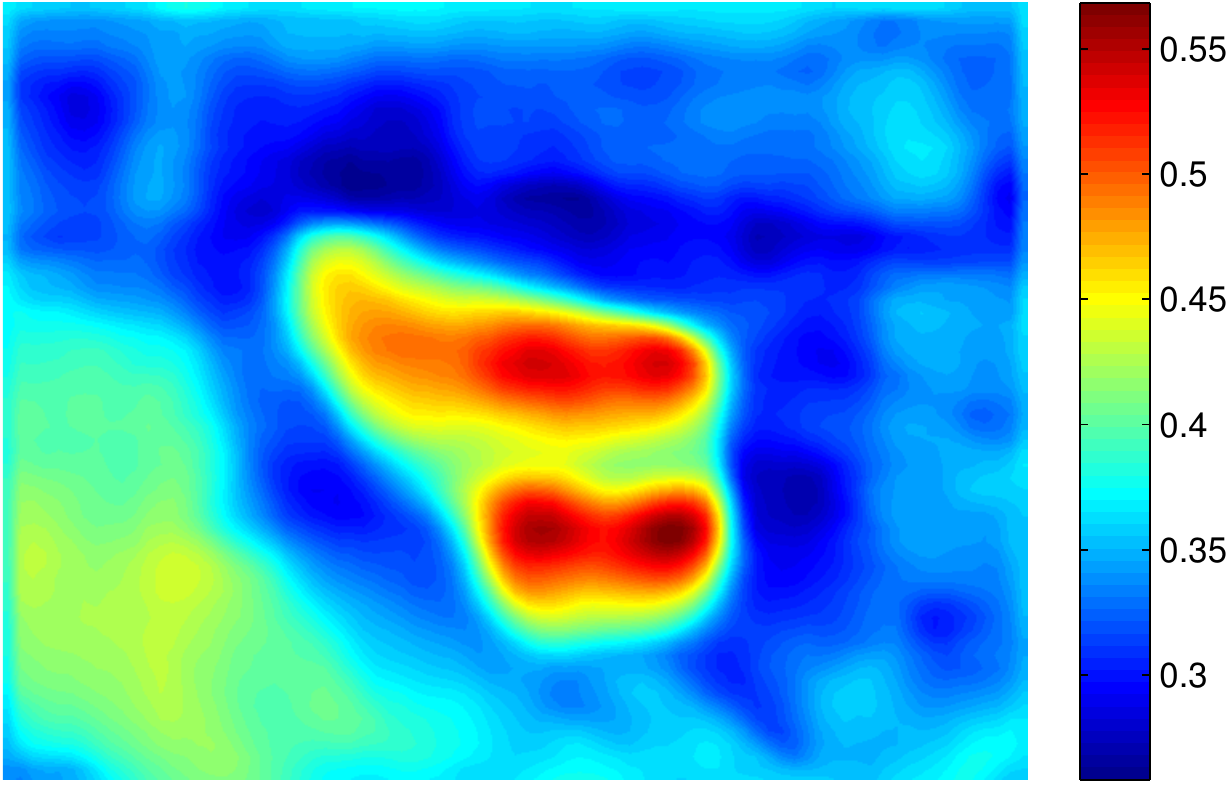} &
   \includegraphics[height=0.1\linewidth]{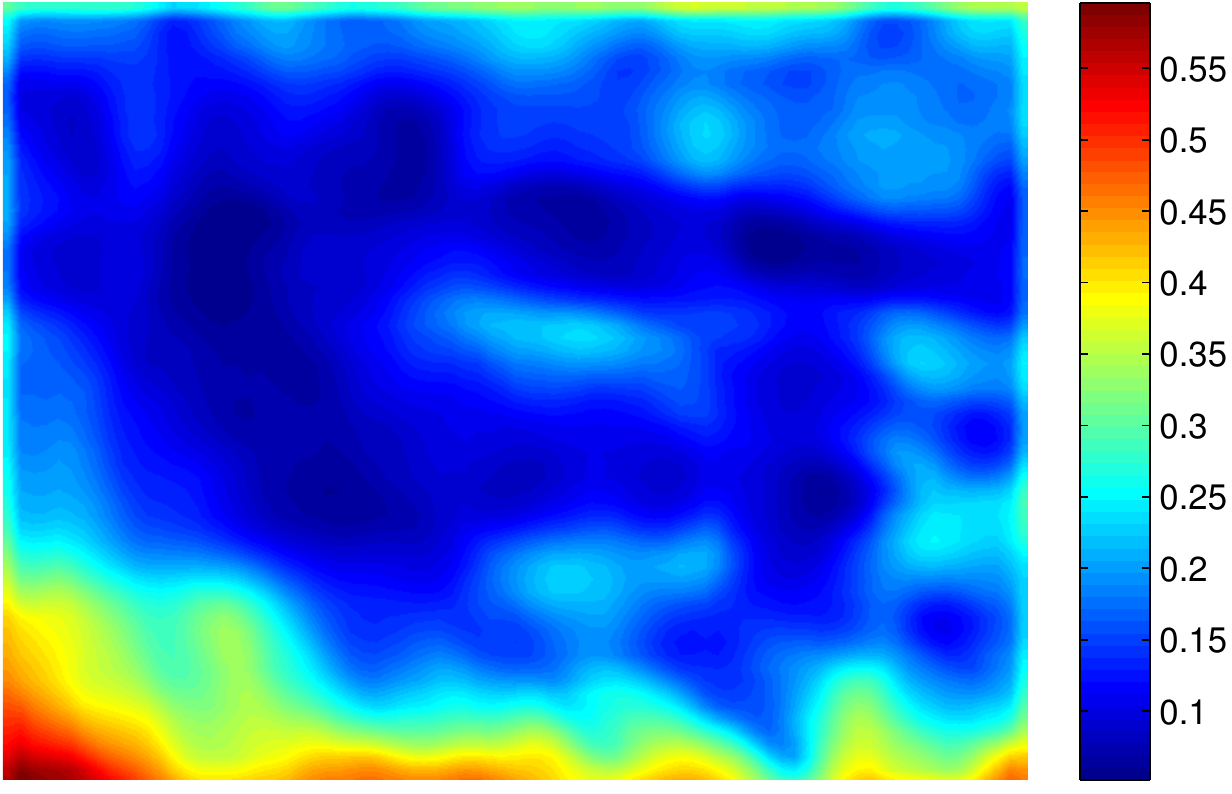} \\
   \includegraphics[height=0.09\linewidth]{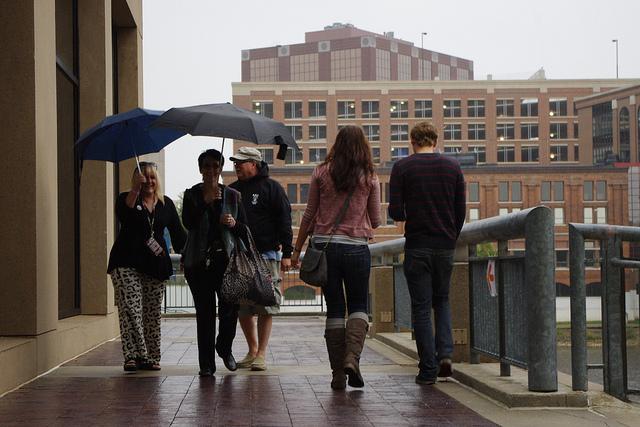} &
   \includegraphics[height=0.09\linewidth]{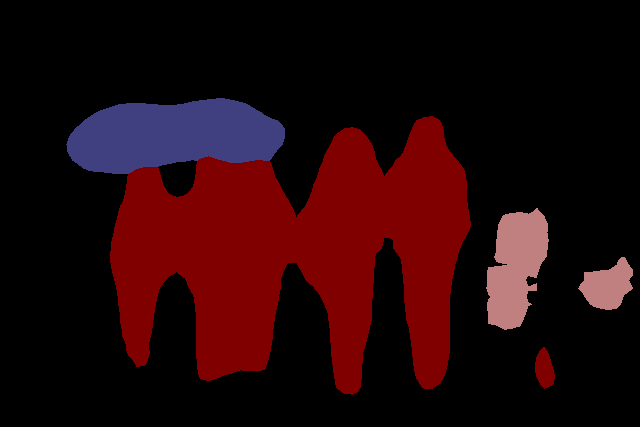} &
   \includegraphics[height=0.09\linewidth]{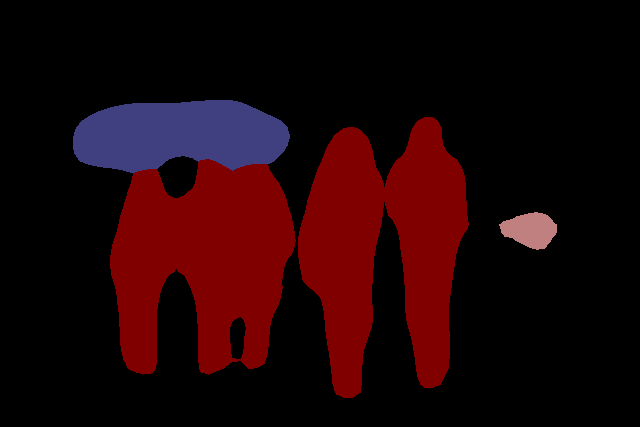} &
   \includegraphics[height=0.09\linewidth]{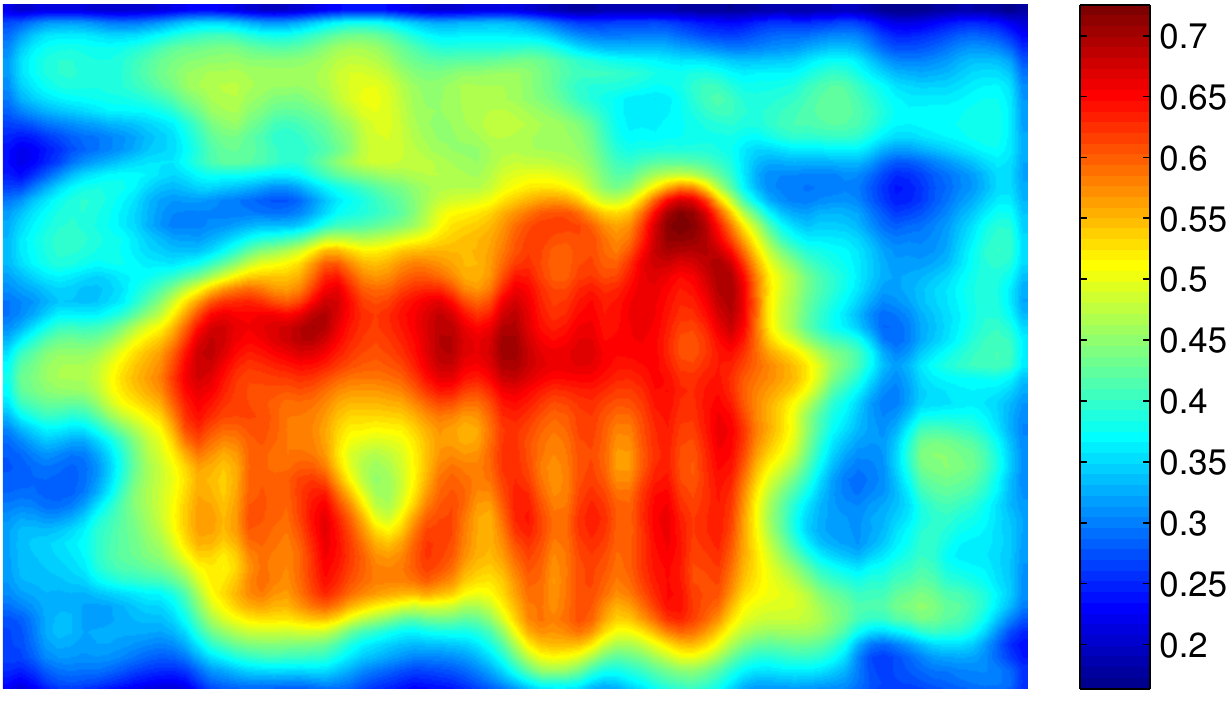} &
   \includegraphics[height=0.09\linewidth]{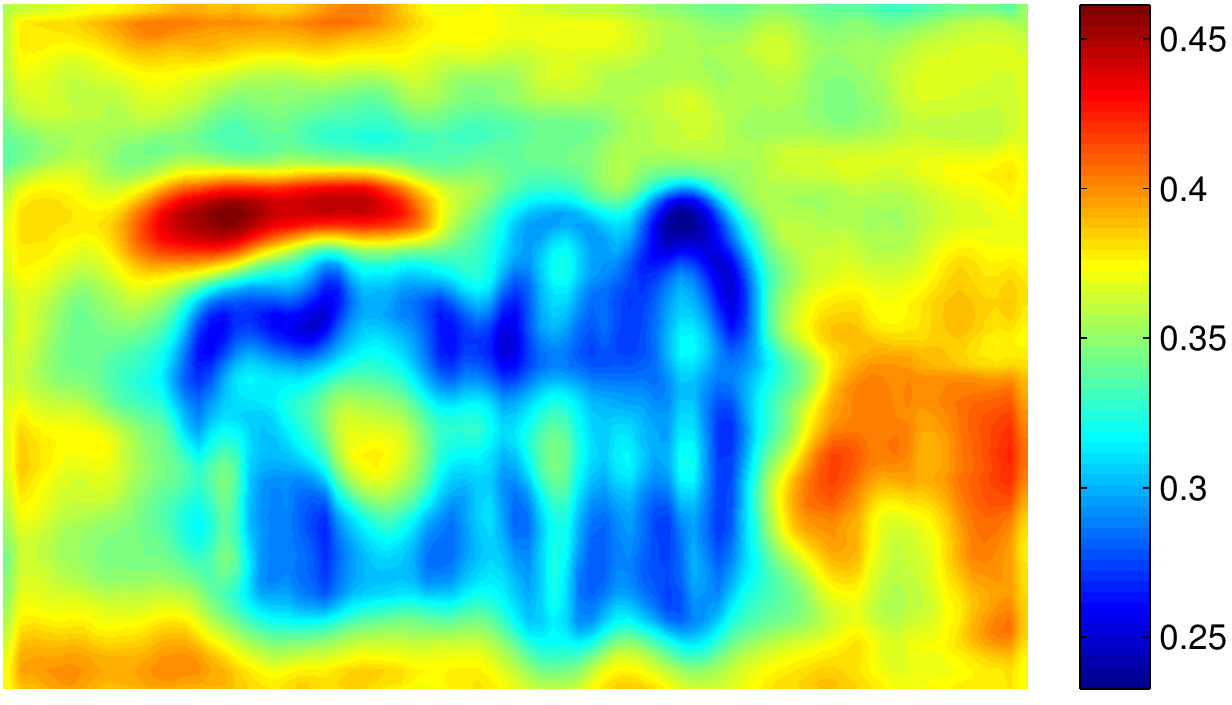} &
   \includegraphics[height=0.09\linewidth]{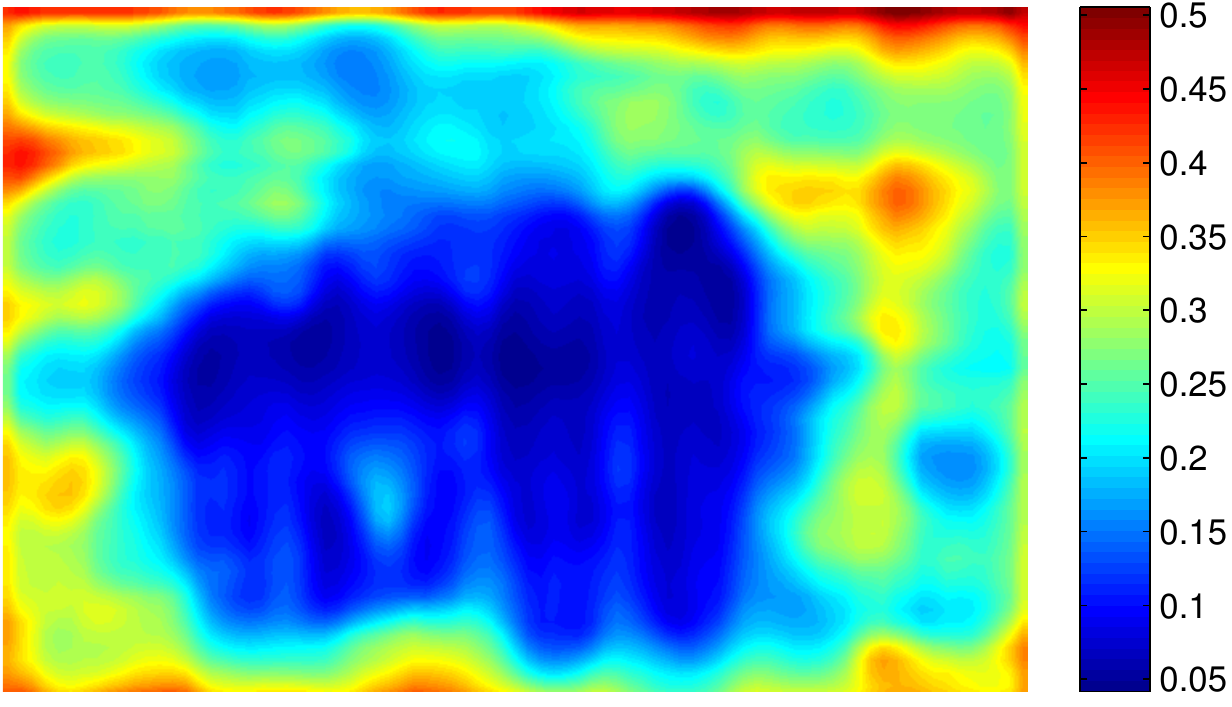} \\
   \includegraphics[height=0.09\linewidth]{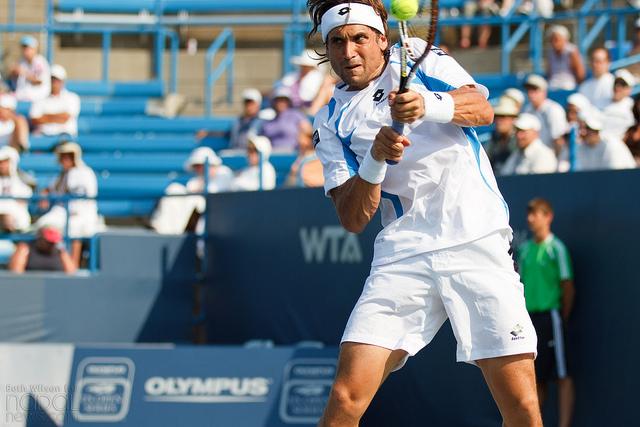} &
   \includegraphics[height=0.09\linewidth]{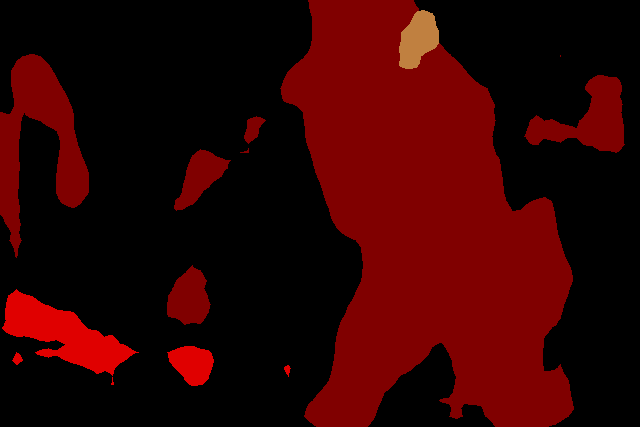} &
   \includegraphics[height=0.09\linewidth]{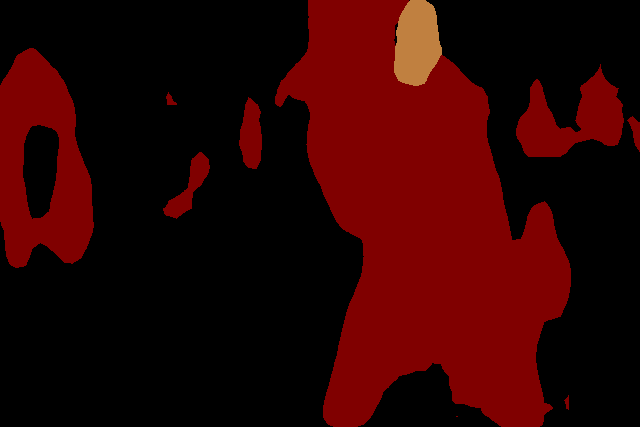} &
   \includegraphics[height=0.09\linewidth]{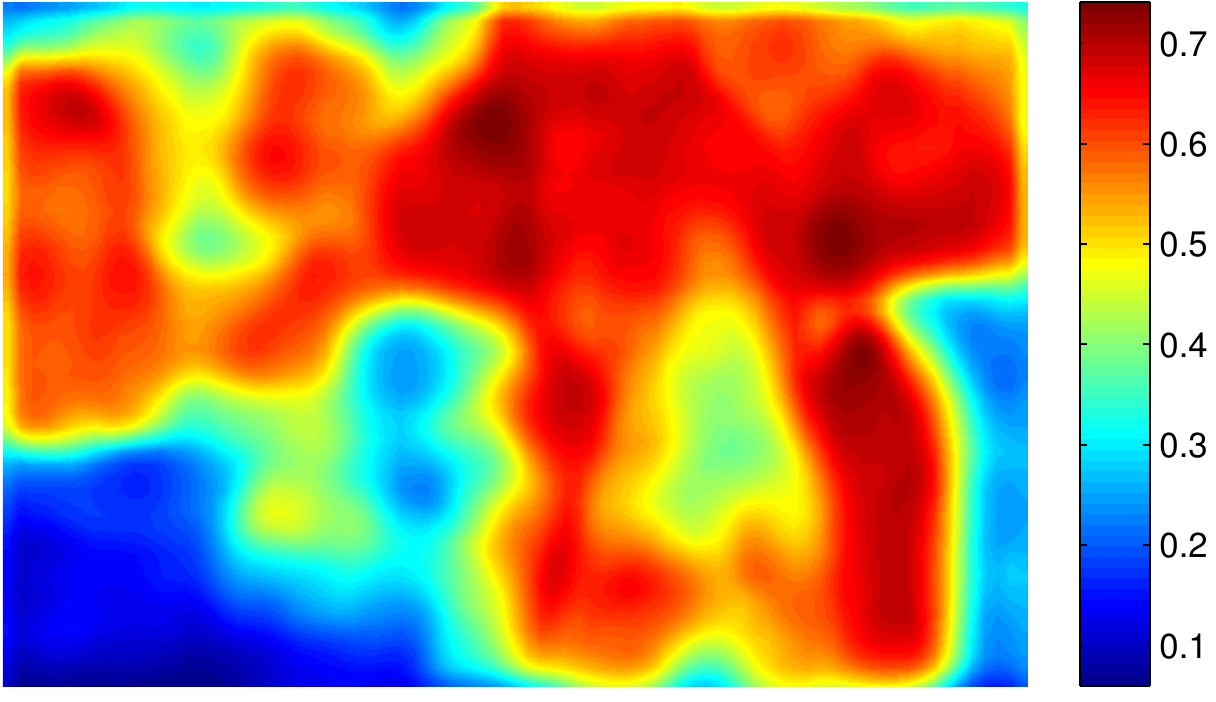} &
   \includegraphics[height=0.09\linewidth]{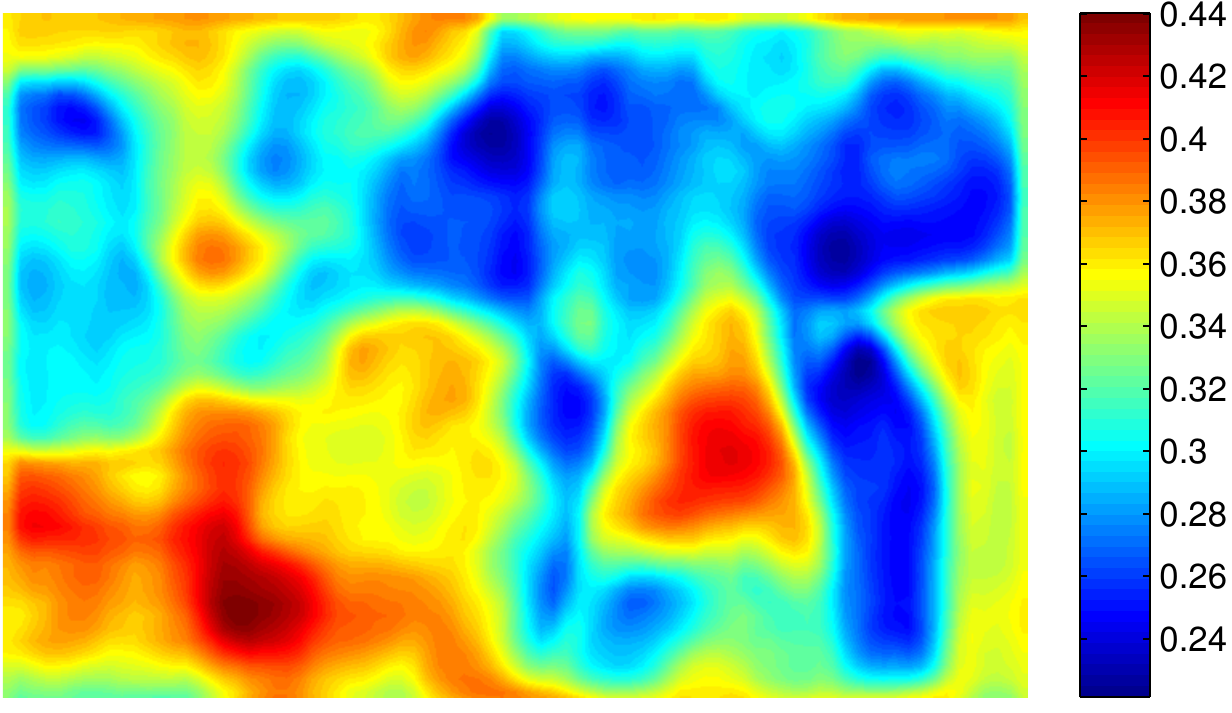} &
   \includegraphics[height=0.09\linewidth]{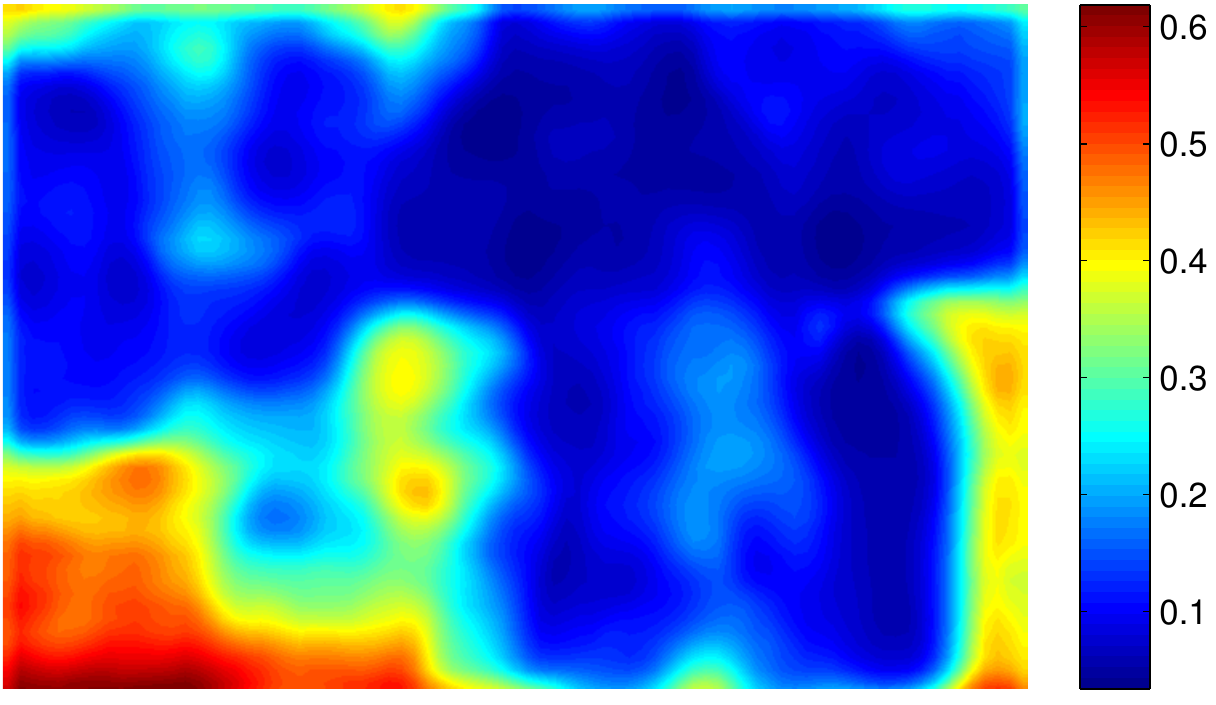} \\
   {\scriptsize (a) Image} & 
   {\scriptsize (b) Baseline} & 
   {\scriptsize (c) Our model} & 
   {\scriptsize (d) Scale-1 Attention} & 
   {\scriptsize (e) Scale-0.75 Attention} &
   {\scriptsize (f) Scale-0.5 Attention} \\
  \end{tabular}
  \caption{Qualitative segmentation results on subset of MS-COCO 2014 validation set.}
  \label{fig:pascal_coco_results2_app}  
\end{figure*}

We show more qualitative results on PASCAL-Person-Part \cite{chen_cvpr14} in \figref{fig:pascal_part_results_app}, on PASCAL VOC 2012 \cite{everingham2014pascal} in \figref{fig:pascal_voc12_results_app}, and on subset of MS-COCO 2014 \cite{lin2014microsoft} in \figref{fig:pascal_coco_results2_app}.

{\small
\bibliographystyle{ieee}
\bibliography{egbib}
}

\end{document}